\documentclass{article}

\usepackage[main, final]{neurips_2026}

\usepackage[utf8]{inputenc} 
\usepackage[T1]{fontenc}    
\usepackage{hyperref}       
\usepackage{url}            
\usepackage{booktabs}       
\usepackage{amsfonts}       
\usepackage{nicefrac}       
\usepackage{microtype}      
\usepackage{arydshln}      
\usepackage{hyperref}
\usepackage{bbm}
\usepackage{multirow}
\usepackage[dvipsnames,table]{xcolor}
\usepackage{tcolorbox}
\usepackage{subfig}
\usepackage{tikz}
\usepackage{chemfig}
\usepackage[version=4]{mhchem}

\tcbuselibrary{skins, breakable}
\usepackage{xurl}
\usepackage{iftex}\ifXeTeX\else\usepackage{breakurl}\fi  
\usepackage[export]{adjustbox}
\usepackage{wrapfig}
\usepackage{booktabs}
\usepackage{array}

\usepackage[toc,page,header]{appendix}

\usepackage{amsmath}
\usepackage{amssymb}
\usepackage{mathtools}
\usepackage{amsthm}
\usepackage{algorithm}
\usepackage{algorithmic}    
\usepackage[capitalize,noabbrev]{cleveref}

\usepackage{enumitem}
\usepackage{amsmath}
\usepackage{chemfig}
\usepackage{chemmacros}
\usepackage[toc,page,header]{appendix}
\usepackage{minitoc}

\definecolor{royalblue}{RGB}{65,105,225}
\definecolor{royalblue2}{RGB}{135, 160, 240}
\definecolor{lightblue}{RGB}{248,250,255}
\definecolor{contentbg}{RGB}{241,245,249}
\definecolor{scalebg}{RGB}{219,234,254}
\definecolor{emphasisyellow}{RGB}{254,243,199}
\definecolor{emphasisyellow2}{RGB}{253, 224, 107} 
\definecolor{moleculered}{RGB}{254,226,226}
\definecolor{moleculered2}{RGB}{220,38,38}
\definecolor{navyblue}{RGB}{30,58,138}
\definecolor{mediumblue}{RGB}{52, 160, 164}
\tcbuselibrary{skins,breakable}

\definecolor{promptnavy}{RGB}{30,58,110}
\definecolor{promptrule}{RGB}{203,213,225}
\definecolor{promptbg}{RGB}{250,251,253}
\definecolor{promptscalebg}{RGB}{235,241,250}
\definecolor{promptamber}{RGB}{217,119,6}
\definecolor{promptamberbg}{RGB}{255,248,235}
\definecolor{promptslate}{RGB}{100,116,139}

\newtcolorbox{promptbox}[1]{
    enhanced,
    arc=2pt,
    boxrule=0.6pt,
    colframe=promptrule,
    colback=promptbg,
    title={\textsc{Prompt}\enspace\textbar\enspace #1},
    fonttitle=\bfseries\small,
    coltitle=white,
    colbacktitle=promptnavy,
    attach boxed title to top left={xshift=8pt, yshift=-\tcboxedtitleheight/2},
    boxed title style={arc=1.5pt, boxrule=0pt, left=5pt, right=5pt, top=2pt, bottom=2pt},
    top=\tcboxedtitleheight/2+4pt,
    bottom=6pt,
    left=9pt,
    right=9pt,
    fontupper=\small,
    before upper={\setlist[itemize]{leftmargin=14pt, itemsep=1pt, topsep=3pt, parsep=0pt, label=\textcolor{promptnavy}{\scriptsize$\blacktriangleright$}}%
                  \setlist[enumerate]{leftmargin=14pt, itemsep=1pt, topsep=3pt, parsep=0pt}%
                  \setlength{\parskip}{4pt}},
    width=\textwidth,
    before skip=4pt,
    after skip=6pt,
}

\newtcolorbox{clusterscale}{
    enhanced,
    frame hidden,
    arc=0pt,
    colback=promptscalebg,
    borderline west={2.5pt}{0pt}{promptnavy},
    left=7pt, right=6pt, top=3pt, bottom=3pt,
    before skip=4pt, after skip=4pt,
    fontupper=\small
}

\newtcolorbox{instruction}{
    enhanced,
    frame hidden,
    arc=0pt,
    colback=promptamberbg,
    borderline west={2.5pt}{0pt}{promptamber},
    left=7pt, right=6pt, top=3pt, bottom=3pt,
    before skip=6pt, after skip=6pt,
    fontupper=\small
}

\newtcolorbox{molecules}{
    enhanced,
    frame hidden,
    arc=1.5pt,
    colback=white,
    borderline={0.6pt}{0pt}{promptslate, dashed},
    left=7pt, right=6pt, top=3pt, bottom=3pt,
    before skip=4pt, after skip=0pt,
    halign=center,
    fontupper=\ttfamily\footnotesize,
    coltext=promptslate
}

\theoremstyle{plain}
\newtheorem{theorem}{Theorem}[section]
\newtheorem{proposition}[theorem]{Proposition}

\theoremstyle{definition}

\newtheorem{assumption}[theorem]{Assumption}
\theoremstyle{remark}

\usepackage{amsmath,amsfonts,bm}

\def\eqref#1{equation~\ref{#1}}

\def\1{\bm{1}}

\DeclareMathAlphabet{\mathsfit}{\encodingdefault}{\sfdefault}{m}{sl}
\SetMathAlphabet{\mathsfit}{bold}{\encodingdefault}{\sfdefault}{bx}{n}

\newcommand{\KL}{D_{\mathrm{KL}}}

\DeclareMathOperator*{\argmax}{arg\,max}

\newcommand{\ie}{{\em i.e.\/}}
\newcommand{\eg}{{\em e.g.\/}}

\newcommand{\oneb}{{\mathbf{1}}}

\newcommand{\Ncal}{{\mathcal N}}
\newcommand{\Pcal}{{\mathcal P}}

\newcommand{\Fcal}{{\mathcal F}}

\newcommand{\Lcal}{{\mathcal L}}

\newcommand{\Ocal}{{\mathcal O}}

\newcommand{\Xcal}{{\mathcal X}}
\newcommand{\Ycal}{{\mathcal Y}}

\newcommand{\Scal}{{\mathcal S}}

\newcommand{\Reals}{{\mathbb{R}}}

\usepackage[utf8]{inputenc} 
\usepackage[T1]{fontenc}    
\usepackage{hyperref}       
\usepackage{url}            
\usepackage{booktabs}       
\usepackage{amsfonts}       
\usepackage{nicefrac}       
\usepackage{microtype}      
\usepackage{xcolor}         

\title{Informing Acquisition Functions via Foundation Models for Molecular Discovery}

\author{
Qi Chen$^{3,4}\thanks{The initial work was conducted at the University of Toronto, with subsequent improvements at Aalto University.\\ Correspondence to: qi.chen.inf@aalto.fi, florian@cs.toronto.edu.}$, 
Fabio Ramos$^{8,9}$,
Al\'an Aspuru-Guzik$^{1,6,7,10}$,
Florian Shkurti$^{1,2,5,6,7}\footnotemark[1]$\\
$^1$ University of Toronto (UofT),
$^2$ UofT Data Science Institute,\\
$^3$ ELLIS Institute Finland,
$^4$ Aalto University\\
$^5$ UofT Robotics Institute, $^6$ Acceleration Consortium,
$^7$ Vector Institute\\
$^8$ NVIDIA Research Seattle Robotics Lab (SRL),
$^9$ University of Sydney\\
$^{10}$ Senior Fellow, Canadian Institute for Advanced Research (CIFAR)
}

\begin{document}
\doparttoc 
\faketableofcontents 

\maketitle

\begin{abstract}
Bayesian optimization (BO) is widely used to accelerate molecular discovery by
reducing costly oracle evaluations. Foundation models provide a promising source
of prior knowledge for BO, but simply using their representations in standard
surrogate-based pipelines is often unreliable in low-data regimes with
high-dimensional features and vast discrete candidate spaces. Rather than
asking whether LLMs are universally useful for molecular BO, we ask how
Acquisition Functions (AFs) should be designed to exploit weak, high-dimensional, and
partially informative foundation-model priors. We propose \emph{LLM-guided Acquisition Tree} (LLMAT), a surrogate-free BO
framework that reformulates likelihood-free acquisition estimation as recursive
local acquisition learning. LLMAT trains binary classifiers on
LLM representations, where each classifier jointly induces a promising/non-promising partition and defines a local AF within the corresponding region. This produces a hierarchy of localized AFs and enables efficient candidate selection via Monte Carlo Tree Search. To stabilize learning with few observations, LLMAT meta-learns shared classifier parameters and initializations across tree nodes. An optional LLM-guided clustering module further reduces AF evaluation cost by restricting search to statistically promising coarse property clusters. Extensive experiments and ablations demonstrate consistent improvements in scalability, robustness, and sample efficiency for LLM-guided BO in molecular discovery.
\end{abstract}

\section{Introduction}

Discovering molecules with desirable properties is crucial for drug design, materials science, and chemical engineering. Since exhaustive evaluation over
the vast chemical space \citep{restrepo_chemical_2022} is infeasible, Bayesian optimization (BO, \citet{frazier_tutorial_2018, garnett_bayesian_2023}) has become a standard tool for reducing the number of costly oracle evaluations, such as DFT simulations \citep{parr_density_1989} and laborious experimental measurements.
BO typically relies on a probabilistic surrogate model (\eg, Gaussian Processes (GPs) and Bayesian Neural Networks (BNNs)) to define an Acquisition Function (AF), which selects the next molecule by balancing exploration and exploitation.

Foundation models offer a promising source of prior knowledge for molecular BO, as they encode information from large-scale text and molecular corpora. However, recent studies \citep{kristiadi_sober_2024} suggest that simply plugging LLM representations into standard BO pipelines does not reliably lead to better molecular optimization: fixed LLM features or Parameter-Efficient Fine-Tuning (PEFT)-enhanced \citep{hu_lora_2021} neural surrogates can still be hampered by high-dimensional representations, low-data overfitting \citep{xie_bogrape_2025}, and the cost of repeatedly fitting Bayesian surrogates over large discrete candidate pools. This raises a more fundamental question: not whether foundation-model features are informative, but how AFs should be designed to exploit weak, high-dimensional, and partially informative priors in the low-data regime.

Our central hypothesis is that the bottleneck lies not only in the representation itself, but also in the global nature of conventional AFs. When only a handful of labeled molecules are available, fitting a single global surrogate or AF over high-dimensional foundation-model embeddings can obscure local structure and lead to unstable candidate selection.

\begin{figure*}[ht]
\centering
\includegraphics[height=4.8cm]{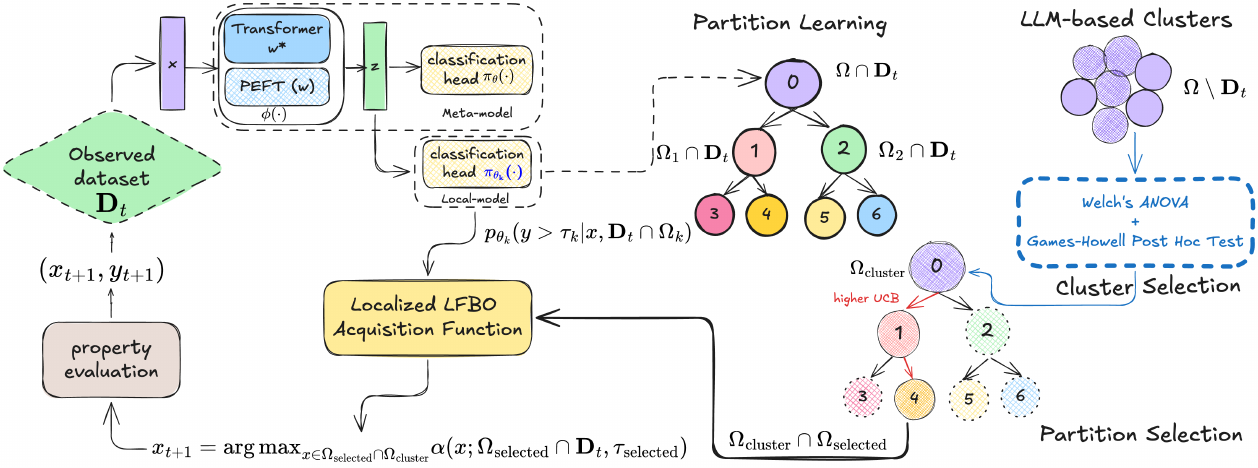}
\caption{Overview of LLMAT: Binary classifiers on the LLM features recursively partition the candidate space to a tree structure and local AFs are learned from observations in each partition $\Omega_k \cap \mathbf{D}_t$. Promising partition $\Omega_{\text{selected}}$ is selected via UCT. Unseen candidates are clustered via LLM prompts, and promising clusters $\Omega_{\text{cluster}}$ are selected via statistical tests. Within $\Omega_{\text{cluster}} \cap \Omega_{\text{selected}}$, the property of the candidate with the highest local AF value is evaluated and added to the dataset.}
\label{fig:bo} 
\end{figure*}

To address this, we propose \texttt{LLM-guided Acquisition Tree} (LLMAT), illustrated in Fig.~\ref{fig:bo}: a surrogate-free BO framework that turns likelihood-free AF estimation into a recursive local search procedure. Instead of fitting a global Bayesian surrogate, LLMAT learns binary classifiers on top of foundation-model representations. The key novelty is not the use of tree search or likelihood-free BO separately, but the identification of the likelihood-free acquisition classifier as a single object that simultaneously estimates a local acquisition function and induces the next search-space partition. Each classifier therefore serves two coupled roles: partitioning the candidate space into more and less promising regions, and defining a local likelihood-free AF within each induced region.

This formulation is especially beneficial in the low-data regime, where each tree node may contain only a small number of observations. We therefore meta-learn shared classifier and PEFT parameters across nodes, yielding a stable initialization for local AF estimation. We further incorporate an optional LLM-guided clustering module that uses foundation models not as property predictors, but to identify coarse, property-relevant regions of the molecular search space, focusing AF evaluation on statistically promising clusters. The consistency and utility of these clusters across different LLM APIs are exemplified and statistically validated in Sec.~\ref{sec:llm-ranking}.

To summarize, the main contributions of our paper are:
\begin{itemize}
    \item We introduce a foundation-model-informed AF approach for molecular BO. Unlike standard BO methods that estimate a single global AF, LLMAT recursively learns local AFs over classifier-induced partitions of the molecular search space.
    \item We show that likelihood-free acquisition classifiers can serve as a unified algorithmic primitive for both density-ratio acquisition estimation and adaptive search-space partitioning. This is the central algorithmic distinction of LLMAT: the same classifier both scores candidates through a local AF and defines the next partition, rather than combining a separately learned partitioning policy with a separately defined acquisition function. The resulting formulation naturally supports fixed foundation-model features and PEFT-based adaptation.
    
    \item We stabilize local AF learning in the low-data regime through meta-learned shared classifier and PEFT parameters, enabling reliable use of high-dimensional foundation-model representations.
    \item We introduce an optional LLM-guided clustering module for scalable AF evaluation and validate LLMAT on six molecular discovery benchmarks, where it improves sample efficiency, robustness, and computational cost over previous LLM-based BO baselines.
\end{itemize}

\section{Related Work}
Due to space limitations, a more detailed discussion of related works is provided in Appendix~\ref{sec:additional_related}.

\textbf{LLMs for Molecule Discovery.} 
Recent studies have explored the potential of LLMs in a variety of chemistry-related tasks, including molecular property prediction \citep{lu_unified_2022, christofidellis_unifying_2023, jablonka_14_2023}, property-related molecule optimization \citep{ramos_bayesian_2023, kristiadi_sober_2024}, and molecular generation with desired properties \citep{liu_conversational_2024, flam-shepherd_language_2023, jablonka_leveraging_2024}. However, many of these LLM-based approaches \citep{ramos_bayesian_2023, guo_what_2023} rely heavily on in-context learning (ICL) and prompt engineering. This dependency poses challenges for tasks involving strict numerical objectives, where LLMs often fall short in satisfying precise quantitative constraints or optimizing for specific target values \citep{ai4science_impact_2023}. To address this limitation, recent efforts have augmented LLMs with optimization frameworks such as BO or Evolutionary Algorithms (EA). For example, \citet{kristiadi_sober_2024} integrate Laplace approximations with BNN for property prediction, while \citet{wang_efficient_2024} combine EA to generate and search molecules with desired property. 

\textbf{LLMs for Bayesian Optimization over Molecules.}
LLMs and chemistry foundation models encode useful priors from text and molecular corpora, but exploiting them for BO remains challenging in low-data regimes. Prior work has used LLMs either through adaptive in-context prompting~\citep{ramos_bayesian_2023}, which is costly and context-limited, or as fixed/PEFT feature extractors with Bayesian neural surrogates~\citep{kristiadi_sober_2024}, which require expensive Laplace approximations over large candidate sets. Moreover, a single global surrogate can be unreliable with few observations and high-dimensional LLM embeddings~\citep{wang_learning_2020}. In contrast, LLMAT uses foundation-model priors to directly learn localized acquisition functions without surrogate fitting, and optionally restricts AF evaluation to promising LLM-induced clusters for scalability.

\textbf{Difference from tree-partitioned black-box optimization.}~The closest algorithmic line is tree-partitioned black-box optimization \citep{wang_learning_2020,wang_sample-efficient_2021,li_navigating_2025}, which adaptively partitions the search space and uses tree search or
partition-level weighting to guide candidate selection. However, these methods typically decouple partition learning from acquisition estimation: partitions are induced by K-means clustering and auxiliary classifiers, while candidates are scored by separate surrogate models. In contrast, LLMAT uses the same likelihood-free acquisition classifier to both estimate a local density-ratio AF and induce the next partition. Thus, LLMAT differs from prior tree-partitioned BO not by merely adding MCTS to LFBO, but by making the LFBO classifier itself the mechanism for both local acquisition estimation and recursive partition learning. This unifies search-space refinement and acquisition learning, turning tree search into a mechanism for recursive local AF learning, which is particularly suitable for LLM-based BO with PEFT-based adaptation.


\section{Background}
\subsection{Bayesian Optimization}\label{sec:BO}
Bayesian Optimization (BO) aims to effectively maximize some unknown function $f:\Xcal \rightarrow \Ycal$ over the candidate space $\Xcal$, \ie, find $x^* = \arg\max_{x\in\mathcal{X}}f(x)$ \footnote{It can be extended to minimization without loss of generality.}, given a dataset $\mathbf{D}_t=\{x_i, y_i\}_{i=1}^t$ of $t$ observations, where $x_i$ is a molecule, $y_i=f(x_i)+\epsilon, \forall i \in [t]$ is a property value, and $\epsilon$ is noise conventionally assumed to be a Gaussian with $\epsilon \sim \Ncal(0, \sigma^2)$. Due to the intractability of $f$, a probabilistic surrogate model $\hat{f}$ learned from $\mathbf{D}_t$ is often used to approximate $f$ with consideration of epistemic uncertainty, reflected in the variance of the posterior $p(\hat{f}|\mathbf{D}_t)$. Hence, for any unobserved data $(x, y)$, we have the predictive posterior $p(y|x, \mathbf{D}_t)=\int p(y|\hat{f}(x)) p(\hat{f}(x)| \mathbf{D}_t) d \hat{f}(x)$. 

\textbf{Acquisition Function and Expected Utility.} To address problems in the sequential decision-making setting, BO methods typically select the next query $x_{t+1}$ by maximizing an acquisition function $\alpha: \mathcal{X} \to \mathbb{R}$, defined as:  
\begin{equation}\label{eq:af}
\alpha(x; \mathbf{D}_t, \tau) := \mathbb{E}_{y \sim p(y | x, \mathbf{D}_t)}[u(y; \tau)],    
\end{equation}
where $x_{t+1} = \arg\max_{x\in \Xcal} \alpha(x;\mathbf{D}_t, \tau)$, $u(y; \tau)$ is a chosen utility function and $\tau$ is a threshold on $y$. The selection of $\tau$ controls the exploration-exploitation trade-off, and sometimes is set as $\tau = \max_t y_t$, the best observed value so far (the density-ratio form below instead sets $\tau$ as a quantile of the observed $y$). Common choices of the utility function $u(y; \tau)$ satisfying Eq.(1) include: $u^{EI}(y;\tau)=\max(y-\tau, 0)$ for Expected Improvement (EI, \citet{mockus_toward_1978}) that measures how much $y$ exceeds the threshold $\tau$;  $u^{PI}(y;\tau)=\oneb(y-\tau>0)$, indicating whether $y$ exceeds $\tau$ for Probability of Improvement (PI, \citet{kushner_new_1964}); Many other AFs can be expressed in this form with a more complex utility function, such as Upper Confidence Bound (UCB, \citet{srinivas_gaussian_2009}), Entropy Search (ES, \cite{hennig_entropy_2012}), Knowledge Gradient (KG, \citet{frazier_knowledge-gradient_2008}), Thompson Sampling (TS, \citet{thompson_likelihood_1933}).

\textbf{Direct Estimation of Acquisition Functions.} 
Typical BO instantiations can be characterized as \emph{indirect}: they first approximate the predictive posterior 
\(p(y \mid x, \mathbf{D}_t)\) via a surrogate model based on Gaussian Processes (GPs) 
or Bayesian Neural Networks (BNNs), and then compute the acquisition functions via Eq.~(\ref{eq:af}) for 
a given utility function. We provide an introduction to this in Appendix~\ref{sec: indirect}. However, these 
methods often face scalability issues due to the high computational cost when combining large, deep models, which limits their flexibility.
Our proposed method on the other hand directly models AFs, forgoing the need for surrogate model. Let $l(x)=p(x|y\leq\tau,\mathbf{D}_t)$ and $g(x)=p(x|y>\tau,\mathbf{D}_t)$ be two densities that respectively characterize the data distribution on non-promising and promising region of candidate space.  Instead of explicitly modeling the predictive posterior $p(y|x,\mathbf{D}_t)$, \citet{bergstra_algorithms_2011} and \citet{tiao_bore_2021} directly model the AF as $\gamma$-relative density ratio:
\begin{equation}
r_{\gamma}(x):=\frac{g(x)}{\gamma g(x) + (1-\gamma)l(x)}=h_{\gamma}(r_0(x)) \,,
\end{equation}
 where $h_{\gamma}:u\rightarrow(\gamma + u^{-1}(1-\gamma))^{-1}, u>0$ is strictly non-decreasing, $r_0(x)=\frac{g(x)}{l(x)}$ is the ordinary density ratio, and $\gamma=p(y>\tau|\mathbf{D}_t)$ indicates $\tau$ is set as the $(1-\gamma)$-quantile of all observed $y$, i.e., $\tau=F^{-1}(1-\gamma)$ with $F$ the CDF of $y$, so that a fraction $\gamma$ of observations exceed $\tau$. This density ratio has been proven by \citet{song_general_2022} to be equivalent to PI.
Tree Parzen Estimator (TPE, \citep{bergstra_algorithms_2011}) estimates the density ratio $r_0(x)$ by separately estimating $l(x)$ and $g(x)$ with tree variant of Kernel Density Estimation (KDE).
BORE \citep{tiao_bore_2021} avoids such indirect estimation by formulating the problem as a binary classification and showing the classifier $\pi_{\theta}(x)\approx p(c=1|x,\mathbf{D}_t)=\gamma r_{\gamma}(x)$, where $c=\oneb(y > \tau)$ is the class label and $p(c=1|\mathbf{D}_t) = p( y> \tau|\mathbf{D}_t) =\gamma$. Likelihood-free Bayes Optimization (LFBO, \citet{song_general_2022}) adopts variational $\phi-$divergence estimation to directly estimate expected utility AFs for complex $p(y|x,\mathbf{D}_t)$ and $u(y;\tau)$, not limited to PI. Then, the likelihood-free AF is defined as:   
\begin{equation}
\alpha(x;\mathbf{D}_t,\tau):=\pi_{\theta^*}(x)/(1-\pi_{\theta^*}(x))\,,
\end{equation}
where $\pi_{\theta}(x)\in(0,1)$ is the output of a generic binary classifier with parameters $\theta$ (it need not equal $p(c=1|x,\mathbf{D}_t)$, which is only the minimizer of BORE's unweighted cross-entropy), and $\theta^*= \arg\min_{\theta} \Lcal^{\tau}_{\mathbf{D}_t}(\theta)$ with the utility-weighted loss
$$\Lcal^{\tau}_{\mathbf{D}_t}(\theta):= \mathbb{E}_{(x,y)\sim \mathbf{D}_t}[-u(y;\tau)\log \pi_{\theta}(x) - \log(1-\pi_{\theta}(x))]\,.$$
\subsection{BO for Molecular Discovery}

In molecular discovery, the goal is to identify a novel molecule $x \in \Xcal$ with desirable properties $y \in \Ycal \subseteq \Reals$, often requiring exploration of a \textbf{vast} and \textbf{discrete} search space $\Xcal$ (estimated to contain over $10^{100}$ unique molecules). Exhaustive property evaluation is infeasible and costly, e.g., when relying on Density Functional Theory (DFT). In practice, we often restrict the search to a smaller candidate subset $\Omega \subset \Xcal$. At each round $t+1$, given prior observations $\mathbf{D}_t=\{x_i, y_i\}_{i=1}^{t} \cup \mathbf{D}_0$ with $n$ initial points $\mathbf{D}_0=\{\tilde{x}_j, \tilde{y}_j\}_{j=1}^n$, a new candidate $x_{t+1}$ is selected from $\Omega \setminus \mathbf{D}_t :=\{x_i \in \Omega \mid x_i \notin \mathrm{supp}_X(\mathbf{D}_t)\}$
for evaluation, aiming to find the optimal molecule $x^*$ in as few rounds as possible. This naturally fits the sequential BO framework.  

To input molecules into machine learning models, common representations \citep{griffiths_gauche_2023,janakarajan_language_2024} include: (1) text-based formats such as SMILES \citep{weininger_smiles_1988}, SELFIES \citep{krenn_self-referencing_2020}, and IUPAC names; (2) feature-based fingerprints such as Morgan \citep{morgan_generation_1965} and ECFPs \citep{rogers_extended-connectivity_2010}; and (3) graph-based encodings \citep{duvenaud_convolutional_2015}. In this work, we focus on text-based representations, which can be directly generated by modern NLP models, including general-purpose LLMs, transformer-based chemical foundation models, and domain-specific LLMs specialized for chemical data.  

\section{Methodology}
Early BO rounds with few observations, high-dimensional molecular features, and large discrete candidate spaces make global surrogate fitting unreliable and full-space AF evaluation costly. LLMAT addresses these challenges by coupling \emph{where to search} with \emph{how to estimate acquisition values}. \textbf{(1)} It learns local AFs over classifier-induced partitions, using a meta-learned
initialization to stabilize local classifiers in the low-data regime. \textbf{(2)} It uses LLM-based clustering and statistical testing to prune the candidate pool to property-relevant regions before AF estimation. Unlike conventional BO, which optimizes a global AF over the full candidate set, and unlike prior MCTS-based BO methods where the tree mainly serves as a search policy over externally defined partitions and surrogate scores, LLMAT makes the tree itself an acquisition-learning structure. Each likelihood-free classifier both estimates a local density-ratio AF and induces the next promising/non-promising partition, thereby coupling candidate-space refinement and acquisition learning in a single objective. This yields a hierarchy of localized AFs traversed by tree search, as illustrated in Fig.~\ref{fig:bo}. A non-rigorous theoretical analysis is given in Appendix~\ref{sec:theory} (without a full convergence guarantee): it interprets the node-wise classifiers as density-ratio estimators, shows why recursive local classifiers can outperform a global one, decomposes the error under PI into approximation, estimation and routing terms, and decomposes the cumulative regret.
\subsection{Meta-Learning Candidate Partitions and Local AFs}

We next describe the training of shared classifiers during tree construction, which simultaneously induce search-space partitions and define local AFs within each partition.



\textbf{Recursive Candidate Space Partitioning.} At any iteration $t$, we have the currently observed dataset $\mathbf{D}_t$ defined in Sec.~\ref{sec:BO}. As shown in Fig.~\ref{fig:bo}, we hierarchically partition the candidate set $\Omega$ by combining Monte Carlo Tree Search (MCTS), where each node in the tree corresponds to a subset of $\Omega$. The tree construction begins at the root node (node 0) with $\Omega_0 = \Omega$ being the entire space.

\begin{wrapfigure}{r}{0.54\textwidth}
\vspace{-20pt}
\centering
\scalebox{0.8}{
\begin{minipage}[t]{0.68\textwidth}
\begin{algorithm}[H]
   \caption{Meta-learning Tree Partitions and local AFs}
   \label{alg:MCTS}
\begin{algorithmic}[1]
   \STATE {\bfseries Input:} Observed dataset $\mathbf{D}_t$,  quantile $\gamma$, candidate set $\Omega$, tree depth $L$, classification head $\theta$, PEFT parameters $w$, learning rate $\eta$, SGD steps $K$, finetuning flag $\mathbb{F}$, minimal leaf sample size $m$. 
   \STATE Nodes $\Ncal = \{\}$;
   \STATE $\Omega_0=\Omega, \mathbf{D}_t\cap \Omega_0 = \mathbf{D}_t, \mathbf{D}_t\cap \Omega_k=\emptyset, \forall k> 0$;
   \FOR{ $k=0$ {\bfseries to} $2^{L+1}-2$}
   \IF{$|\Omega_k \cap \mathbf{D}_t| < m$ or $\Omega_k$ not splitable}
   \STATE continue;
   \ENDIF
   \STATE $\tau_k=\Phi_k^{-1}(\gamma)$; $\tilde{\theta} = \theta_{\lceil k/2\rceil -1}$ if $k>0$ else $\tilde{\theta}=\theta$;
   \STATE Fix $w$, update $\theta_k = \text{SGD}(\Lcal^{\tau_k}_{\mathbf{D}_t\cap \Omega_k}(\tilde{\theta}, w), K)$;
   \STATE $\theta \leftarrow \theta + \eta(\theta_k - \theta)$, $\Ncal \gets \Ncal \cup \{k\}$;
   \FOR{$(x, y)$ in $\mathbf{D}_t\cap \Omega_k$}
   \IF{$\pi_{\theta_k}(x) > 0.5$}
   \STATE $\mathbf{D}_t\cap\Omega_{2k+1} \leftarrow (\mathbf{D}_t\cap\Omega_{2k+1}) \cup \{(x, y)\}$;
   \ELSE
   \STATE $\mathbf{D}_t\cap\Omega_{2k+2} \leftarrow (\mathbf{D}_t\cap\Omega_{2k+2}) \cup \{(x, y)\}$; 
   \ENDIF
   \ENDFOR
\ENDFOR
\IF{$\mathbb{F}$}
    \STATE Fix $\theta$, update $w = \text{PEFT}(\frac{1}{|\Ncal|}\sum_{k\in \Ncal} \Lcal^{\tau_k}_{\mathbf{D}_t\cap \Omega_k}(\theta, w))$;
\ENDIF
\STATE {\bfseries Output:} $\pi_{\theta_k}(x), \alpha(x;\Omega_k\cap\mathbf{D}_t, \tau_k), \forall k \in \Ncal$;
\end{algorithmic}
\end{algorithm}
\end{minipage}
}
\vspace{-15pt}
\end{wrapfigure}

Each node $\Omega_k$ is recursively bifurcated into two subsets using a binary classifier $\pi_{\theta_k}$. Let $\Omega_{2k+1} := \{x \in \Omega_k \mid \pi_{\theta_k}(x) > 0.5\}$ and $\Omega_{2k+2}:= \{x \in \Omega_k \mid \pi_{\theta_k}(x) \leq 0.5\}$ be the left and right children respectively, representing the more promising (i.e., more likely to contain high-property candidates) and less promising regions. The tree expansion continues until a predefined maximum depth $L$ or the minimum number of samples in each node is reached, yielding a total maximum of $2^{L+1} - 1$ nodes. To train the binary classifier at each node $\Omega_k$, we use the subset of observed samples within that node, i.e., $\mathbf{D}_t \cap \Omega_k := \{(x_i,y_i) \in \mathbf{D}_t \mid x_i \in \Omega_k \}$, where their class labels can be obtained by directly thresholding $y$ as we described in the following. For unobserved samples in $\Omega_k$, the learned classifier $\pi_{\theta_k}$ assigns them to the left or right child. When estimating the AFs, rather than classifying all samples across the whole tree, we only route the search from the root to the most promising leaf, using classifiers at each node along that path.

\textbf{LFBO Classifier Shared for Bifurcating.} In LFBO, the AF at each iteration $t$ is estimated directly by learning a (weighted) binary classifier $\pi_{\theta}(x)$ whose odds estimate the AF: pointwise, the minimizer of the weighted loss satisfies $\pi^*(x)/(1-\pi^*(x))=\mathbb{E}[u(y;\tau)|x]=\alpha(x)$, instead of training surrogate models for typical BO algorithms. This classifier can be reused as a natural choice for the aforementioned candidate space partitioning during the construction of the tree. 

\begin{wrapfigure}{r}{0.54\textwidth}
\vspace{-10pt}
\centering
\scalebox{0.8}{
\begin{minipage}[t]{0.65\textwidth}
\begin{algorithm}[H]
   \caption{Partition Selection for Refined and Reduced AFs}
   \label{algo:BO}
\begin{algorithmic}[1]
 \STATE {\bfseries Input:} Dataset $\mathbf{D}_t$, tree depth $L$, candidates in selected  clusters $\Omega_{\text{cluster}}$, classifier $\theta$, learning rate $\eta$, SGD steps $K$, quantile $\gamma$, UCT hyper-parameter $\lambda$, time budget $T$, all nodes $\Ncal$.
\FOR{$t=0$ {\bfseries to} $T$}
  \STATE Run Algo.~\ref{alg:MCTS};
  \STATE \textcolor{blue}{\#Select the search tree path.}
  \STATE Set $k=0$, $\tau_k=\Phi_k^{-1}(\gamma)$;
  \STATE Path $\Pcal \gets []$;
  \WHILE{$k \in \Ncal$}
    \STATE $\Pcal \gets \Pcal + [k]$;
    \STATE Compute partition score for $2k+1$ and $2k+2$;
    \STATE $k = \arg\max_{j \in \{2k+1, 2k+2\}} \Scal_j$;
  \ENDWHILE
  \STATE \textcolor{blue}{\#Backtrack and select local AF up the path.}
  \STATE $\Pcal \gets \text{Reverse}(\Pcal)$;
  \STATE selected = 0;
  \FOR{each $k$ in $\Pcal$}
      \IF{$|\Omega_k\cap\Omega_{\text{cluster}}| = 0$}
          \STATE continue;
    \ELSE
        \STATE selected = k;
        \STATE break;
    \ENDIF
  \ENDFOR
  \STATE Sample $x_{t+1} = \argmax\limits_{x\in\Omega_{\text{selected}}\cap\Omega_{\text{cluster}}} \alpha(x;\Omega_{\text{selected}}\cap \mathbf{D}_t, \tau_{\text{selected}})$;
  \STATE Evaluate $y_{t+1} = f(x_{t+1})$;
  \STATE Update $\mathbf{D}_{t+1} \gets \mathbf{D}_t \cup \{(x_{t+1}, y_{t+1})\}$;
\ENDFOR
\end{algorithmic}
\end{algorithm}
\end{minipage}
}
\vspace{-15pt}
\end{wrapfigure}
At each node $k$, we set $\tau_k=\Phi_k^{-1}(\gamma)$ under the same $\gamma$, where $\Phi_k(\tau):=p(y>\tau|\mathbf{D}_t\cap\Omega_k)=1-F_k(\tau)$ is the survival function of $y$ in node $k$ and $F_k$ its CDF. Hence $\tau_k=F_k^{-1}(1-\gamma)$ is the $(1-\gamma)$-quantile of the observed $y$ in $\mathbf{D}_t\cap\Omega_k$, i.e., a fraction $\gamma$ of the samples in the node lie above $\tau_k$. Then, the corresponding binary classifier is $\pi_{\theta_k}(x)=p_{\theta_k}(c=1|x, \mathbf{D}_t\cap\Omega_k)$ with $c = \oneb(y>\tau_k)$, where the binary labels $c$ for training the classifier are assigned by thresholding the $y$ for all training samples in node $k$ instead of using the binary clusters in~\citet{wang_learning_2020}.

\textbf{Meta-learning Shared Binary Classifiers.}
Using a shared classification head connected after the transformer feature extractor, we propose a meta-learning approach that works for both fixed features and parameter efficient fine-tuning (PEFT). Since the tree is built recursively, we use the sequential version of Reptile \citep{nichol2018first}, meta-training the classification head. For a maximal $L$-depth Tree, we store one meta-model $\theta$ and $|\Ncal|$ local models $\theta_k, \forall k \in \Ncal$.The detailed algorithm for learning the partitions and AFs at each iteration is presented in Algo.~\ref{alg:MCTS}.

\subsection{Candidate Selection on Promising Subsets}
With the built tree partition and learned local AFs, we can perform refined AF optimization via both partition and cluster selection to identify promising candidates efficiently.

\textbf{Partition Selection.} Given the aforementioned partition rule, a greedy strategy that always selects the left node would exploit the most promising leaf but risks overfitting to suboptimal partitions without sufficient exploration. To balance exploration and exploitation, we select partitions using some score metrics. For example, as in \citet{wang_learning_2020}, let 
$n_k = |\mathbf{D}_t \cap \Omega_k|$ denote the visit count for each node $k$, and let the node value 
$
v_k = \frac{1}{n_k} \sum_{i=1}^{n_k} y_i
$ 
be the average property value of observed samples at node $k$. Then, the Upper Confidence Bound for Trees (UCT) score for a node $k \in \Ncal$ is defined as:
$\Scal^{\text{UCT}}_k:= v_k + 2 \lambda \sqrt{2\log(n_{\lceil \frac{k}{2} \rceil-1})/n_k}$,
where $n_{\lceil \frac{k}{2} \rceil-1}$ is the number of visits at the parent node of $k$-th node and $\lambda$ is a hyper-parameter that controls the exploration-exploitation trade-off. Another possible choice is $\Scal^{\text{Var}}_k:= v_k + 2 \lambda \sqrt{\text{var}_k}$, where $\text{var}_k = \frac{1}{n_k} \sum_{i=1}^{n_k} (y_i -v_k)^2$. When $\lambda=0$, both recover the greedy policy if the classification is accurate enough (so that the left child has the larger $v_k$). Finally, we select the child node with the highest score from the root to a leaf. If no candidates reach a leaf node, we use a backtracking approach to select the parent node instead. The selected partition is denoted as $\Omega_{selected}$, as shown in Algo.~\ref{algo:BO}.

\textbf{Clustering for Efficient Estimation of AFs.}
To reduce the high GPU memory and computation cost of calculating AFs for all candidates at each BO round, we propose a clustering-based approach. Molecular feature representations are precomputed once and grouped into clusters. Candidates within the selected clusters then undergo the partition selection process, and the optimal candidate is determined by comparing AFs only for $\Omega_{\text{cluster}}\cap \Omega_{selected}$. 

\textbf{(1) Cluster Selection via Statistical Test.} We propose a statistical filtering approach on $\mathbf{D}_t$ to identify promising clusters for reduced AF estimation. Welch's ANOVA (Appendix~\ref{sec: welch}) serves as an omnibus test of equal mean property across clusters, and the Games-Howell post-hoc test (Appendix~\ref{sec:posthoc}) compares all pairs of clusters: for each pair that differs significantly at p-value $p$, the cluster with the lower mean is excluded. To support this, the BO initialization $\mathbf{D}_0$ is constructed by uniformly sampling from each cluster.


\textbf{(2) LLMs for Property-Related Clustering.} Unlike unsupervised methods (e.g., k-means), general LLMs (e.g., ChatGPT) classify molecules into extreme high/high/medium/low/extreme low property groups via tailored prompts (Appendix~\ref{sec:promtps}), leveraging chemical knowledge from their training data. This yields property-aligned clusters that enhance BO at a cost orders of magnitude lower than ICL-based approaches \citep{ramos_bayesian_2023}, offering a cost-effective way to exploit knowledge from general LLMs.

\section{Experiments}
In this section, we experimentally validate the proposed algorithm across multiple datasets and ablation scenarios, demonstrating its ability to accelerate molecular discovery through superior BO performance and increased computational efficiency. All results are reported as the mean over 15 independent runs with different random seeds, with shaded bands showing $\pm$ one standard error of the mean (SE). Each run starts from $10$ initial observations and uses a budget of $100$ BO rounds. Additional details are provided in Appendix~\ref{sec:exp_detail} and \ref{sec:add_results}. Our code is publicly available at \url{https://github.com/livreQ/LLMAT}.

\subsection{Setting} 

\textbf{Datasets.} We employ the well-known multi-modal Levy function \citep{levy_topics_2006} to generate a synthetic set to demonstrate the effectiveness of the proposed method in achieving more refined AF estimation within localized sub-regions. For molecular data, following \citet{kristiadi_sober_2024}, we initialize BO with $n=10$ points and evaluate our models on six benchmark datasets that capture realistic molecular design challenges across diverse domains: (1) minimizing redox potential for redoxmers (1,407 molecules), (2) minimizing solvation energy for flow battery electrolytes (1,407 molecules) \citep{agarwal_discovery_2021}, (3) minimizing docking scores of kinase inhibitors (10,449 molecules) for drug discovery \citep{graff_accelerating_2021}, (4) maximizing fluorescence oscillator strength for laser materials (10,000 molecules) \citep{strieth-kalthoff_delocalized_2024}, (5) maximizing power conversion efficiency (PCE) in photovoltaic materials (10,000 molecules) \citep{lopez_harvard_2016}, and (6) maximizing $\pi$--$\pi^*$ transition wavelengths for organic photoswitches (392 molecules) \citep{griffiths_data-driven_2022}. Together, these tasks cover a broad spectrum of molecular properties, providing a comprehensive testbed for molecular discovery. For each dataset, the physics-based simulators released by the original authors are used as the ground-truth oracles.
\begin{figure*}[ht]
    \centering
    \includegraphics[width=0.9\linewidth]{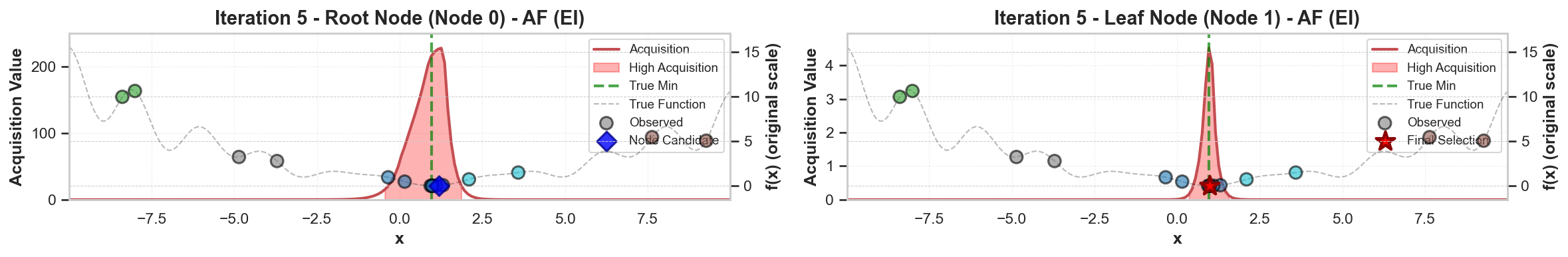}
    \caption{Leaf-node AFs offer more fine-grained candidate suggestions for Levy-1D optimization.}
    \label{fig:LLMAT_refine_AFs}
\vspace{-0.3cm}
\end{figure*}

\textbf{Fixed Features and Foundation Models.} To assess whether the findings of \citet{kristiadi_sober_2024} that LLMs benefit molecular BO only when trained on domain specific data hold for our algorithm, we benchmark LLMAT against several baselines on using different features: (i) Morgan fingerprints \citep{morgan_generation_1965} as a chemistry specific algorithmic representation, (ii) chemistry foundation models including T5-Chem \citep{christofidellis_unifying_2023} and MolFormer \citep{ross_large-scale_2022}, and (iii) general LLMs such as T5 \citep{raffel_exploring_2023}, GPT-2 (medium and large), and Llama-2-7b \citep{touvron_llama_2023}. Our comparison includes (1) Bayesian optimization using fixed feature representations extracted from these models, and (2) Parameter-Efficient Fine-Tuning (PEFT) of the above chemistry foundation models.

\textbf{Evaluation Metrics.}\label{sec:metric}
 We report the trajectory of the best observed property value over time and use the \textbf{GAP} metric defined in \citet{jiang_binoculars_2020}:
\(
\text{GAP}_t := \frac{y_t^* - y_0}{y^* - y_0},
\)
which normalizes the progression of the optimal value, where $y^*$ is the global optimum, $y_0$ is the initial optimum, and $y_t^*$ is the best value observed up to iteration $t$. However, in molecular discovery, identifying a single optimal molecule is often not the only goal. In practice, it is equally important to discover a \emph{set of high-performing candidates} that scientists can further evaluate, balance multiple criteria, or analyze for structure-property relationships. For this objective, we additionally report the \textbf{average regret}:
\(
\text{Regret}_t := \frac{1}{t}\sum_{i=1}^t |y^* - y_i|,
\)
which reflects the overall quality of the molecules discovered so far. When aggregating results across datasets, we normalize the average regrets for comparability. 

\subsection{Performance Analysis}

\textbf{Effectiveness of Localized Search on Synthetic Data.} Beyond LLM adaptation and its application to molecular discovery, our method contributes a novel discrete BO algorithm: we introduce shared binary classifiers that jointly serve both tree partitioning and localized AF approximation. To illustrate this benefit, we sample 1,000 examples from the Levy-1D function and observe that the proposed algorithm yields a more refined AF estimate at the selected leaf node than at the root node used by vanilla LFBO \citep{song_general_2022}. As shown in Fig.~\ref{fig:LLMAT_refine_AFs}, the leaf node exhibits a narrower confidence region and more accurate estimation around the true optimum, and the refinement becomes more pronounced over BO iterations as additional observations accumulate, see Sec.~\ref{sec:toy} in the Appendix.

\begin{figure}[H]
    \vspace{-0.2cm}
    \centering
    \includegraphics[width=0.32\linewidth]{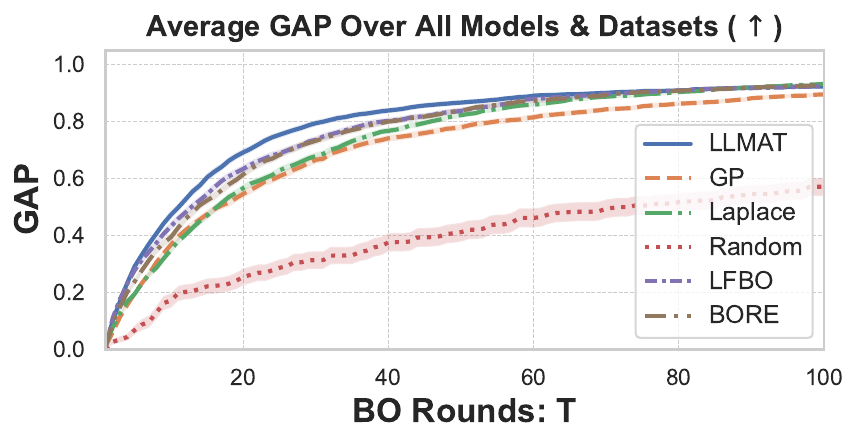}
    \includegraphics[width=0.32\linewidth]{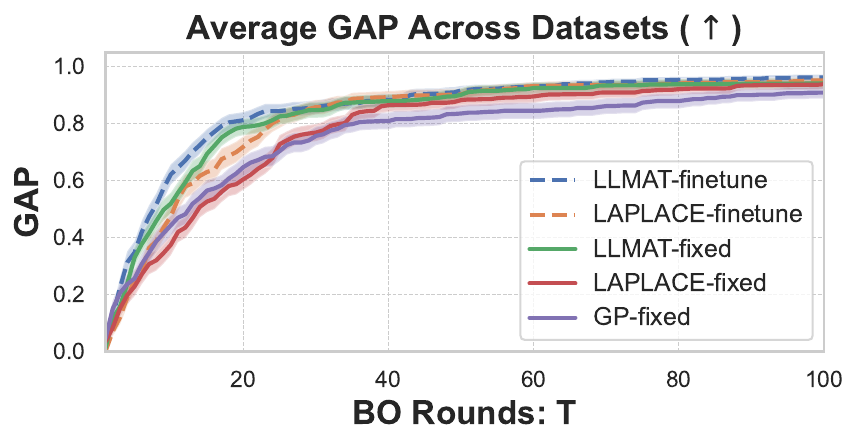}
    \includegraphics[width=0.32\linewidth]{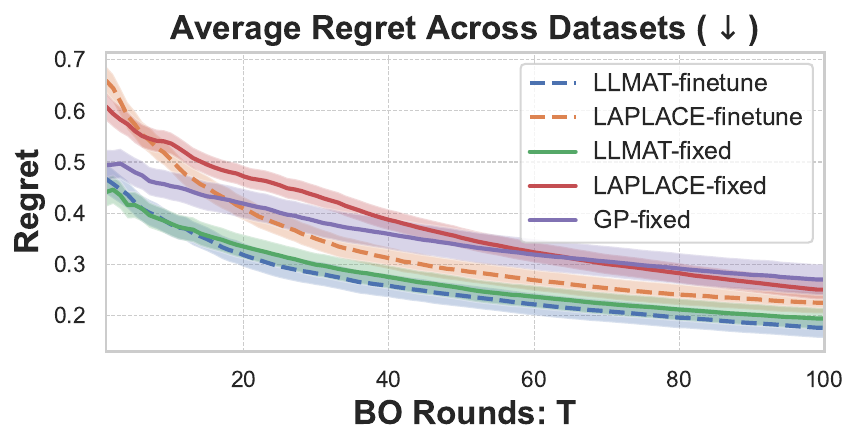}
    \caption{\textbf{Left:} Comparison over baselines on fixed features across all models and datasets. 
    \textbf{Middle and right:} Average GAP and regret, defined in Sec.~\ref{sec:metric}, across datasets for fine-tuned (PEFT on $w$) and fixed (frozen pretrained weights $w^*$) T5-Chem models.
    }
\label{fig:t5-chem-mol-finetuning}
\vspace{-0.5cm}
\end{figure}
  
\textbf{Superior Performance of LLMAT on Molecular Data.~} In the left panel of Fig.~\ref{fig:t5-chem-mol-finetuning}, we show that the proposed LLMAT algorithm outperforms standard discrete BO baselines when applied to fixed features extracted from the aforementioned models: Gaussian Processes (GP), the Laplace approximation (LAPLACE) \citep{kristiadi_sober_2024}, Likelihood-Free BO (LFBO), BORE \citep{tiao_bore_2021}, and random search (RANDOM), as measured by the average GAP metric across all datasets and models.
The middle and right panels of Fig.~\ref{fig:t5-chem-mol-finetuning} further compare LLMAT against LAPLACE across all datasets to assess iterative PEFT effectiveness with T5-Chem backbones. MolFormer PEFT results are in Fig.~25. While iterative PEFT enhances BO performance for both approaches, LAPLACE with PEFT still falls short of the performance achieved by our algorithm without PEFT. The radar chart on the left of Fig.~\ref{fig:molformer_tree_depth} summarizes the benefit of fine-tuning relative to a fixed representation, in magnitude and in speed. With $\Delta\mathrm{GAP}_t=\mathrm{GAP}^{\mathrm{FT}}_t-\mathrm{GAP}^{\mathrm{Fixed}}_t$ and $\Delta\mathrm{Regret}_t=\mathrm{Regret}^{\mathrm{Fixed}}_t-\mathrm{Regret}^{\mathrm{FT}}_t$ averaged over runs (positive values favor fine-tuning), it reports six metrics: the peak (maximum over rounds) and final (mean over the last 10 rounds) improvements of GAP and regret, and for each of them a peak-round score that is larger when the maximum improvement is reached earlier (Appendix~\ref{sec:radar}). It shows (1) faster convergence in early BO rounds, (2) superior final performance, and (3) greater stability across different foundation models. The PEFT improvements are more substantial for Molformer than T5-Chem, reflecting T5-Chem's larger scale and richer domain knowledge.



\begin{figure*}[!ht]
\vspace{-0.2cm}
    \centering
    \includegraphics[width=0.9\linewidth]{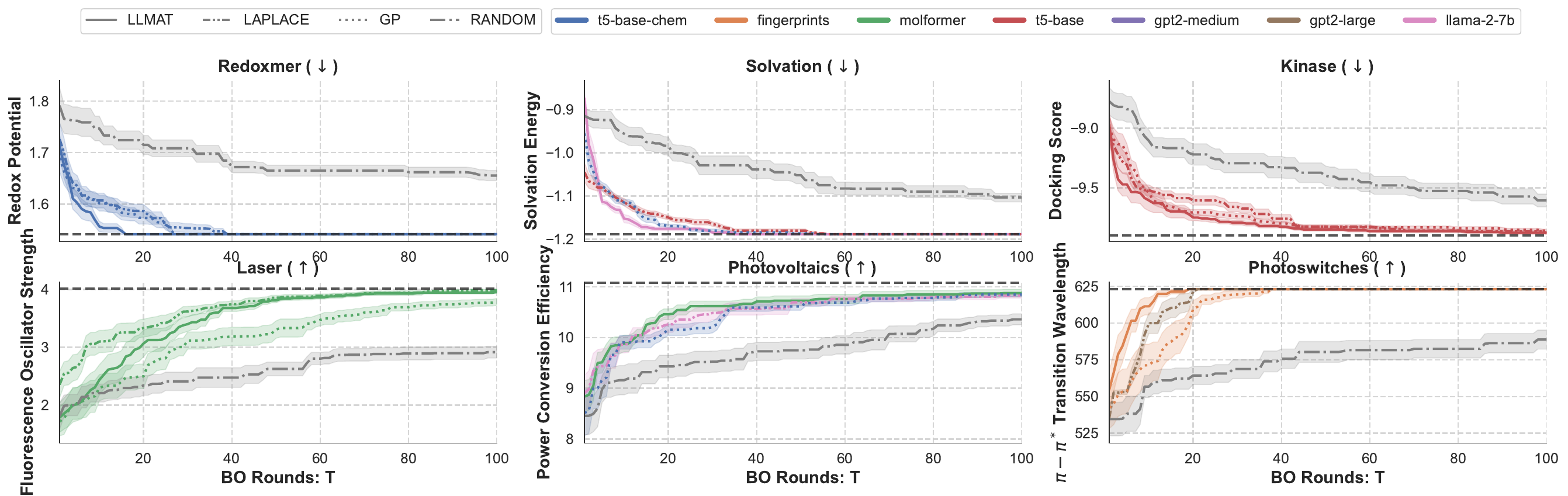}
    \caption{BO with fixed features (best-performing feature per algorithm shown). LLMAT achieves best overall performance, attaining top results on Solvation, Kinase, and Photoswitches using features from non-domain-specific models. The black dashed line marks the true optimum for each dataset.}
    \vspace{-0.2cm}
    \label{fig:fix-best-y}
\end{figure*}

\textbf{Better Exploitation of Priors from Fixed Features.~} In Fig.~\ref{fig:fix-best-y}, we compare LLMAT with maximal tree depth $L = 1$ of $3$ nodes against GP, LAPLACE, and RANDOM, using various fixed features mentioned before. We only plot the feature that achieved the best performance for each algorithm; the full plot is deferred to Fig.~\ref{fig:fix-best-y-all}.
Fig.~\ref{fig:fix-best-y} shows that LLMAT achieves superior performance compared to all baselines on most datasets, with slightly lower performance than LAPLACE on the Laser dataset. Other baselines exhibit substantial performance drops on several datasets while LLMAT remains consistent. LLMAT, LAPLACE, and GP achieve their best performance on the same feature models for the Redox-mer, Kinase, and Laser datasets (T5-Chem, T5, and MolFormer, respectively). However, the best feature models differ for other datasets. The strong performance of LLMAT with general-purpose LLMs such as T5, GPT2, and Llama-2-7b suggests that they can still provide valuable information, and Morgan fingerprints that do not rely on domain-specific knowledge are also competitive on some datasets. This contrasts with \citet{kristiadi_sober_2024}, who argue that LLMs are useful for BO over molecules only when pretrained or finetuned on domain-specific data, highlighting the importance of algorithmic design to fully exploit prior in different models. We also add results with fixed embeddings from a pretrained graph neural network (GIN; \citealp{xu_how_2019, hu_strategies_2020}), comparing LLMAT against the baselines on a graph representation in addition to SMILES-based LLM features; LLMAT also achieves the best average Gap AUC there (Appendix~\ref{sec:gin}).

\textbf{Improved Computational Efficiency.~} We also report training time, AF prediction time, and overall runtime per BO round using T5-Chem features in Fig.~\ref{fig:t5-chem-time-fix}. Our method achieves the shortest training time across all datasets. While prediction time increases slightly for the largest datasets (Kinase, Laser, Photovoltaics), this overhead is offset by GP's poor performance on them. By contrast, LAPLACE incurs the highest overall cost, especially for predicting AFs during PEFT in Fig.~\ref{fig:t5-chem-time-finetune},  making it the least efficient.

\begin{figure*}[ht]
    \centering
    \includegraphics[width=0.2\linewidth]{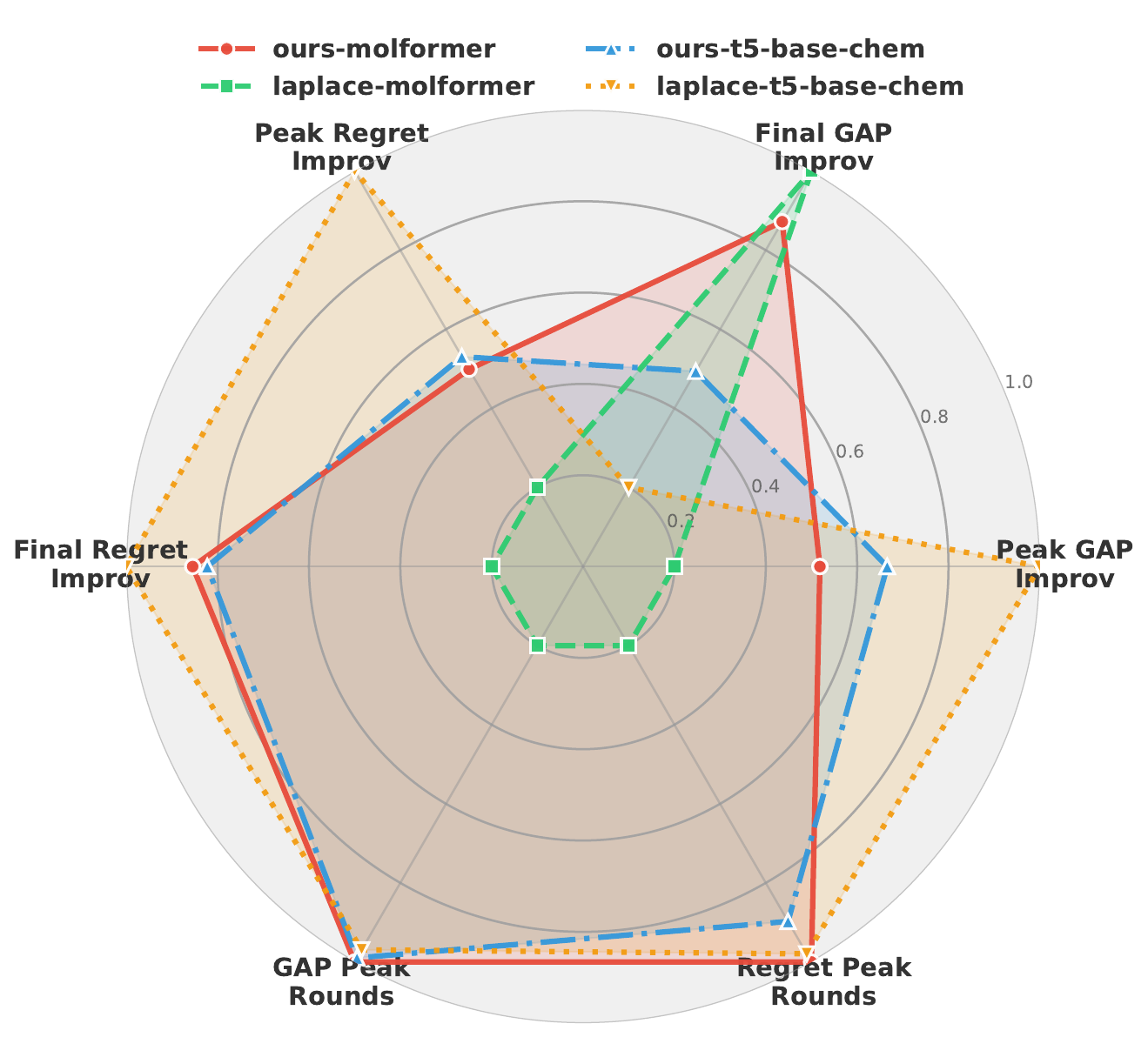}
    \includegraphics[width=0.35\linewidth]{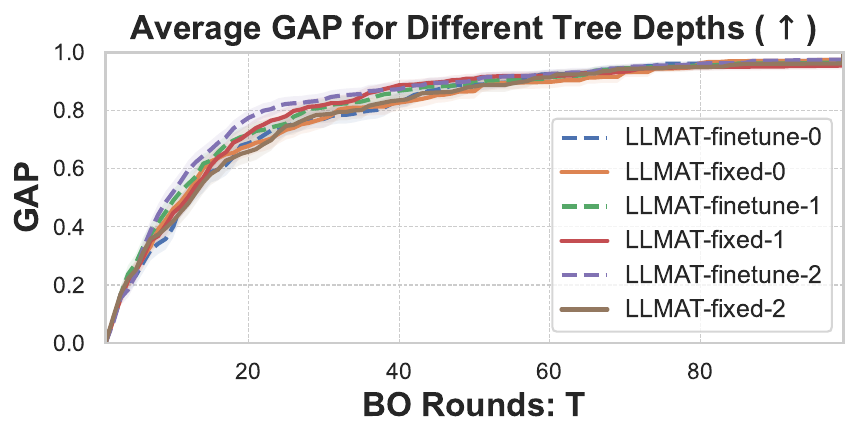}
    \includegraphics[width=0.35\linewidth]{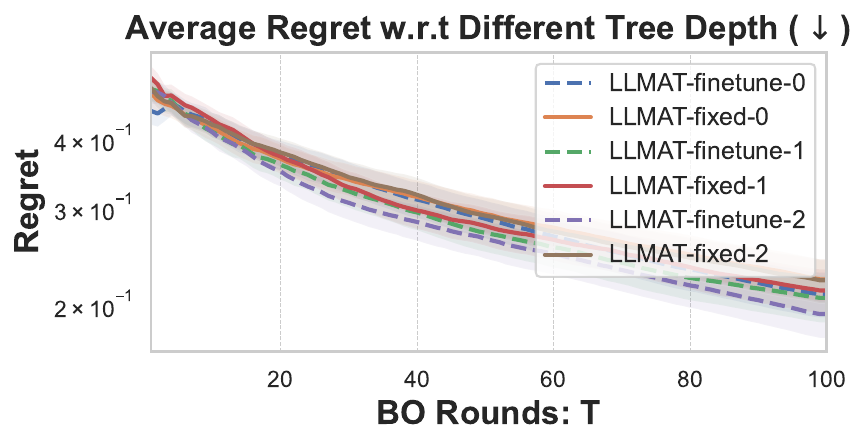}
    \caption{
    \textbf{Left:} Effectiveness of PEFT on T5-Chem and Molformer (radar chart of the improvements of fine-tuning over fixed features; for the peak-round scores, larger means that the maximum improvement is reached earlier, see Appendix~\ref{sec:radar}). 
    \textbf{Middle and right:} Tree-depth ablation on fixed and fine-tuned MolFormer across datasets, showing the importance of partitioning.
    }
    \label{fig:molformer_tree_depth}
    \vspace{-0.6cm}
\end{figure*}


\subsection{Ablation Studies}
To better understand the behavior of LLMAT, evaluate the effectiveness of its individual modules, and assess its robustness to noise, we conduct comprehensive ablation studies. In addition to average performance over all datasets, per-dataset results are provided in the Appendix.

\begin{figure*}[!ht]
    \centering
    \includegraphics[width=0.9\linewidth]{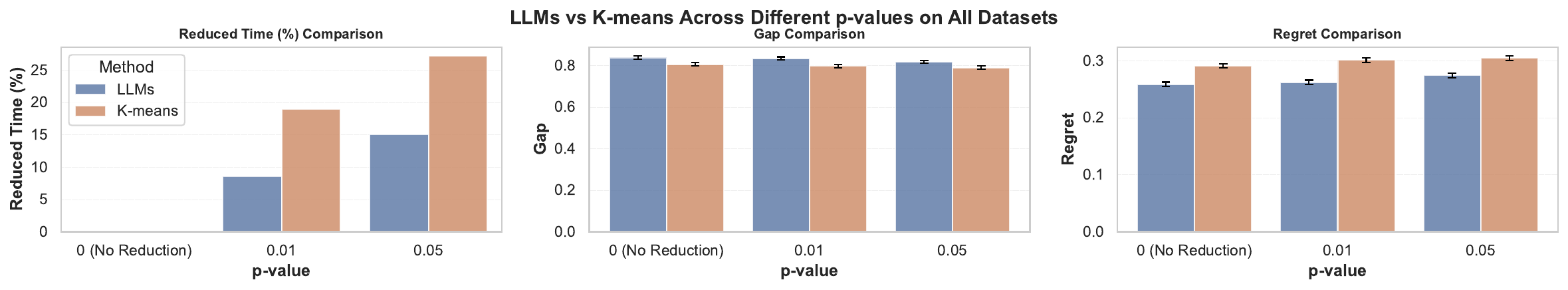}
    \caption{Average reduced prediction time percentage, average GAP, and average regret across six datasets for LLM-based clustering and K-means clustering w.r.t. different p-values. }
    \label{fig:molformer_clustering}
    \vspace{-0.3cm}
\end{figure*}

\textbf{Tree Depth Ablation.~} 
The middle and right plots in Fig.~\ref{fig:molformer_tree_depth} present an ablation on tree depths for LLMAT, which further demonstrates the benefit of refined local AFs. It shows that using only LFBO ($L=0$) performs worse compared to deeper tree depths. With PEFT, performance peaks at depth $2$, demonstrating the benefit of deeper partitions, while with fixed features, performance is highest at depth $1$ in Fig.~\ref{fig:molformer_tree_depth}. In Appendix~\ref{sec:depth-ablation}, we extend the fixed-feature ablation to $L\in\{0,\dots,4\}$ and report Gap AUC: moderate partitioning improves over $L=0$ (depth $3$ is best, $80.08$ vs.\ $78.12$), while deeper trees bring no further gain. 


\textbf{Cluster Selection Ablation.~}  
We compare LLM-based clustering with K-means on various features across different p-values, which determine the removal of clusters with significantly poorer property values. Fig.~\ref{fig:molformer_clustering} shows that LLM-based clustering achieves better average GAP and regret at p-value = 0, highlighting its superior property-awareness for BO initialization. Increasing p-values filters more clusters during AF estimation, reducing prediction time by 10--25\%. While LLM-based clustering removes fewer clusters than K-means at the same p-values, its performance remains stable at 0.01 and degrades only modestly at 0.05. In contrast, K-means is more prone to removing informative clusters, causing larger performance drops. Degradation does not always follow the p-value, as it depends on the property-cluster correlation. Fig.~\ref{fig:molformer_clustering_all} shows LLM clustering removing irrelevant clusters on Redox-mer, reducing computation and improving BO performance simultaneously.

\begin{figure}[!ht]
    \centering
    \includegraphics[width=0.45\linewidth]{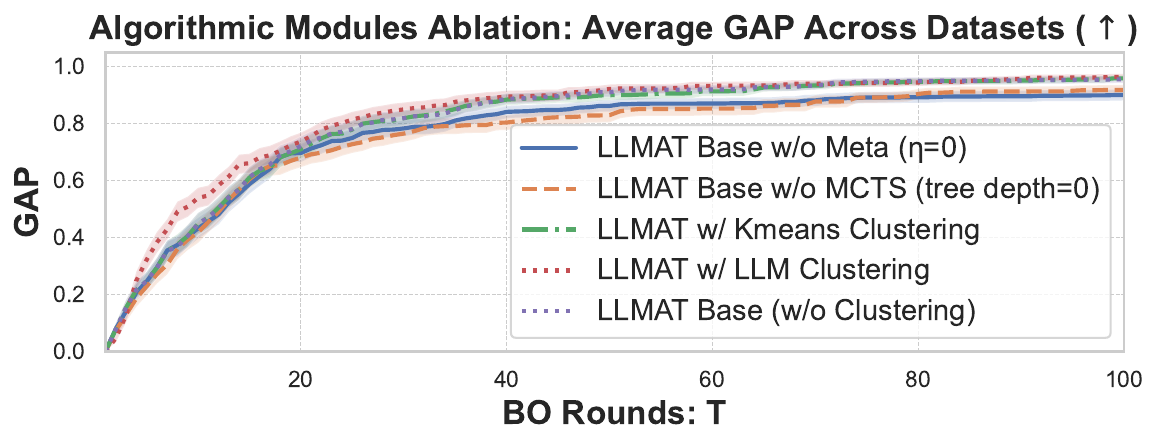}
    \includegraphics[width=0.5\linewidth]{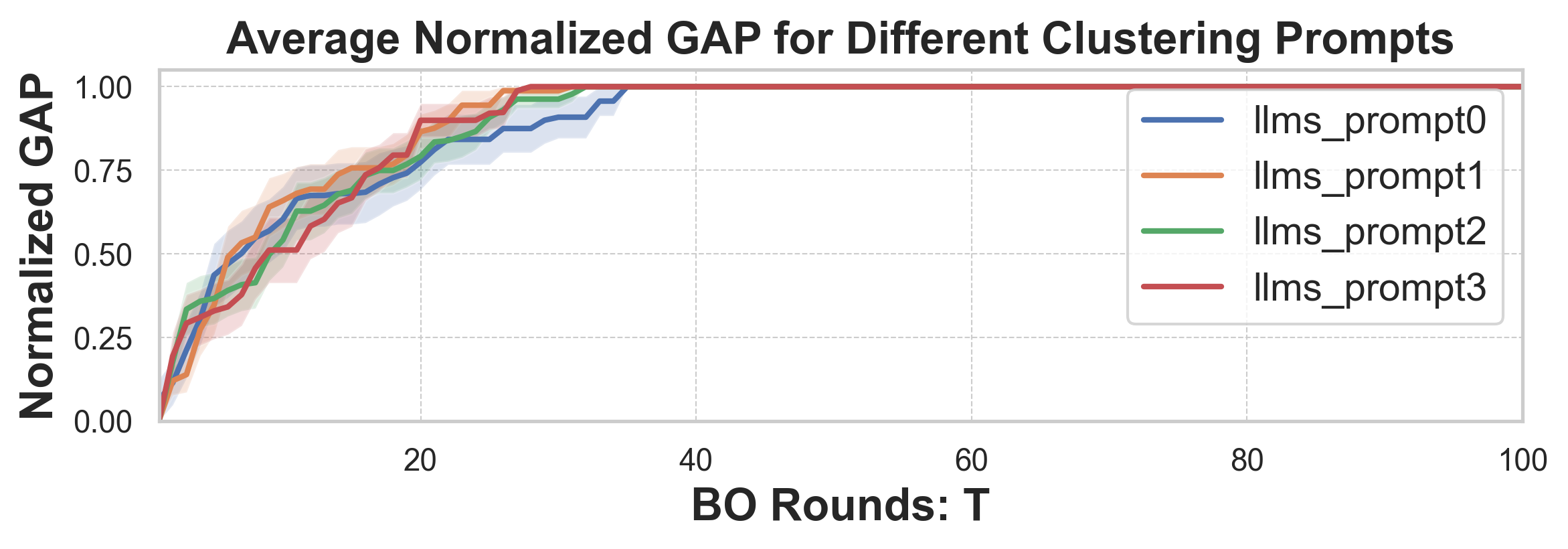}
    \caption{\textbf{Left:} Average GAP across six datasets illustrating the impact of individual algorithmic module ablations. \textbf{Right:} LLM Clustering Prompt Sensitivity in LLMAT on Redox-mer Data.}
    \label{fig:molformer_algo_ablation_remove}
    \vspace{-0.5cm}
\end{figure}

\textbf{Ablation Study for Different Algorithmic Modules.~} To provide a clearer understanding of the contribution of each algorithmic component in LLMAT, we conduct a detailed ablation study by selectively removing or adding modules and evaluating their impact on overall performance, as shown in the left subplot of Fig.~\ref{fig:molformer_algo_ablation_remove}. The results indicate that both MCTS and meta-learning can enhance performance. LLM-based clustering offers a slight improvement, while K-means has a negligible effect. This is expected, as the clustering methods are primarily designed to reduce computational cost rather than boost BO performance.

\textbf{Sensitivity of Clustering prompts.~} To evaluate LLMAT's stability with respect to clustering prompts, we generated four similar prompts and compared their cluster labels agreements and cluster distributions in Fig.~\ref{fig:prompt_ablation} in the Appendix. The right subplots of Fig.~\ref{fig:molformer_algo_ablation_remove} show that the BO performance, measured by the GAP metric, remains largely consistent. This stability can be attributed to the statistical test used during cluster selection, which incorporates observed data to enhance robustness.

Due to space limitations, additional ablation studies on label noise, the quantile $\gamma$ and the extension to multi-objective optimization are provided in Fig.~\ref{fig:Label Noise Ablation}, Fig.~\ref{fig: Ablation of gamma.}, and Fig.~\ref{fig:multi-objective extension} in the Appendix, respectively. 
\section{Conclusion}
In this paper, we introduced LLMAT, a novel surrogate-free BO algorithm that incorporates LLM and foundation model features to inform the design and effectiveness of acquisition functions in BO. Our method also introduces a learned hierarchical partitioning scheme, using Monte Carlo Tree Search with tree nodes defined by shared binary classifiers to learn local acquisition functions that are informed by LLM priors. By incorporating shared parameters and meta-learning across these binary classifiers, LLMAT achieves better AF generalization in low-data regimes and enables efficient exploration of the vast search spaces of molecules. Extensive evaluations on six chemistry datasets demonstrate that LLMAT consistently outperforms baselines on most datasets, highlighting the value of incorporating general LLMs and domain-specific foundation models with principled algorithmic design for molecular discovery. 

\textbf{Limitations.}\label{sec:limitations}
We focus on BO (for post-hoc verification) rather than candidate generation, so a full comparison with evolutionary and generative methods is left for future work. We also consider only 1D/text-based and 2D molecular representations; since LLMAT operates on generic pretrained representations, incorporating 3D molecular foundation models is conceptually straightforward, but evaluating them with localized BO is future work. Finally, we focus on single-objective BO; a preliminary multi-objective extension is in Fig.~\ref{fig:multi-objective extension} (Appendix), while a full treatment with Pareto-front modeling and dedicated acquisition strategies is left for future work.

\begin{ack}
We appreciate the constructive feedback from the anonymous reviewers and meta-reviewers, and thank Jixiang Qing for proofreading before the submission.
The main work was supported by grant number DSI-PDFY3R1P11 from the Data Sciences Institute at the University of Toronto.
The refinement of this work continued after Qi Chen joined the ELLIS Institute Finland and Aalto University.
Florian Shkurti is supported by the Canada First Research Excellence Fund (CFREF) and would also like to acknowledge funding from the AI2050 Fellowship program by Schmidt Sciences.
We thank Compute Canada for computational resources, and acknowledge the computational resources provided by the Aalto Science-IT project through the Triton cluster.
\textbf{Competing interests.} The authors declare no competing interests.
\end{ack}


\bibliographystyle{unsrtnat}
\bibliography{references}

\newpage
\appendix
\addcontentsline{toc}{section}{Appendix} 
\part{Appendix} 
\parttoc 

\newpage
\section{General LLMs Capture Relative Property Ordering, Not Absolute Values}
\label{sec:llm-ranking}
\begin{figure}[H]
\centering
\begin{tcolorbox}[colframe=blue!50!black, colback=white!10, width=0.99\textwidth, valign=center, halign=left]
\centering
\begin{minipage}{0.28\textwidth}
\centering
\begin{tcolorbox}[colframe=RoyalBlue, colback=RoyalBlue!8, width=\textwidth, left=1mm, right=1mm, top=0.5mm, bottom=0.5mm, boxrule=1.5pt]
\centering
\includegraphics[width=1.8cm]{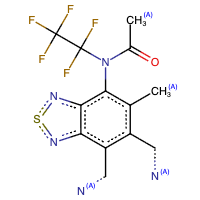}\\[0.03cm]
{\color{RoyalBlue}\small\textbf{Molecule 1}}
\end{tcolorbox}
\end{minipage}%
\hfill%
\begin{minipage}{0.28\textwidth}
\centering
\begin{tcolorbox}[colframe=SeaGreen, colback=SeaGreen!8, width=\textwidth, left=1mm, right=1mm, top=0.5mm, bottom=0.5mm, boxrule=1.5pt]
\centering
\includegraphics[width=1.8cm]{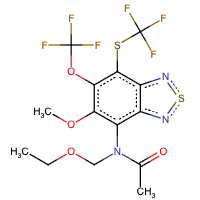}\\[0.03cm]
{\color{SeaGreen}\small\textbf{Molecule 2}}
\end{tcolorbox}
\end{minipage}%
\hfill%
\begin{minipage}{0.28\textwidth}
\centering
\begin{tcolorbox}[colframe=Salmon, colback=Salmon!8, width=\textwidth, left=1mm, right=1mm, top=0.5mm, bottom=0.5mm, boxrule=1.5pt]
\centering
\includegraphics[width=1.8cm]{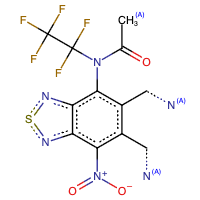}\\[0.03cm]
{\color{Salmon}\small\textbf{Molecule 3}}
\end{tcolorbox}
\end{minipage}
\vspace{0.08cm}
\hrule height 1.5pt\relax
\vspace{0.08cm}
\raggedright
{\small\textcolor{RoyalBlue}{Molecule 1:} CC(=O)N(c1c(C)c(C\#N)c(C\#N)c2nsnc12)C(F)(F)C(F)(F)F\\
\textcolor{SeaGreen}{Molecule 2:} CCOCN(C(C)=O)c1c(OC)c(OC(F)(F)F)c(SC(F)(F)F)c2nsnc12\\
\textcolor{Salmon}{Molecule 3:} CC(=O)N(c1c(C\#N)c(C\#N)c([N+](=O)[O-])c2nsnc12)C(F)(F)C(F)(F)F}\\[0.05cm]
{\small\textbf{Prompt:} Estimate redox potential for \textcolor{RoyalBlue}{\{Molecule 1\}} \textcolor{SeaGreen}{\{Molecule 2\}} and \textcolor{Salmon}{\{Molecule 3\}}. Respond with number and unit.\\
\textbf{DeepSeek:} \textcolor{RoyalBlue}{2.15 V}, \textcolor{SeaGreen}{1.42 V}, \textcolor{Salmon}{2.35 V}. 
\textbf{GPT-4o:} \textcolor{RoyalBlue}{1.42 V}, \textcolor{SeaGreen}{1.10 V}, \textcolor{Salmon}{2.0 V}. \\
\textbf{Ground Truth:} \textcolor{RoyalBlue}{2.80 V}, \textcolor{SeaGreen}{1.54 V}, \textcolor{Salmon}{3.38 V}.}\\[0.02cm]
\noindent\begin{tikzpicture}
\draw[dashed, line width=0.5pt] (0,0) -- (\textwidth,0);
\end{tikzpicture}
\vspace{0.03cm}
{\small\textbf{Prompt:} Group \textcolor{RoyalBlue}{\{Molecule 1\}} \textcolor{SeaGreen}{\{Molecule 2\}} and \textcolor{Salmon}{\{Molecule 3\}} into clusters \{0-4, low-high\} by redox potential. Respond with a number only. \\
\textbf{DeepSeek:} \textcolor{RoyalBlue}{3}, \textcolor{SeaGreen}{1}, \textcolor{Salmon}{4}. 
\textbf{GPT-4o:} \textcolor{RoyalBlue}{2}, \textcolor{SeaGreen}{1}, \textcolor{Salmon}{4}.}
\end{tcolorbox}
\caption{General LLMs provide coarse property ranking of redox potentials rather than accurate numeric value for the three given molecules; responses are color-matched.}
\label{fig:llm_cluster}
\vspace{-0.2cm}
\end{figure}

Fig.~\ref{fig:llm_cluster} illustrates this on three Redox-mer molecules. When asked to estimate the redox potential directly, both DeepSeek and GPT-4o give inaccurate values: they underestimate all three molecules (e.g., DeepSeek predicts $2.15$/$1.42$/$2.35$\,V and GPT-4o $1.42$/$1.10$/$2.0$\,V, versus the ground truth $2.80$/$1.54$/$3.38$\,V). However, both models order the molecules correctly (Molecule 2 $<$ Molecule 1 $<$ Molecule 3), and when asked to group the molecules into clusters from low to high, they again assign them in this same order (DeepSeek: $1$, $3$, $4$; GPT-4o: $1$, $2$, $4$ for Molecules 2, 1, 3). General LLMs therefore cannot be trusted for absolute property values, but can still provide a coarse property ranking, which is the signal that LLM-based clustering uses.

We further run experiments across multiple LLM APIs to assess whether LLMs yield meaningful property rankings, comparing direct clustering (DC), property prediction (PP), and property-derived clustering (PC).

\begin{table}[h]
\centering
\caption{Comparison of LLM APIs across property prediction and clustering metrics.}
\label{tab:llm-ranking}
\resizebox{\textwidth}{!}{%
\begin{tabular}{lccccccc}
\toprule
\textbf{Model} & \textbf{PP $R^2$\,$\uparrow$} & \textbf{DC HunAcc\,$\uparrow$} & \textbf{DC Spearman\,$\uparrow$} & \textbf{DC PairAcc\,$\uparrow$} & \textbf{PC HunAcc\,$\uparrow$} & \textbf{PC Spearman\,$\uparrow$} & \textbf{PC PairAcc\,$\uparrow$} \\
\midrule
GPT-4o          & $-6.0016$  & $0.3141$ & $0.4943$ & $0.7732$          & $0.2459$ & $0.1479$ & $0.5795$ \\
Gemini-2.5-pro  & $-87.9568$ & $0.4471$ & $0.8041$ & $\mathbf{0.9084}$ & $0.2232$ & $0.2569$ & $0.7163$ \\
Qwen3-Plus      & $-0.8572$  & $\mathbf{0.4961}$ & $\mathbf{0.8055}$ & $0.8977$ & $\mathbf{0.3689}$ & $\mathbf{0.6510}$ & $\mathbf{0.7845}$ \\
\bottomrule
\end{tabular}%
}
\end{table}

\paragraph{Evaluation setup.}
We evaluate property prediction using $R^2 = 1 - \frac{\sum_i (y_i - \hat{y}_i)^2}{\sum_i (y_i - \bar{y})^2}$, where $R^2 < 0$ indicates performance worse than a mean predictor. Ground-truth clusters are obtained via \textbf{quantile-based binning into five equal-frequency groups}. DC outputs labels directly, while for PC, predicted values are mapped to clusters using the same bin edges. We assess ranking via Spearman (rank correlation) and \textbf{PairAcc (pairwise ordering accuracy, aligned with BO objectives)}, and clustering via \textbf{Hungarian accuracy (HunAcc)}. Meaningful ranking is defined as consistency with ground-truth ordering under these metrics.

This decomposition isolates where LLM clustering contributes in the pipeline.

\begin{itemize}
    \item While property prediction is unreliable ($R^2 < 0$ for all models), DC achieves strong ranking performance (Spearman up to $0.80$, PairAcc up to $0.91$), \textbf{well above random (${\approx}0$ and $0.5$)}.
    \item Importantly, PC performs substantially worse than DC, indicating that LLMs capture \textbf{relative ordering directly rather than through numerical prediction}.
    \item Overall, these results \textbf{show that} although LLMs fail at absolute prediction, they provide reliable ranking signals sufficient for BO, as evidenced by strong PairAcc and consistent BO performance across LLM APIs (Fig.~\ref{fig:api_ablation}).
\end{itemize}

A related concern is that LLMs may have memorized property values of these public datasets. The negative $R^2$ of property prediction (Table~\ref{tab:llm-ranking}) indicates that the LLMs do not reproduce numerical values, although we cannot rule out that they have seen coarse trends from public data.

\section{Additional Background}

\subsection{Indirect Estimation of Acquisition Functions}\label{sec: indirect}
Based on the definition of the acquisition function in Eq.(\ref{eq:af}), it is natural to first calculate the predictive posterior $p(y|x,\mathbf{D}_t)$, then estimate the AFs given the selected utility functions. In the following, we introduce two typical surrogate models that are non-parametric and parametric, respectively.
\paragraph{Gaussian Process (GP).} A typical choice of a non-parametric surrogate model $\hat{f}(x)$ is a GP with $p(\hat{f}(x)|\mathbf{D}_t) = \Ncal(\mathbf{k}_t^T \mathbf{C}_t^{-1}\mathbf{y}_t, k(x,x) - \mathbf{k}_t^T \mathbf{C}_t^{-1} \mathbf{k}_t)$ being Gaussian given some kernel function $k(\cdot, \cdot)$, where $\mathbf{k}_t[i]=k(x_i,x) \forall i\in [t]$, $\mathbf{K}_t[i,j]=k(x_i, x_j), \forall i,j \in [t]$, and $\mathbf{C}_t = \mathbf{K}_t+\sigma^2 \mathbf{I}_t$. In this case, the predictive posterior is tractable and has the form of $p(y|x, \mathbf{D}_t)= \Ncal(\mathbf{k}_t^T \mathbf{C}_t^{-1}\mathbf{y}_t, k(x,x) + \sigma^2 - \mathbf{k}_t^T \mathbf{C}_t^{-1} \mathbf{k}_t)$.  However, the basic GPs have a $\Ocal(t^3)$ computational complexity for training, even sparse GPs still require $\Ocal(t)$, posing difficulty to scale to a large dataset of observations.

\paragraph{Bayesian Neural Network (BNN).} If we consider parametric surrogate model $\hat{f}(x)= \text{NN}(\theta, x)$ with $\text{NN}(\theta, \cdot)$ being a predefined Neural Network (NN) of parameter $\theta$, then, the predictive posterior of BNN is $p(y|x, \mathbf{D}_t)=\int p(y|x,\theta) p(\theta|\mathbf{D}_t) d\theta$. The posterior $p(\theta|\mathbf{D}_t)$ is often not analytically tractable. Hence, solutions like Laplace Approximation (LA), Variational Inference (VI) and Monte Carlo Sampling are often adopted to approximate the integral.

\subsection{Parameter-Efficient Finetuning}

Fine-tuning large pretrained models on downstream tasks often requires updating all model parameters, which can be computationally expensive and memory-intensive. Parameter-efficient fine-tuning (PEFT) methods aim to reduce this overhead by introducing a small set of trainable parameters while keeping the original pretrained model weights frozen.

Among various PEFT methods, \textit{Low-Rank Adaptation} (LoRA) \cite{hu_lora_2021} has gained popularity due to its efficiency and effectiveness. LoRA hypothesizes that the update to the pretrained weight matrices can be approximated by a low-rank decomposition. Formally, consider a pretrained weight matrix $W_0 \in \mathbb{R}^{d \times k}$ in a neural network (e.g., a linear layer or attention projection). Instead of updating $W_0$ directly, LoRA introduces two trainable matrices $A \in \mathbb{R}^{r \times k}$ and $B \in \mathbb{R}^{d \times r}$, where $r \ll \min(d, k)$ is the rank of the adaptation:
\begin{equation*}
    W = W_0 + \Delta W = W_0 + BA,
\end{equation*}
where $BA$ is a low-rank matrix capturing task-specific updates. During fine-tuning, $W_0$ is kept frozen, and only $A$ and $B$ are optimized. For an input $x \in \mathbb{R}^k$, the layer output becomes
\begin{equation*}
    y = Wx = W_0 x + BA x.
\end{equation*}
To further control the magnitude of updates, a scaling factor $\alpha$ is often introduced:
\begin{equation*}
    W = W_0 + \frac{\alpha}{r} BA,
\end{equation*}
where the factor $\frac{\alpha}{r}$ ensures the update has a similar scale across different rank settings. This formulation significantly reduces the number of trainable parameters from $d \times k$ to $r(d+k)$ while retaining strong adaptation capability. 

\newpage
\subsection{Welch's ANOVA}\label{sec: welch}
  
Suppose we have $k$ clusters, where cluster $i$ has sample size $n_i$, sample mean $\bar{x}_i$, and sample variance $\sigma_i^2$. Define the weights:
\[
w_i = \frac{n_i}{\sigma_i^2}, \quad i = 1, \dots, k,
\]
and the weighted mean:
\[
\bar{x}_w = \frac{\sum_{i=1}^k w_i \bar{x}_i}{\sum_{i=1}^k w_i}.
\]

The Welch F-statistic is then
\[
F = \frac{\sum_{i=1}^k w_i (\bar{x}_i - \bar{x}_w)^2}{(k-1) \left[ 1 + \frac{2(k-2)}{k^2-1} \sum_{i=1}^k \frac{(1 - w_i / \sum_j w_j)^2}{n_i - 1} \right]}.
\]

The approximate degrees of freedom for the denominator is given by the Satterthwaite approximation:
\[
\nu \approx \frac{k^2 - 1}{3 \sum_{i=1}^k \frac{(1 - w_i / \sum_j w_j)^2}{n_i - 1}}.
\]

Finally, $F$ is compared against an $F$-distribution with $(k-1, \nu)$ degrees of freedom to compute the p-value.

\subsection{Games-Howell Post-hoc Test} \label{sec:posthoc}
Suppose we have $k$ clusters, where cluster $i$ has sample size $n_i$, sample mean $\bar{x}_i$, and sample variance $\sigma_i^2$. For a pair of groups $(i,j)$, the test statistic is

\[
t_{ij} = \frac{\bar{x}_i - \bar{x}_j}{\sqrt{\frac{\sigma_i^2}{n_i} + \frac{\sigma_j^2}{n_j}}}.
\]

The degrees of freedom for this comparison are approximated using the Welch--Satterthwaite equation:

\[
\nu_{ij} = \frac{\left(\frac{\sigma_i^2}{n_i} + \frac{\sigma_j^2}{n_j}\right)^2}
                 {\frac{(\sigma_i^2 / n_i)^2}{n_i - 1} + \frac{(\sigma_j^2 / n_j)^2}{n_j - 1}}.
\]

The p-value is computed from the studentized range distribution with $k$ groups and $\nu_{ij}$ degrees of freedom:

\[
p_{ij} = 1 - F_q\Big(\sqrt{2}\,|t_{ij}|;\, k, \nu_{ij}\Big),
\]

where $F_q$ is the cumulative distribution function of the studentized range distribution. This accounts for multiple comparisons across the $k$ groups, so no additional correction is needed. Our implementation uses the Games-Howell test from the \texttt{pingouin} package.

\newpage
\section{Related Works}\label{sec:additional_related}

\subsection{LLMs for Optimization.} 
Recent research has increasingly leveraged pretrained Large Language Models (LLMs) as informative priors in optimization tasks across a variety of domains. For instance, LLMs have been used to improve prompt optimization strategies by conditioning on query-dependent representations \citep{sun_query-dependent_2024, yang_large_2024, guo_connecting_2024}. Other works have integrated LLMs into evolutionary algorithms, using them as generative operators to evolve prompts or candidate solutions \citep{meyerson_language_2024, lehman_evolution_2022, chen_evoprompting_2023}.~In the context of Bayesian Optimization (BO), LLMs have also been employed to guide acquisition functions, either for hyperparameter tuning \citep{liu_large_2024, zhang_using_2023} or for scientific discovery tasks such as molecular optimization \citep{kristiadi_sober_2024, ramos_bayesian_2023}.

\subsection{High-dimensional BO} 
Several approaches have been developed to scale BO to high-dimensional search spaces, including structural assumptions, subspace embedding, variable selection, and local modeling and space partitioning.  \citet{wang_learning_2020, wang_sample-efficient_2021} progressively bifurcate the search space into more promising and less promising subspaces, starting with the full candidate space $\Omega$. K-means clustering is applied to observed data (features $x$ and values $y$) within each space to split the data into ``good'' and ``bad'' clusters based on the average values of the clusters. These labeled clusters are then used to train a Support Vector Machine (SVM) classifier, which produces a non-linear decision boundary to define latent actions for bifurcating the search space. Finally, local surrogate models for BO are trained on observed data in the selected subspace via Upper Confidence Bound (UCB). Based on \citet{wang_learning_2020}, \citet{li_navigating_2025} propose to reweigh the samples in different partitions by their UCB values without constraining sampling to certain partitions. In contrast, we do not separately use SVMs for partitioning and additional models for acquisition function estimation.

We focus on leveraging LLM-derived priors to enhance high-dimensional Bayesian Optimization (BO) for molecular discovery by using LLMs for feature extraction, finetuning and prompting for clustering. Unlike prior work, we introduce a shared neural network classifier that simultaneously partitions the candidate space and estimates acquisition functions at each tree node, effectively amortizing both tasks. To further improve data efficiency, we adopt a meta-learning \citep{chen_generalization_2021, chen_stability-plasticity_2023} approach that amortizes classifier training across nodes, enabling more stable learning in early data-scarce stages. Moreover, while previous methods assume a continuous search space and rely on rejection sampling, they fail in discrete settings where selected regions may contain no valid candidates. We address this limitation via a backtracking strategy. Our method traverses a UCT-guided path through the search tree, estimating local acquisition functions from leaf to parent until valid candidates are found. In the worst case, it returns to the original Likelihood-Free BO (LFBO) approach.

\newpage

\raggedbottom 
\section{Clustering Prompts}\label{sec:promtps}

\subsection{Redoxmer}

\begin{promptbox}{Redoxmer}
You are a chemist, please use the molecules provided, group them into five clusters based on their redox potential.

\begin{clusterscale}
\textbf{Clustering Scale:} extremely low: 0, low: 1, medium: 2, high: 3, extremely high: 4
\end{clusterscale}

\textbf{Analyze the following features for each molecule:}

\begin{itemize}
    \item Number and type of electron-withdrawing groups (e.g., \ce{CF3}, \ce{NO2}, \ce{CN}, halogens)
    \item Number and type of electron-donating groups (e.g., alkyl, methoxy, hydroxyl)
    \item Positioning of substituents on aromatic rings (meta, para, ortho)
    \item Presence of sulfur-containing functional groups (e.g., \ce{SCF3}, \ce{S=O})
    \item Degree of molecular polarity (based on F, O, or N atoms)
\end{itemize}

Use these criteria to evaluate the redox potential qualitatively and cluster the molecules accordingly.

\begin{instruction}
\textbf{Response Format:} Respond strictly with the counter and numerical cluster labels only. Do not include any additional text.
\end{instruction}

\textbf{Now please cluster the following molecules:}

\begin{molecules}
\{molecules to be inserted here\}
\end{molecules}
\end{promptbox}

\subsection{Solvation}

\begin{promptbox}{Solvation}
You are a chemist, please use the molecules provided, group them into five clusters based on predicted solvation energy.

\begin{clusterscale}
\textbf{Clustering Scale:} extremely low: 0, low: 1, medium: 2, high: 3, extremely high: 4
\end{clusterscale}

\textbf{Consider these molecular features:}

\begin{itemize}
    \item Polarity and hydrogen bonding capability
    \item Molecular size and surface area
    \item Number and type of charged/ionizable groups
    \item Hydrophilic/hydrophobic balance
\end{itemize}

Use these criteria to evaluate the solvation energy qualitatively and cluster the molecules accordingly.

\begin{instruction}
\textbf{Response Format:} Respond strictly with the counter and numerical cluster labels only. Do not include any additional text.
\end{instruction}

\textbf{Now please cluster the following molecules:}

\begin{molecules}
\{molecules to be inserted here\}
\end{molecules}
\end{promptbox}

\subsection{Kinase}

\begin{promptbox}{Kinase}
Act as a computational chemist. You have a list of SMILES strings for molecules, and I want to group them into 5 clusters (0 to 4) where cluster 0 has the lowest predicted kinase docking affinity and cluster 4 has the highest.

\begin{clusterscale}
\textbf{Clustering Scale:} extremely low: 0, low: 1, medium: 2, high: 3, extremely high: 4
\end{clusterscale}

\textbf{Prioritize these criteria:}

\begin{itemize}
    \item Presence of kinase-binding motifs (e.g., hinge-binding heterocycles, hydrophobic pockets)
    \item Functional groups (e.g., hydrogen bond donors/acceptors, aromatic rings)
    \item Molecular weight and polarity (smaller/lipophilic molecules often bind kinases better)
    \item Similarity to known kinase inhibitors (e.g., ATP analogs, tyrosine kinase inhibitors)
\end{itemize}

\begin{instruction}
\textbf{Response Format:} Respond strictly with the counter and numerical cluster labels only. Do not include any additional text.
\end{instruction}

\textbf{Now please cluster the following molecules:}

\begin{molecules}
\{molecules to be inserted here\}
\end{molecules}
\end{promptbox}

\subsection{Photovoltaics}

\begin{promptbox}{Photovoltaics}
You are a chemist, please use the molecules provided, group them into five clusters based on predicted photovoltaic conversion efficiency.

\begin{clusterscale}
\textbf{Clustering Scale:} extremely low: 0, low: 1, medium: 2, high: 3, extremely high: 4
\end{clusterscale}

\textbf{Consider these molecular features:}
\begin{itemize}
    \item Electronic structure indicators (conjugation extent, aromatic systems, electron-rich/deficient regions, push-pull molecular design)
    \item Molecular architecture (planarity potential, $\pi$-system connectivity, structural rigidity, molecular size)
    \item Light harvesting features (conjugated backbone length, donor-acceptor patterns, chromophore presence, substituent effects)
\end{itemize}

Use these criteria to evaluate the photovoltaic conversion efficiency qualitatively and cluster the molecules accordingly.

\begin{instruction}
\textbf{Response Format:} Respond strictly with the counter and numerical cluster labels only. Do not include any additional text.
\end{instruction}

\textbf{Now please cluster the following molecules:}

\begin{molecules}
\{molecules to be inserted here\}
\end{molecules}
\end{promptbox}

\subsection{Laser}

\begin{promptbox}{Laser}
Act as a computational chemist with expertise in photophysics. Group these SMILES strings into 5 clusters based on predicted fluorescence oscillator strength (relevant for lasers).

\begin{clusterscale}
\textbf{Clustering Scale:} very low: 0, low: 1, moderate: 2, high: 3, very high: 4
\end{clusterscale}

\textbf{Consider these factors:}
\begin{itemize}
    \item Conjugation length: Longer conjugation increases oscillator strength
    \item Aromaticity: Aromatic systems often have strong $\pi$-$\pi^*$ transitions
    \item Functional groups: Electron-donating/withdrawing groups alter oscillator strength
    \item Molecular rigidity: Rigid molecules tend to have higher oscillator strength
\end{itemize}

\begin{instruction}
\textbf{Response Format:} Respond strictly with the counter and numerical cluster labels only. Do not include any additional text.
\end{instruction}

\textbf{Now please cluster the following molecules:}

\begin{molecules}
\{molecules to be inserted here\}
\end{molecules}
\end{promptbox}

\subsection{Photoswitches}

\begin{promptbox}{Photoswitches}
Act as a computational chemist with expertise in photochemistry. You have a list of SMILES strings for organic molecules, and want to group them into 5 clusters based on their predicted $\pi$-$\pi^*$ transition wavelengths.

\begin{clusterscale}
\textbf{Clustering Scale:}
\begin{enumerate}
    \item Cluster 0: deep UV range (200--300 nm)
    \item Cluster 1: UV range (300--400 nm)
    \item Cluster 2: blue/visible range (400--500 nm)
    \item Cluster 3: green/red/visible range (500--700 nm)
    \item Cluster 4: near-infrared range (700--1000 nm)
\end{enumerate}
\end{clusterscale}

Use your knowledge of molecular structure and photochemistry to analyze the SMILES strings and assign cluster labels. \textbf{Consider the following factors:}

\begin{itemize}
    \item \textbf{Conjugation length:} Longer conjugation typically shifts the $\pi$-$\pi^*$ transition to longer wavelengths
    \item \textbf{Aromaticity:} Aromatic systems often have $\pi$-$\pi^*$ transitions in the UV/visible range
    \item \textbf{Functional groups:} Electron-donating or electron-withdrawing groups can alter the HOMO-LUMO gap
    \item \textbf{Molecular planarity:} Planar molecules tend to have stronger $\pi$-$\pi^*$ transitions
\end{itemize}

\begin{instruction}
\textbf{Response Format:} Respond strictly with the counter and numerical cluster labels only. Do not include any additional text.
\end{instruction}

\textbf{Now please cluster the following molecules:}

\begin{molecules}
\{molecules to be inserted here\}
\end{molecules}
\end{promptbox}

\newpage

\subsection{Variations in Prompt}
We ask ChatGPT to generate 3 alternative prompts (Prompt1--3; Prompt0 is the original) based on our original prompt, the cluster label disagreements, the label distributions, and the generated prompts are listed below.
\begin{figure}[t]
    \centering
    \includegraphics[width=0.45\linewidth]{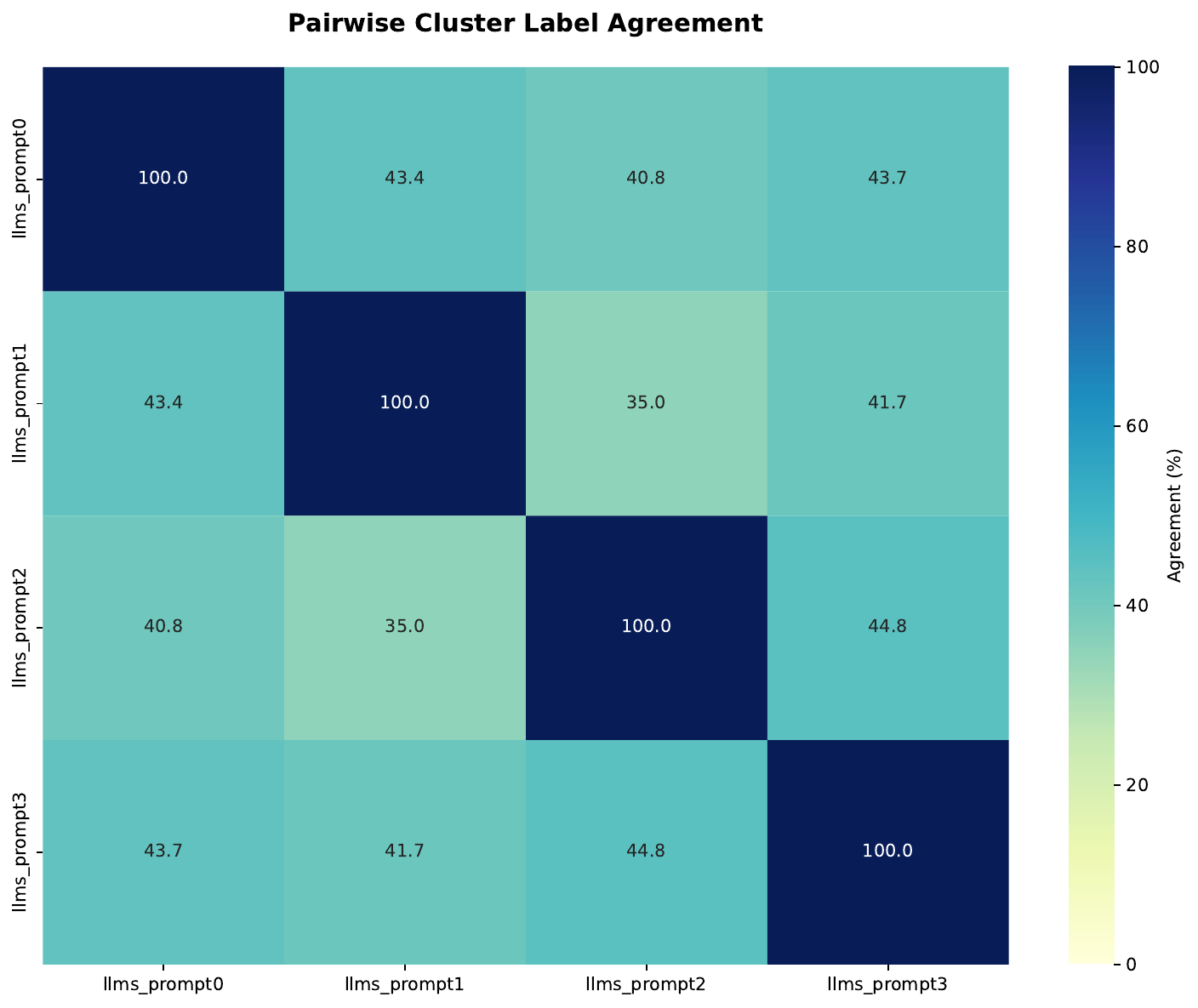}
    \includegraphics[width=0.54\linewidth]{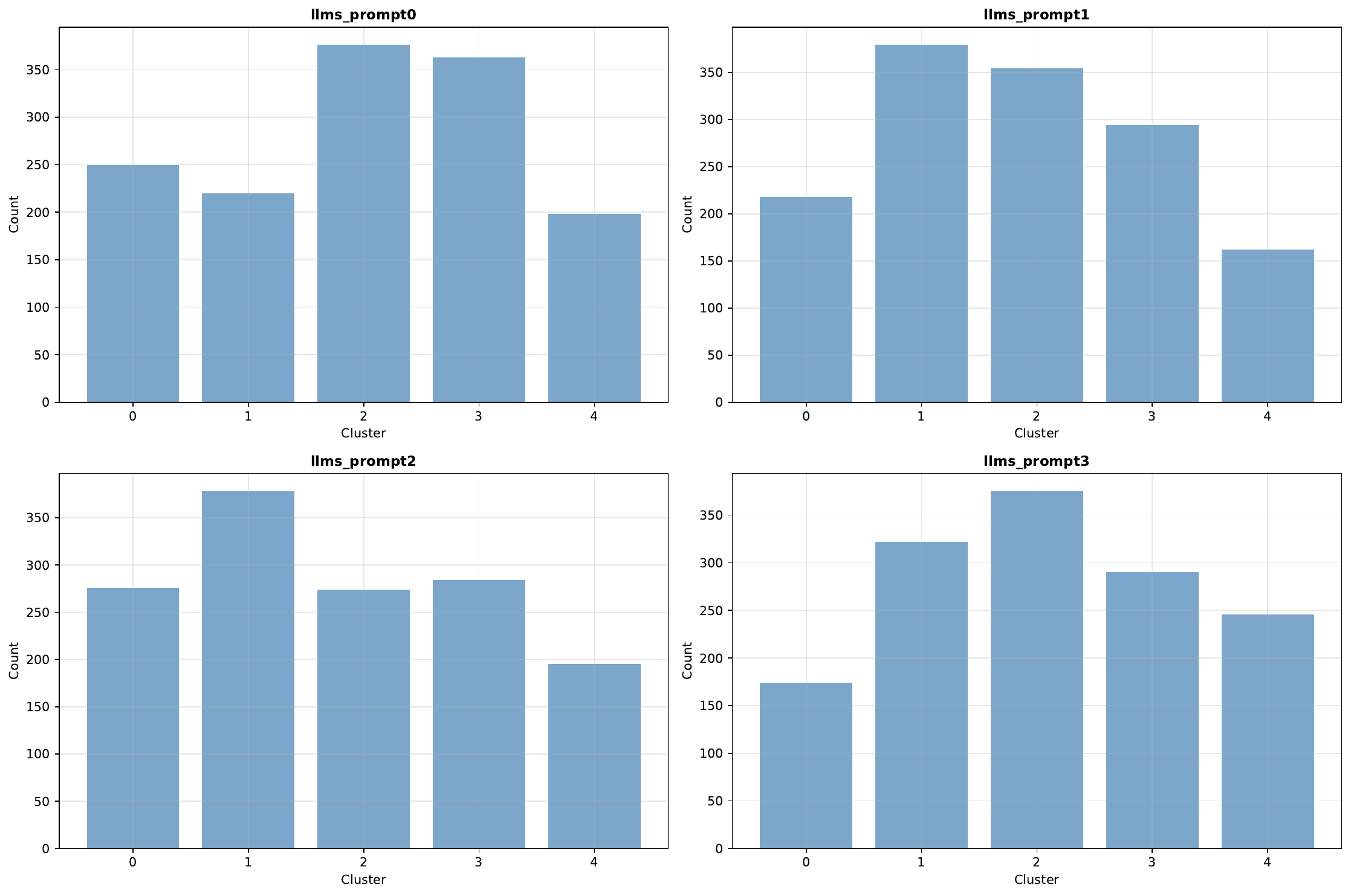}\\
    \includegraphics[width=\linewidth]{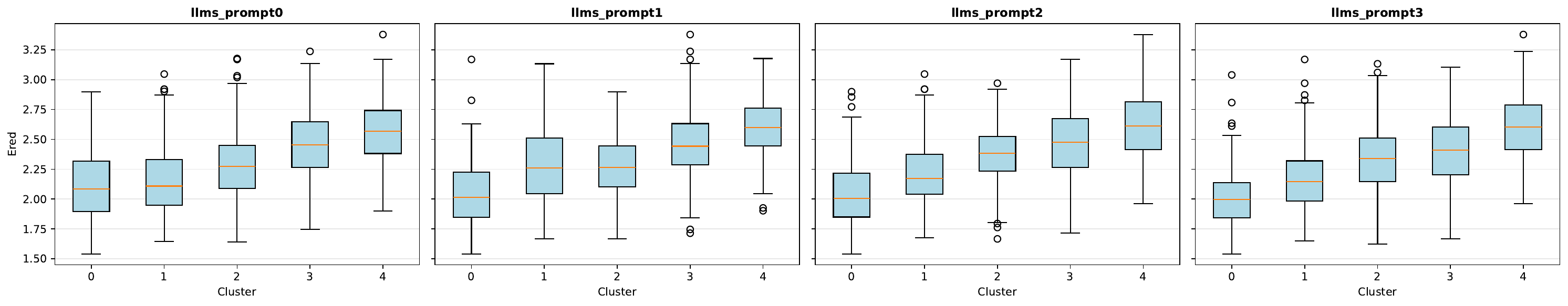}
    \caption{Prompt variation's effect on cluster labels.}
    \label{fig:prompt_ablation}
\end{figure}

\begin{promptbox}{Prompt0}

You are a chemist, please use the molecules provided, group them into five clusters based on their redox potential (extremely low : 0, low: 1, medium: 2, high: 3, extremely high :4). 

\textbf{Analyze the following features for each molecule:} 
 
\begin{itemize}
    \item Number and type of electron-withdrawing groups (e.g., CF$_3$, NO$_2$, CN, halogens).
    \item  Number and type of electron-donating groups (e.g., alkyl, methoxy, hydroxyl).
    \item  Positioning of substituents on aromatic rings (meta, para, ortho).
    \item  Presence of sulfur-containing functional groups (e.g., SCF$_3$, S=O).
    \item Degree of molecular polarity (based on F, O, or N atoms). 
\end{itemize}

Use these criteria to evaluate the redox potential qualitatively and cluster the molecules accordingly.

\begin{instruction}
\textbf{Response Format:} Respond strictly with the counter and numerical cluster labels only. Do not include any additional text.
\end{instruction}

\textbf{Now please cluster the following molecules:}

\begin{molecules}
\{molecules to be inserted here\}
\end{molecules}
\end{promptbox}

\begin{promptbox}{Prompt1}

Act as a chemist. Using the given molecules, assign each one to a redox-potential cluster: 0 = extremely low, 1 = low, 2 = medium, 3 = high, 4 = extremely high.

\textbf{Evaluate each molecule based on:} 
    \begin{itemize}
    \item Electron-withdrawing substituents (CF$_3$, NO$_2$, CN, halogens, etc.)
    \item Electron-donating substituents (alkyl, OMe, OH, etc.)
    \item Substituent positions on aromatic systems (ortho/meta/para)
    \item Sulfur-containing functional groups (SCF$_3$, sulfoxides, etc.)
    \item Overall polarity from heteroatoms (F/O/N) 
\end{itemize}

Use these qualitative features to estimate redox strength and cluster the molecules.

\begin{instruction}
\textbf{Response Format:} Respond strictly with the counter and numerical cluster labels only. Do not include any additional text.
\end{instruction}

\textbf{Now please cluster the following molecules:}

\begin{molecules}
\{molecules to be inserted here\}
\end{molecules}
\end{promptbox}



\begin{promptbox}{Prompt2}
You are an expert computational chemist. For each provided molecule, perform the following steps: Identify electron-withdrawing and electron-donating groups. Determine substituent positions on aromatic cores. Note any sulfur-based functionalities (e.g., SCF$_3$, S=O). Assess molecular polarity based on heteroatom composition. Infer the molecule's qualitative redox potential from these structural features. Then group all molecules into five redox-potential clusters (0--4) from extremely low to extremely high.
\begin{instruction}
\textbf{Response Format:} Respond strictly with the counter and numerical cluster labels only. Do not include any additional text.
\end{instruction}

\textbf{Now please cluster the following molecules:}

\begin{molecules}
\{molecules to be inserted here\}
\end{molecules}
\end{promptbox}

\begin{promptbox}{Prompt3}
As a chemist specializing in structure--property relationships, analyze each molecule and assess its qualitative redox potential. Consider:
    \begin{itemize}
        \item Strength and number of EWG vs EDG
        \item Aromatic substitution pattern and resonance effects
        \item Sulfur-functional group contributions
        \item Polarity arising from heteroatoms
    \end{itemize} 
Based on the identified features, explain your reasoning and categorize each molecule into one of five redox classes (0: extremely low $\rightarrow$ 4: extremely high).
\begin{instruction}
\textbf{Response Format:} Respond strictly with the counter and numerical cluster labels only. Do not include any additional text.
\end{instruction}

\textbf{Now please cluster the following molecules:}

\begin{molecules}
\{molecules to be inserted here\}
\end{molecules}
\end{promptbox}

\newpage
\section{Experimental Details}\label{sec:exp_detail}

The code for LLMAT and all experiments is available at \url{https://github.com/livreQ/LLMAT}.

We follow most of the settings in \citet{kristiadi_sober_2024} for using the foundation models and baselines: GP and Laplace, while we refactored their code (repo \url{https://github.com/wiseodd/lapeft-bayesopt}) for better extension. The LFBO and BORE baselines are LLMAT with only the root classifier (no tree), using the same MLP and $\eta=0.005$; LFBO uses EI weighting and BORE uses PI weighting. All methods run for 100 BO rounds.

\paragraph{Radar Chart Metrics.}\label{sec:radar} The radar chart (Fig.~\ref{fig:molformer_tree_depth}, left) summarizes the benefit of fine-tuning the representation relative to a fixed one, considering both the magnitude of the improvement and how early it is achieved. At BO round $t$, we define $\Delta\mathrm{GAP}_t=\mathrm{GAP}^{\mathrm{FT}}_t-\mathrm{GAP}^{\mathrm{Fixed}}_t$ and $\Delta\mathrm{Regret}_t=\mathrm{Regret}^{\mathrm{Fixed}}_t-\mathrm{Regret}^{\mathrm{FT}}_t$, such that a positive value indicates an advantage of fine-tuning, and denote their means across runs by $\overline{\Delta\mathrm{GAP}}_t$ and $\overline{\Delta\mathrm{Regret}}_t$. The six axes are: \emph{Peak GAP Improv.}, $\max_t\overline{\Delta\mathrm{GAP}}_t$; \emph{Final GAP Improv.}, the mean of $\overline{\Delta\mathrm{GAP}}_t$ over the final 10 BO rounds; \emph{Peak Regret Improv.}, $\max_t\overline{\Delta\mathrm{Regret}}_t$; \emph{Final Regret Improv.}, the mean of $\overline{\Delta\mathrm{Regret}}_t$ over the final 10 BO rounds; \emph{GAP Peak Rounds}, a score of how early the maximum GAP improvement is achieved; and \emph{Regret Peak Rounds}, the analogous score for regret. For the last two, we first find the round of the maximal mean improvement, e.g., $t_{\mathrm{peak}}=\arg\max_t\overline{\Delta\mathrm{GAP}}_t$. Since reaching the maximum improvement earlier indicates better sample efficiency, we transform the raw peak round into a reversed 1--10 score $s_{\mathrm{peak}}=\max\big(1,\min(10,\,11-\tfrac{t_{\mathrm{peak}}}{10})\big)$, so that peaks at rounds 10, 50 and 100 receive scores of 10, 6 and 1, respectively. Hence ``larger is better'' refers to this transformed score, not to the raw number of BO rounds. For visualization, all axes are further normalized to $[0.2,1.0]$; the lower bound of $0.2$ only keeps the smallest values visible.

\paragraph{Computational Resource} The experiments were conducted on multiple server nodes, each equipped with 6 CPUs and a single GPU with 32 GB of memory. To obtain results faster, the experiments were run across different clusters. We ensured that, for each dataset, all algorithms were tested on the same type of CPU and GPU. LLM-based clustering with closed-source APIs costs about \$2.59 (GPT-4o), \$1.81 (Gemini-2.5-Pro), or \$0.47 (Qwen3-Plus) for one clustering pass over all six datasets (Sec.~\ref{sec:llm-cost}).

\subsection{Foundation Models}

\paragraph{Features and Prompts for Foundation Model.} 
For the LLM features, we average the last transformer embeddings along the sequence dimension, ignoring padding and EOS tokens. All LLM-related components in this work were implemented using the Hugging Face Transformers library \cite{wolf_huggingfaces_2019}. We use the single SMILES string as the prompt input for feature extraction and parameter-efficient finetuning of these foundation models.

\subsection{BO on Fixed-features}

We use the same batch size 256 for AFs estimation (\ie, prediction) for all the algorithms. 

\subsubsection{Training Details}
\paragraph{GP} For GP, we use BoTorch \citep{balandat_botorch_2020}, with the Tanimoto kernel from Gauche \citep{griffiths_gauche_2023} for Morgan fingerprints and a Mat\'ern kernel for the other features.  The marginal likelihood is optimized using Adam \citep{kingma_adam_2014} with a learning rate of 0.01 for 500 epochs.

\paragraph{Laplace} A 2-hidden-layer multilayer perceptron with 50 units per layer is used. The network is optimized with Adam at a learning rate of $1\times10^{-3}$ and weight decay $5\times10^{-4}$ for 500 epochs with batch size 20, using cosine annealing for the learning rate \citep{loshchilov_sgdr_2016}. The Laplace approximation is applied post hoc, with prior precision tuned via marginal likelihood for 100 iterations. The Hessian is approximated using a Kronecker structure \citep{ritter_scalable_2018} (a diagonal approximation is used for Llama-2-7b).

\paragraph{LLMAT} We build the MCTS trees that satisfy the tree depth and minimal leaf sample size constraints. A node is split only if it holds more than \texttt{leaf\_sample\_size}$=2$ observations, so nodes deeper in the tree can have very few training points for their classifiers; Appendix~\ref{sec:meta-dynamics} examines training at such deeper nodes. Then we train classifiers that are multilayer perceptrons with one hidden layer of 50 units using ReLU activation. The classifiers are trained with Adam of learning rate $1\times10^{-2}$ and weight decay $5\times10^{-4}$ for 50 epochs with batch size $256$. Other hyperparameters like the quantile $\gamma$, the $\lambda$ for UCT, the meta-learning rate $\eta$, the threshold for partitioning, the tree depth, and the p-value are summarized in Tab.~\ref{tab:ours_fix_hyperparameters}.

\begin{table}[H]
\centering
\caption{Hyperparameter configuration of LLMAT with fixed features. Where several values are listed, they are used per feature model: T5-Chem $\eta{=}0.01$, $p{=}0$; MolFormer/fingerprints/GIN $\eta{=}0.005$, $p{=}0$; T5/GPT-2/Llama $\eta{=}0.005$, $p{=}0.01$. \texttt{tree\_depth}=3 in the code corresponds to $L{=}1$ in the paper (3 nodes).}
\label{tab:ours_fix_hyperparameters}
\begin{tabular}{>{\bfseries}l p{4cm} p{4cm}}
\toprule
\rowcolor{gray!20}
\textbf{Category} & \textbf{Parameter} & \textbf{Value} \\
\midrule

\multirow{6}{*}{\rotatebox{90}{\textbf{General}}} 
& $\gamma$ & 0.5 \\
& $\lambda$ & 0.5 \\
& $\eta$ & 0.005, 0.01 \\
& tree\_depth & 3 \\
& p\_val & 0, 0.01, 0.05 \\
& threshold & 0.5 \\
\midrule
\rowcolor{gray!10}
\multirow{6}{*}{\rotatebox{90}{\textbf{Fix Args}}} 
& batch\_size & 256 \\
\cmidrule{2-3}
& \textit{Head Parameters:} & \\
& \quad n\_epochs & 50 \\
& \quad learning rate & 1e-2 \\
& \quad batch\_size & 256 \\
& \quad leaf\_sample\_size & 2 \\
\bottomrule
\end{tabular}
\end{table}

\subsection{BO with PEFT}
We keep the same LoRA configuration for our algorithm and the Laplace baseline. The batch size for acquisition function estimation was set to $16$ for T5-Chem and $32$ for Molformer. Same batch size was also used for LoRA training.  

\paragraph{LoRA Configuration.} 
We used LoRA with rank 4, applied without bias on the query and value attention weights. The scaling factor $\alpha$ was set to 16, and dropout with probability 0.1 was applied. We followed the implementation from HuggingFace's PEFT library \citep{mangrulkar_peft_2022}.




\subsubsection{Training details of Laplace} 
In \citet{kristiadi_sober_2024}, the following setting was applied to PEFT with Laplace Approximation.
\paragraph{LoRA Training.} 
The LoRA parameters and the regression head were jointly trained using AdamW with learning rates of $3\times10^{-4}$ and $1\times10^{-3}$ for the LoRA and regression head weights, respectively (except for the Photoswitch dataset, where they used $3\times10^{-3}$ and $1\times10^{-2}$). Training was performed for 50 epochs with weight decay 0.01. Subsequently, the regression head was optimized for 100 epochs under the same hyperparameters.

\paragraph{Laplace Approximation.} 
The Laplace approximation was applied to both the LoRA and regression head weights. We used a Kronecker-factored Hessian and optimized the layerwise prior precisions with post hoc marginal likelihood for 200 iterations, following \citet{daxberger_laplace_2021}.

\subsubsection{Training details of our algorithm}

Following the fixed-feature setting, we construct MCTS trees subject to tree-depth and minimum leaf-sample constraints, and train MLP classifiers. To enable larger classifier batch sizes, batching is applied after the feature extractor and LoRA layers. Classifiers are trained with Adam (lr $1\times10^{-2}$, weight decay $5\times10^{-4}$, batch size 256, 50 epochs). LoRA learning rates are task-specific: $3\times10^{-5}$ (Solvation, Kinase), $5\times10^{-5}$ (Redoxmer, Laser), and $1\times10^{-7}$ (Photovoltaics, Photoswitch). Additional hyperparameters, including $\gamma$, $\lambda$, $\eta$, partitioning threshold, tree depth, and $p$-value, are listed in Tabs.~\ref{tab:ours_fix_hyperparameters} and~\ref{tab:peft_hyperparameters}.

\begin{table}[H]
\centering
\caption{Hyperparameter configuration of LLMAT with PEFT.}
\label{tab:peft_hyperparameters}
\begin{tabular}{>{\bfseries}l p{4cm} p{4cm}}
\toprule
\rowcolor{gray!20}
\textbf{Category} & \textbf{Parameter} & \textbf{Value} \\
\midrule


\multirow{10}{*}{\rotatebox{90}{\textbf{PEFT}}} 
& batch\_size & 16, 32 \\
\cmidrule{2-3}
& \textit{Head Parameters:} & \\
& \quad n\_epochs & 50 \\
& \quad learning rate & 1e-2 \\
& \quad batch\_size & 256 \\
& \quad leaf\_sample\_size & 2 \\
& \textit{LoRA Parameters:} & \\
& \quad n\_epochs & 50 \\
& \quad learning rate & 3e-5, 5e-5, 1e-7 \\
& \quad batch\_size & 16, 32 \\
\bottomrule
\end{tabular}
\end{table}

\newpage
\section{Additional Experimental Results}\label{sec:add_results}

\paragraph{Gap AUC.}\label{sec:gap-auc}
To summarize a whole trajectory in one number, the tables in this appendix report the \textbf{Gap AUC}, the area under the GAP curve (Sec.~\ref{sec:metric}) over all BO rounds: \(
\text{Gap AUC} := \sum_{t=1}^{T-1}\tfrac{1}{2}\big(\text{GAP}_t+\text{GAP}_{t+1}\big),
\)
computed per run with $T=100$ rounds (trapezoidal rule), where $y_0$ is the best value of the initial design and GAP$_t$ is clipped to $[0,1]$, and then averaged over runs. It lies in $[0,T-1]$, and higher is better.

In this section, we present additional experimental results on best observed values, GAP metrics, regret, and computational costs for each dataset, covering all fixed feature models, finetuned T5-Chem and MolFormer, clustering and p-value ablations, toy data, the Llama-2-7b results, the multi-objective extension, the LLM-API ablation, and further analyses of representations, tree depth, and cost.
\subsection{Fixed-feature results for more foundation models}
\subsubsection{Historical optimums}

\begin{figure}[H]
    \centering
    \includegraphics[width=\linewidth]{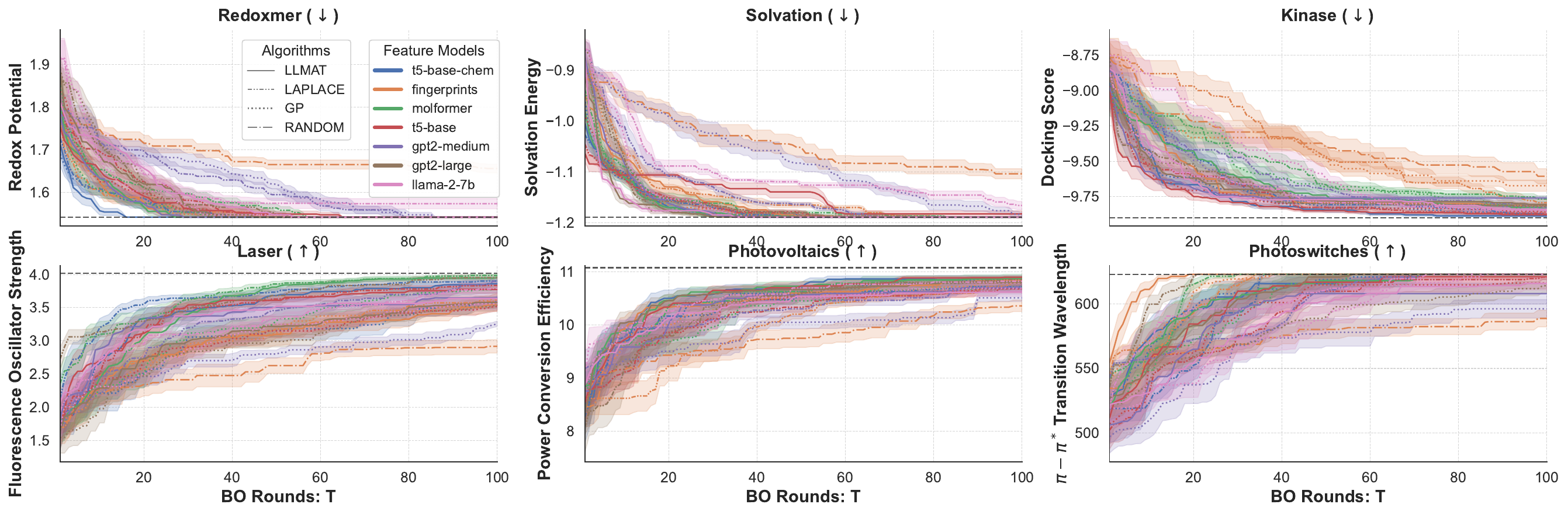}
    \caption{BO process for different datasets on features extracted from various foundation models. }
    \label{fig:fix-best-y-all}
\end{figure}

\subsubsection{Computation Time}

\begin{figure}[H]
    \centering
    \includegraphics[width=0.32\linewidth]{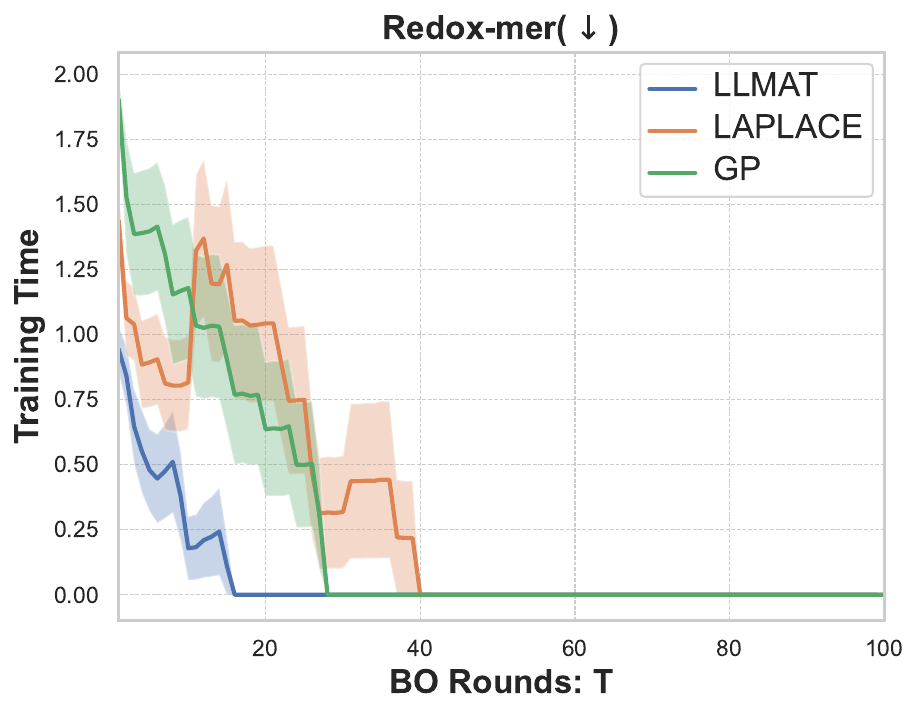}
        \includegraphics[width=0.32\linewidth]{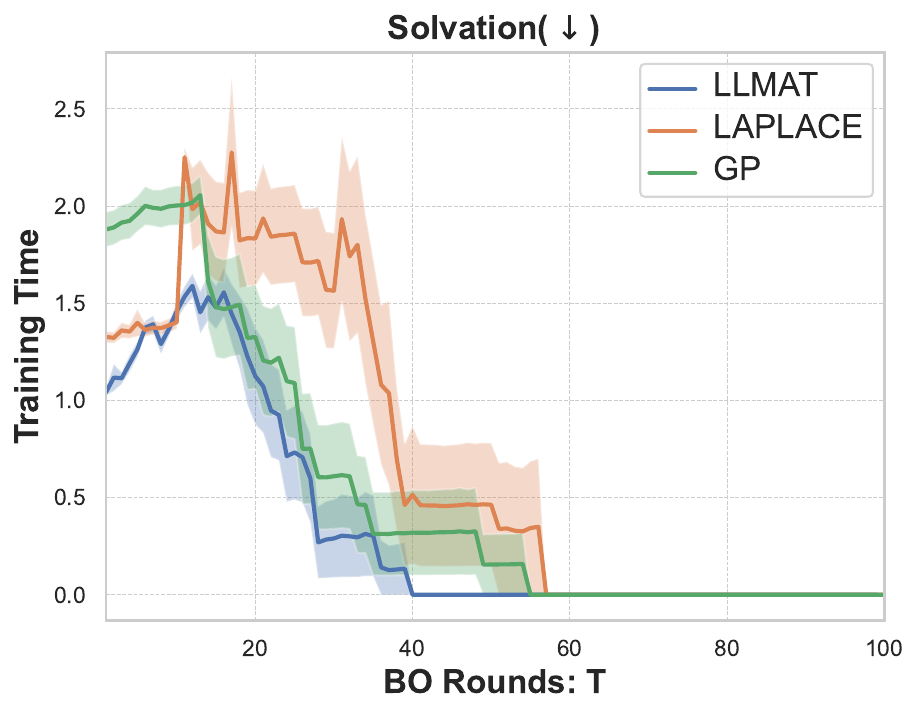}
        \includegraphics[width=0.32\linewidth]{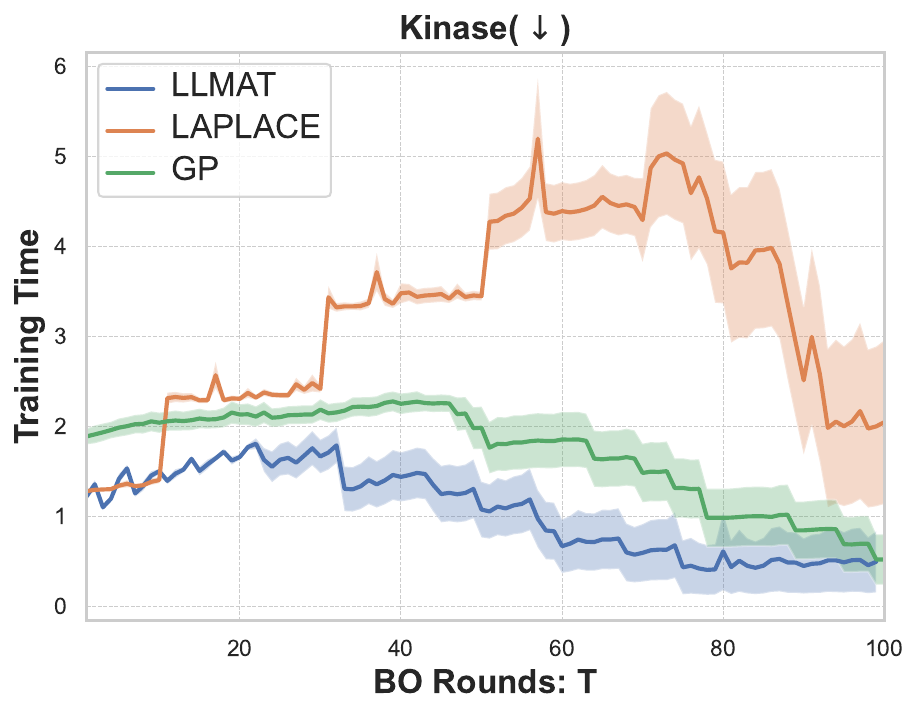}\\
         \includegraphics[width=0.33\linewidth]{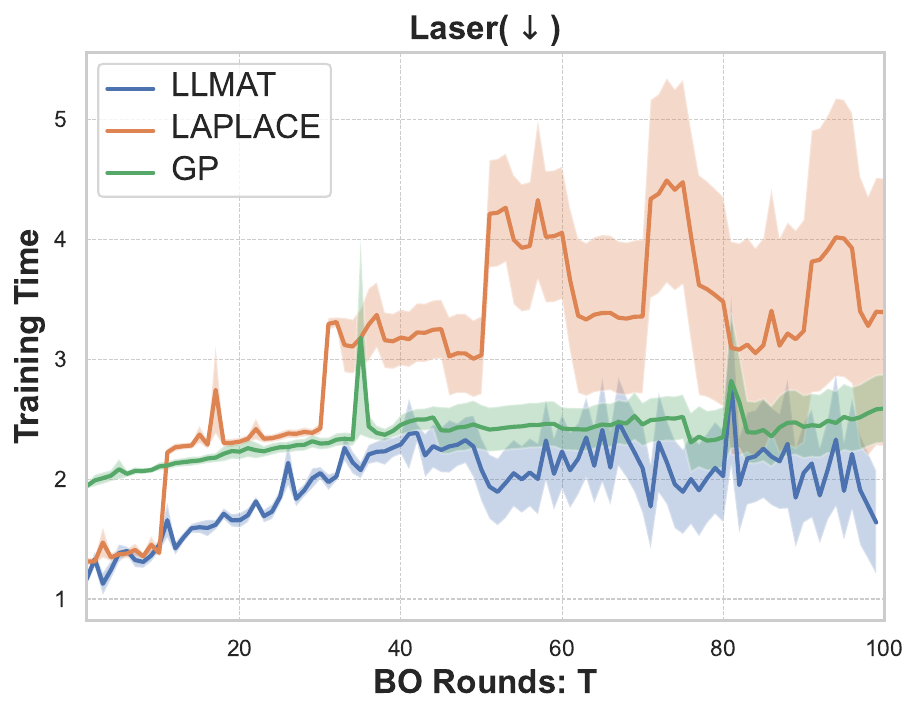}
      \includegraphics[width=0.32\linewidth]{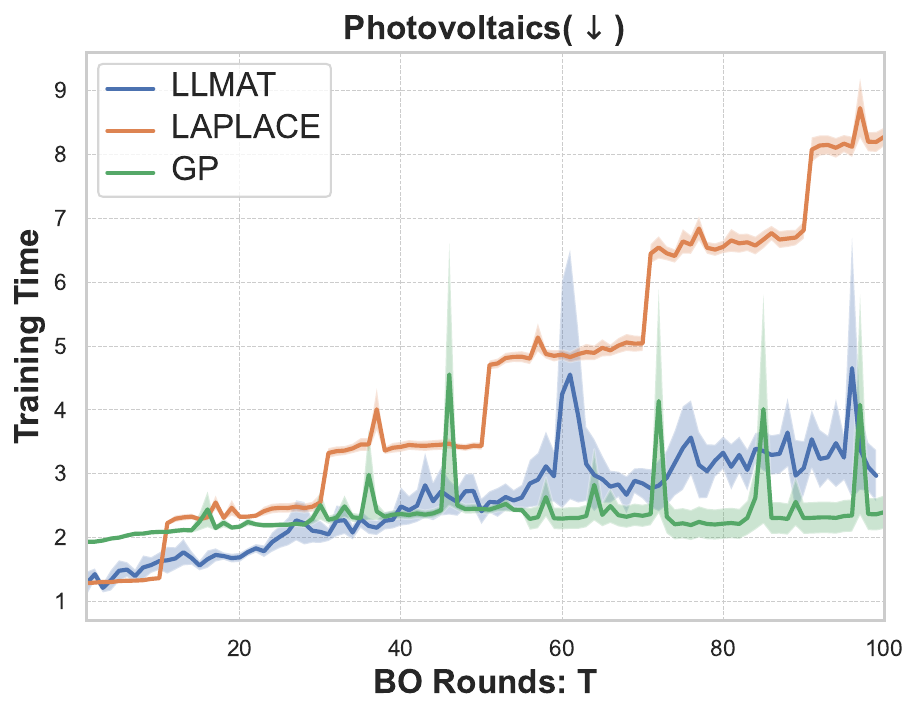}
        \includegraphics[width=0.32\linewidth]{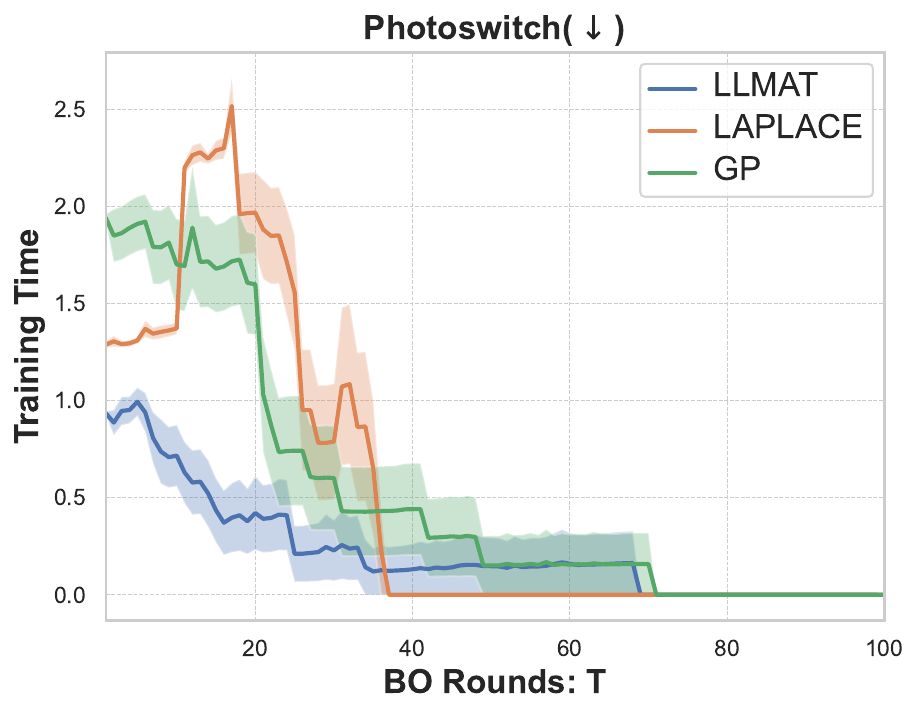}
    \caption{Training time comparison on different chemistry datasets with fixed T5-Chem features. }
    \label{fig:t5-chem-fixed-train-time}
\end{figure}


\begin{figure}[H]
    \centering
    \includegraphics[width=0.32\linewidth]{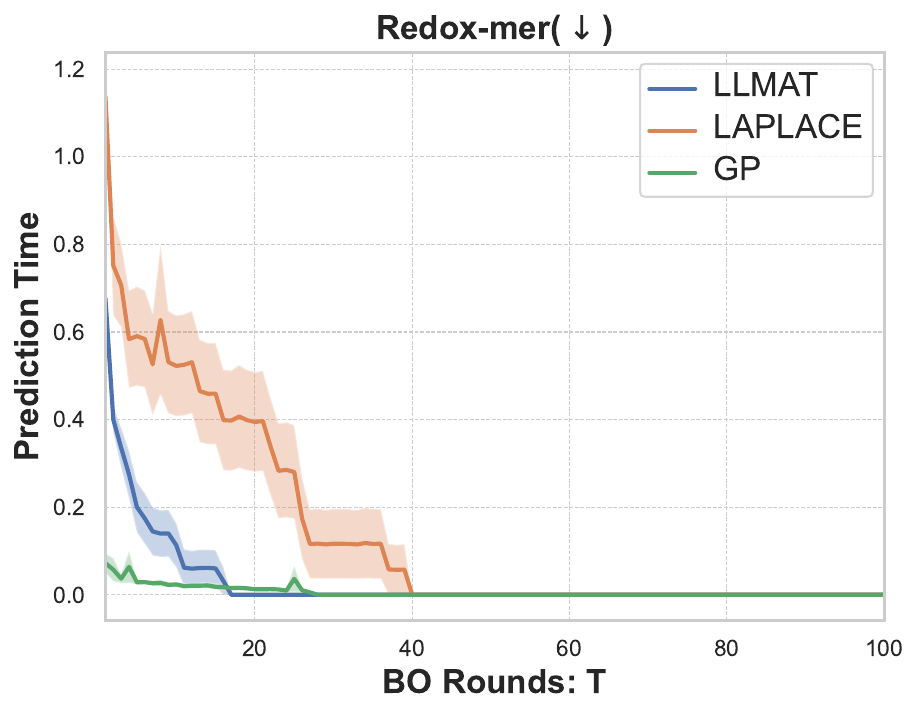}
        \includegraphics[width=0.32\linewidth]{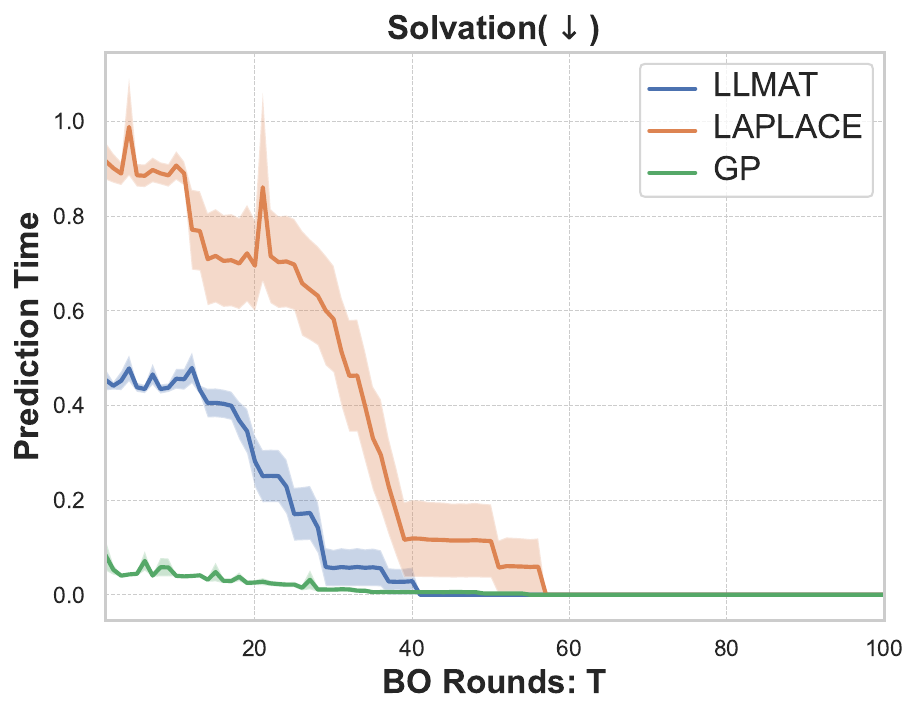}
        \includegraphics[width=0.32\linewidth]{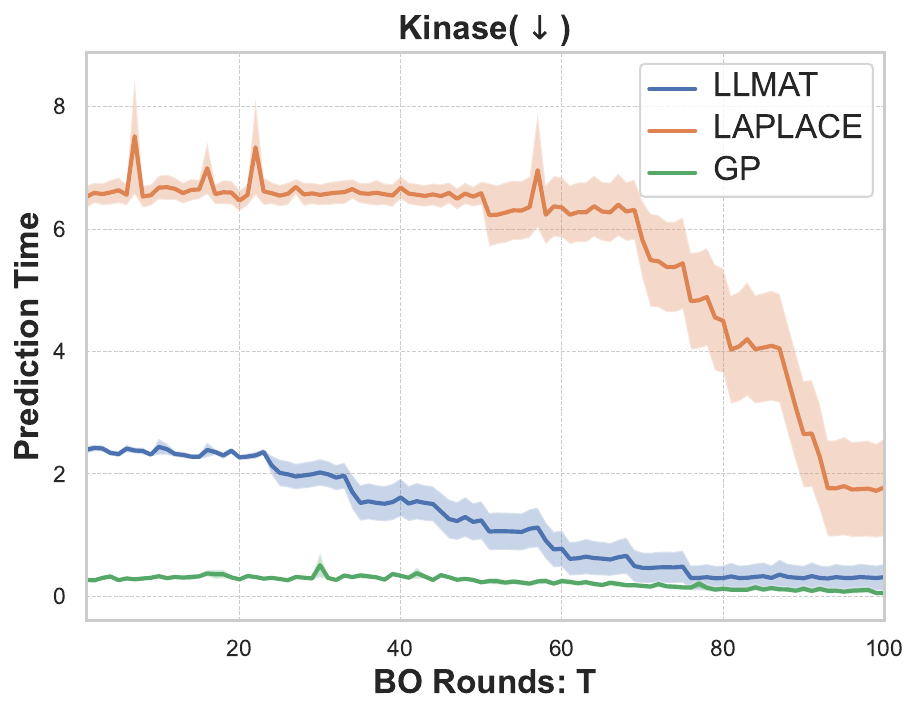}\\
         \includegraphics[width=0.32\linewidth]{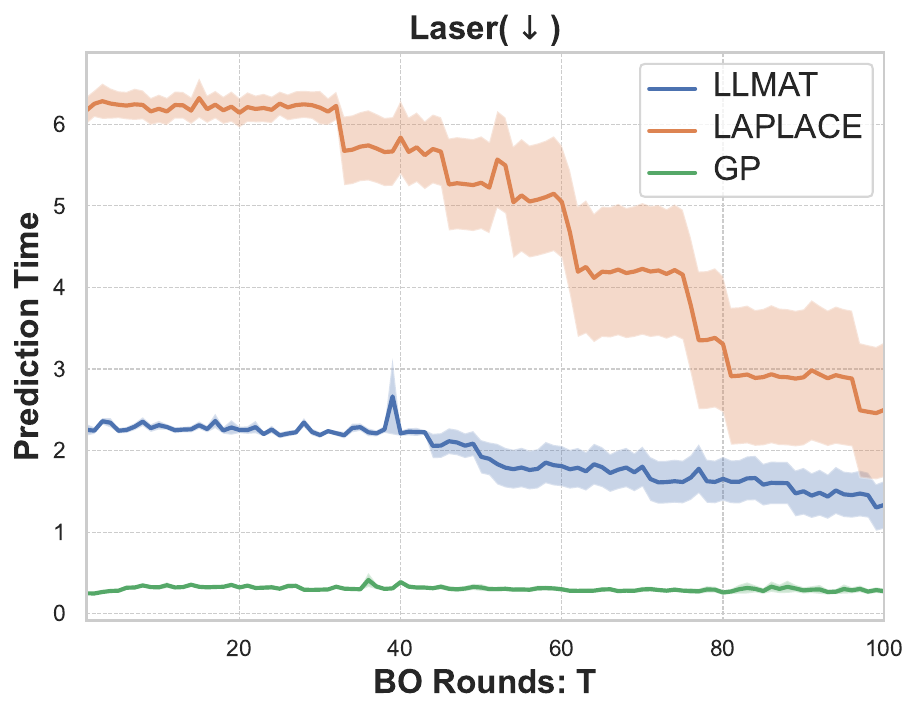}
      \includegraphics[width=0.32\linewidth]{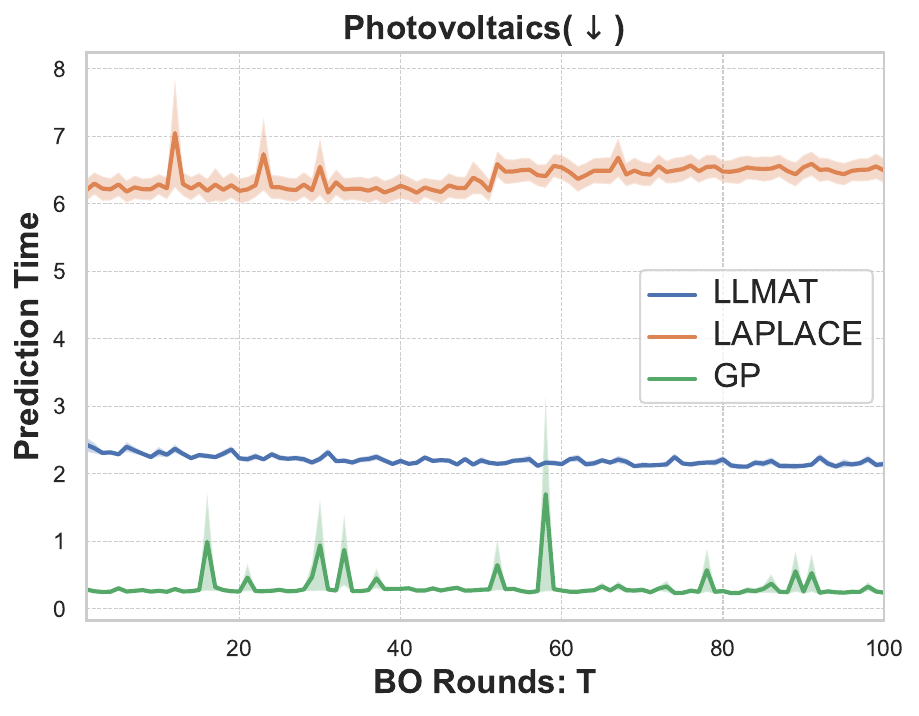}
        \includegraphics[width=0.32\linewidth]{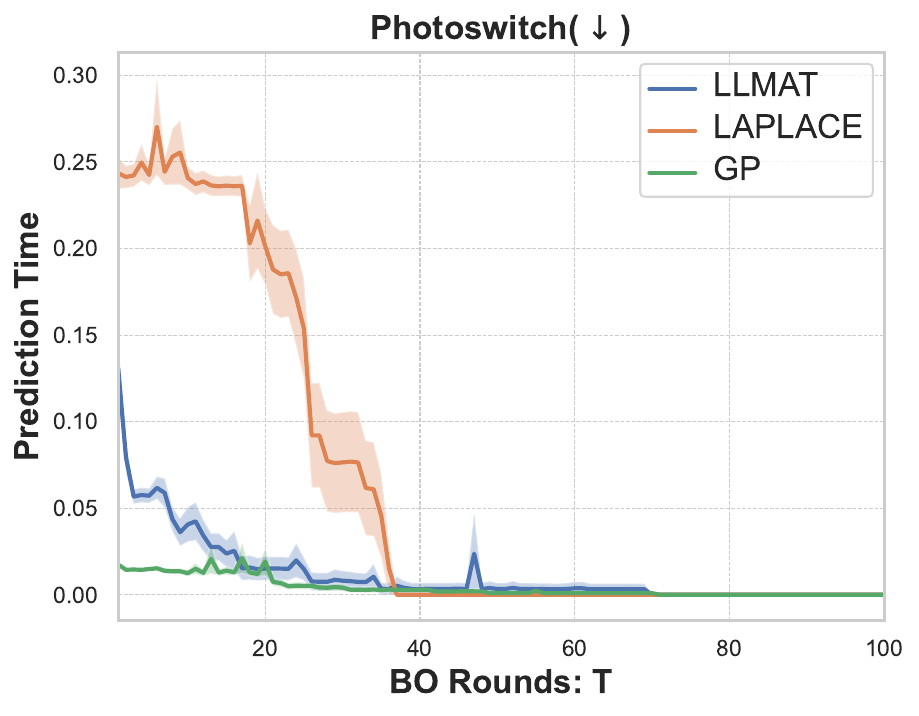}
    \caption{Prediction time comparison on different chemistry datasets with fixed T5-Chem features. }
    \label{fig:t5-chem-fixed-pred-time}
\end{figure}
\begin{figure}[H]
    \centering
    \includegraphics[width=0.32\linewidth]{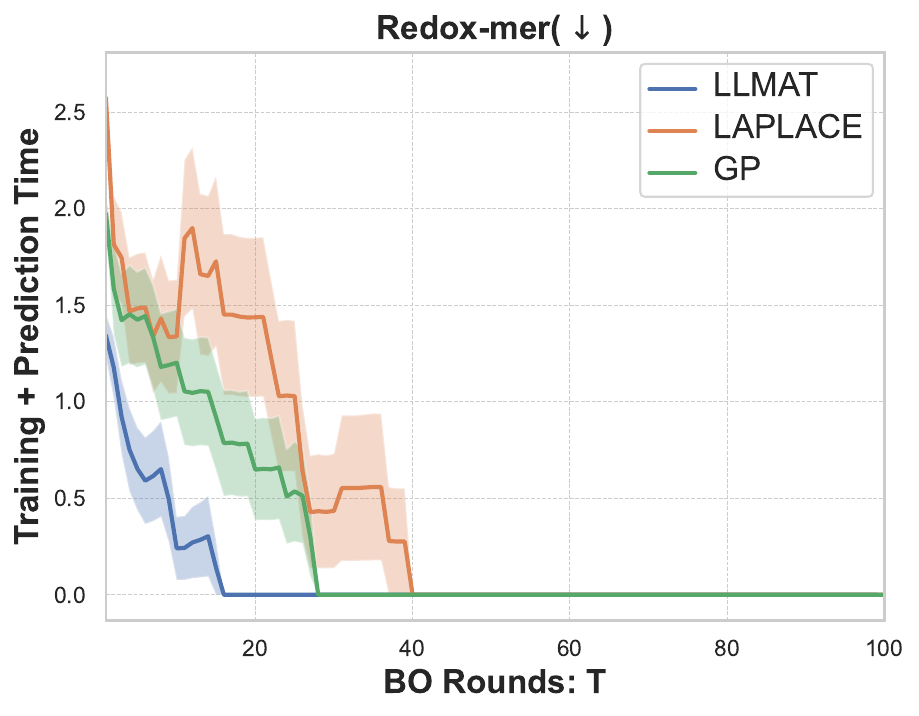}
        \includegraphics[width=0.32\linewidth]{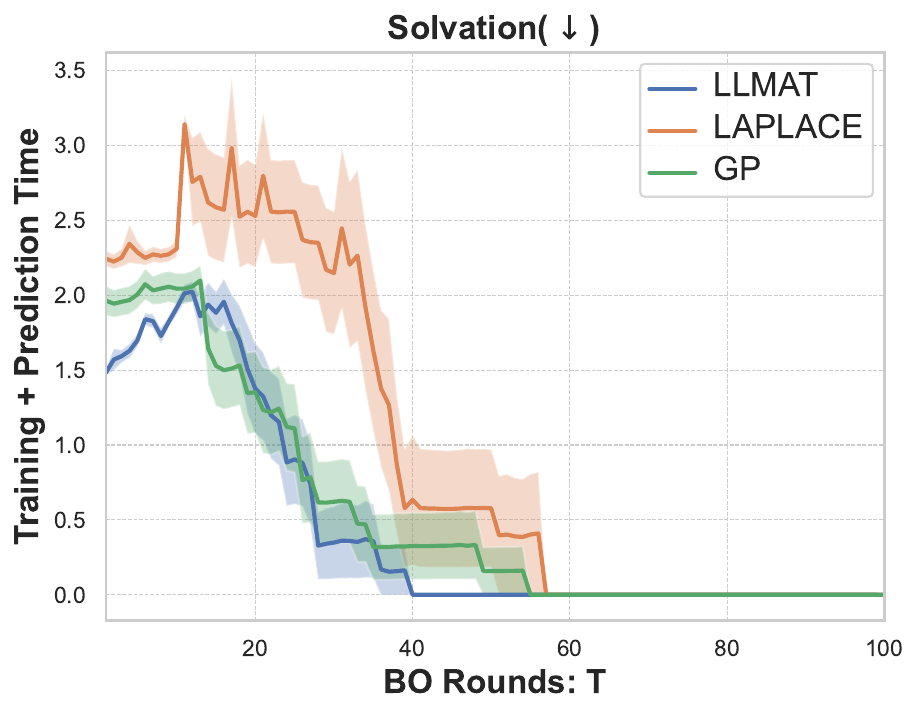}
        \includegraphics[width=0.32\linewidth]{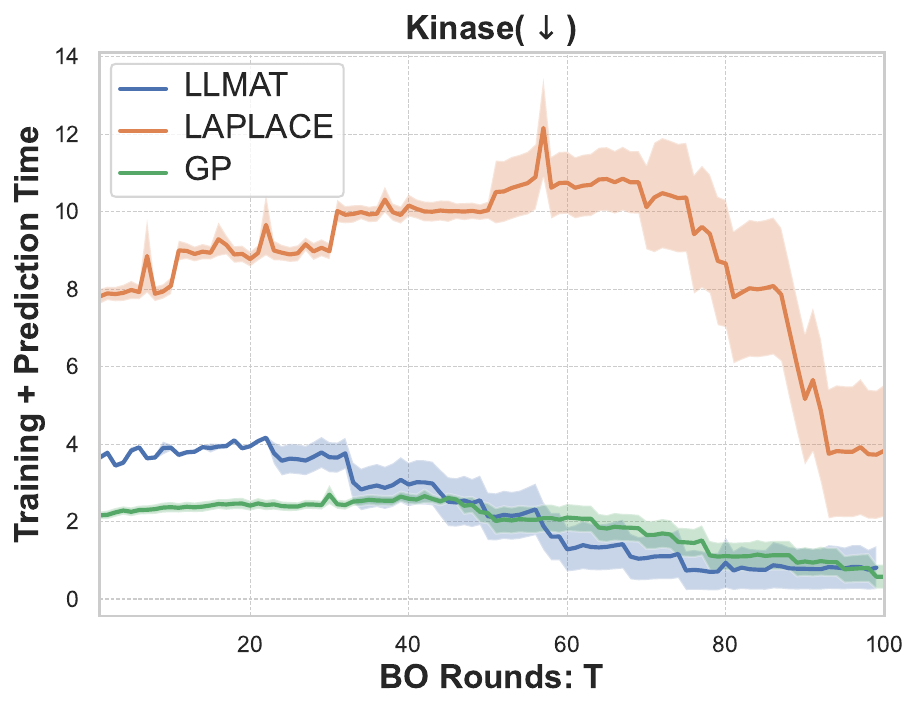}\\
         \includegraphics[width=0.32\linewidth]{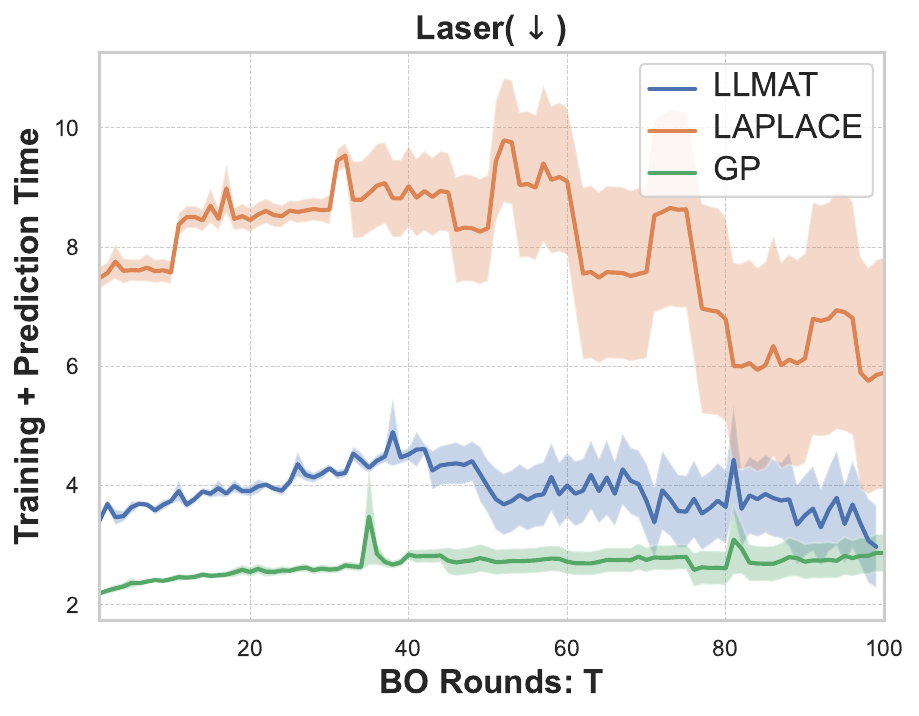}
      \includegraphics[width=0.32\linewidth]{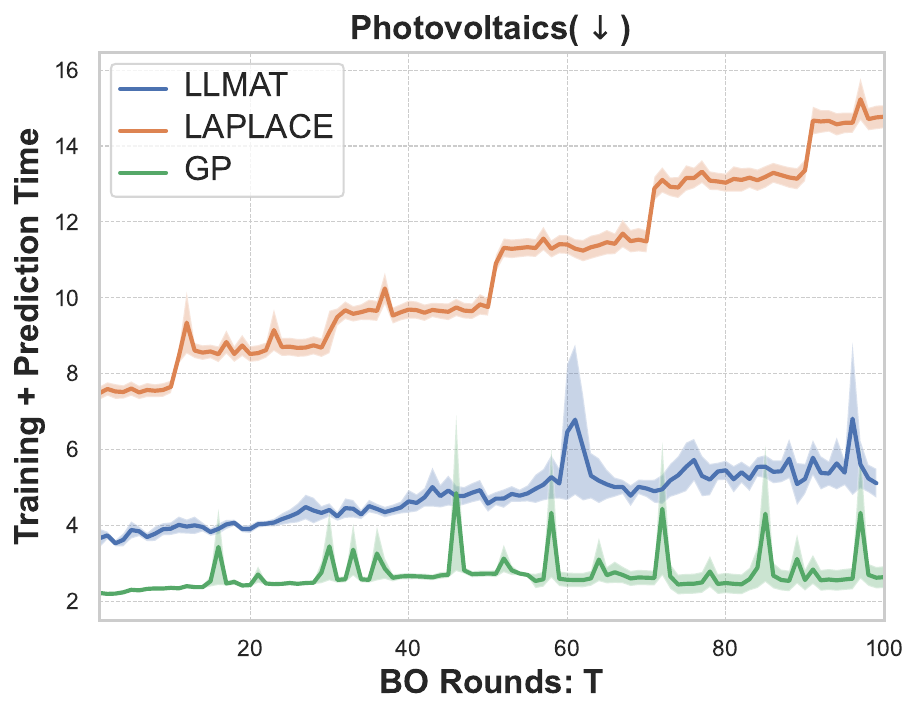}
        \includegraphics[width=0.32\linewidth]{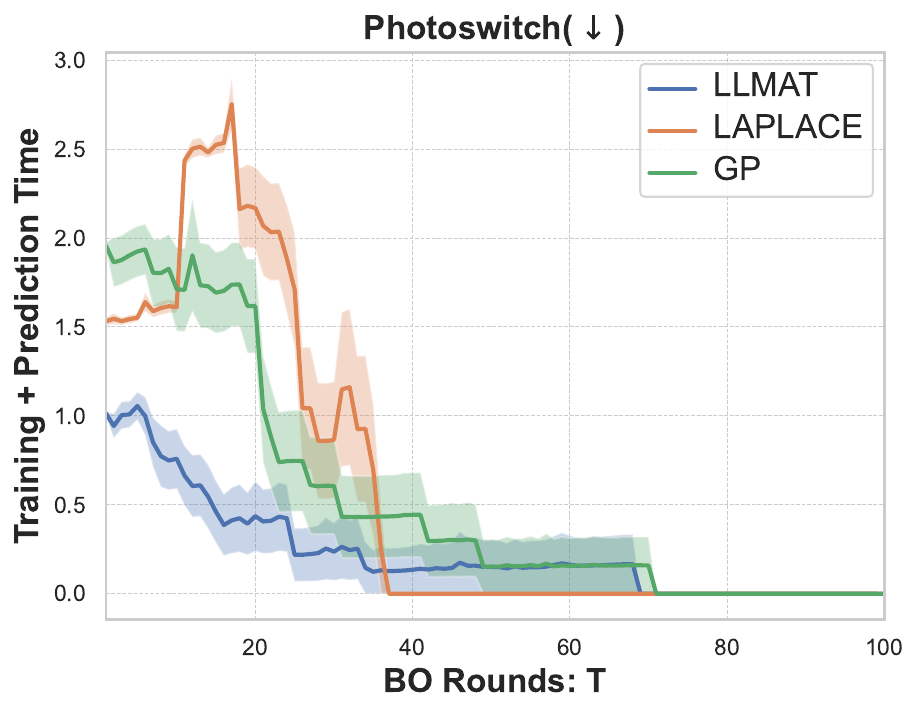}
    \caption{Overall time comparison on different chemistry datasets with fixed T5-Chem features. }
    \label{fig:t5-chem-fixed-overall-time}
\end{figure}
\subsection{Fixed and finetuned results for \texorpdfstring{T5-Chem}{T5-Chem} Model}

\subsubsection{Historical optimums}

\begin{figure}[H]
    \centering
    \includegraphics[width=0.32\linewidth]{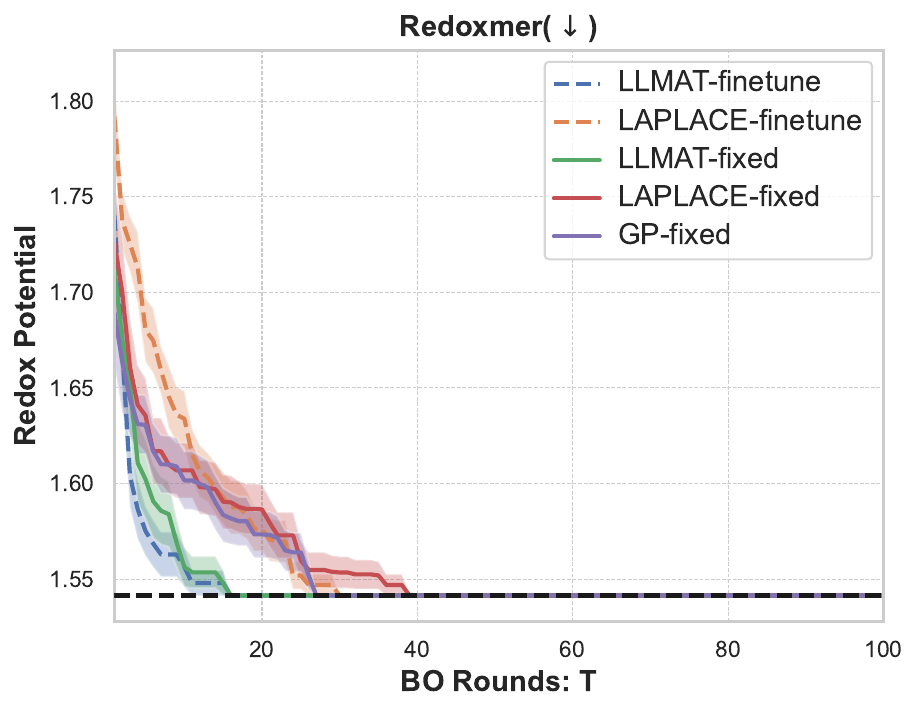}
    \includegraphics[width=0.32\linewidth]{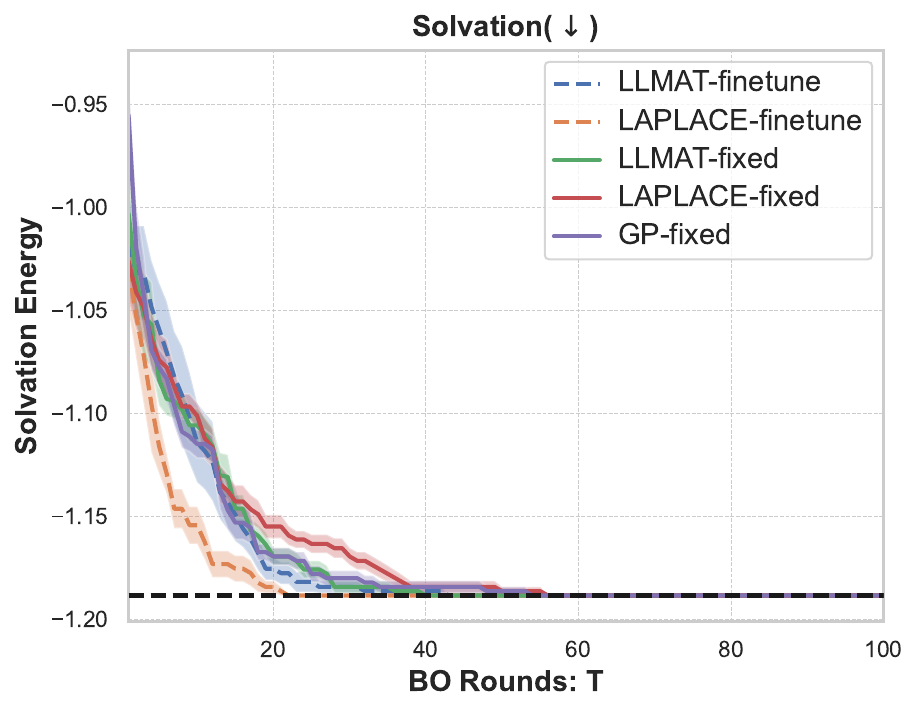}
    \includegraphics[width=0.32\linewidth]{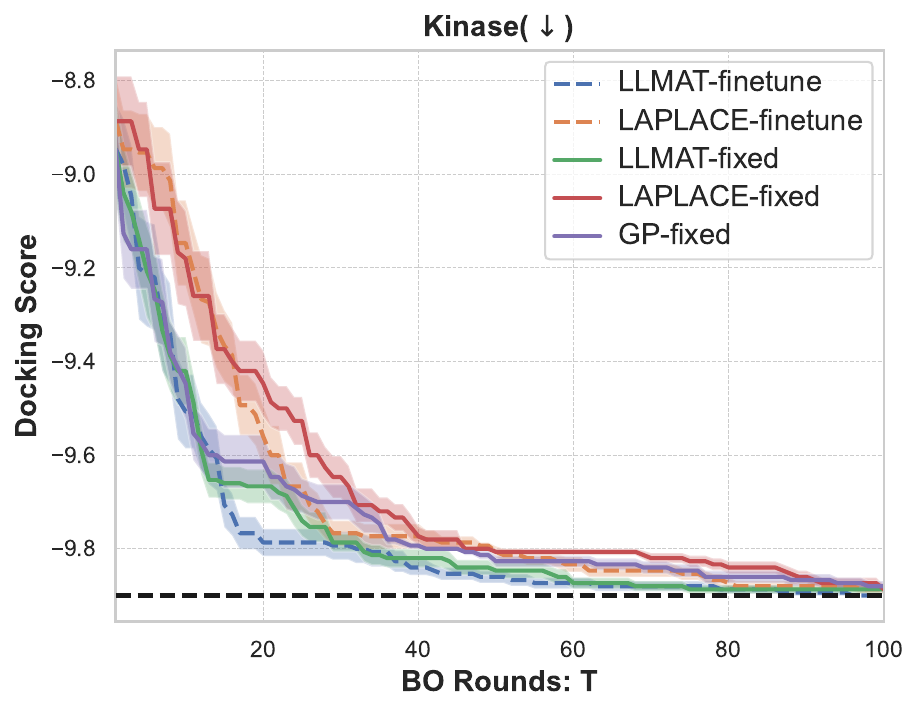}\\
    \includegraphics[width=0.32\linewidth]{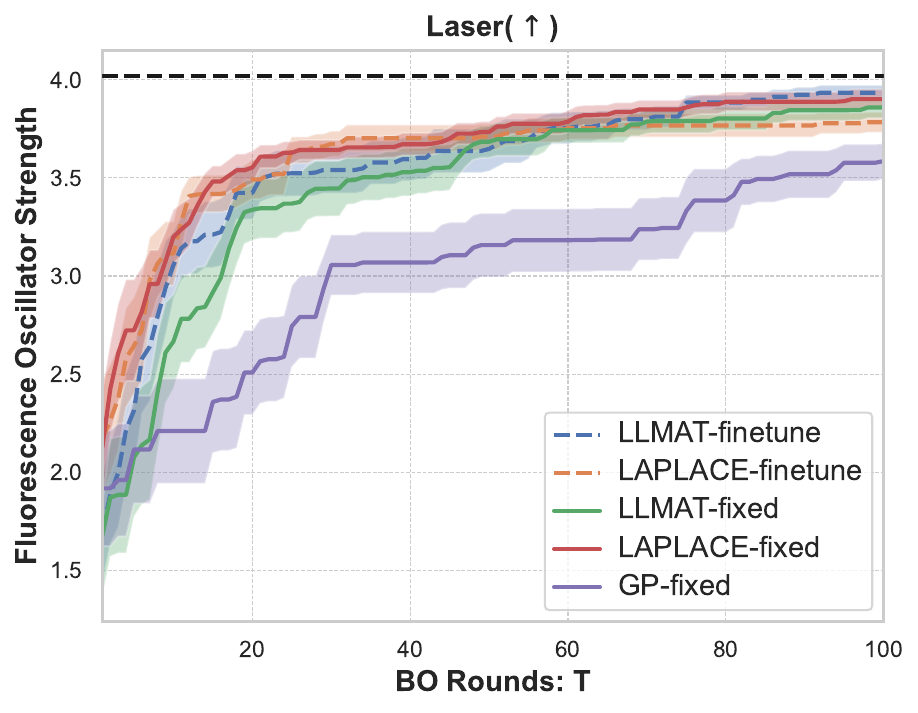}
     \includegraphics[width=0.32\linewidth]{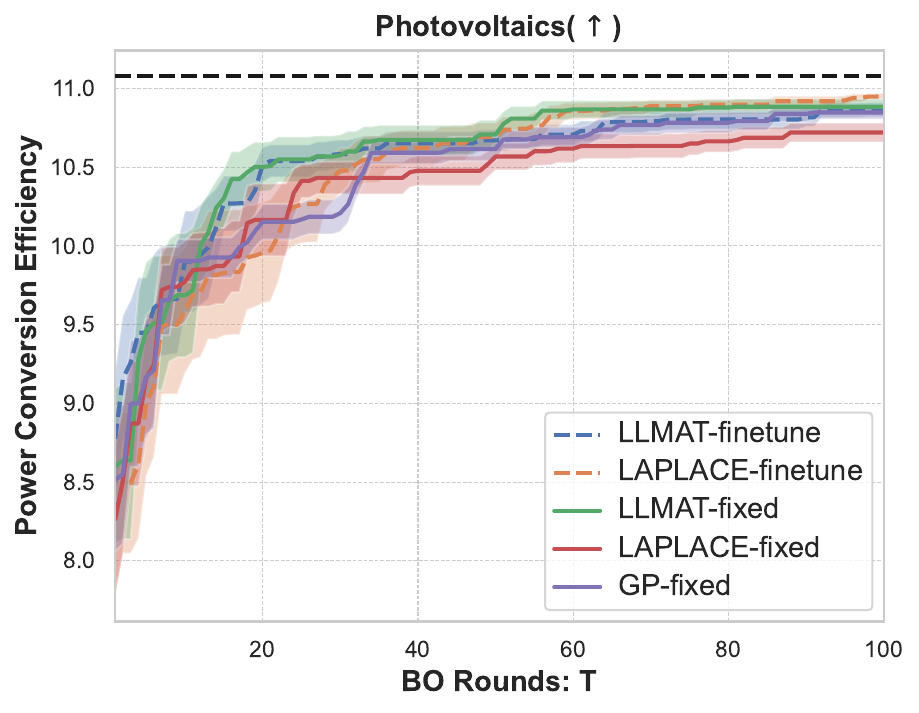}
    \includegraphics[width=0.32\linewidth]{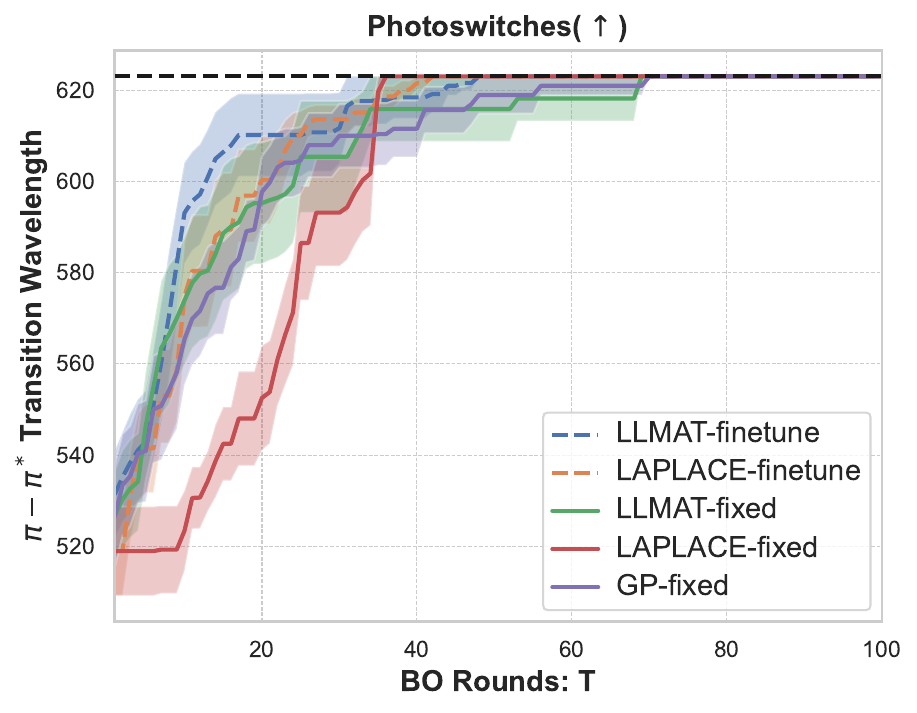}
    \caption{Historical optimums comparison on different chemistry datasets using T5-Chem model. }
    \label{fig:t5-chem-finetune-his}
\end{figure}

\subsubsection{GAP metric}

\begin{figure}[H]
    \centering
    \includegraphics[width=0.32\linewidth]{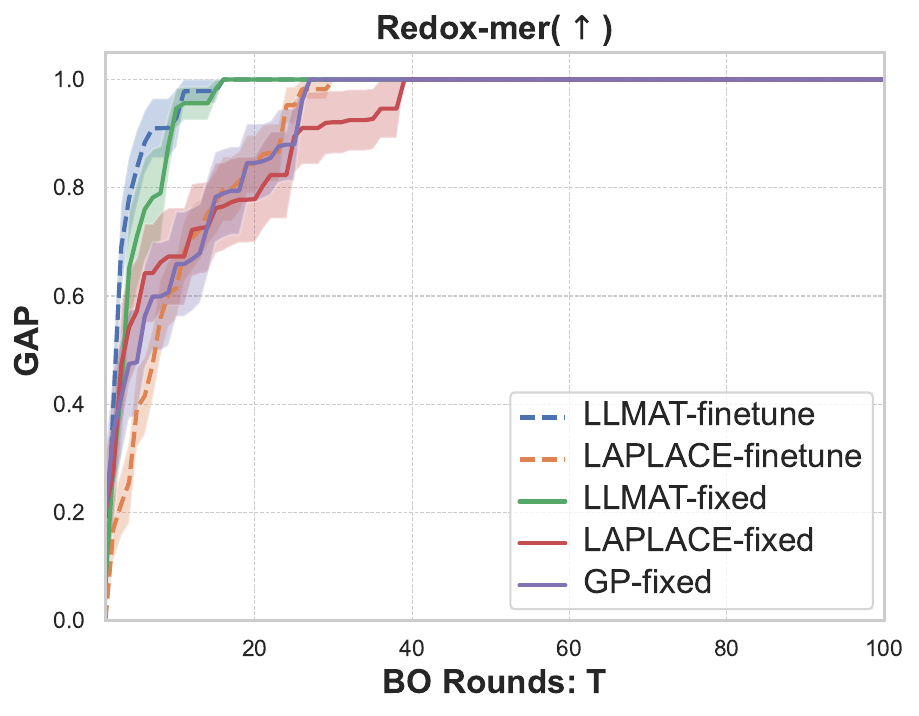}
    \includegraphics[width=0.32\linewidth]{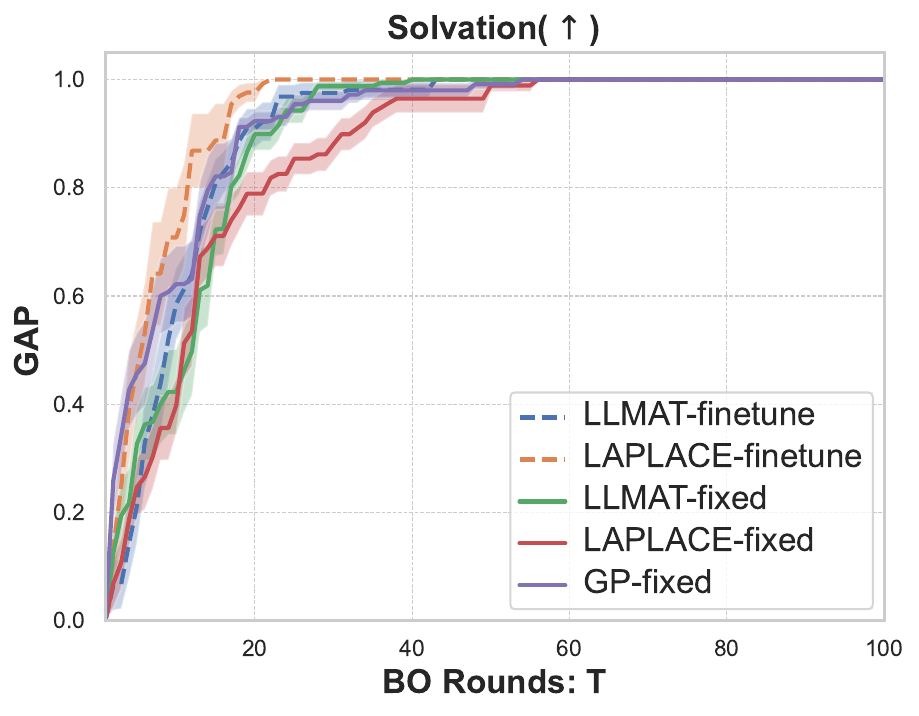}
    \includegraphics[width=0.32\linewidth]{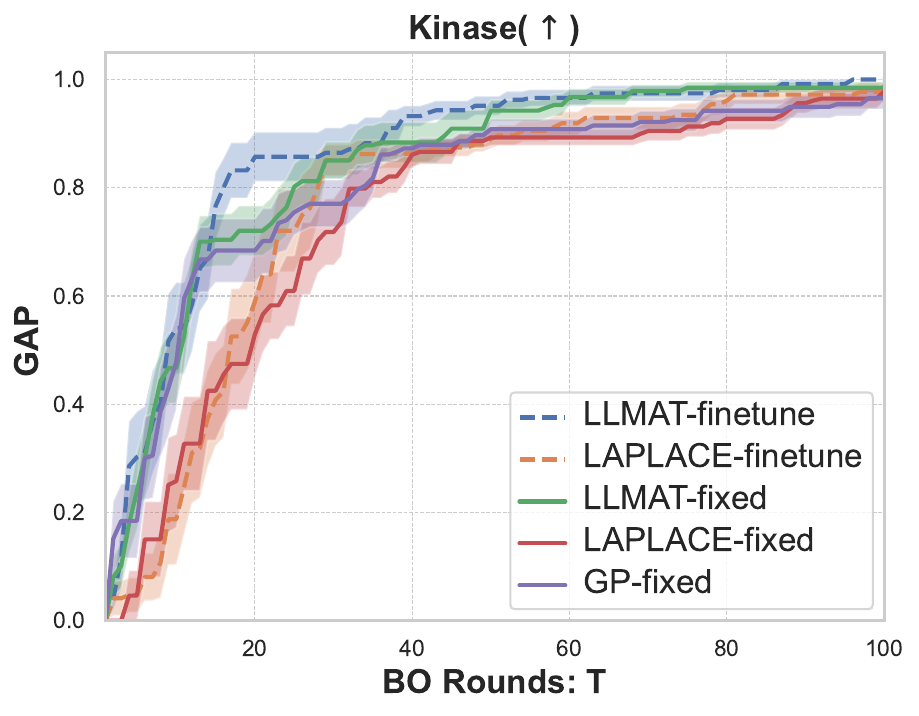}\\
    \includegraphics[width=0.32\linewidth]{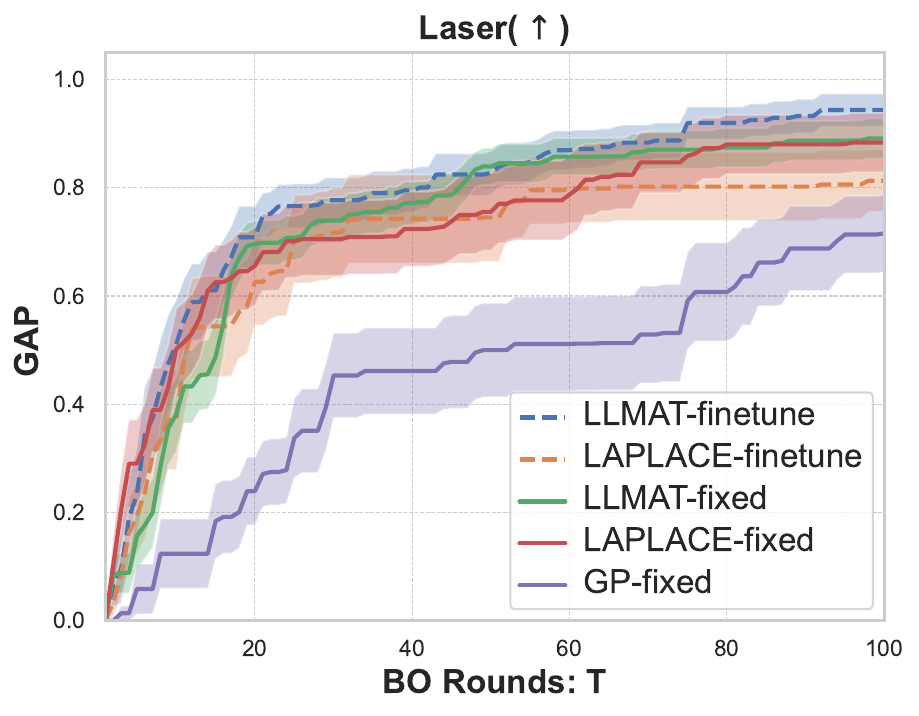}
     \includegraphics[width=0.32\linewidth]{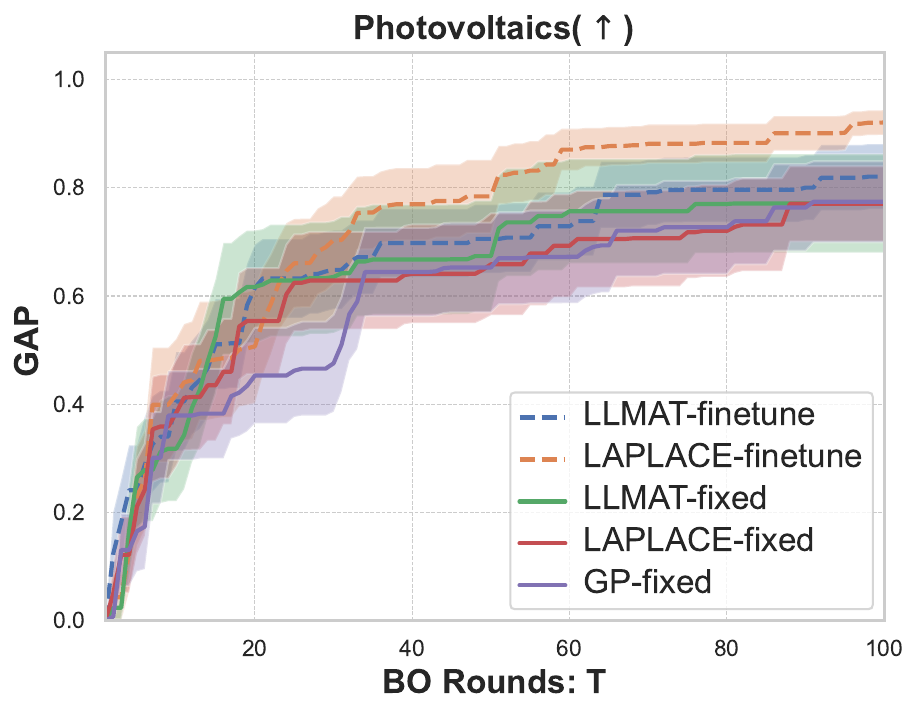}
    \includegraphics[width=0.32\linewidth]{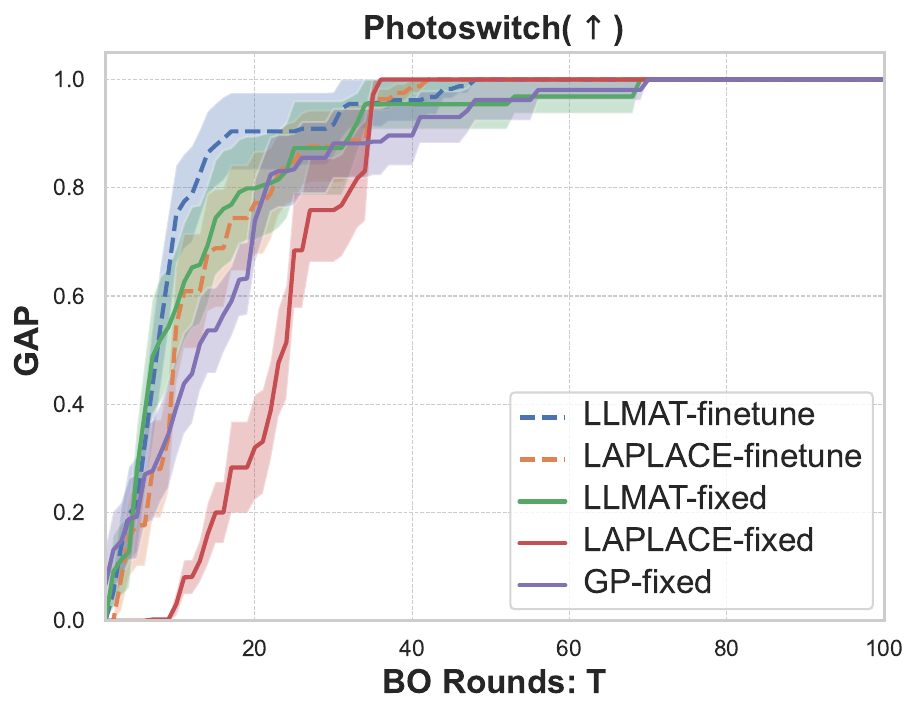}
    \caption{GAP comparison on different chemistry datasets with T5-Chem model. }
    \label{fig:t5-chem-finetune-gap}
\end{figure}

\subsubsection{Average Regret}

\begin{figure}[H]
    \centering
    \includegraphics[width=0.32\linewidth]{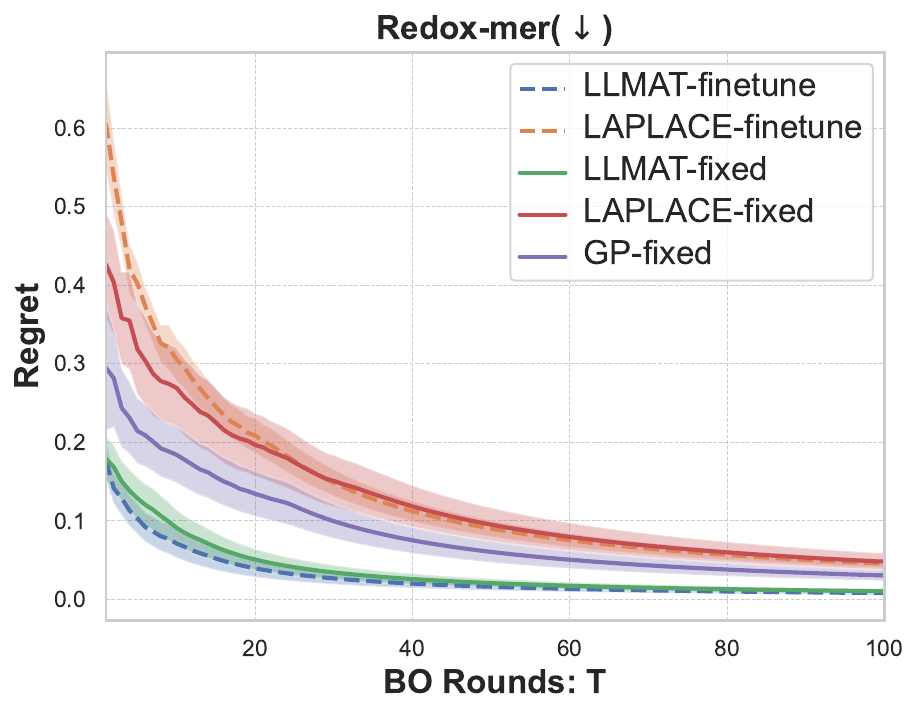}
    \includegraphics[width=0.32\linewidth]{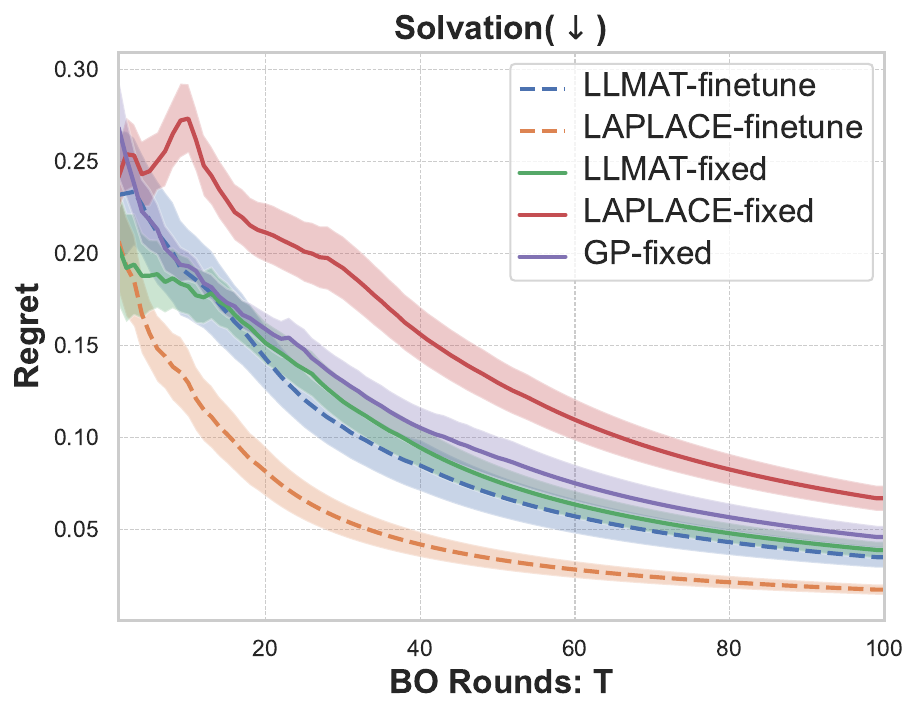}
    \includegraphics[width=0.32\linewidth]{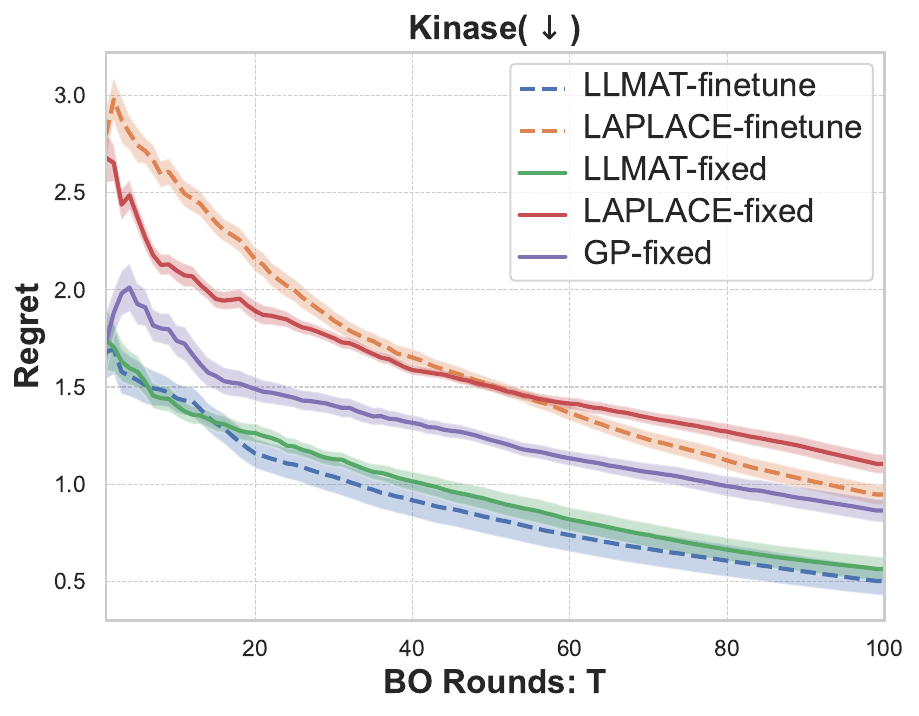}\\
     \includegraphics[width=0.32\linewidth]{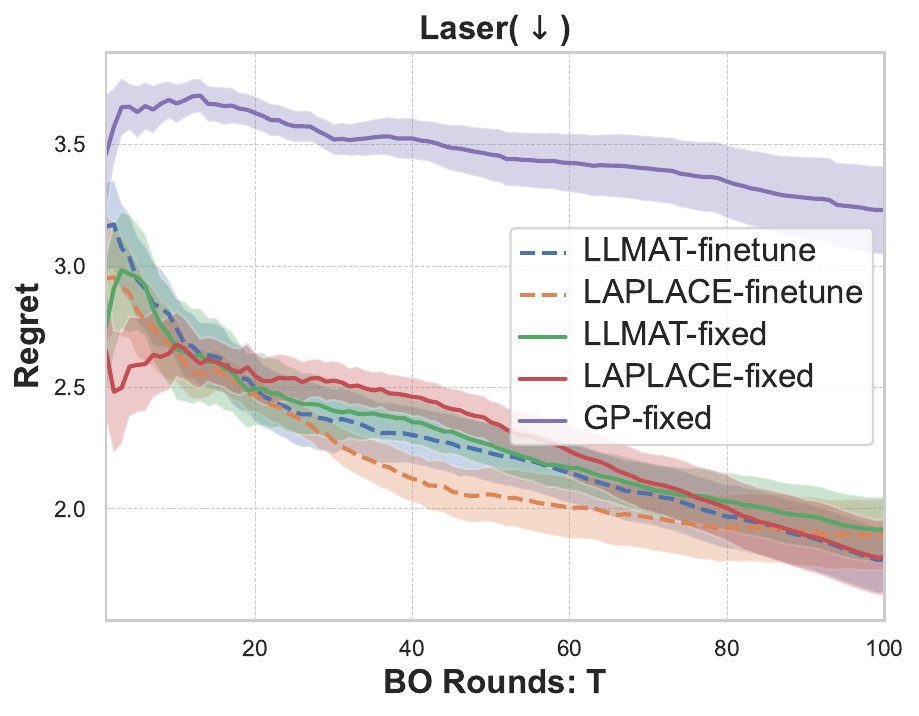}
    \includegraphics[width=0.32\linewidth]{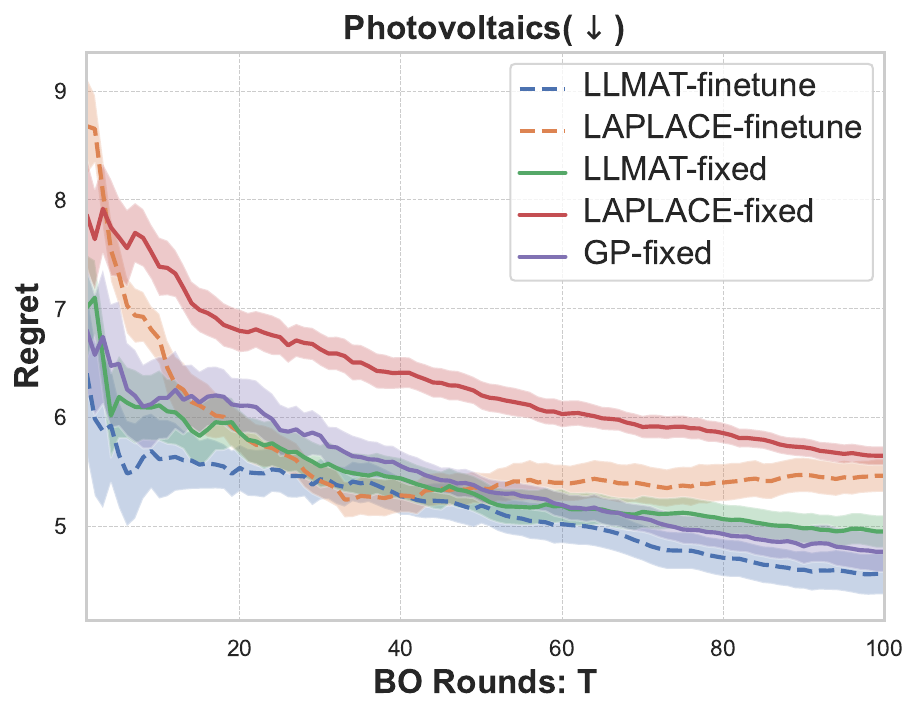}
    \includegraphics[width=0.32\linewidth]{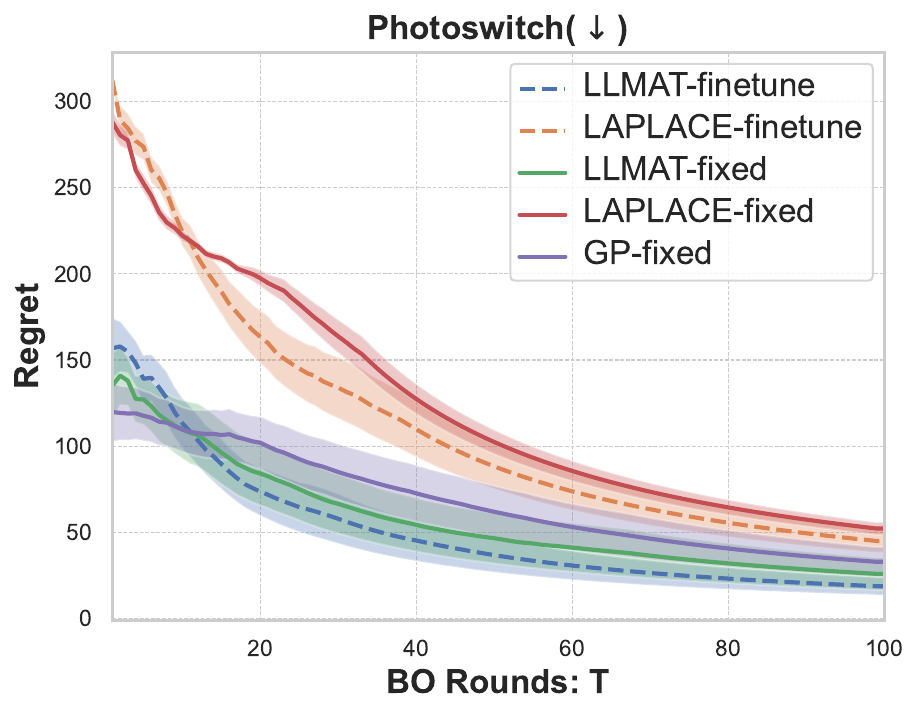}
    \caption{Regrets comparison on different chemistry datasets using T5-Chem model. }
    \label{fig:t5-chem-finetune-regret}
\end{figure}



\subsubsection{Computation time for PEFT}

\begin{figure}[H]
    \centering
    \includegraphics[width=0.32\linewidth]{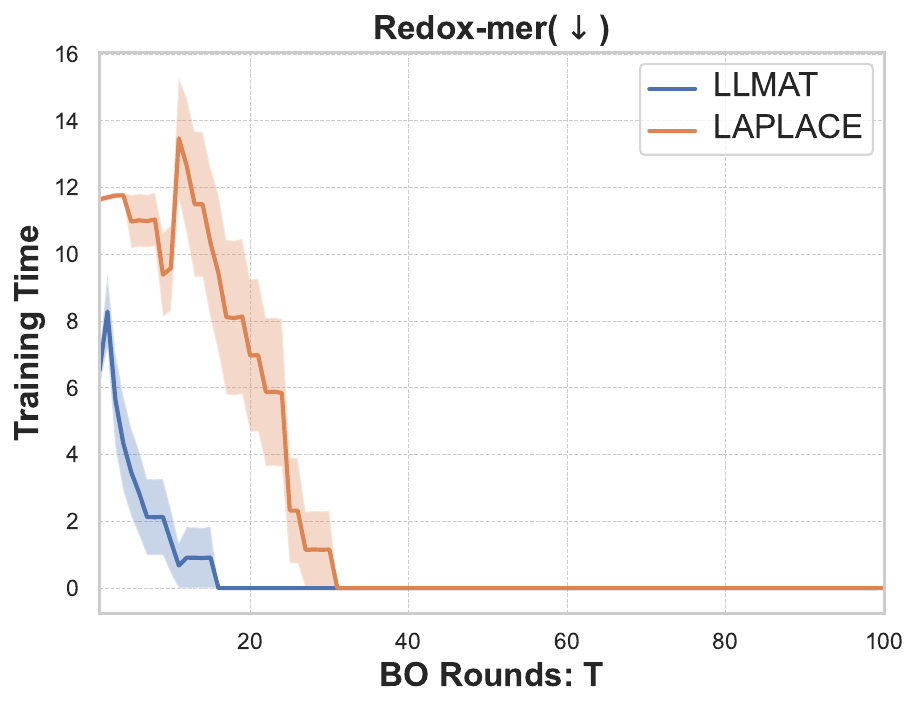}
        \includegraphics[width=0.32\linewidth]{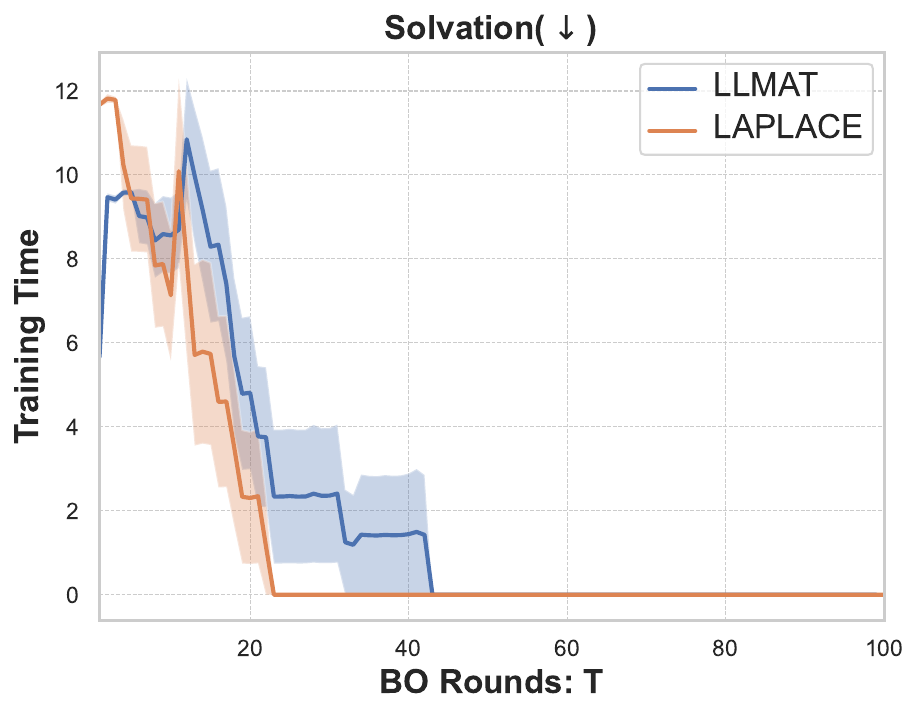}
        \includegraphics[width=0.32\linewidth]{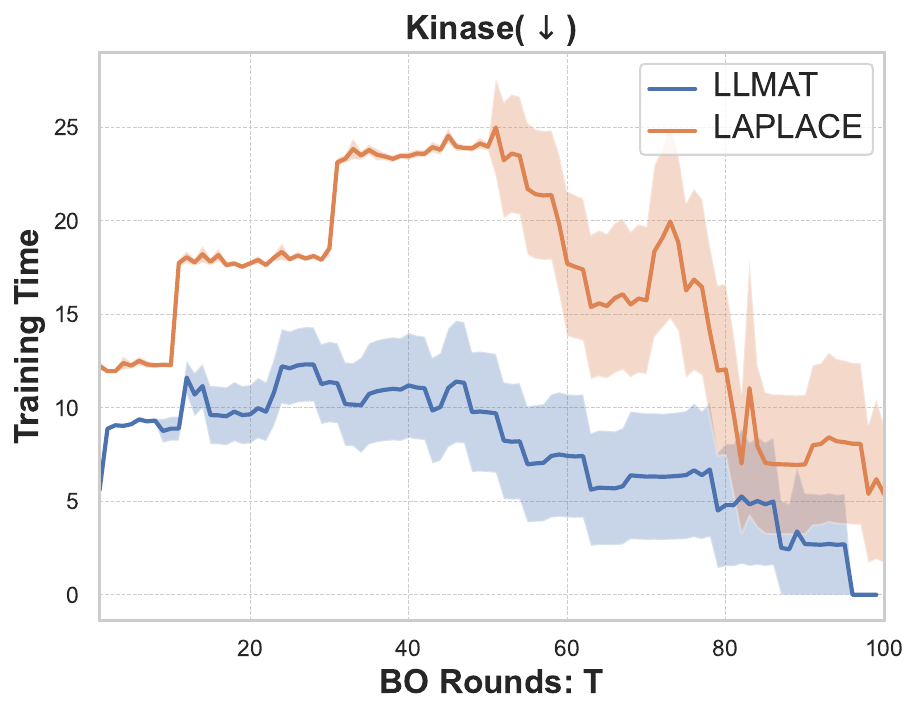}\\
         \includegraphics[width=0.32\linewidth]{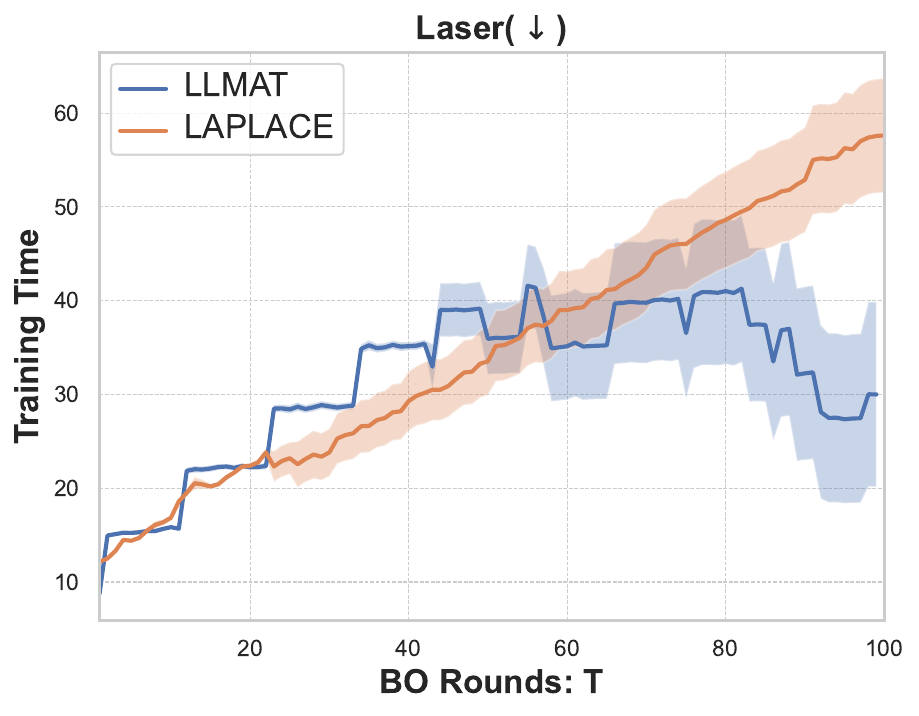}
      \includegraphics[width=0.32\linewidth]{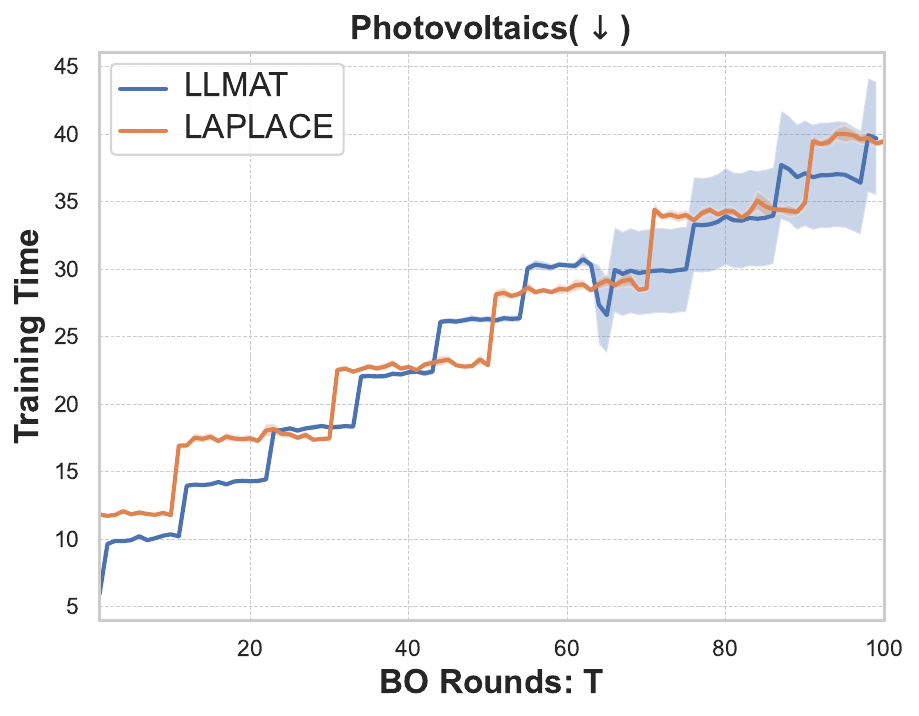}
        \includegraphics[width=0.32\linewidth]{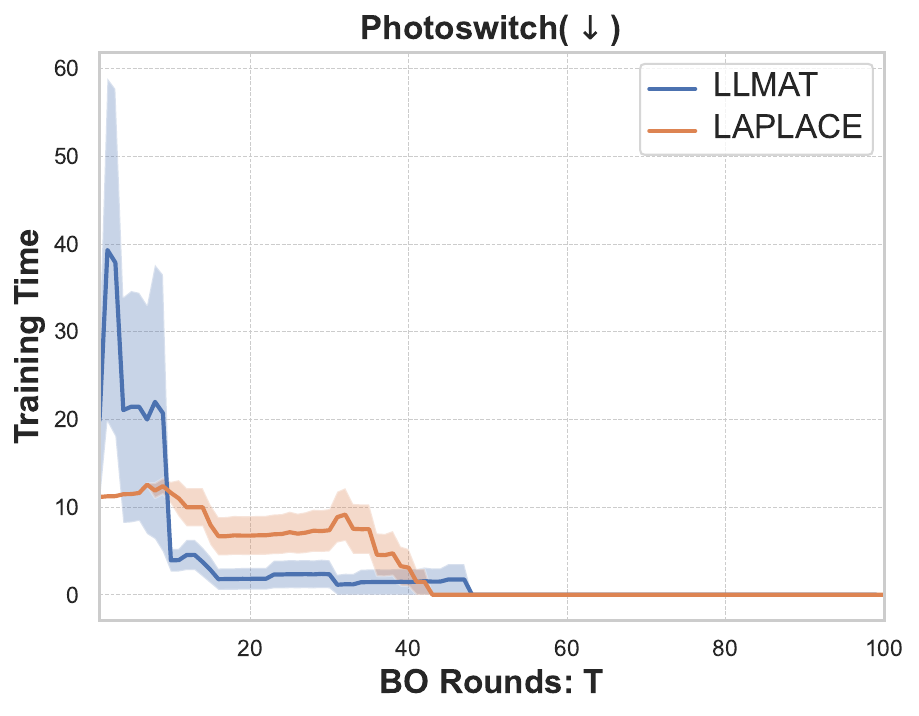}
    \caption{Training time comparison on different chemistry datasets with PEFT on the T5-Chem model. }
    \label{fig:t5-chem-finetune-train-time}
\end{figure}
\begin{figure}[H]
    \centering
    \includegraphics[width=0.32\linewidth]{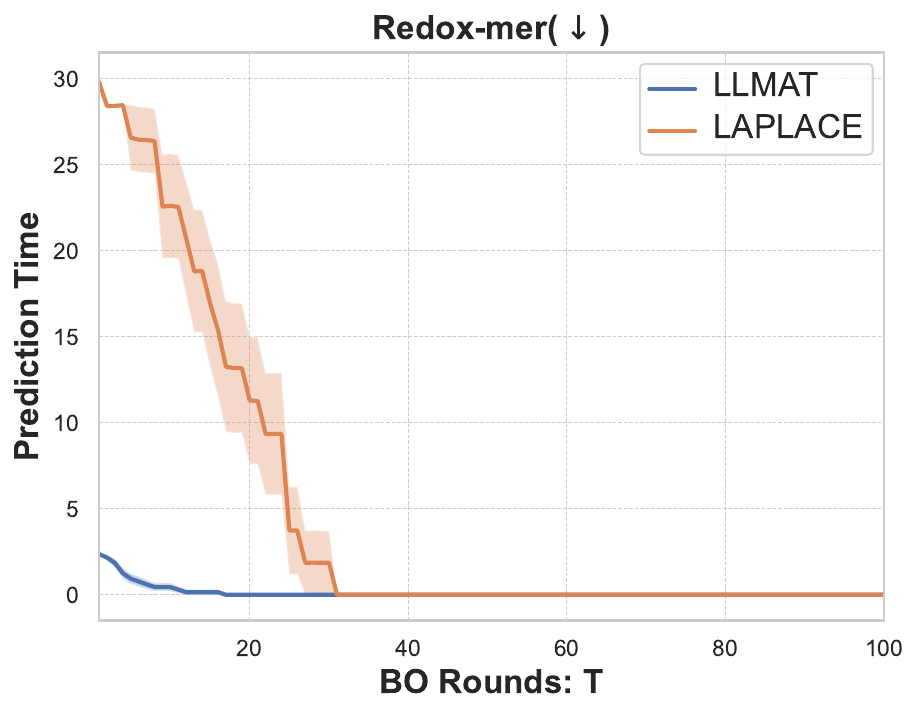}
        \includegraphics[width=0.32\linewidth]{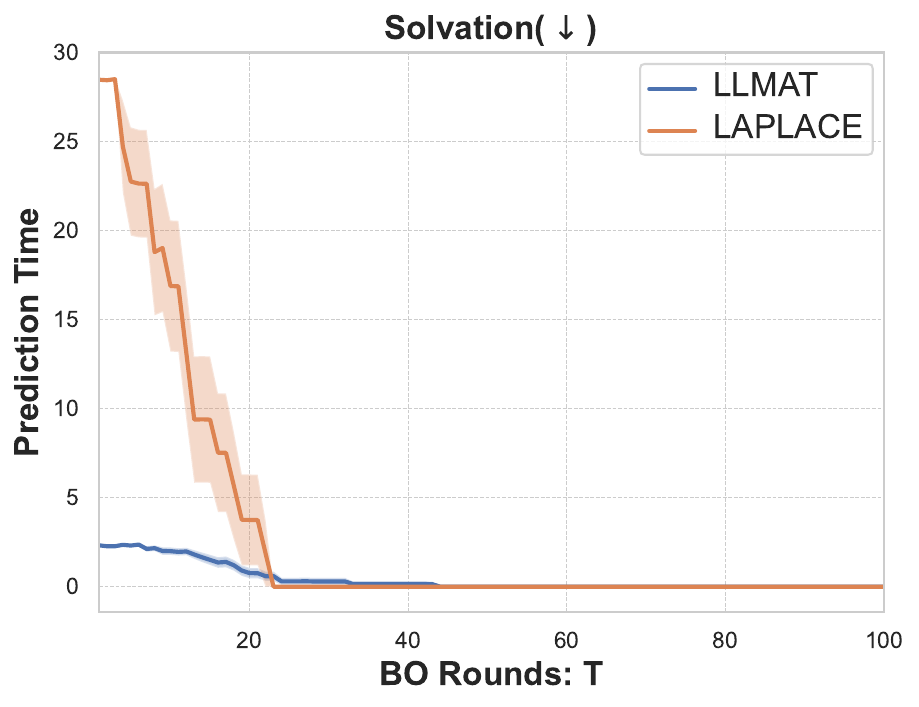}
        \includegraphics[width=0.32\linewidth]{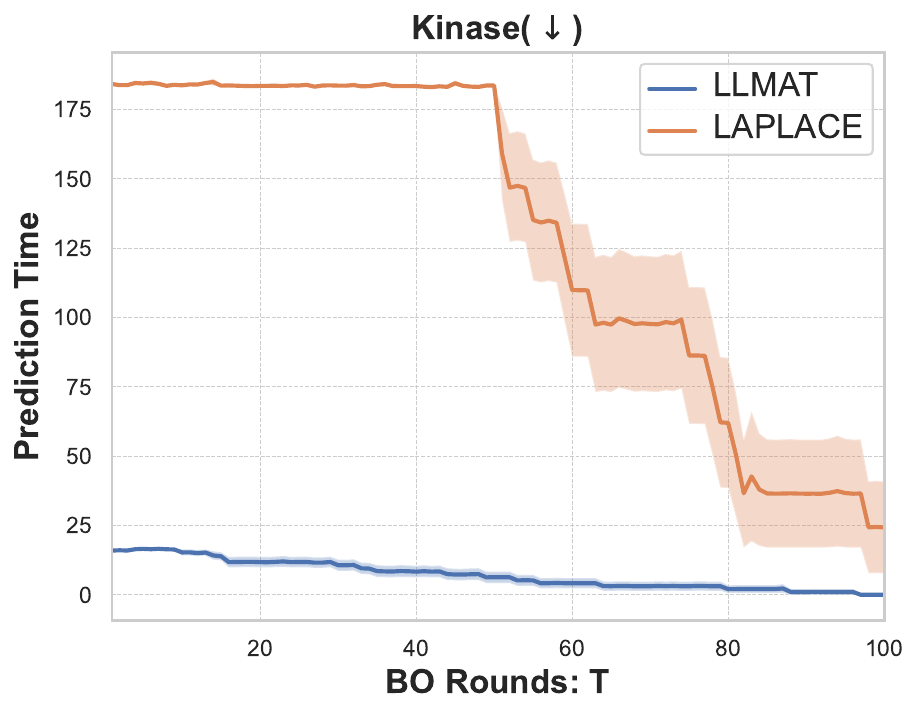}\\
         \includegraphics[width=0.32\linewidth]{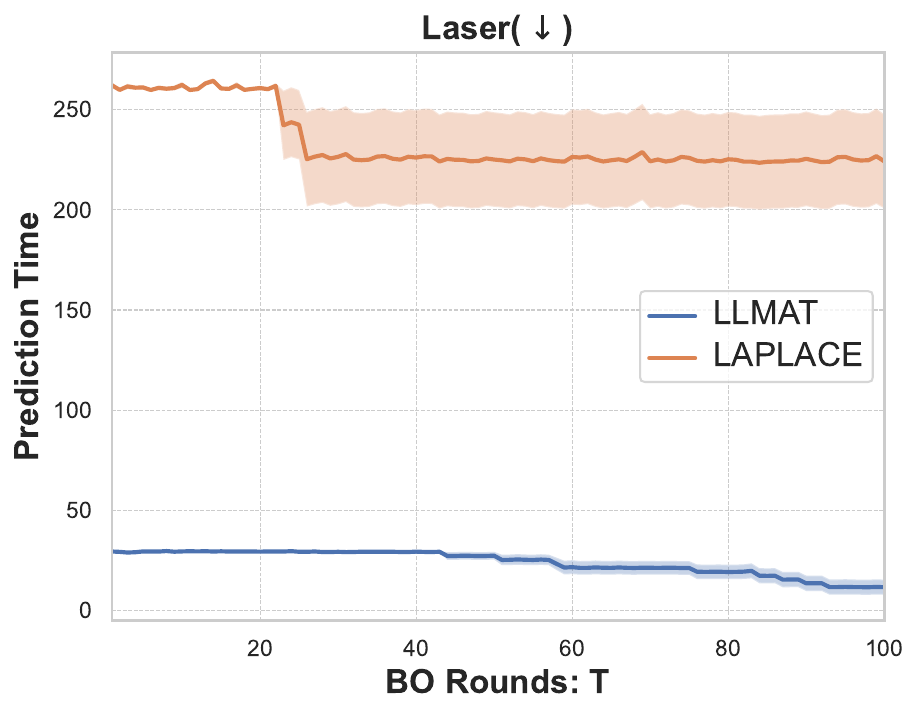}
      \includegraphics[width=0.32\linewidth]{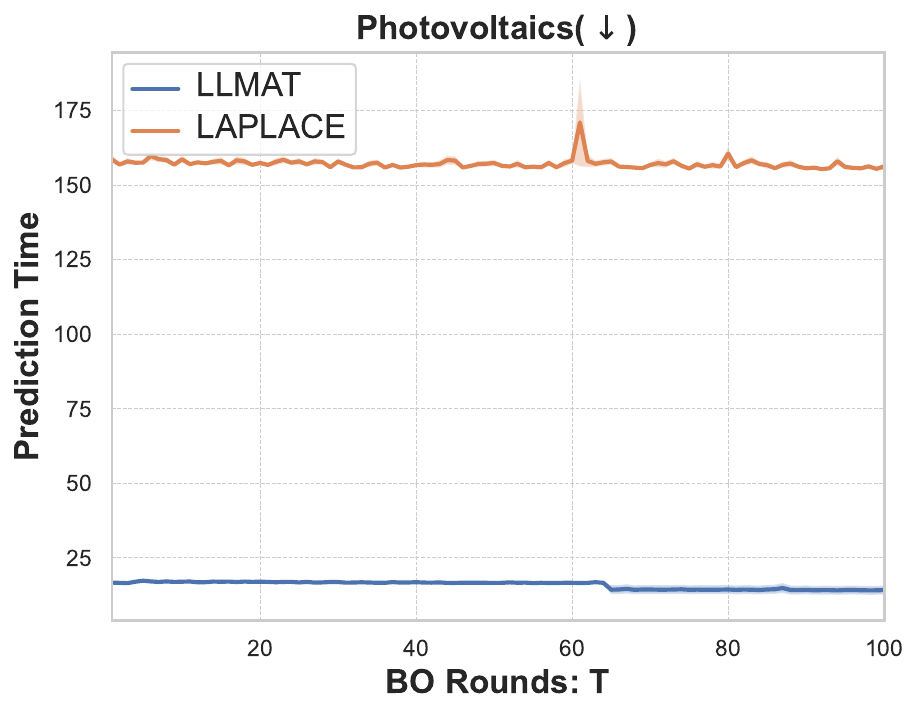}
        \includegraphics[width=0.32\linewidth]{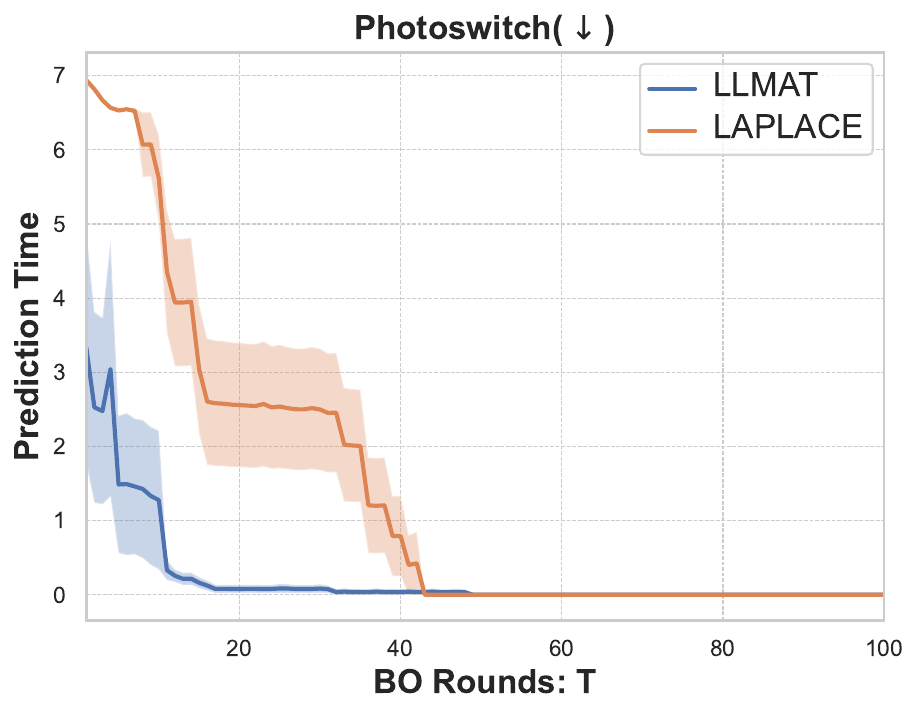}
    \caption{Prediction time comparison on different chemistry datasets with PEFT on the T5-Chem model. }
    \label{fig:t5-chem-finetune-pred-time}
\end{figure}

\begin{figure}[H]
    \centering
    \includegraphics[width=0.3\linewidth]{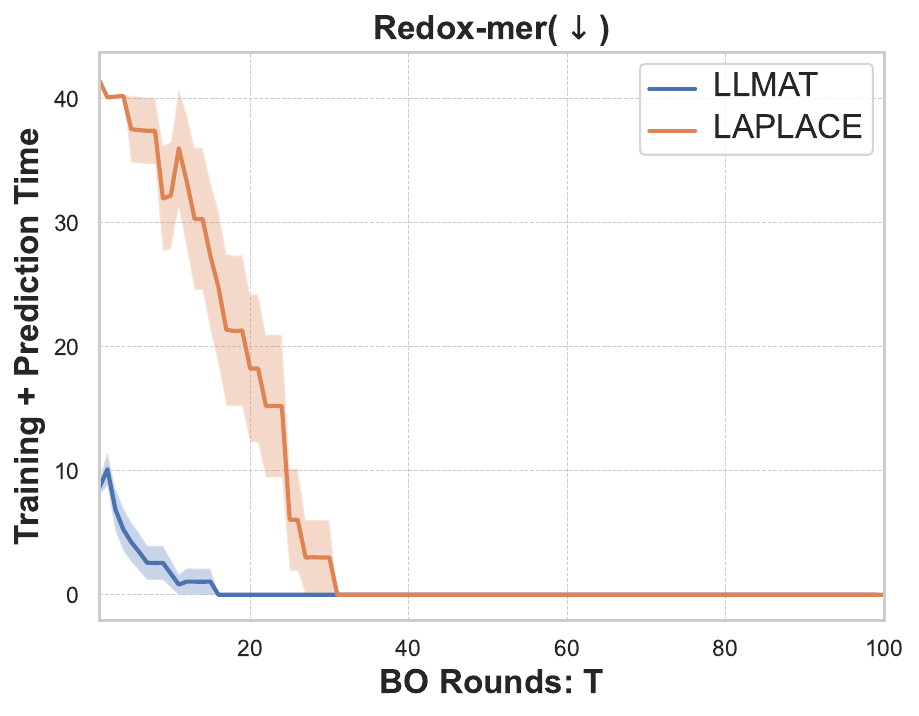}
        \includegraphics[width=0.3\linewidth]{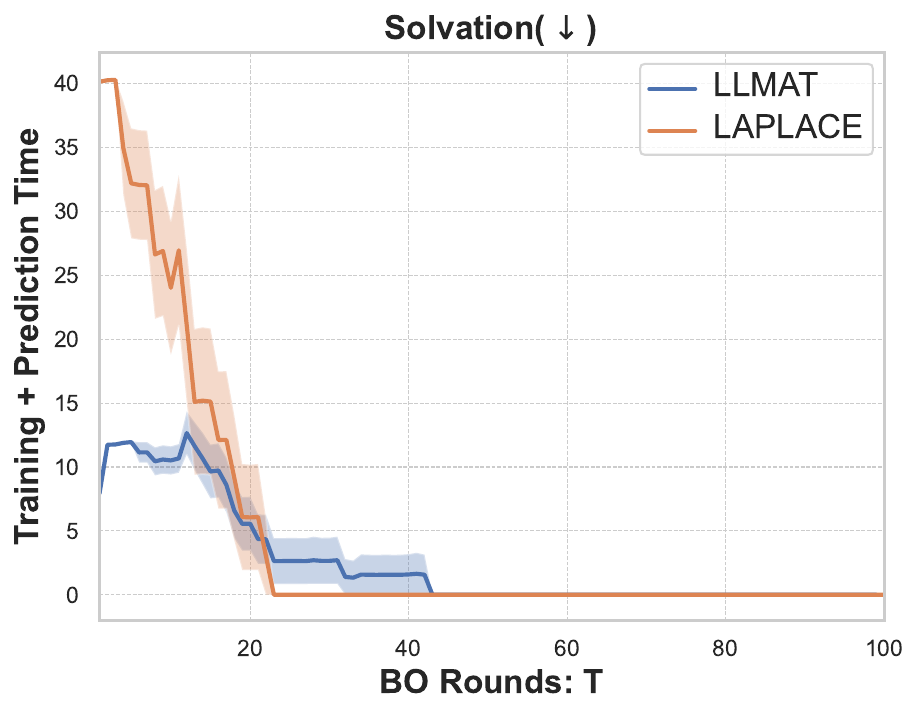}
        \includegraphics[width=0.3\linewidth]{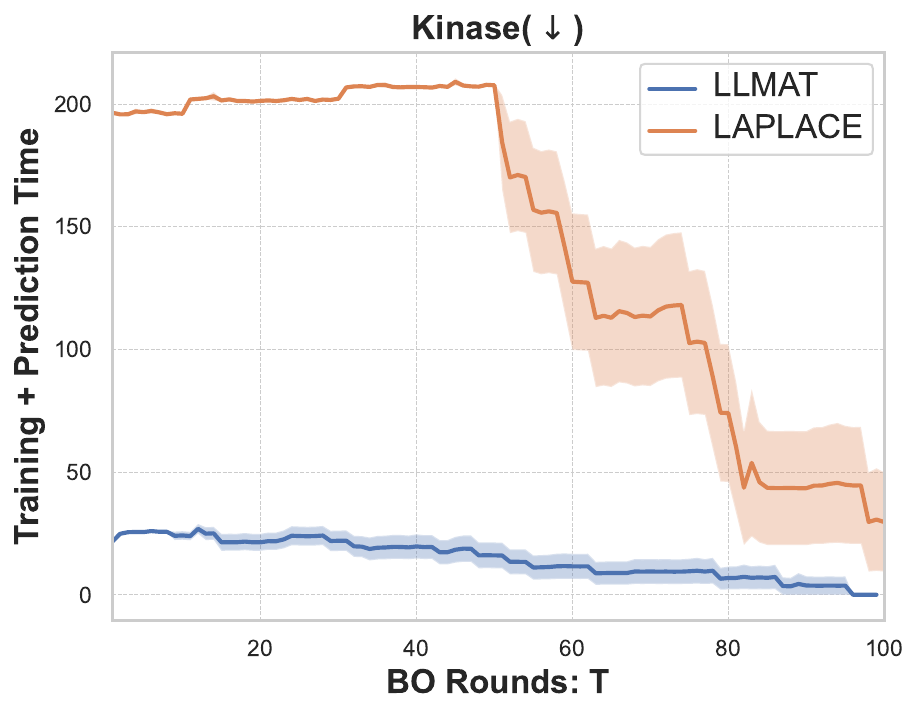}\\
         \includegraphics[width=0.3\linewidth]{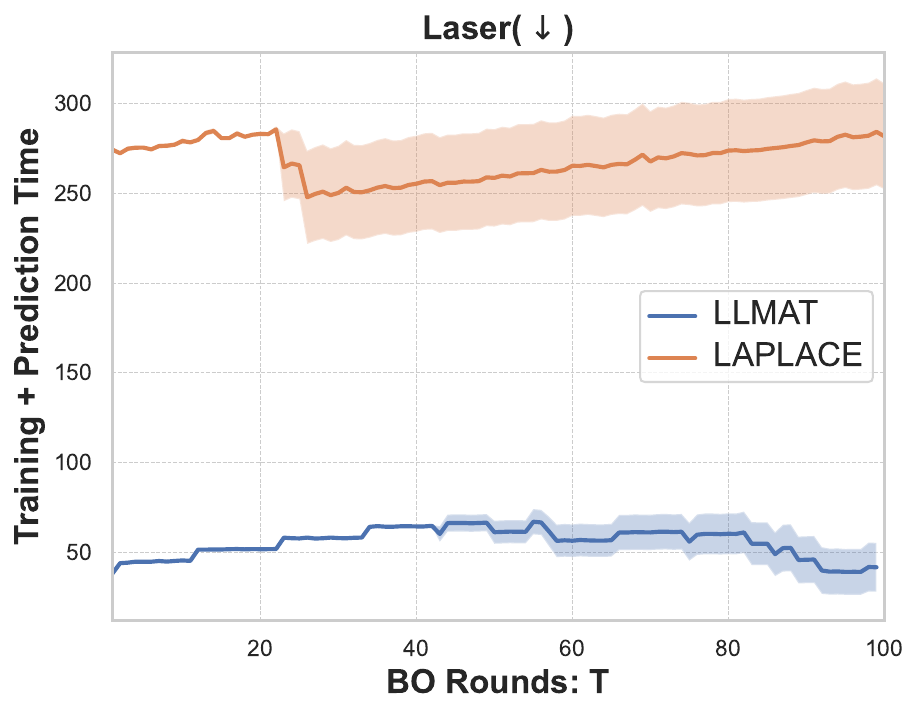}
      \includegraphics[width=0.3\linewidth]{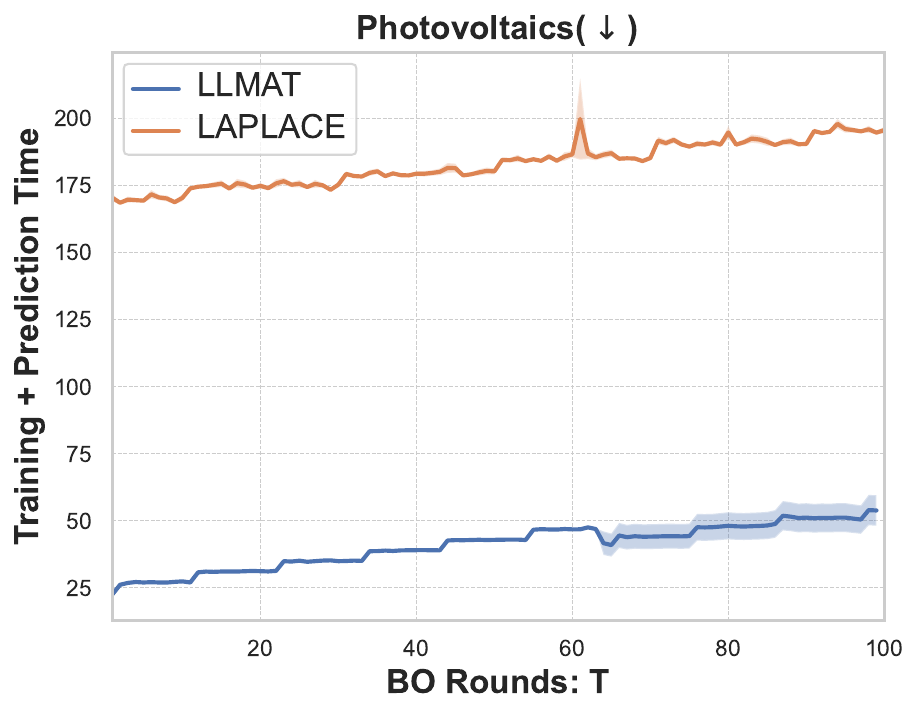}
        \includegraphics[width=0.3\linewidth]{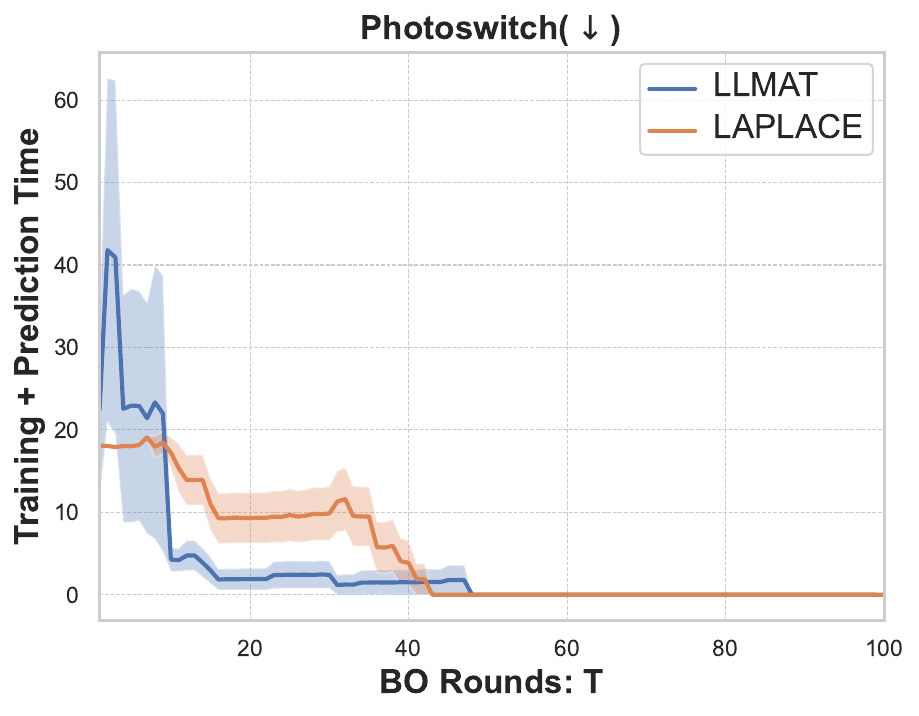}
    \caption{Overall time comparison on different chemistry datasets with PEFT on the T5-Chem model. }
    \label{fig:t5-chem-finetune-overall-time}
\end{figure}

\begin{figure}[H]
    \centering
    \includegraphics[width=\linewidth]{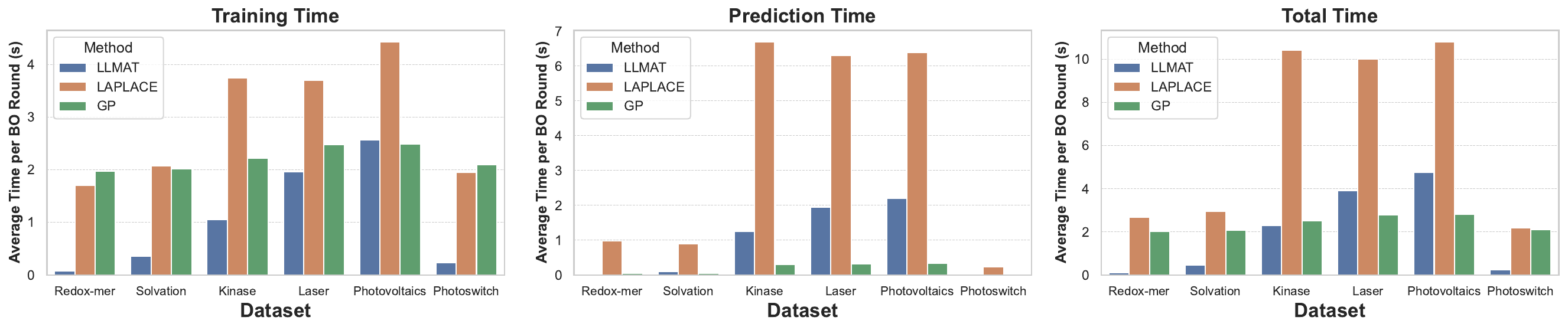}
    \caption{Computation time of different algorithms with fixed T5-Chem features across datasets.}
    \label{fig:t5-chem-time-fix}
\end{figure}
\begin{figure}[H]
    \centering
    \includegraphics[width=\linewidth]{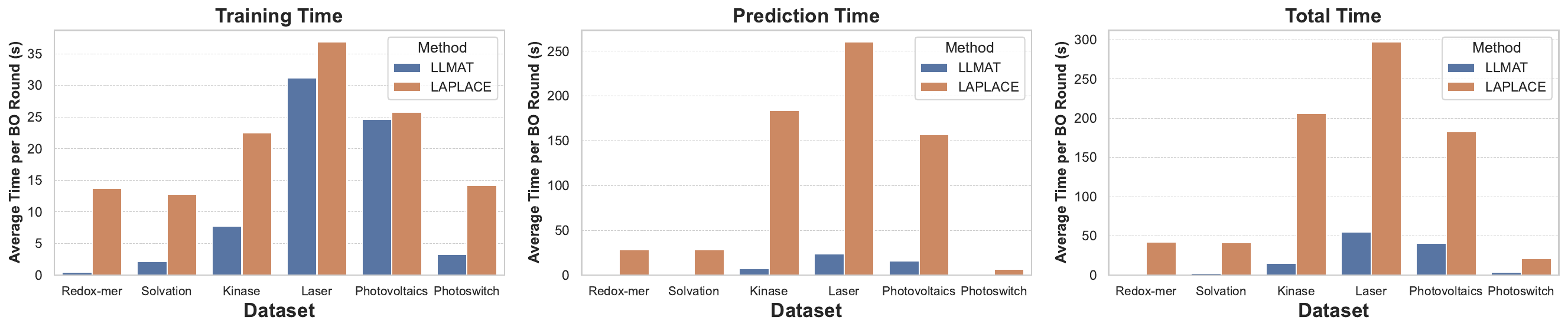}
    \caption{BO time for different datasets on finetuning T5-Chem model with LoRA. }
    \label{fig:t5-chem-time-finetune}
\end{figure}

\subsection{Fixed and finetuned results for Molformer}

\subsubsection{Historical optimums}

\begin{figure}[H]
    \centering
    \includegraphics[width=0.32\linewidth]{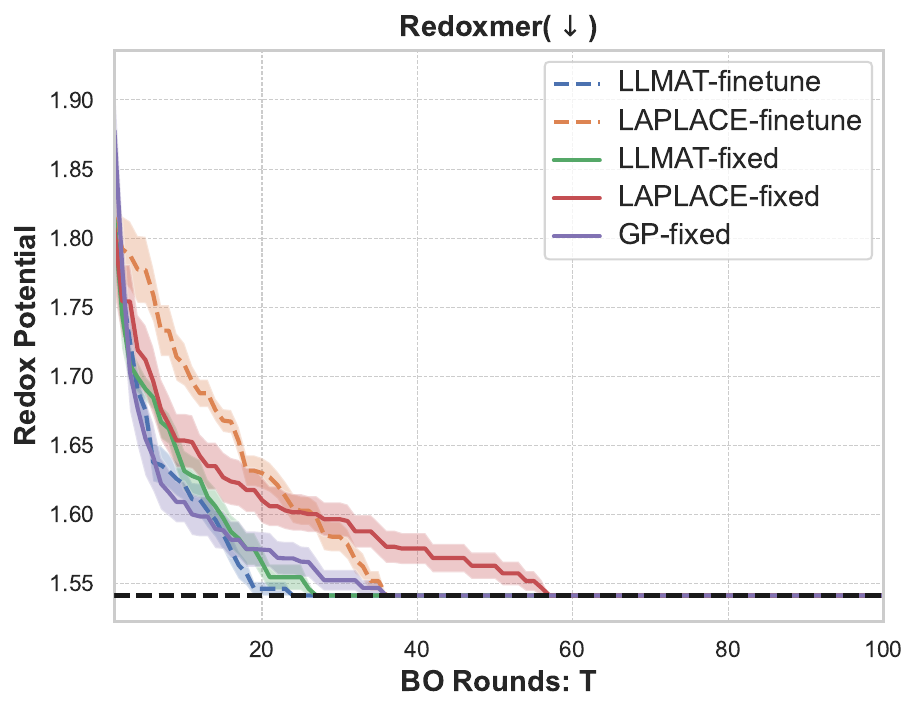}
    \includegraphics[width=0.32\linewidth]{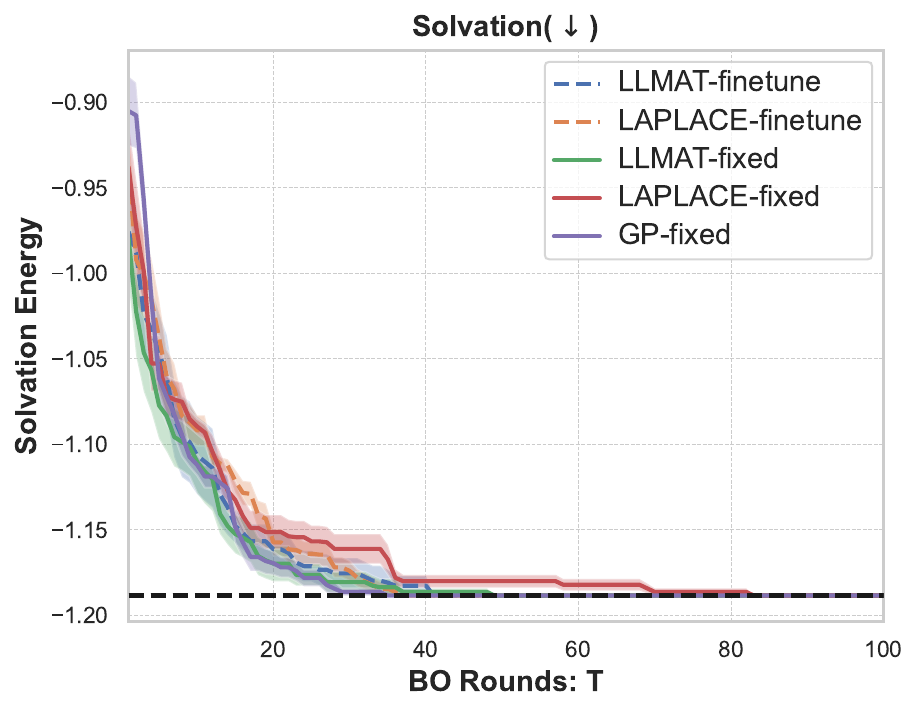}
    \includegraphics[width=0.32\linewidth]{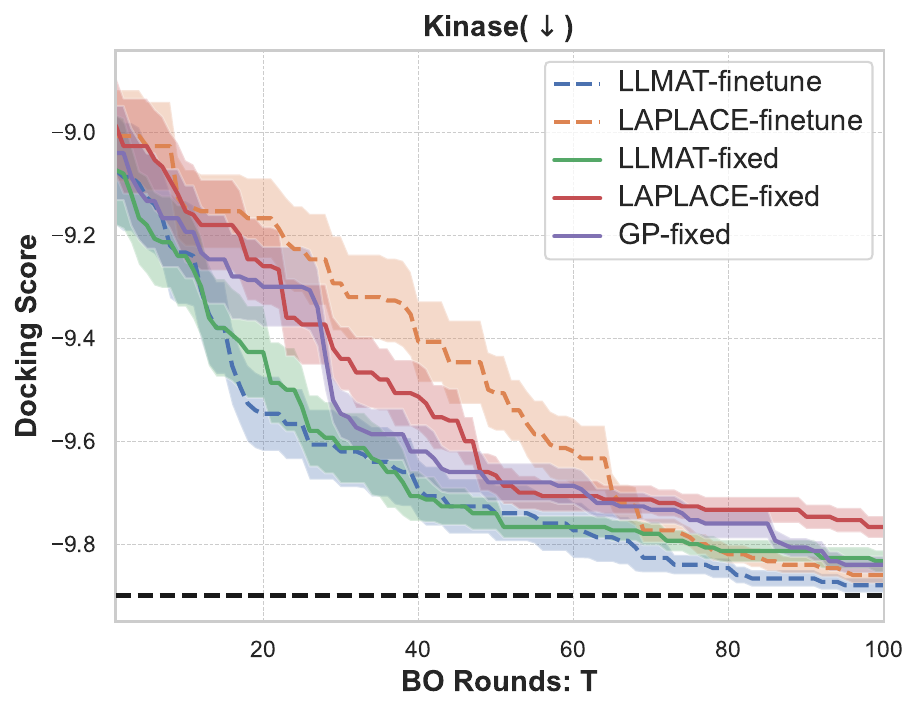}\\
    \includegraphics[width=0.32\linewidth]{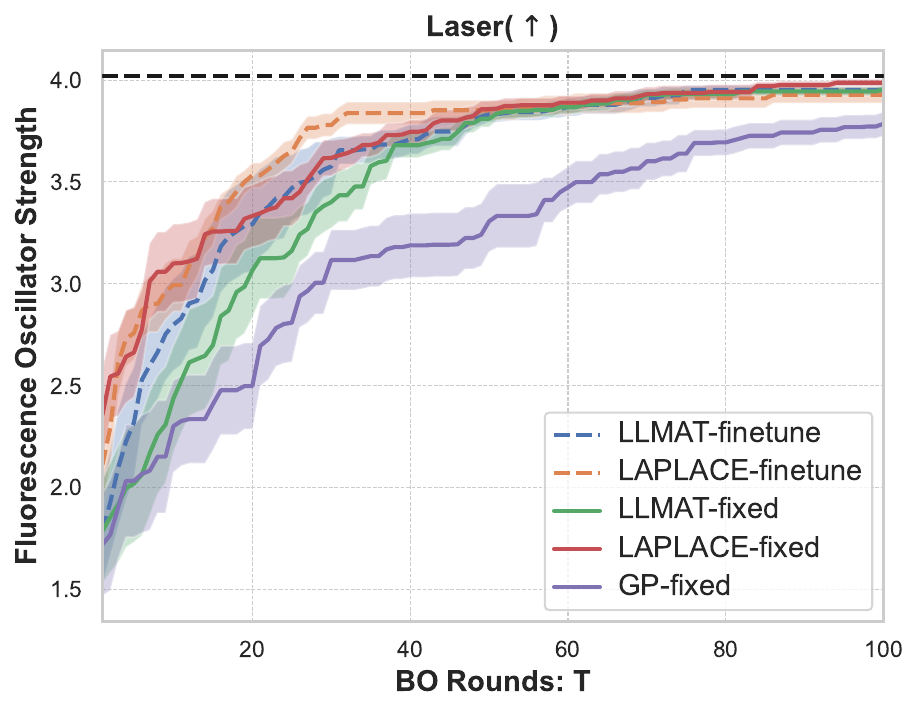}
     \includegraphics[width=0.32\linewidth]{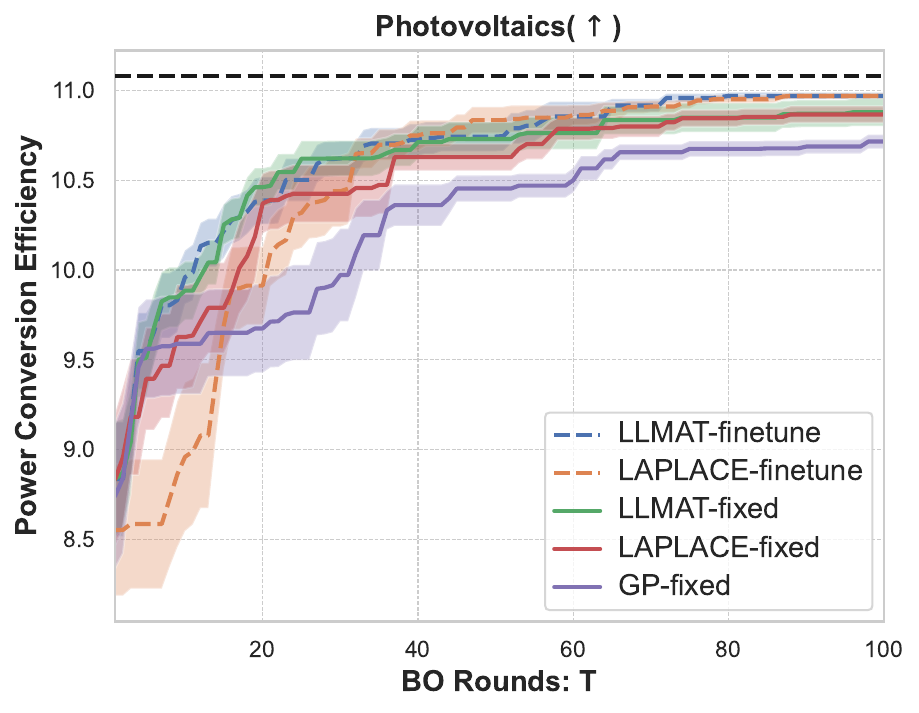}
    \includegraphics[width=0.32\linewidth]{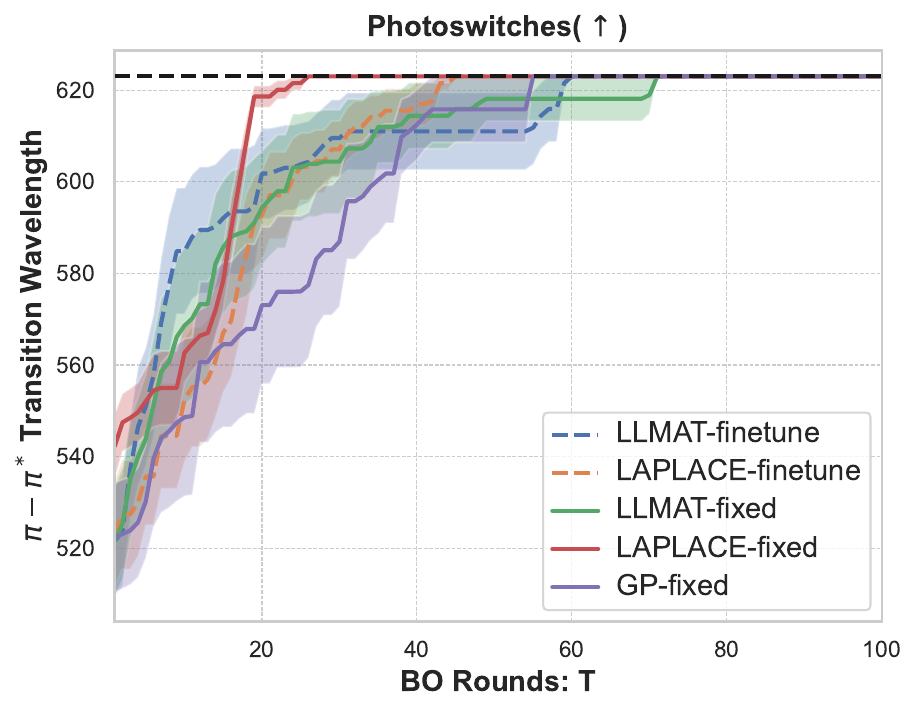}
    \caption{Historical optimums comparison on different chemistry datasets using Molformer.}
    \label{fig:molformer-finetune-his}
\end{figure}

\subsubsection{Average regrets}
\begin{figure}[H]
    \centering
    \includegraphics[width=0.32\linewidth]{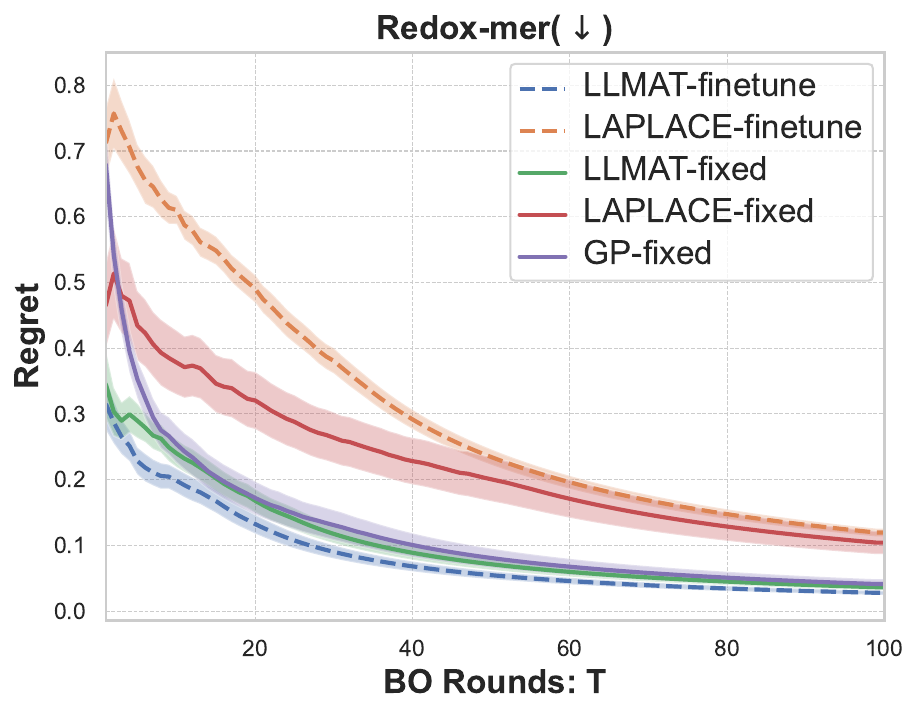}
    \includegraphics[width=0.32\linewidth]{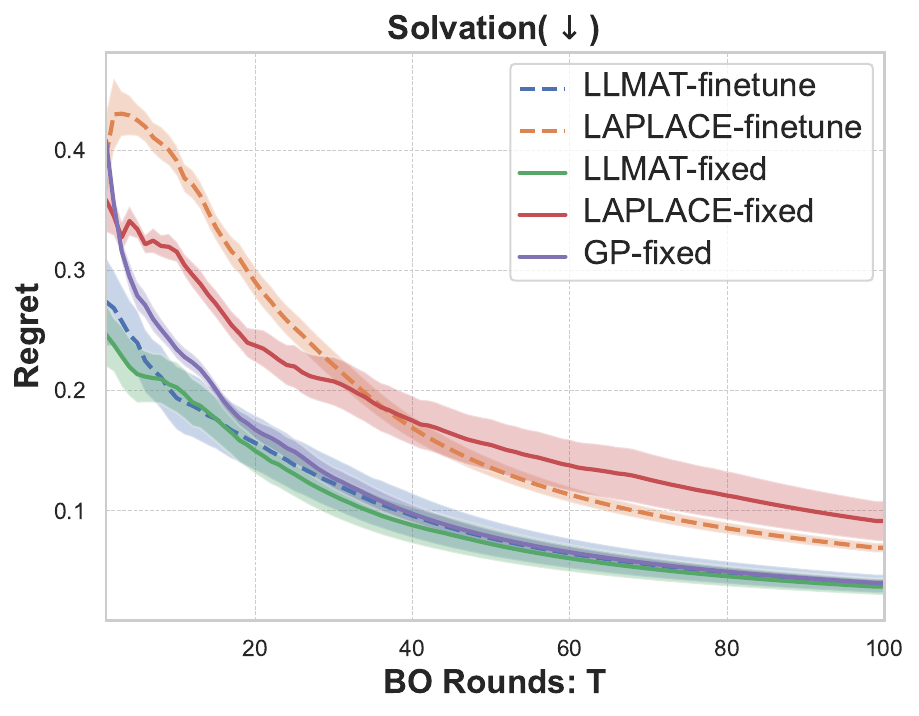}
    \includegraphics[width=0.32\linewidth]{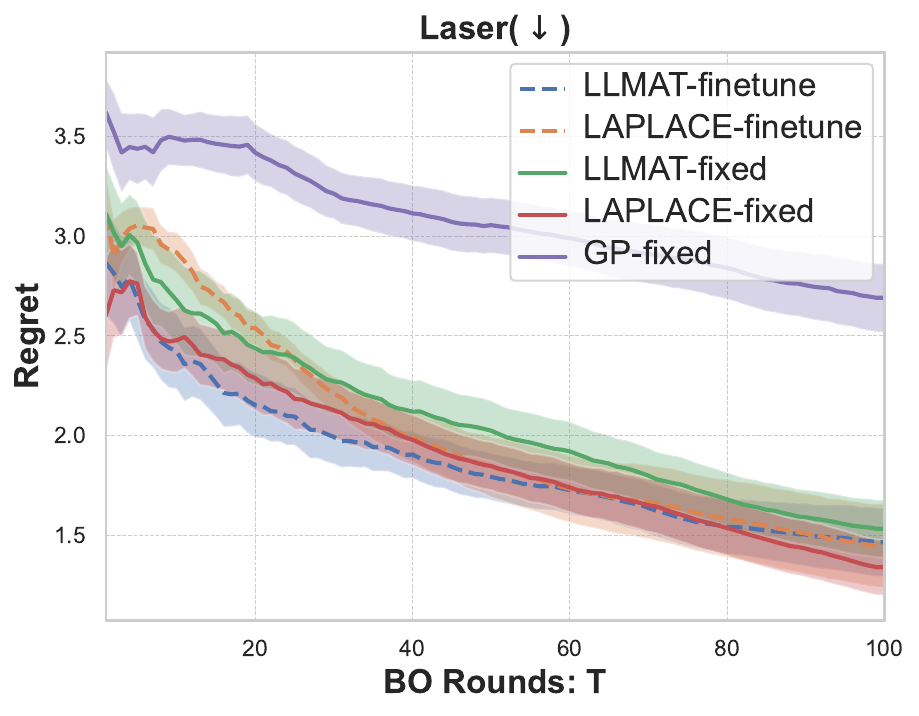}\\
     \includegraphics[width=0.32\linewidth]{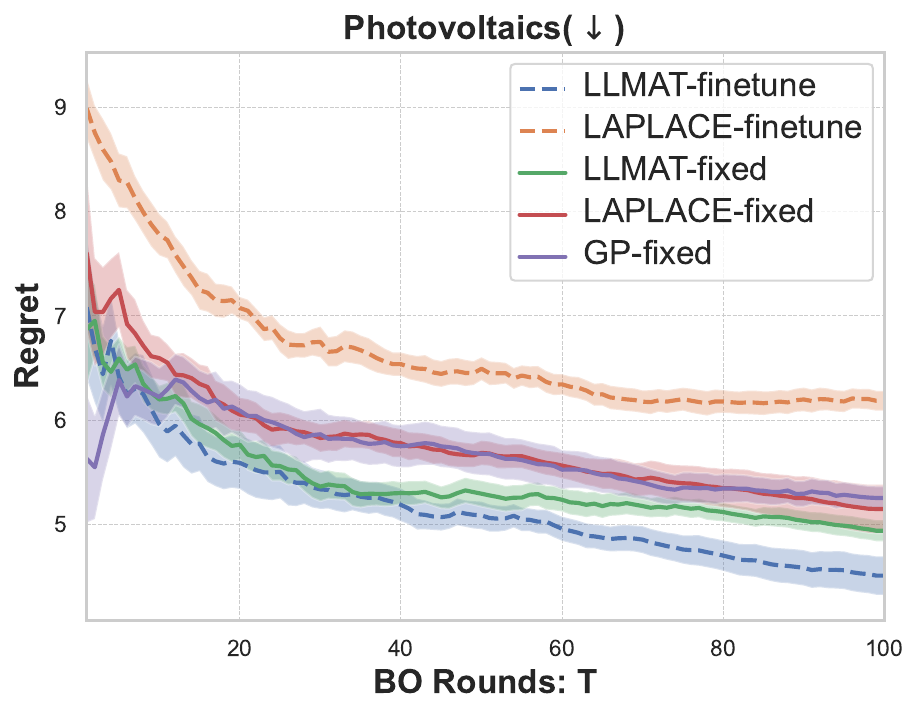}
    \includegraphics[width=0.32\linewidth]{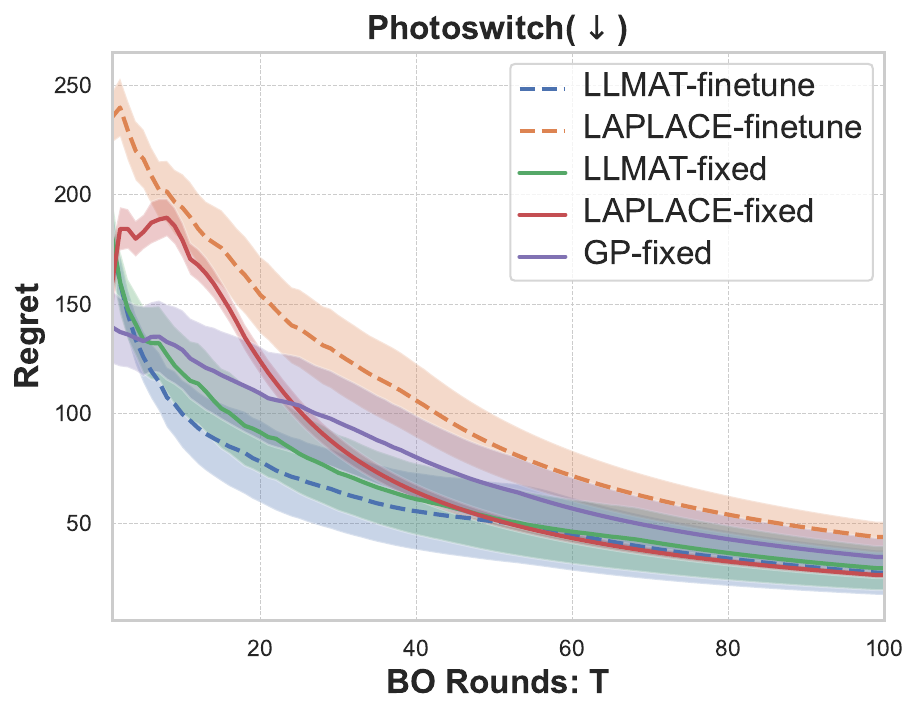}
    \includegraphics[width=0.32\linewidth]{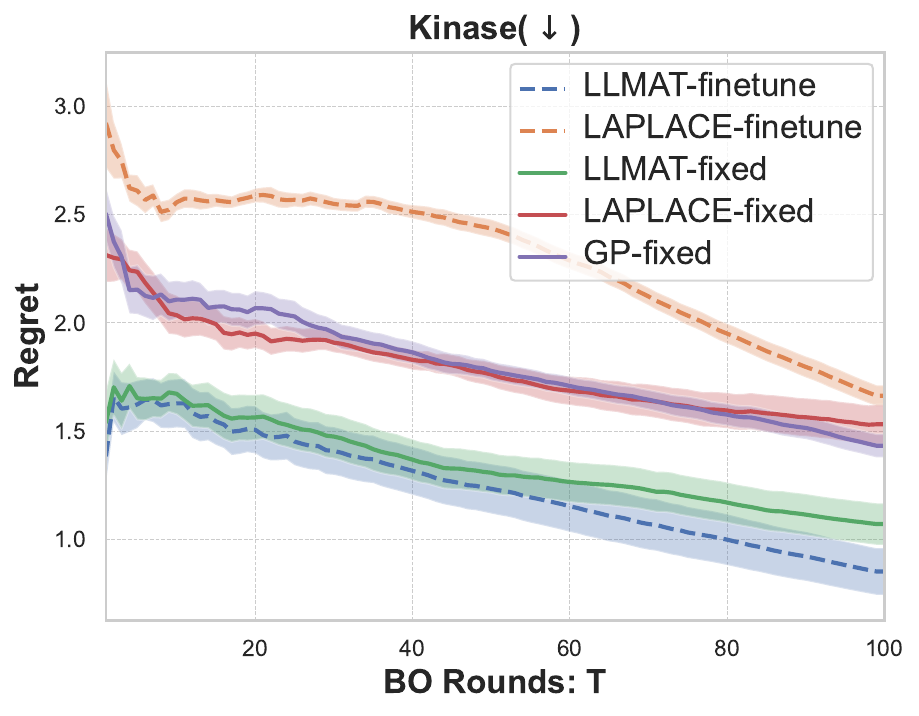}
    \caption{Regrets comparison on different chemistry datasets using Molformer. }
    \label{fig:molformer-regret}
\end{figure}

\subsubsection{GAPs}
\begin{figure}[H]
    \centering
    \includegraphics[width=0.32\linewidth]{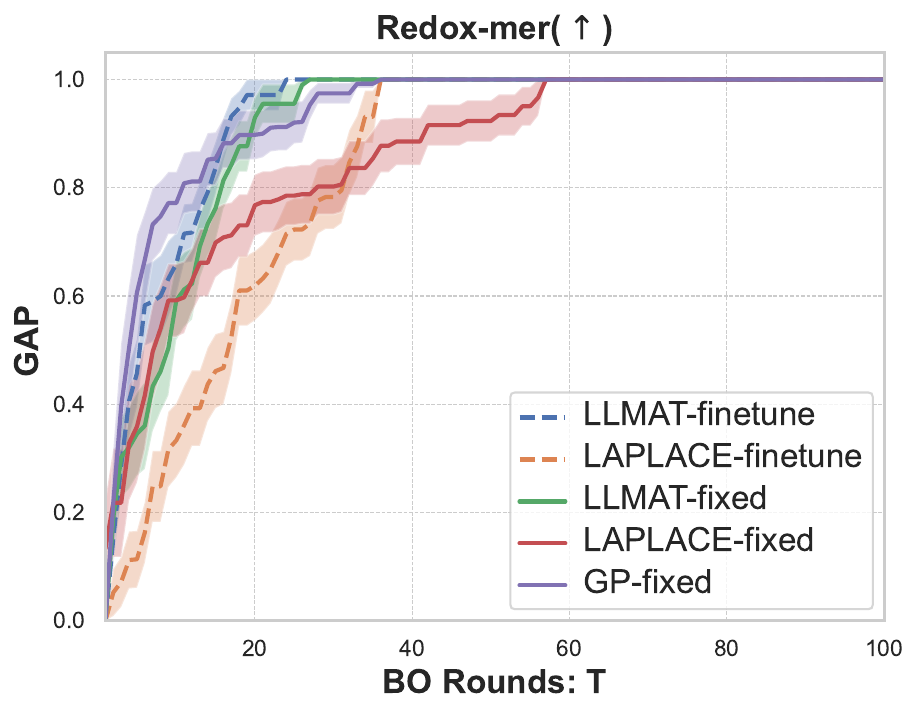}
    \includegraphics[width=0.32\linewidth]{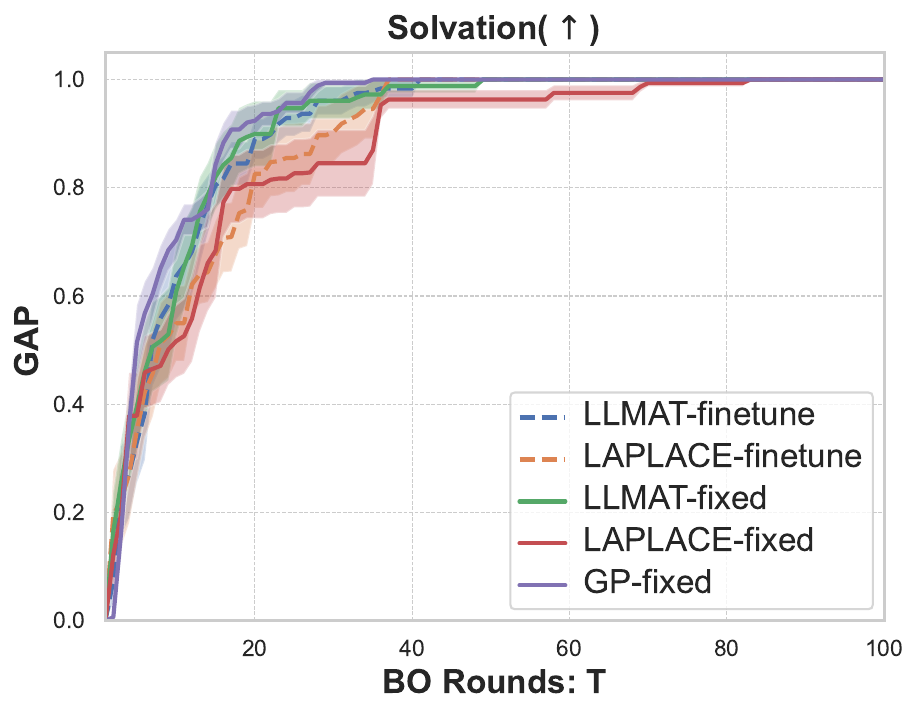}
    \includegraphics[width=0.32\linewidth]{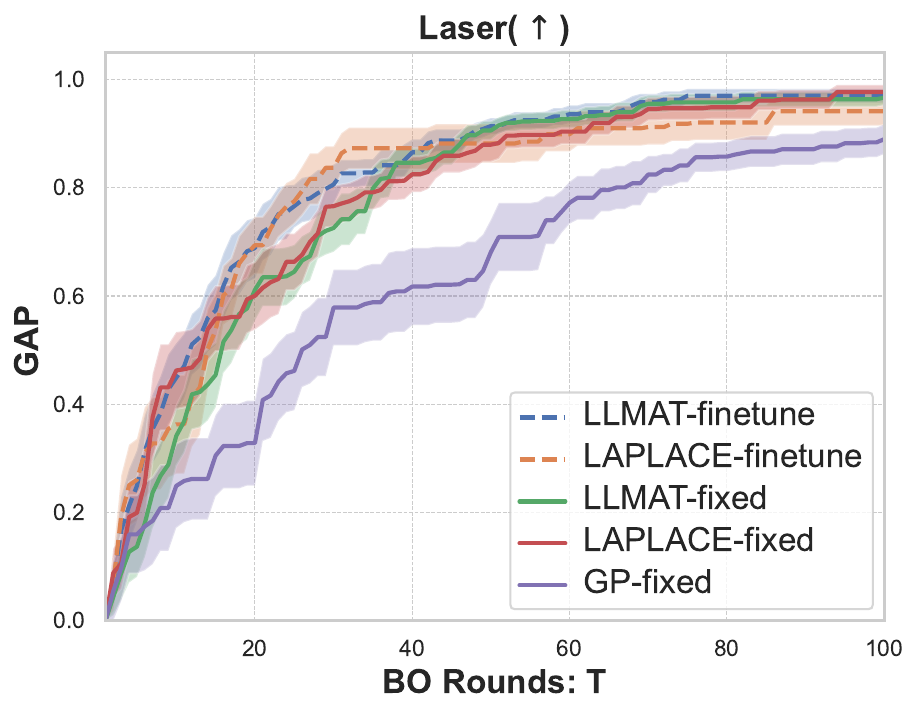}\\
     \includegraphics[width=0.32\linewidth]{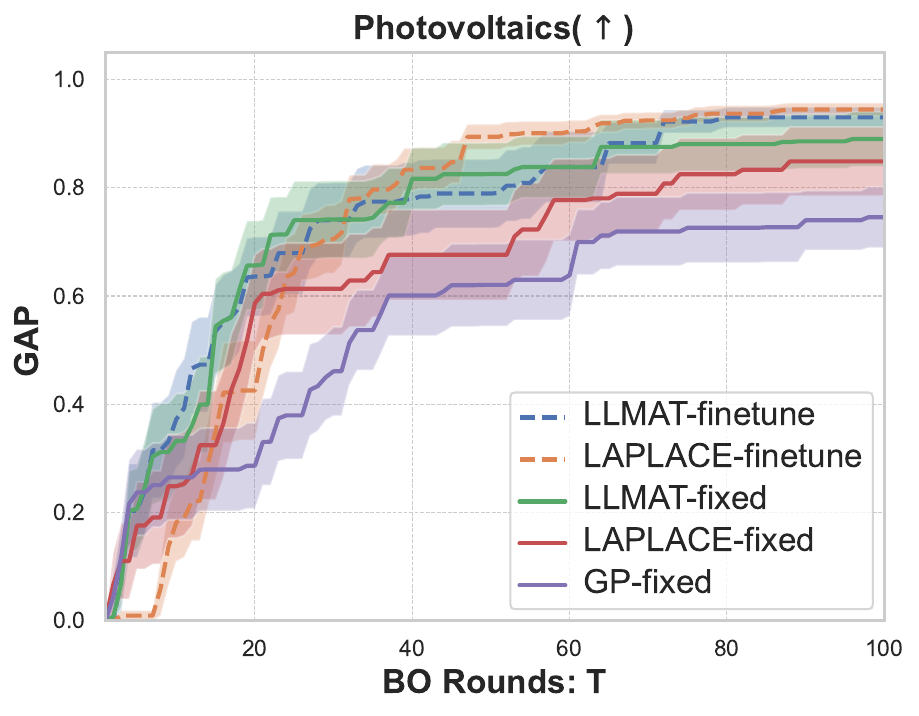}
    \includegraphics[width=0.32\linewidth]{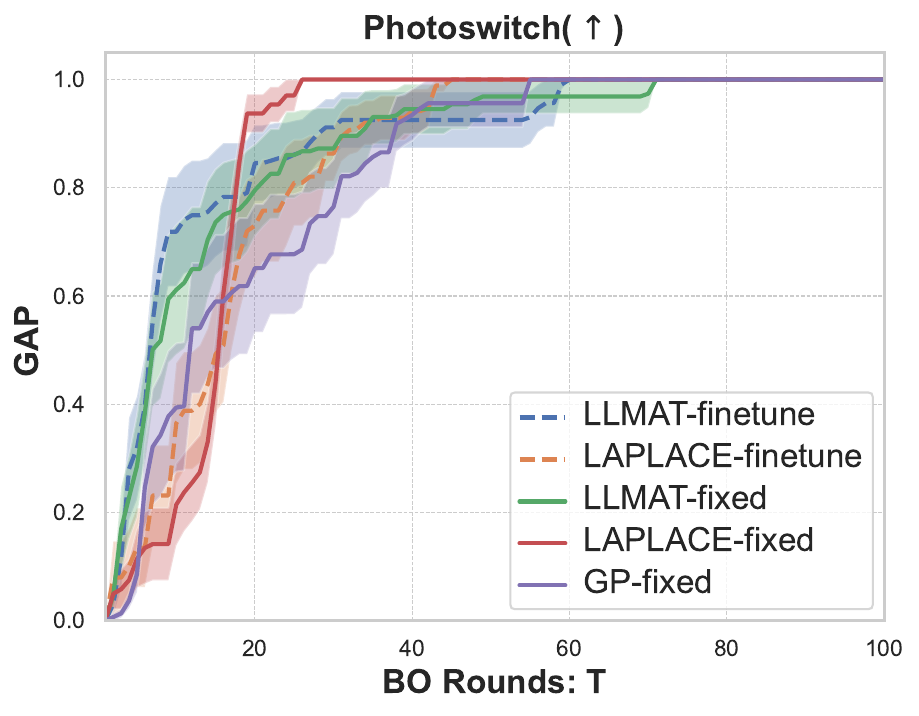}
    \includegraphics[width=0.32\linewidth]{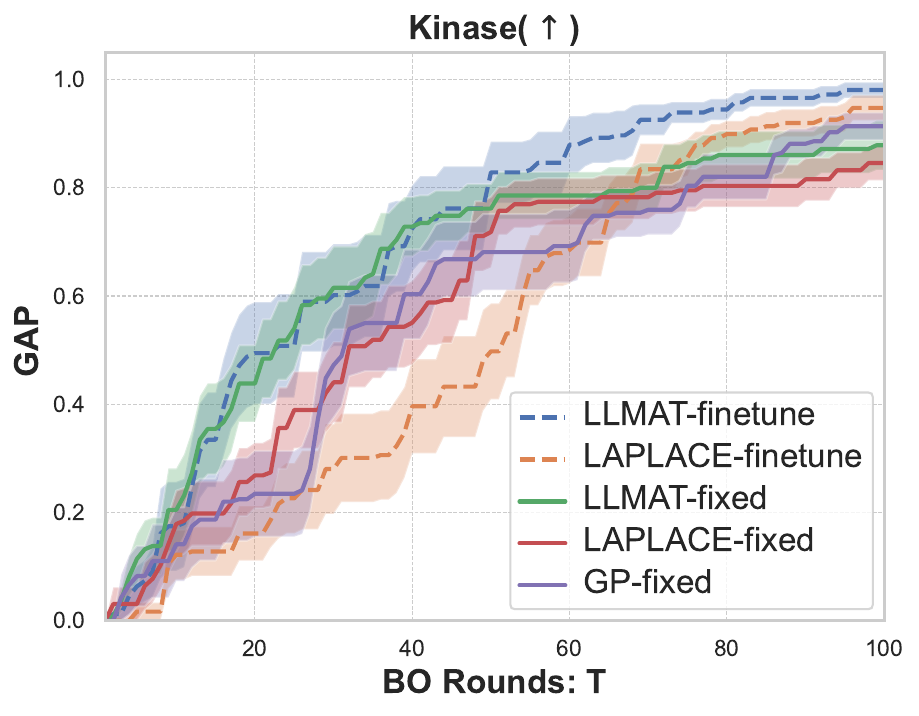}
    \caption{GAPs comparison on different chemistry datasets using Molformer}
    \label{fig:molformer-gap}
\end{figure}


\begin{figure}[H]
    \centering
    \includegraphics[width=0.4\linewidth]{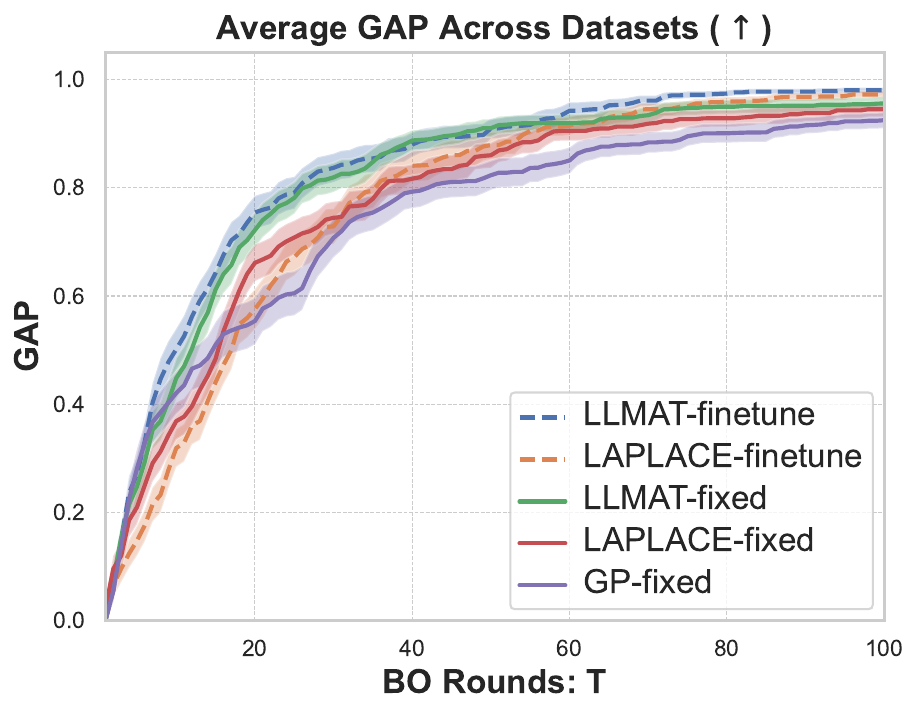}
    \includegraphics[width=0.4\linewidth]{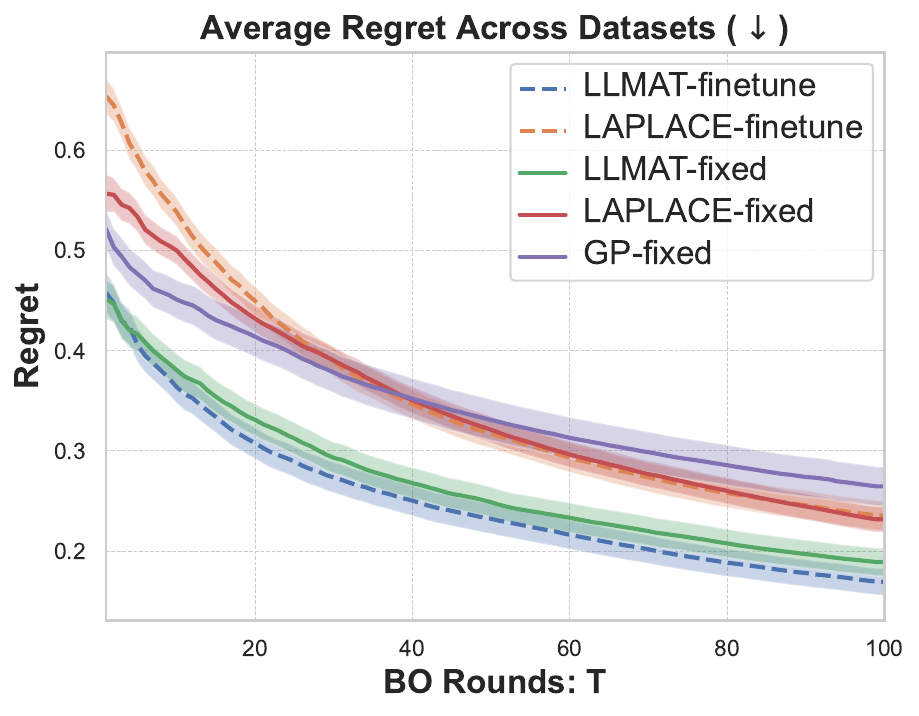}
    \caption{Average GAPs and regrets across all the datasets on finetuning and fixed Molformer. }
    \label{fig:molformer}
\end{figure}

\subsection{Clustering results}\label{Sec:clusters}

\subsubsection{K-means clustering}
\begin{figure}[H]
    \centering
    \includegraphics[width=0.25\linewidth]{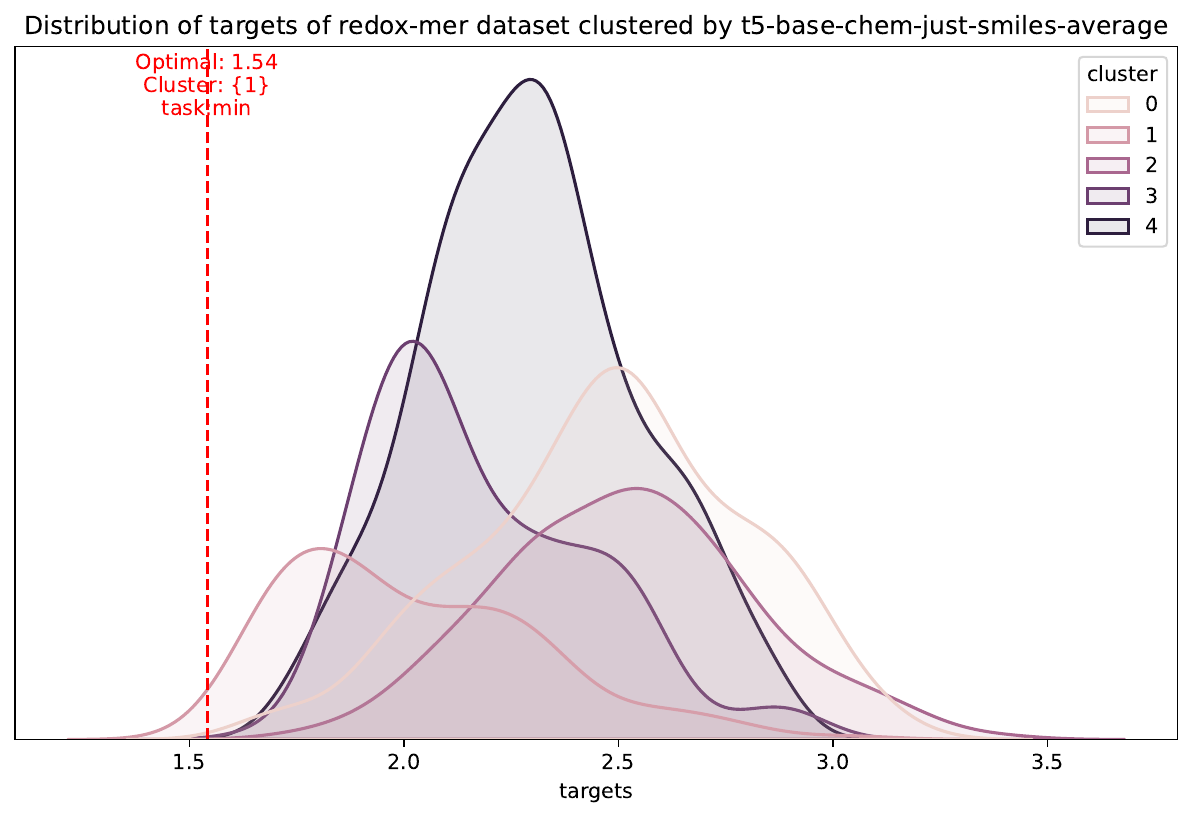}
     \includegraphics[width=0.25\linewidth]{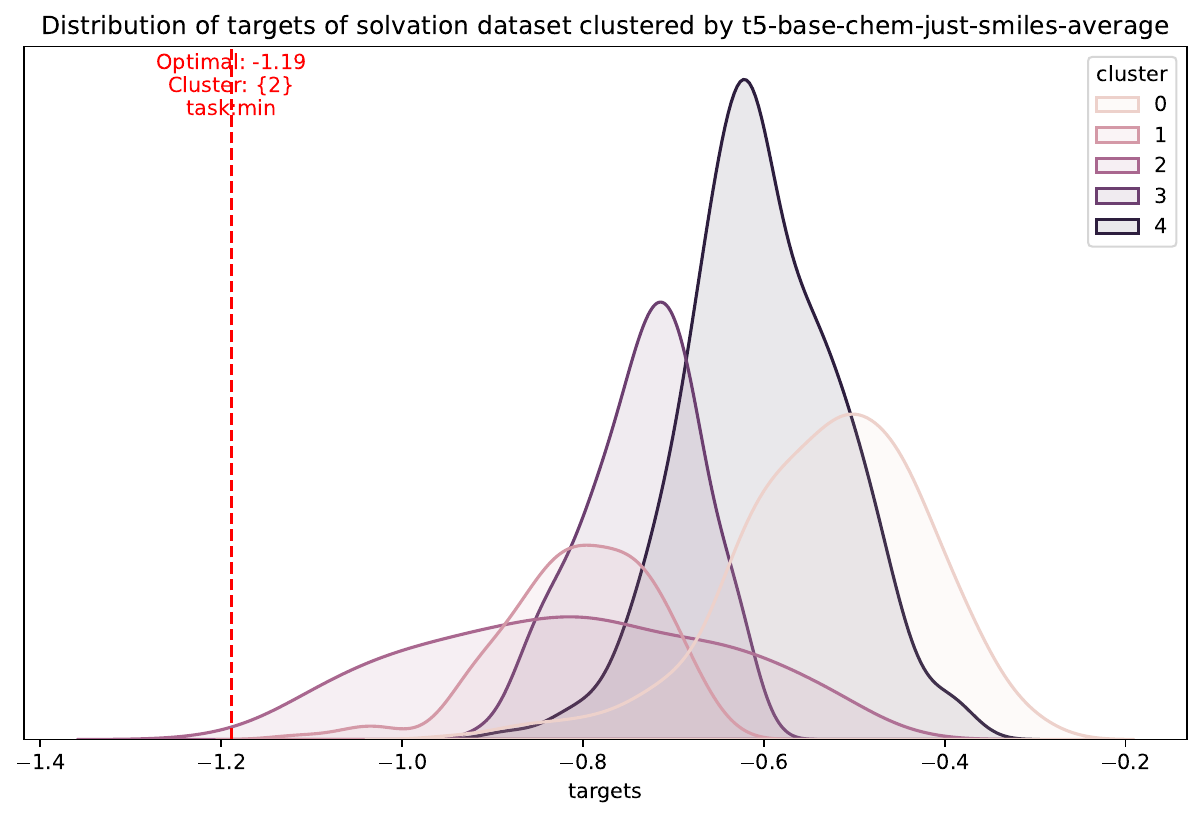}
     \includegraphics[width=0.25\linewidth]{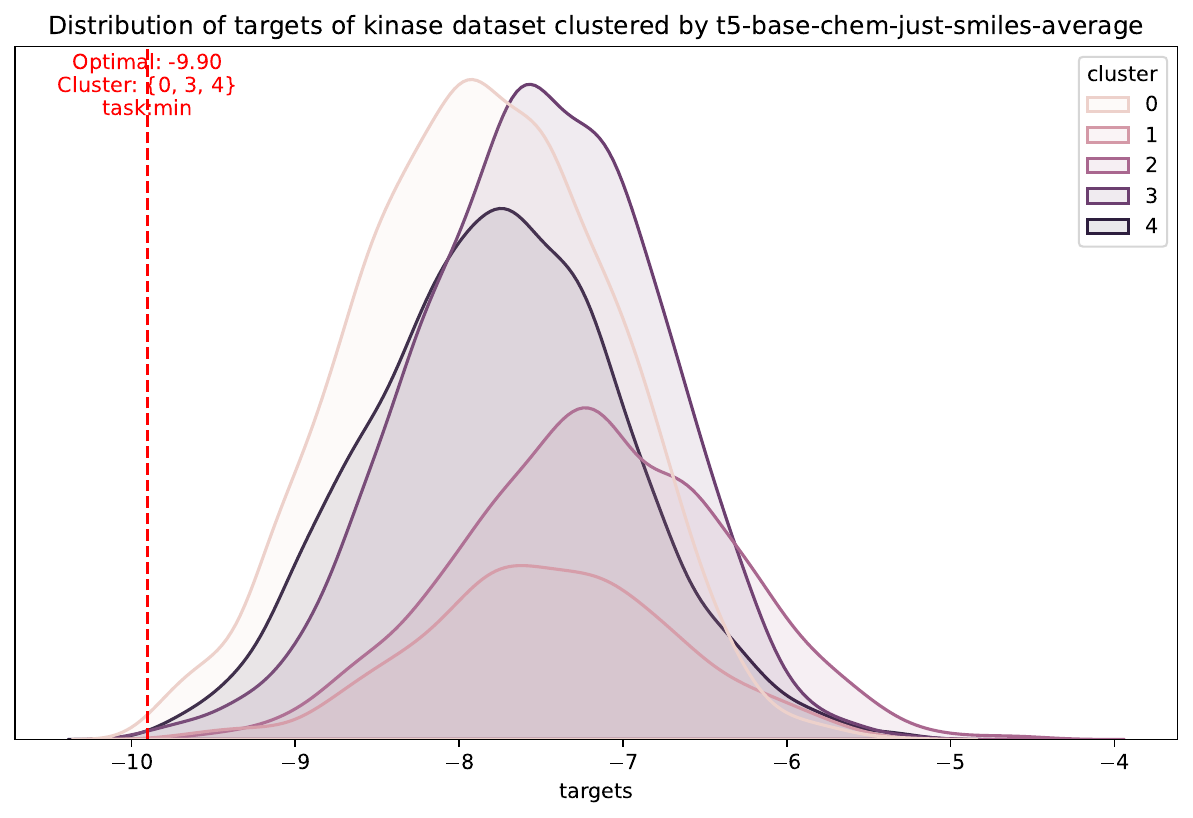}\\
     \includegraphics[width=0.25\linewidth]{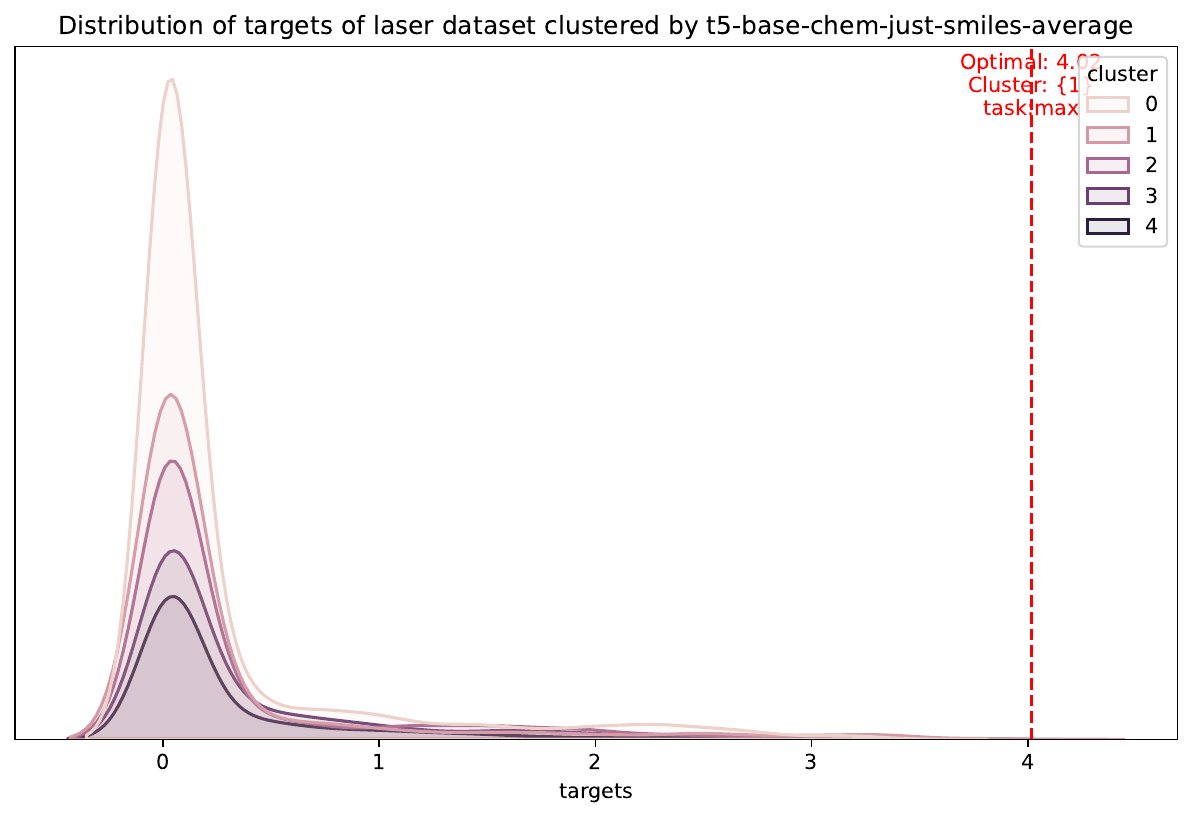}
     \includegraphics[width=0.25\linewidth]{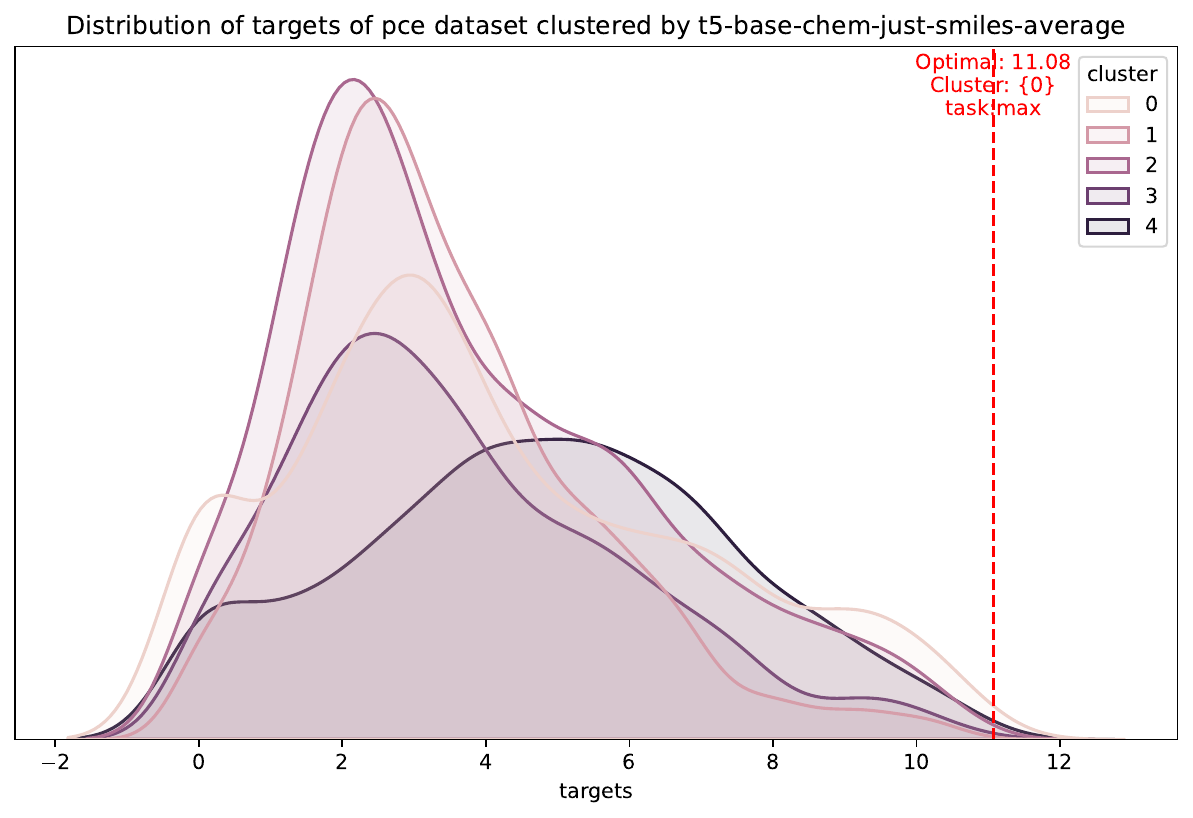}
     \includegraphics[width=0.25\linewidth]{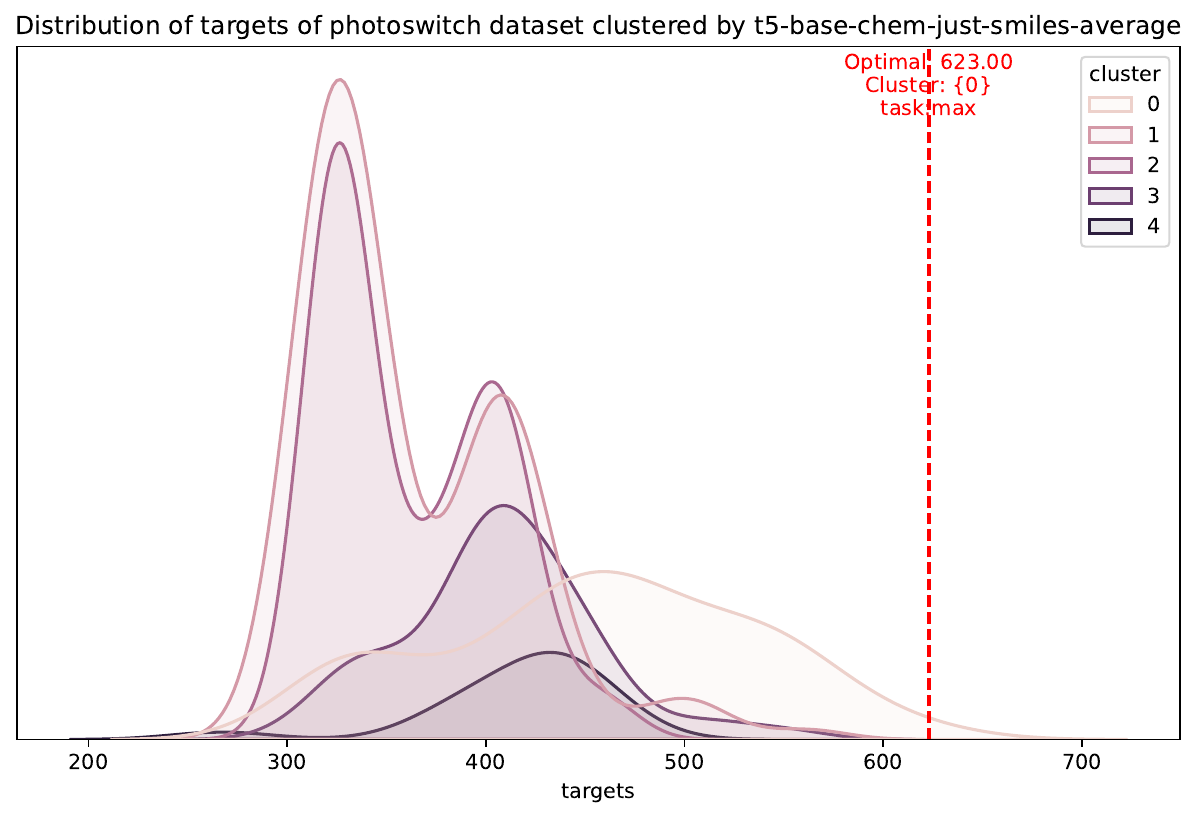}
    \caption{Distribution of property values for K-means clusters on features extracted from T5-Chem. }
    \label{fig:t5-chem_clustering_kmeans}
\end{figure}
\begin{figure}[H]
    \centering
    \includegraphics[width=0.25\linewidth]{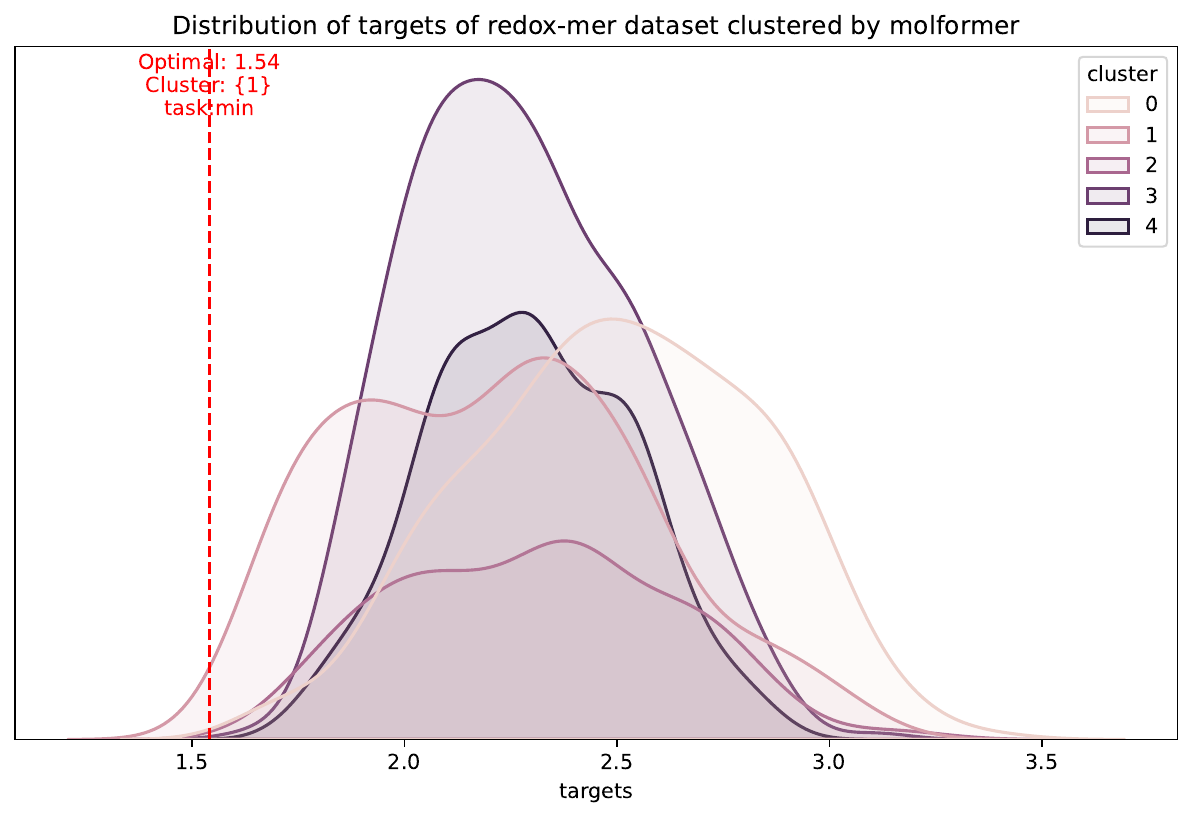}
     \includegraphics[width=0.25\linewidth]{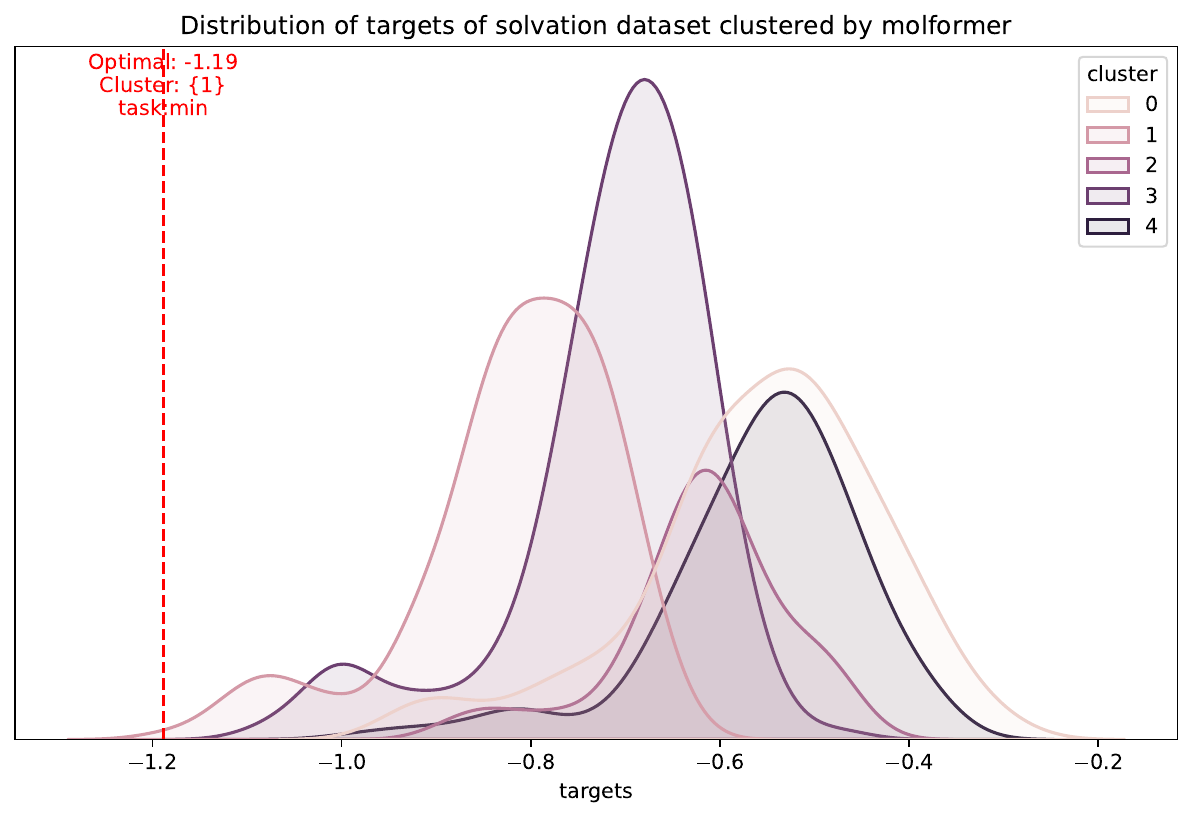}
     \includegraphics[width=0.25\linewidth]{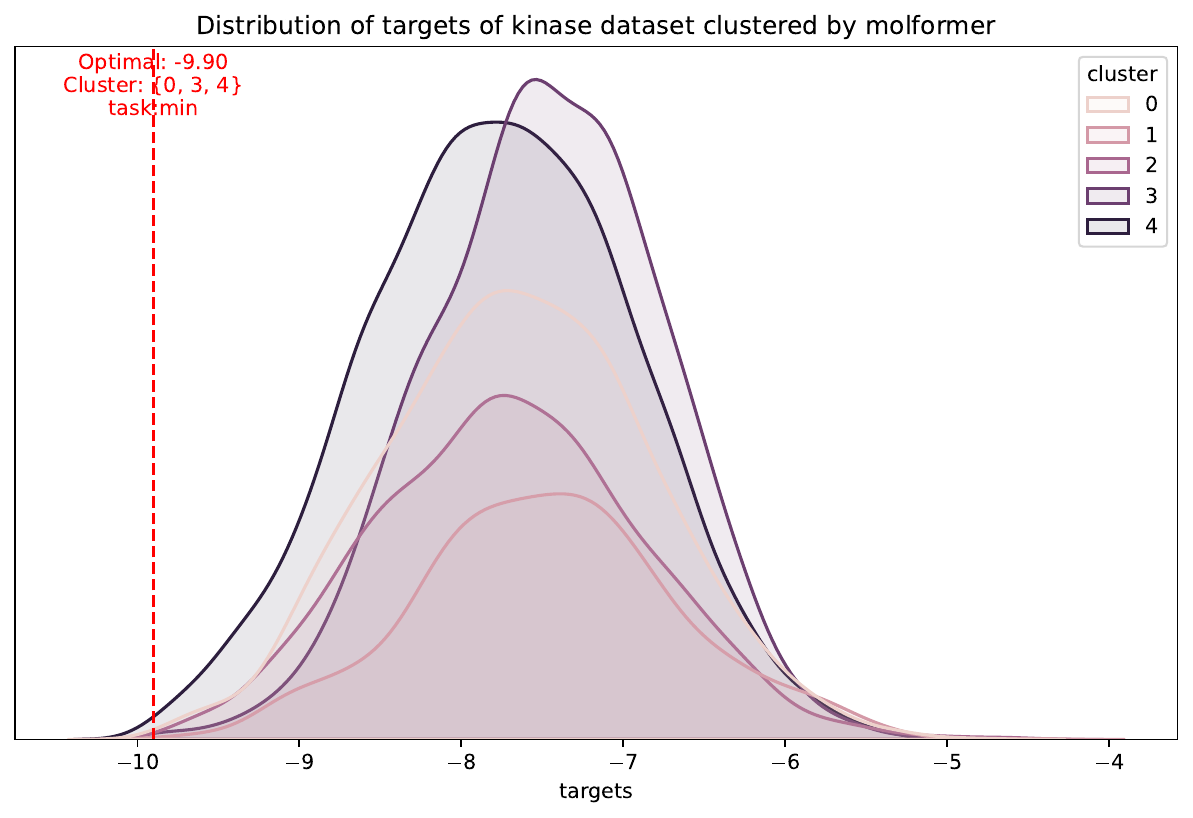}\\
     \includegraphics[width=0.25\linewidth]{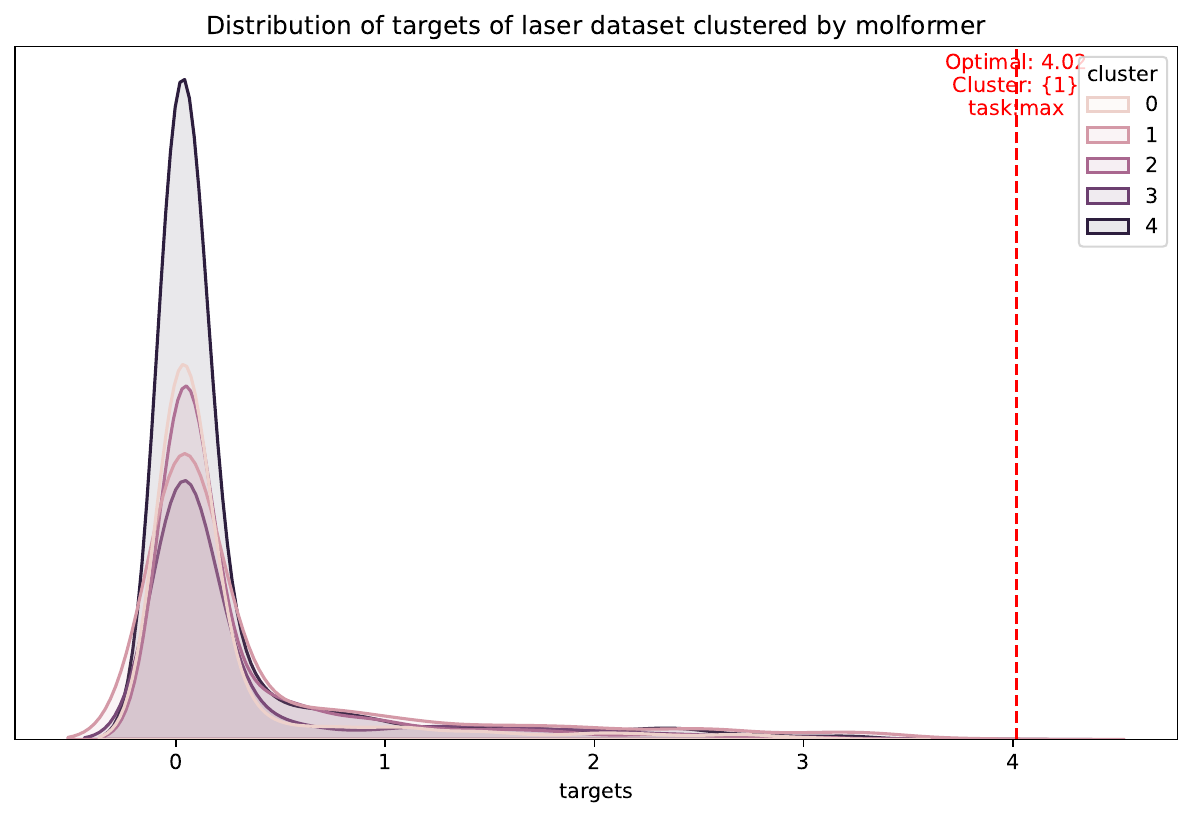}
     \includegraphics[width=0.25\linewidth]{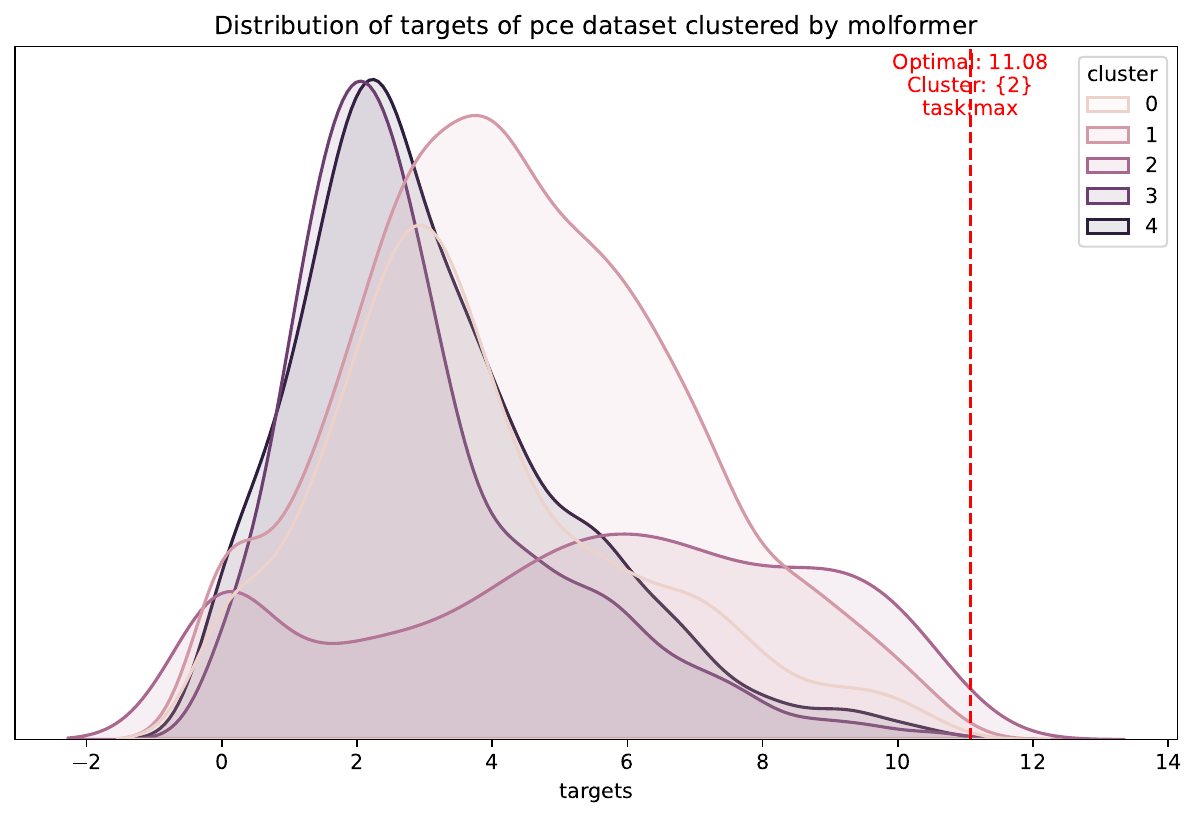}
     \includegraphics[width=0.25\linewidth]{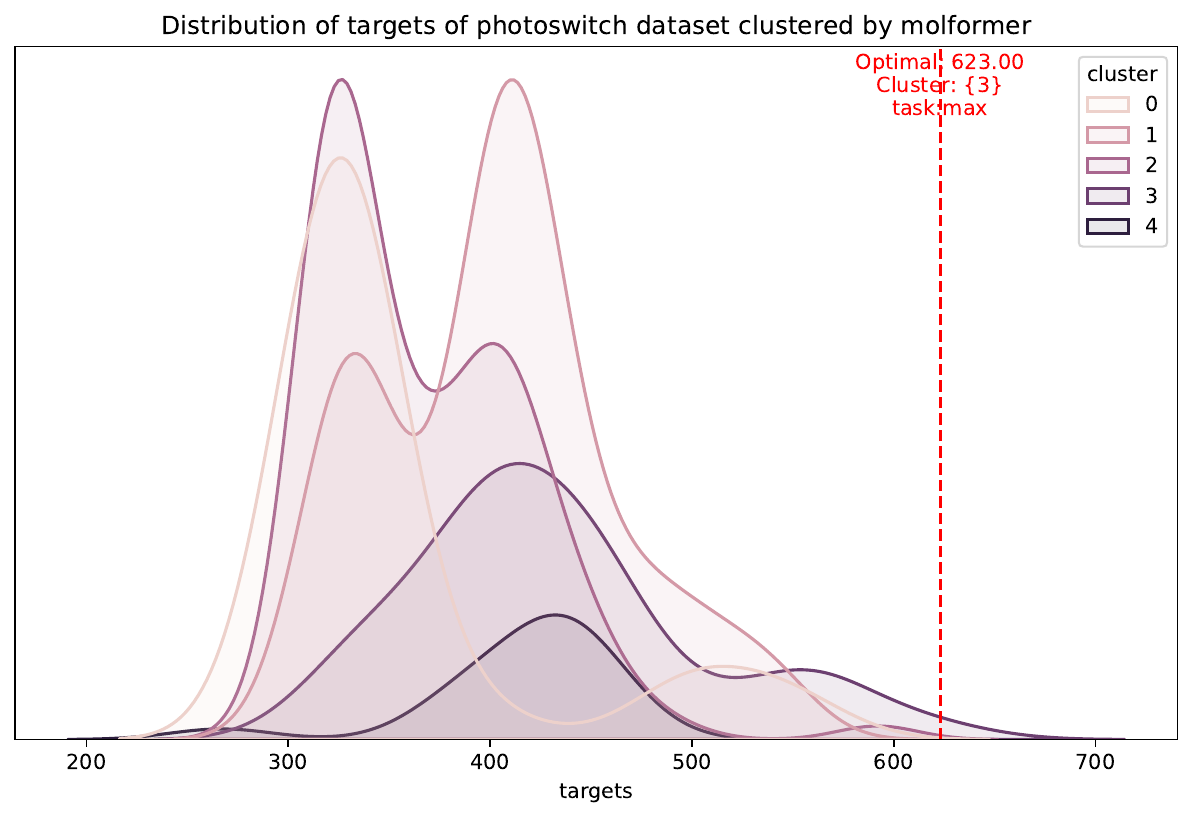}
    \caption{Distribution of property values for K-means clusters on features extracted from Molformer. }
    \label{fig:molformer_clustering_kmeans}
\end{figure}
\subsubsection{LLM-based clustering}
\begin{figure}[H]
    \centering
    \includegraphics[width=0.25\linewidth]{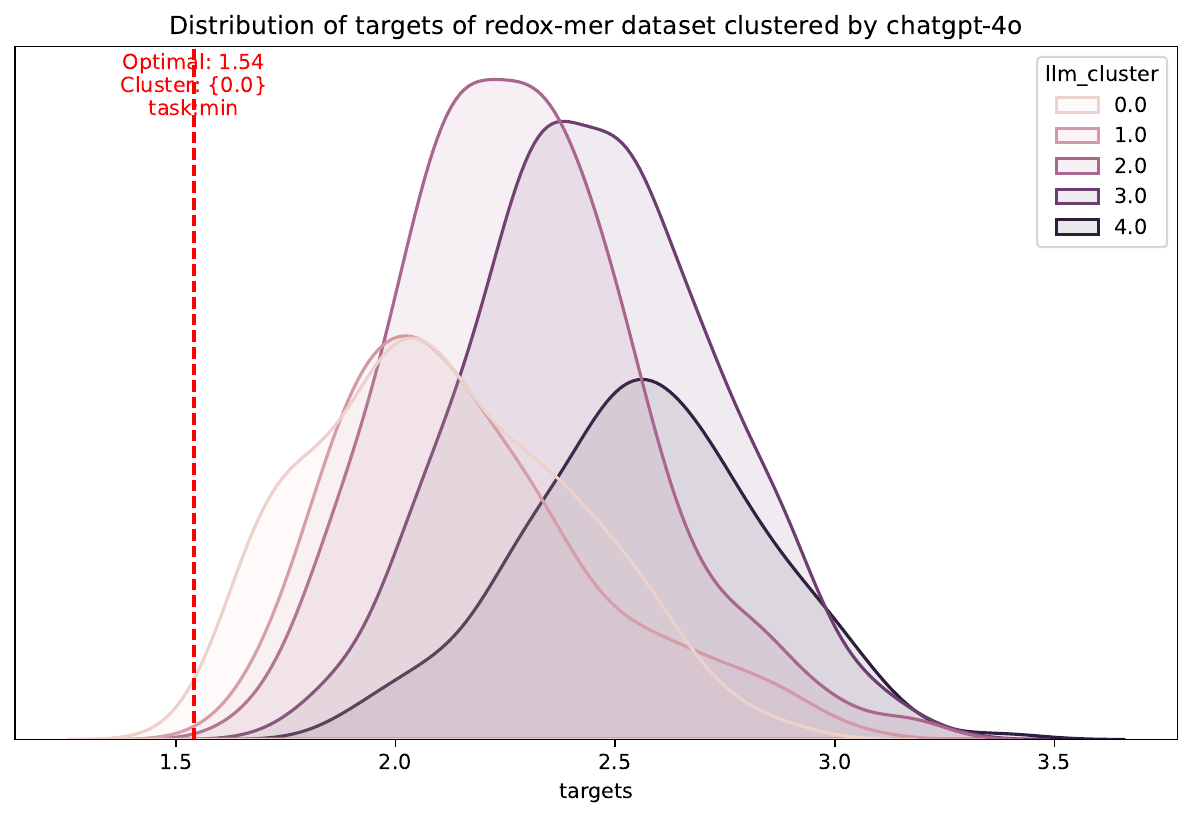}
     \includegraphics[width=0.25\linewidth]{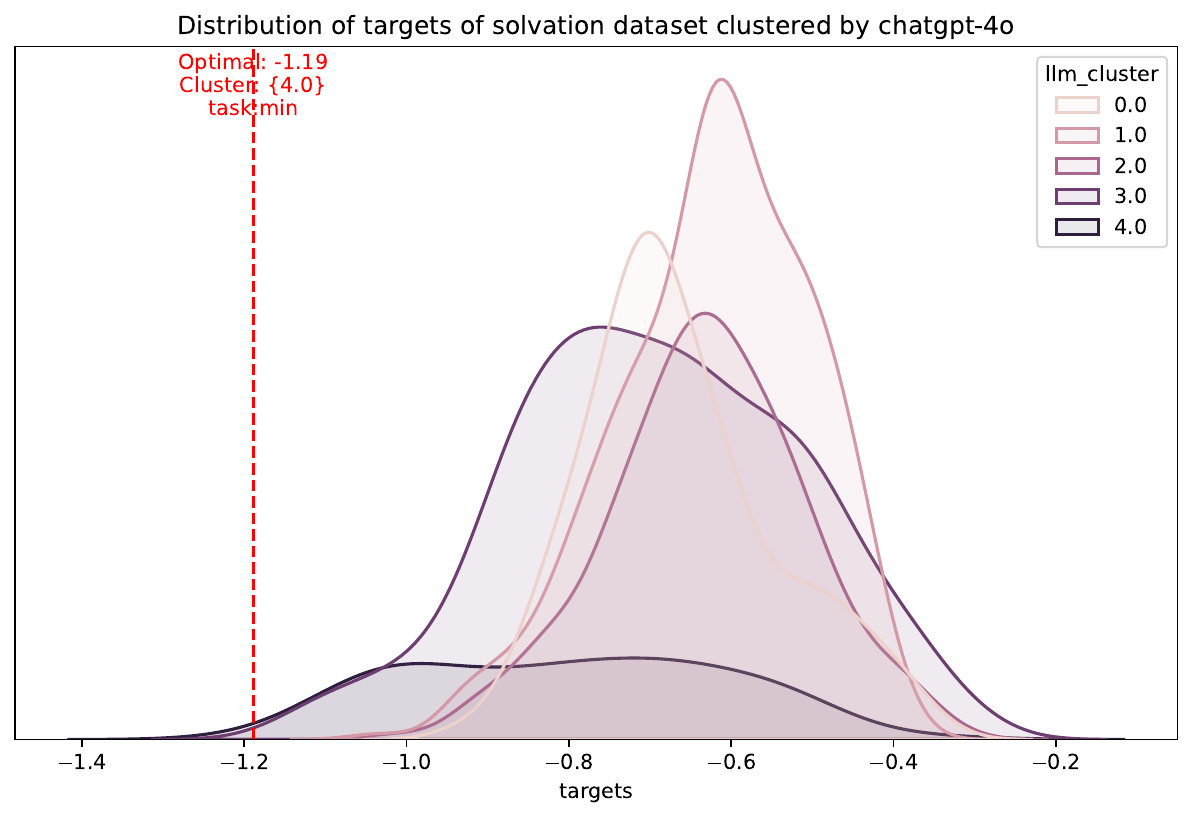}
     \includegraphics[width=0.25\linewidth]{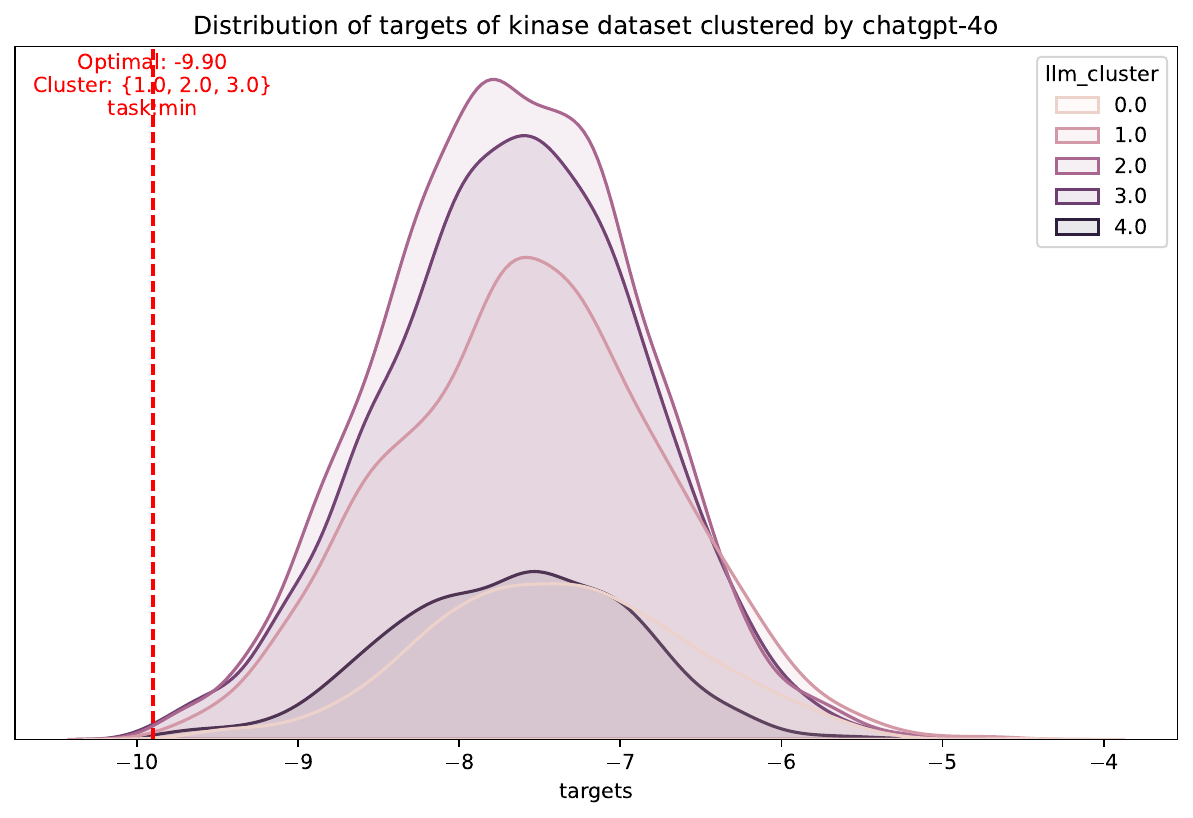}\\
     \includegraphics[width=0.25\linewidth]{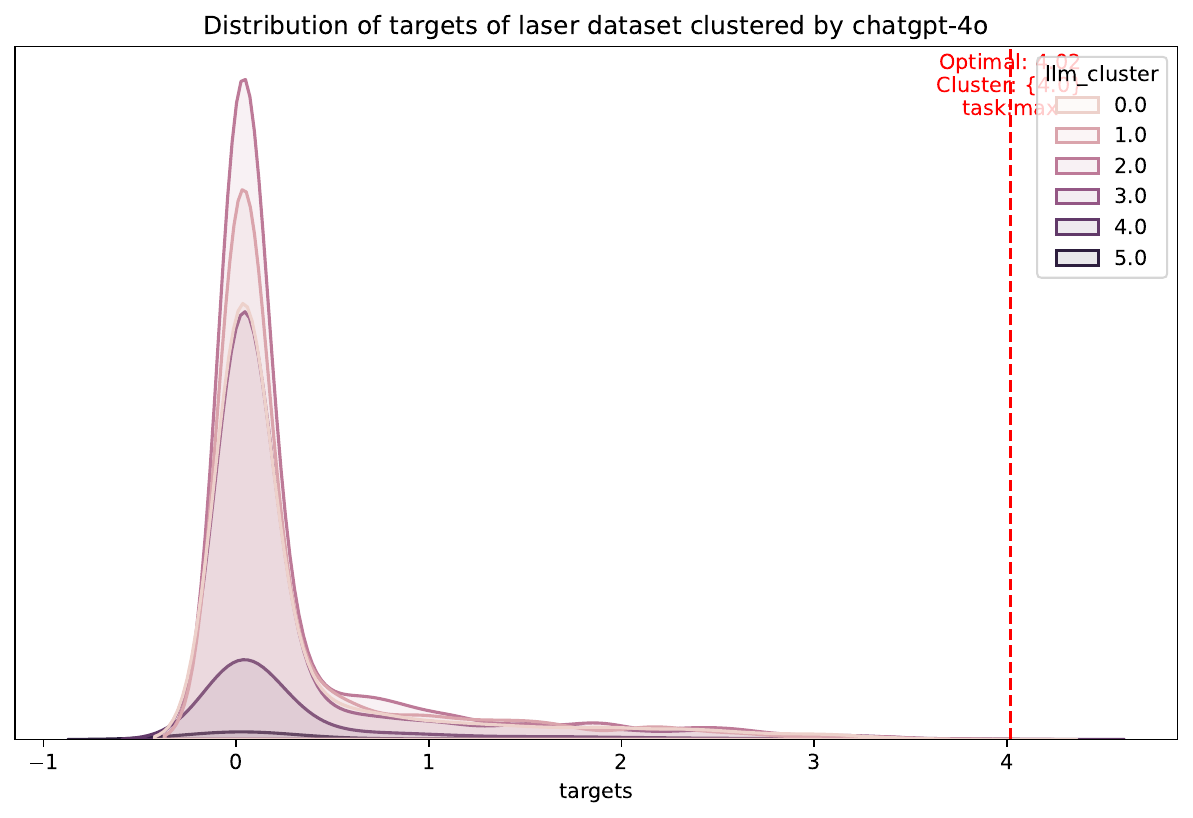}
     \includegraphics[width=0.25\linewidth]{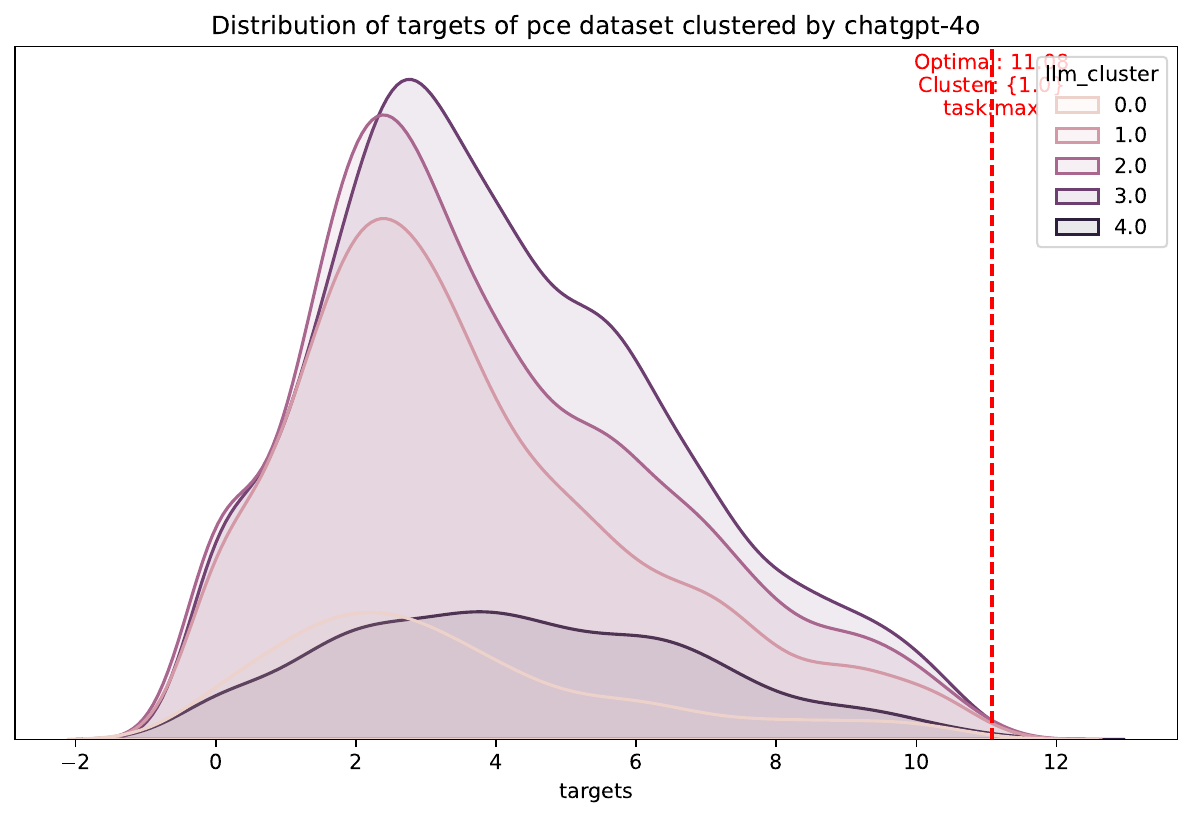}
     \includegraphics[width=0.25\linewidth]{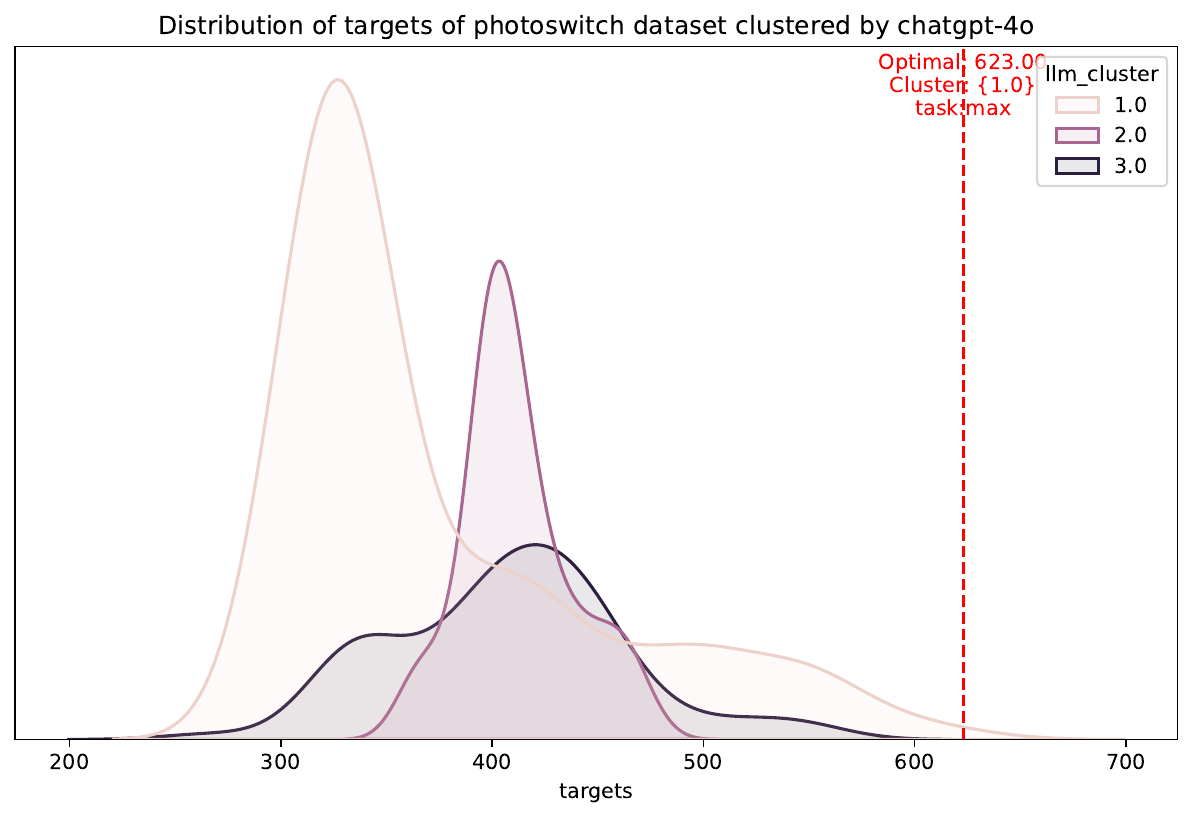}
    \caption{Distribution of property values for GPT-4o-based clustering results.}
    \label{fig:t5-chem_clustering_llms}
\end{figure}


\subsection{Ablation on p-values}\label{Sec:pval_ablation}

In this section, we provide additional details on how varying the p-values impacts our proposed algorithm in terms of the GAP metric, average regret, and prediction time per BO round for each dataset, using both K-means and LLM-based clustering. The results suggest that the LLM-based clustering approach is significantly effective in reducing prediction time while at the same time maintaining (within noise) the GAP and regret performance for Redoxmer and Solvation and improving it for Photovoltaics, reflecting that GPT-4o has informative knowledge on these datasets. This is consistent with its clustering visualization in Fig.~\ref{fig:t5-chem_clustering_llms}, where the mean property values for clusters have an obvious difference. K-means clustering is effective for Redoxmer and Solvation in reducing prediction time without degrading the performance.

\subsubsection{GAPs' change w.r.t different p-values}
\begin{figure}[H]
    \centering
    \includegraphics[width=0.25\linewidth]{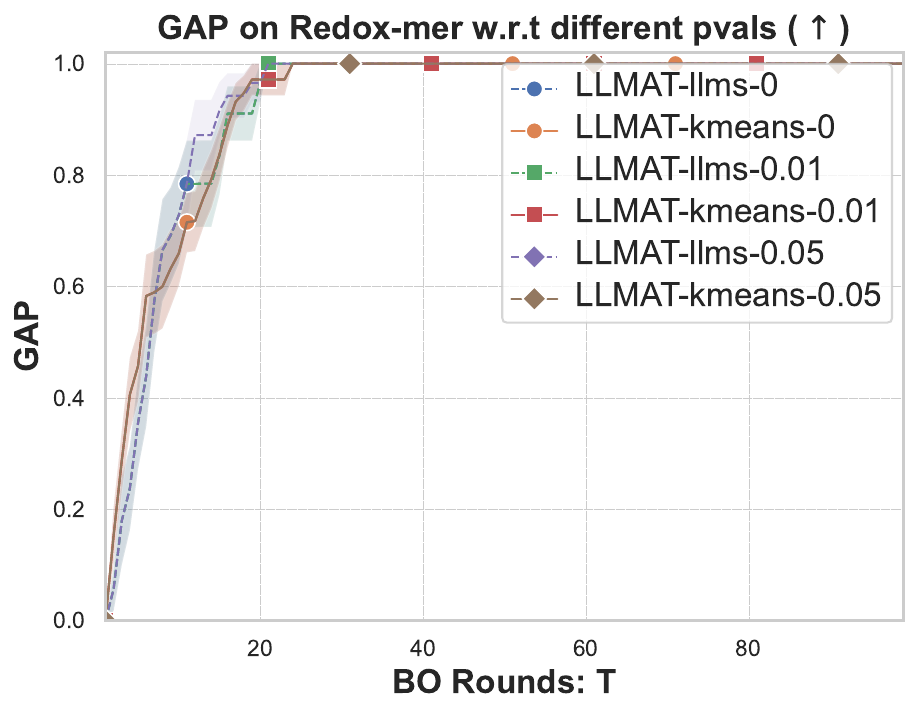}
    \includegraphics[width=0.25\linewidth]{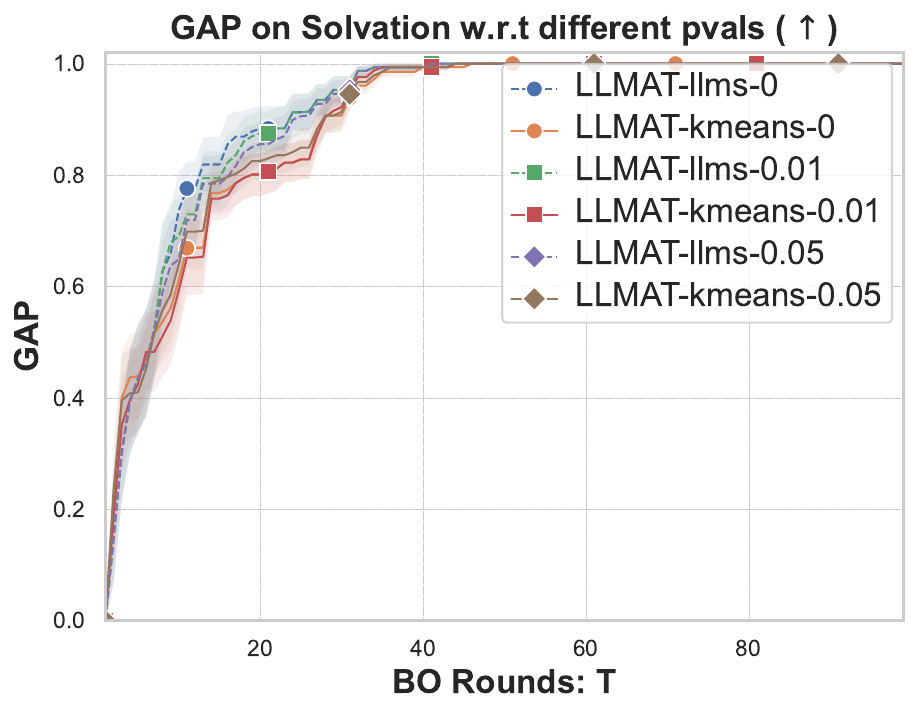}
    \includegraphics[width=0.25\linewidth]{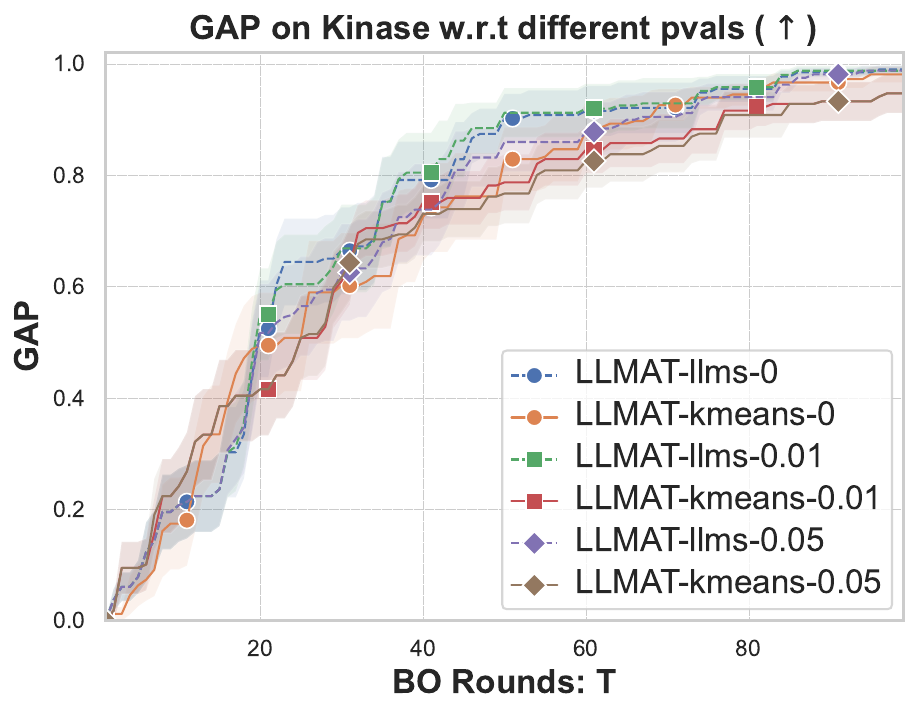}\\
     \includegraphics[width=0.25\linewidth]{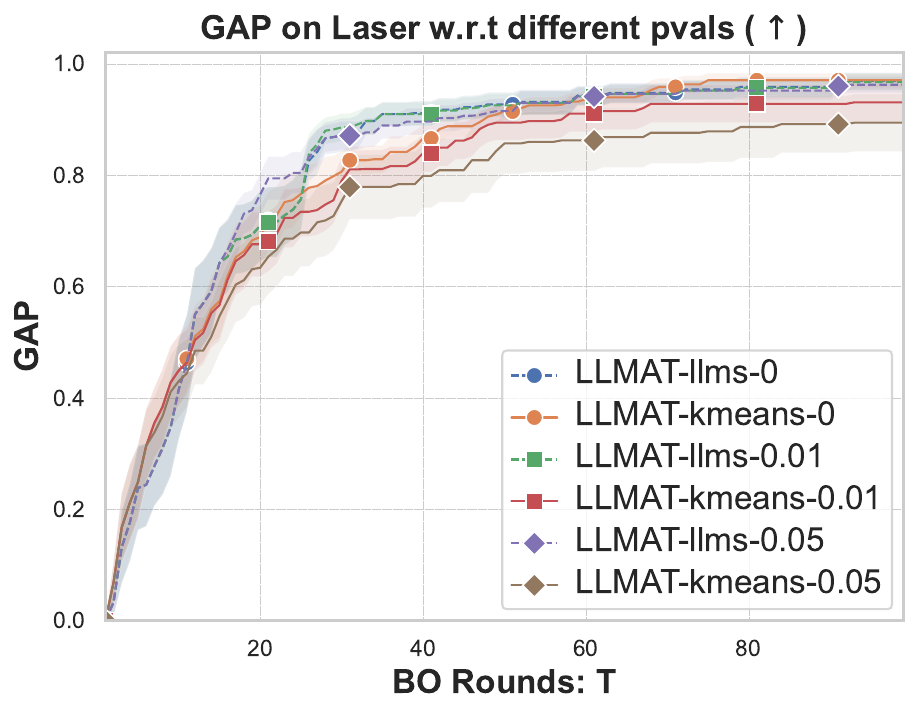}
    \includegraphics[width=0.25\linewidth]{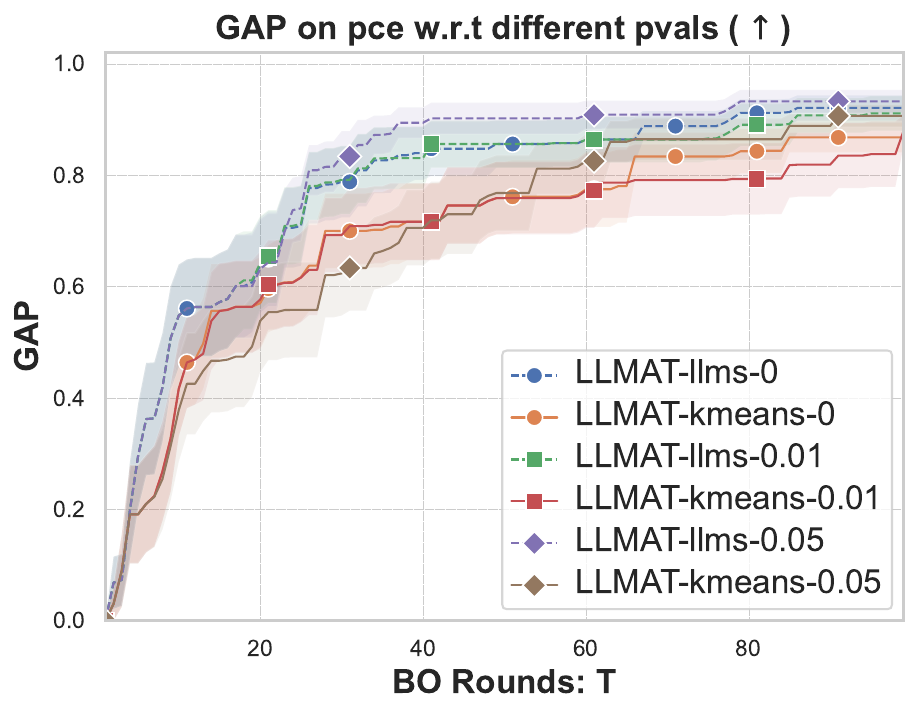}
    \includegraphics[width=0.25\linewidth]{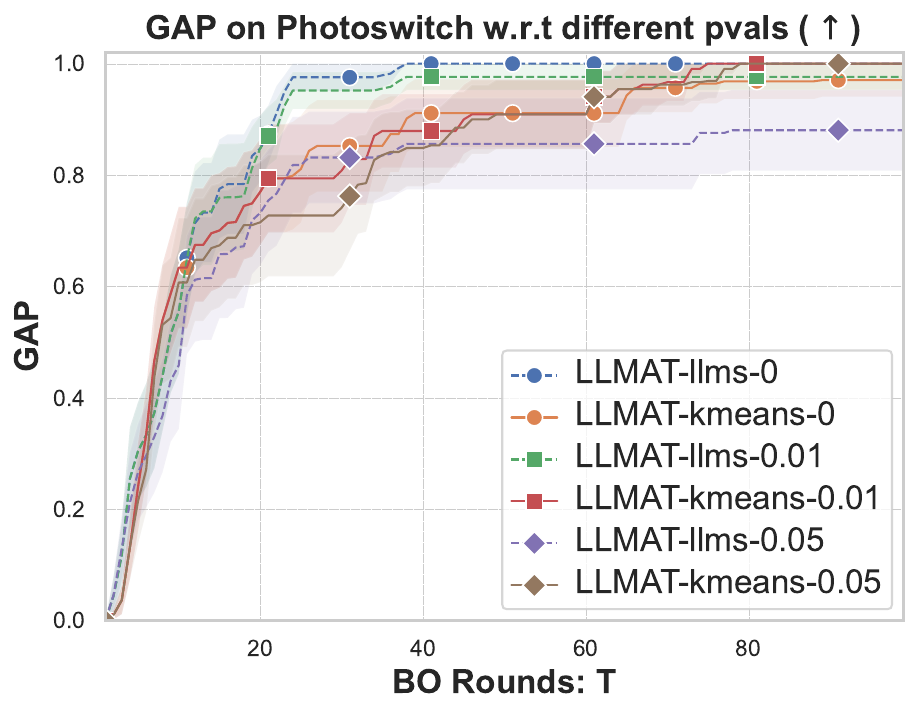}
    \caption{GAPs' change for different clustering approaches and p-values using Molformer. }
    \label{fig:molformer-gap-paval-ablation}
\end{figure}
\subsubsection{Regrets change w.r.t different p-values}
\begin{figure}[H]
    \centering
    \includegraphics[width=0.25\linewidth]{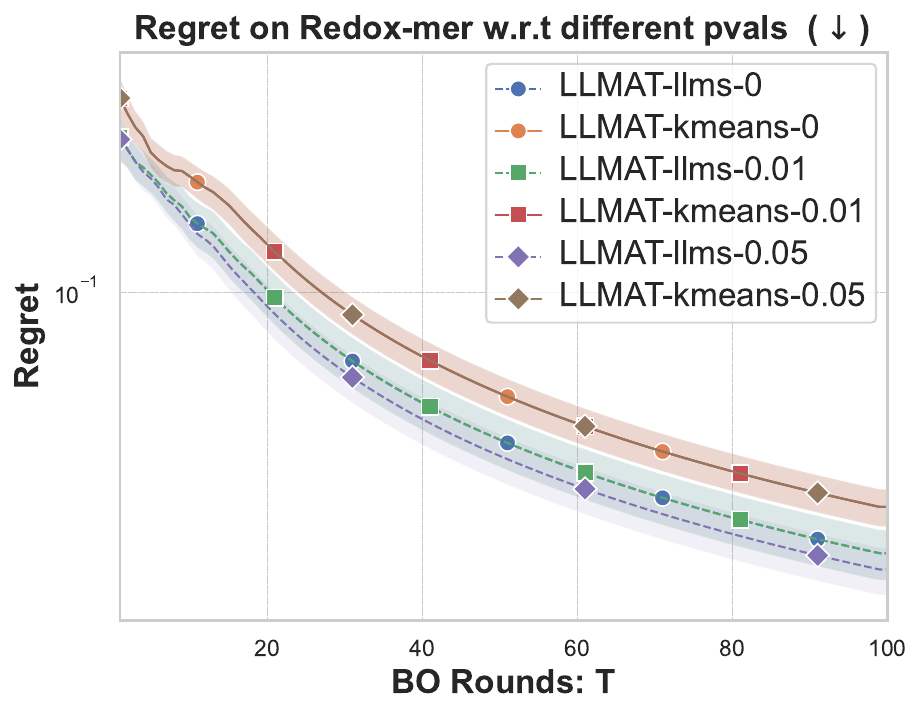}
    \includegraphics[width=0.25\linewidth]{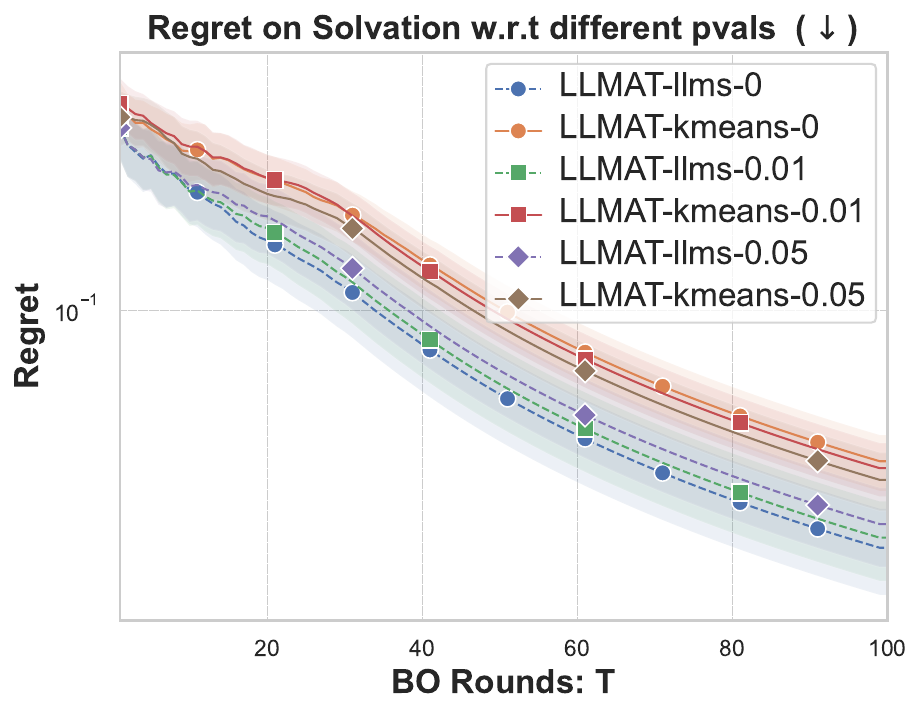}
    \includegraphics[width=0.25\linewidth]{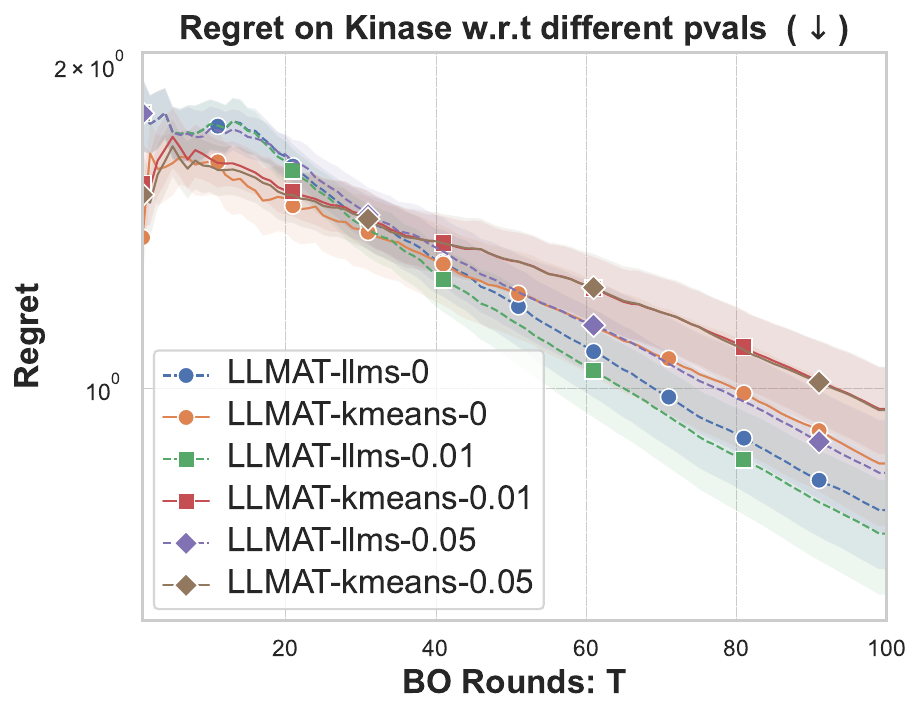}\\
     \includegraphics[width=0.25\linewidth]{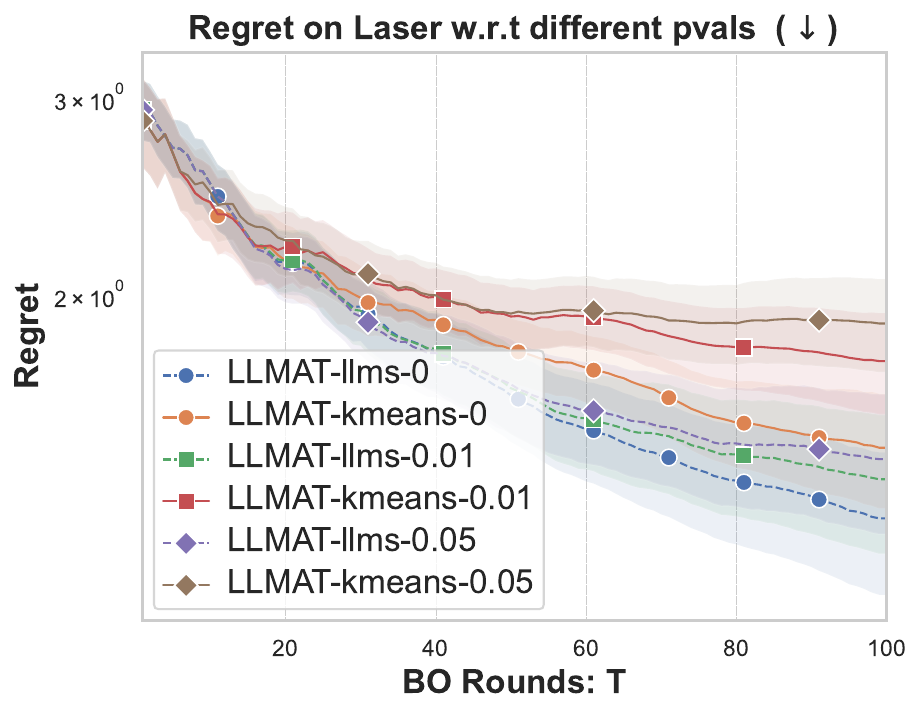}
    \includegraphics[width=0.25\linewidth]{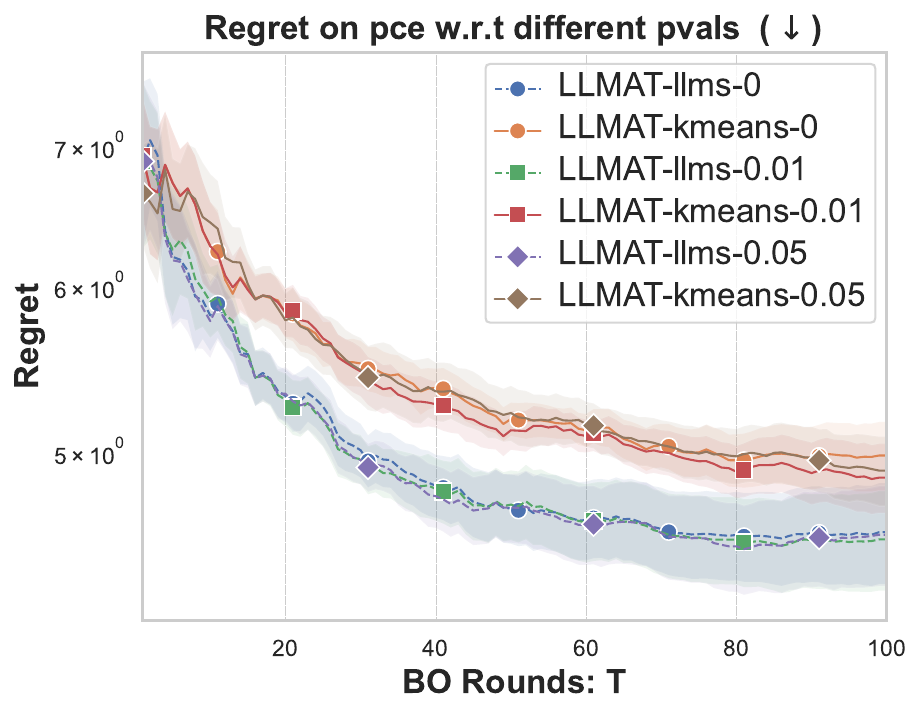}
    \includegraphics[width=0.25\linewidth]{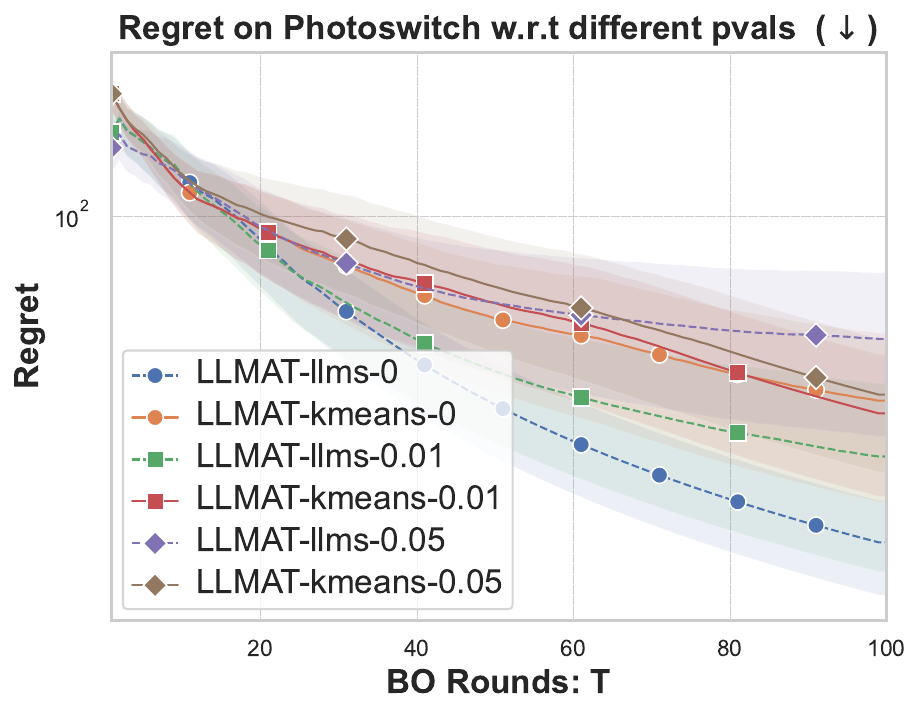}
    \caption{Regrets' change for different clustering approaches and p-values with Molformer. }
    \label{fig:molformer-regret-paval-ablation}
\end{figure}
\subsubsection{Effects on reducing prediction time w.r.t different p-values}
\begin{figure}[H]
    \centering
    \includegraphics[width=0.8\linewidth]{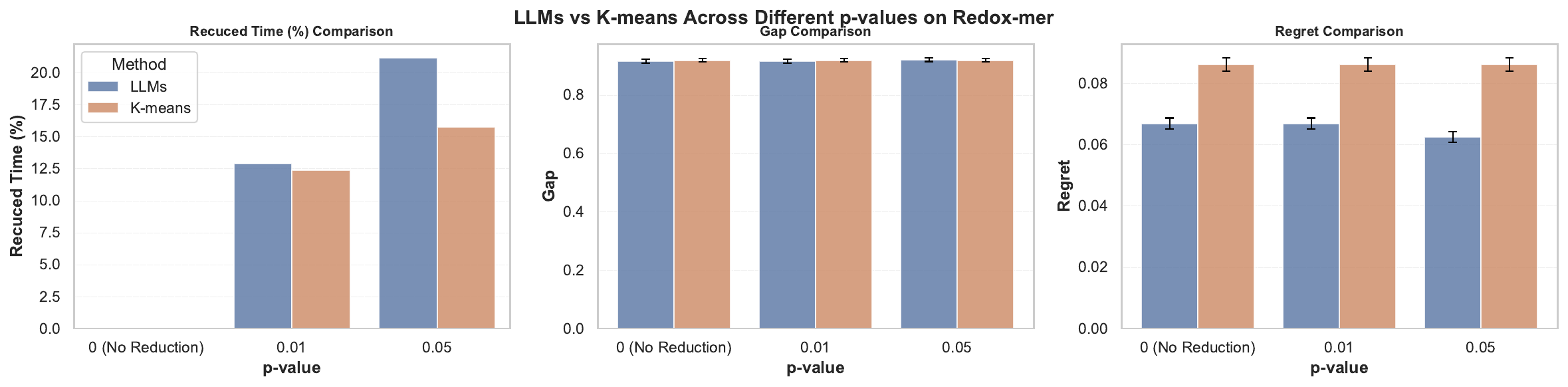}
    \includegraphics[width=0.8\linewidth]{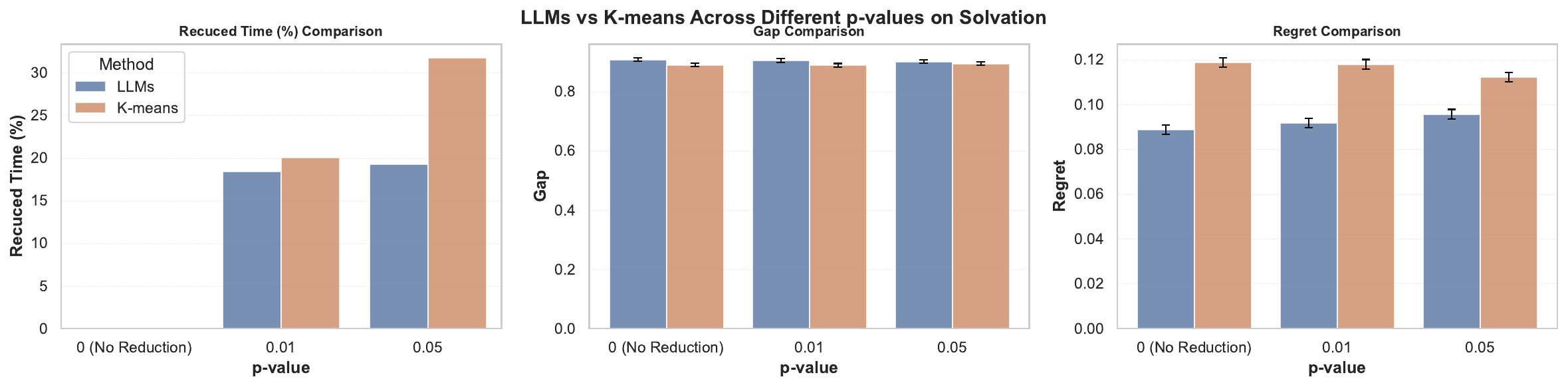}
    \includegraphics[width=0.8\linewidth]{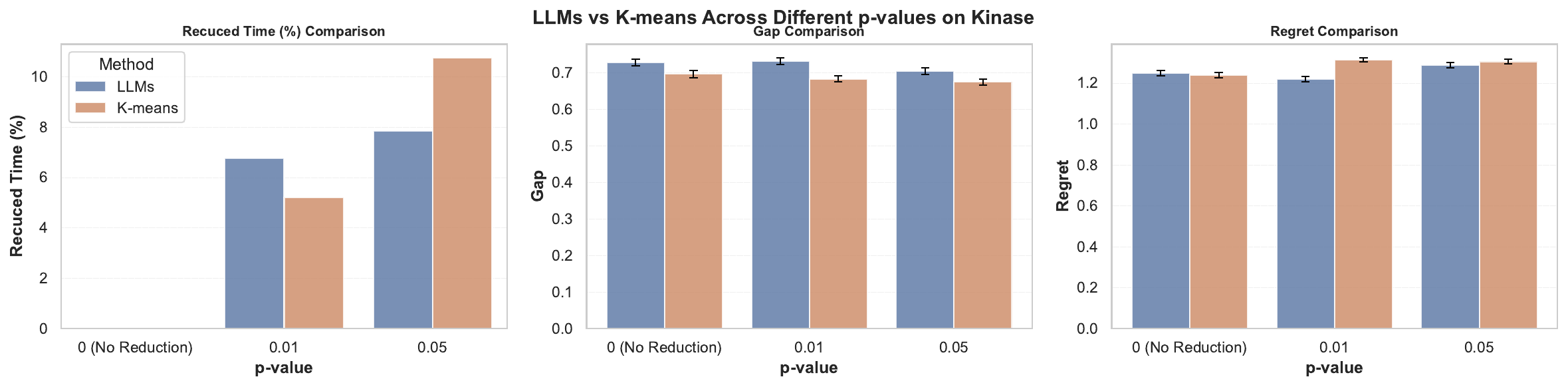}
    \includegraphics[width=0.8\linewidth]{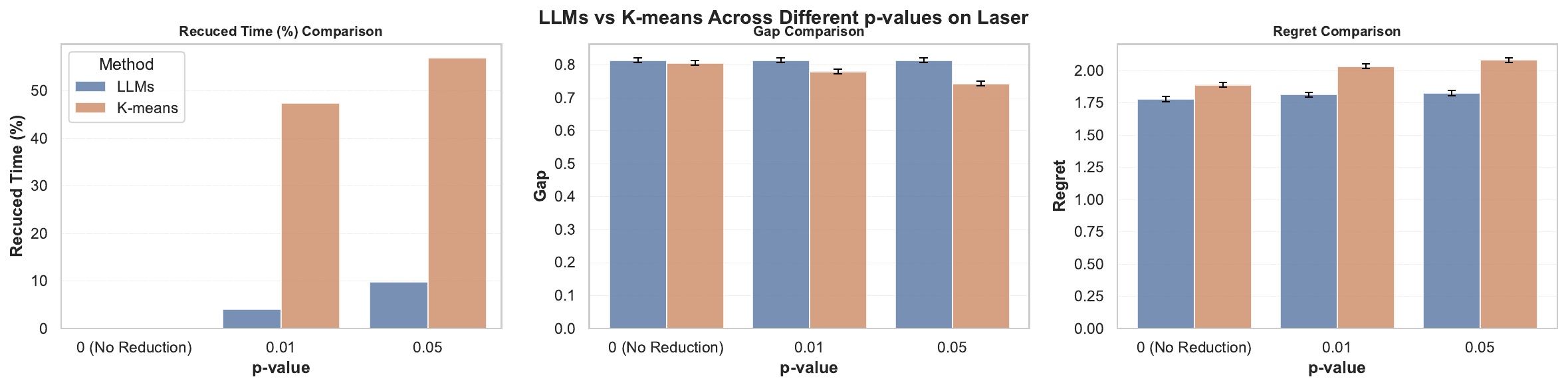}
    \includegraphics[width=0.8\linewidth]{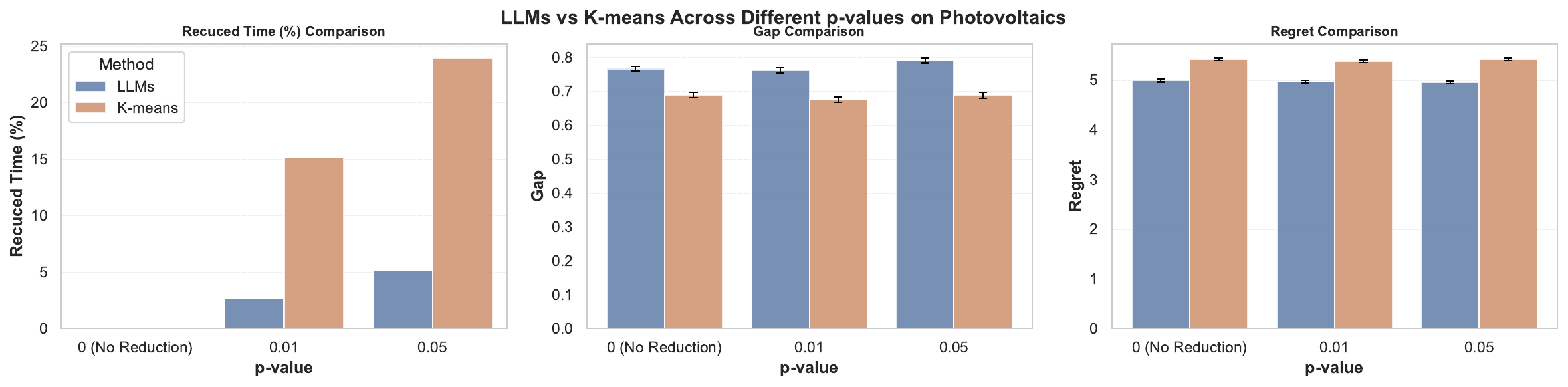}
    \includegraphics[width=0.8\linewidth]{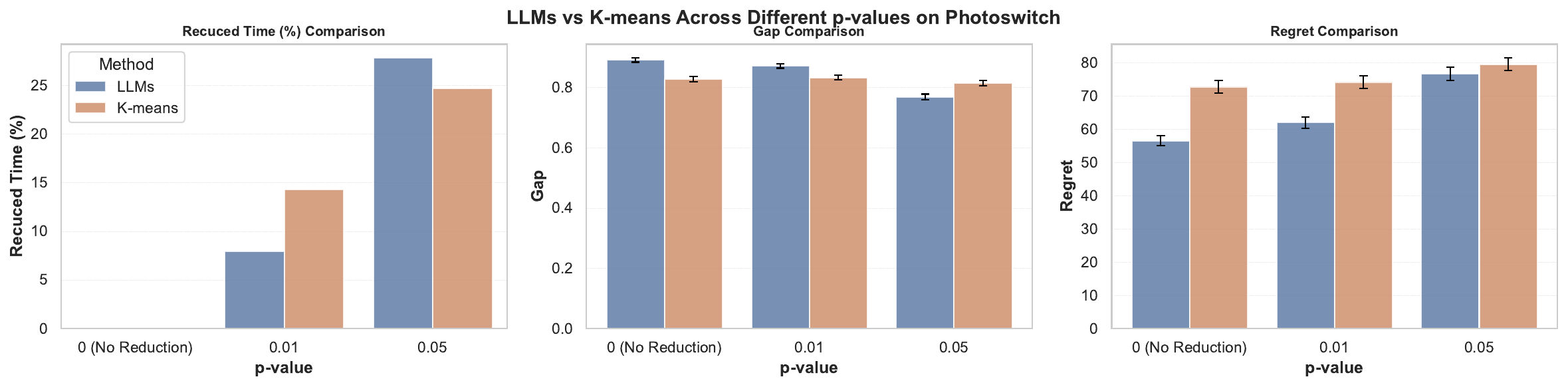}
    \caption{Prediction time reduction percentage, AUC of GAPs, and Regrets across all the datasets for different clustering approaches and p-values. }
    \label{fig:molformer_clustering_all}
\end{figure}

\subsection{Toy Data}\label{sec:toy}

\subsubsection{Levy-1D Data}
\begin{figure}[H]
    \centering
    \includegraphics[width=0.8\linewidth]{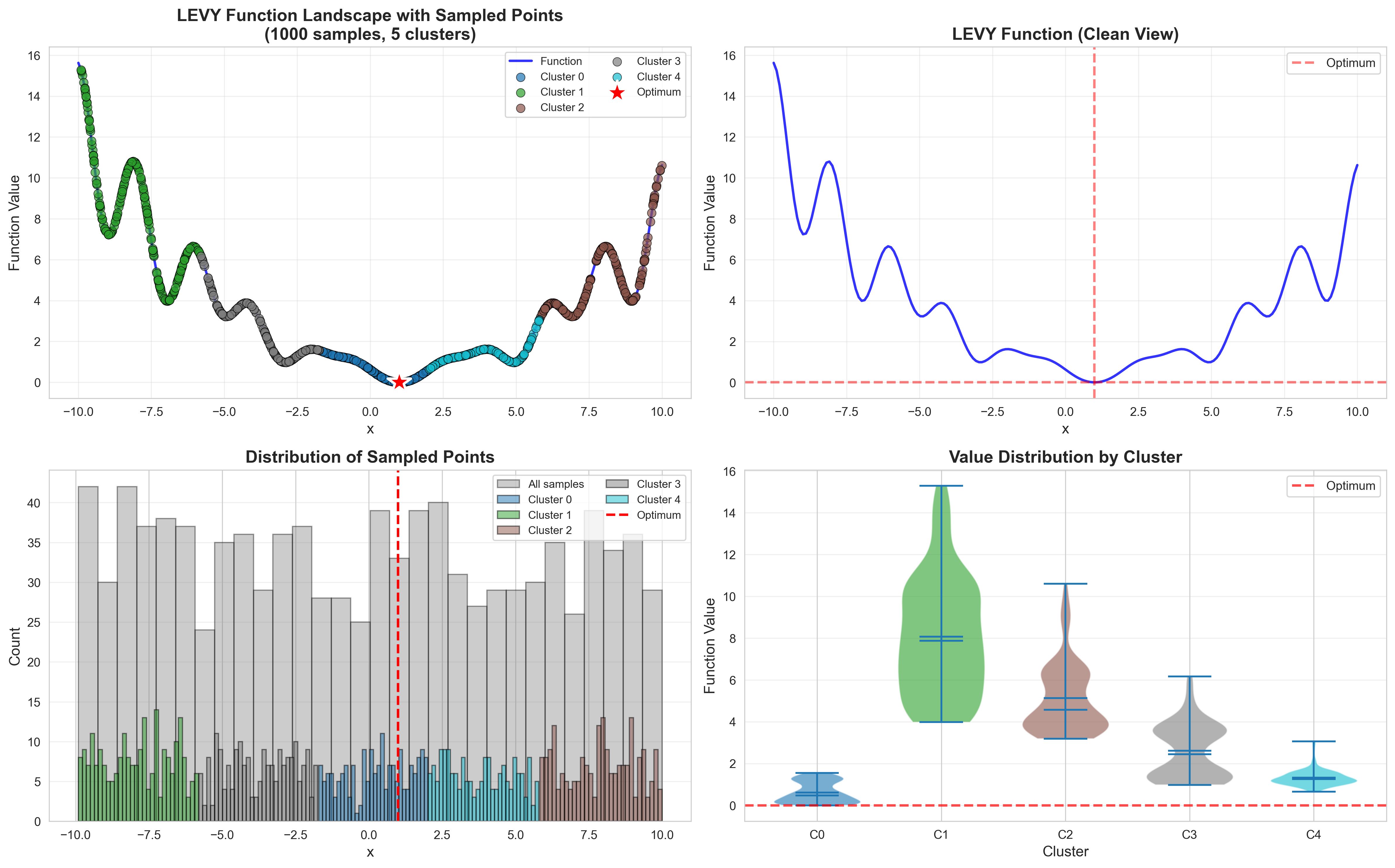}
\caption{Illustration of 1,000 samples of Levy-1D function with K-means clustering.}
    \label{fig:Levey_1D}
\end{figure}


\begin{figure}[H]
    \centering
    \begin{minipage}[t]{0.49\linewidth}
        \centering
        \includegraphics[width=\linewidth]{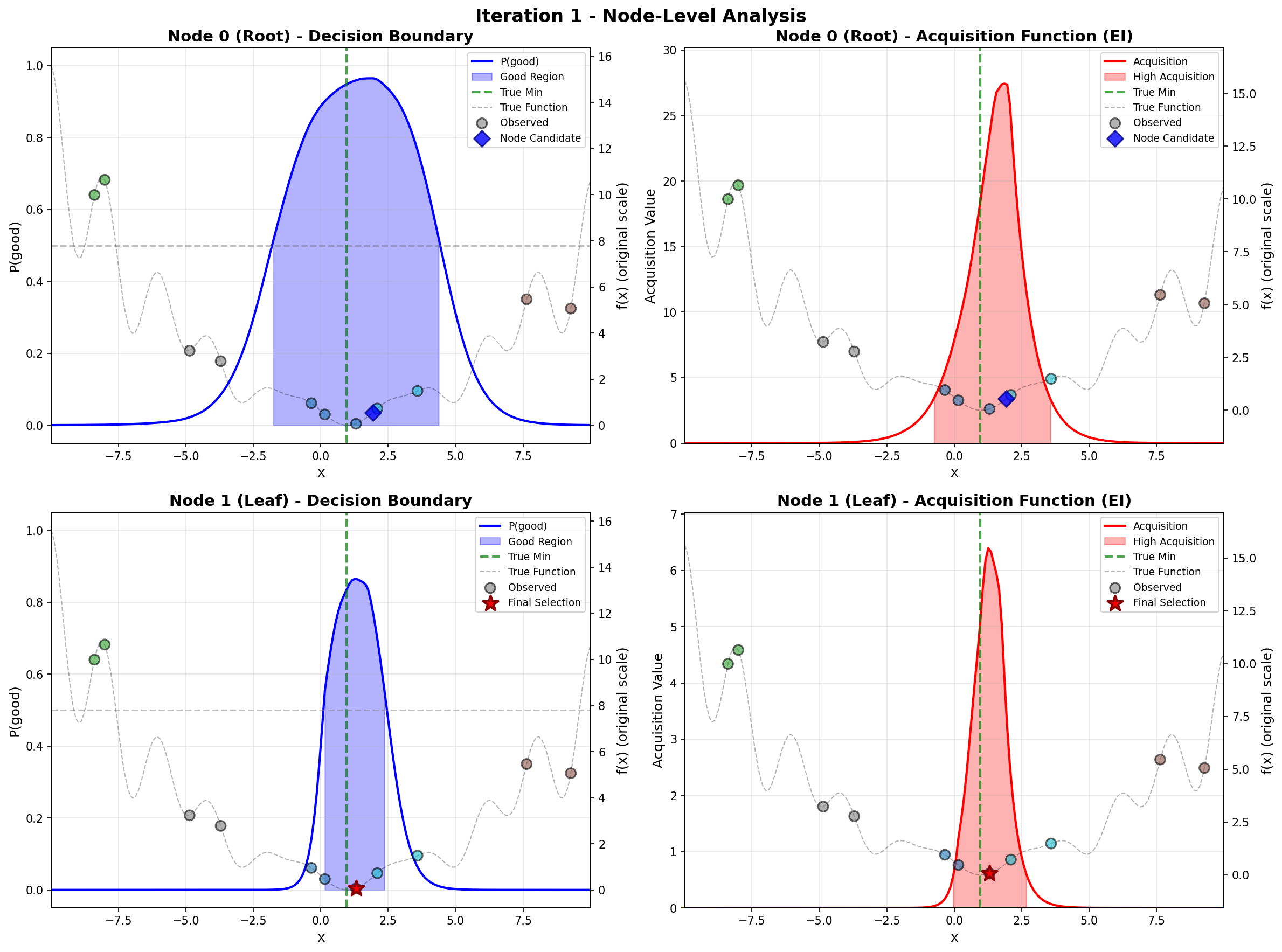}
        \caption{Refined AFs in LLMAT enable more efficient BO (Iteration 1).}
        \label{fig:LLMAT_refine_AFs_iteration1}
    \end{minipage}\hfill
    \begin{minipage}[t]{0.49\linewidth}
        \centering
        \includegraphics[width=\linewidth]{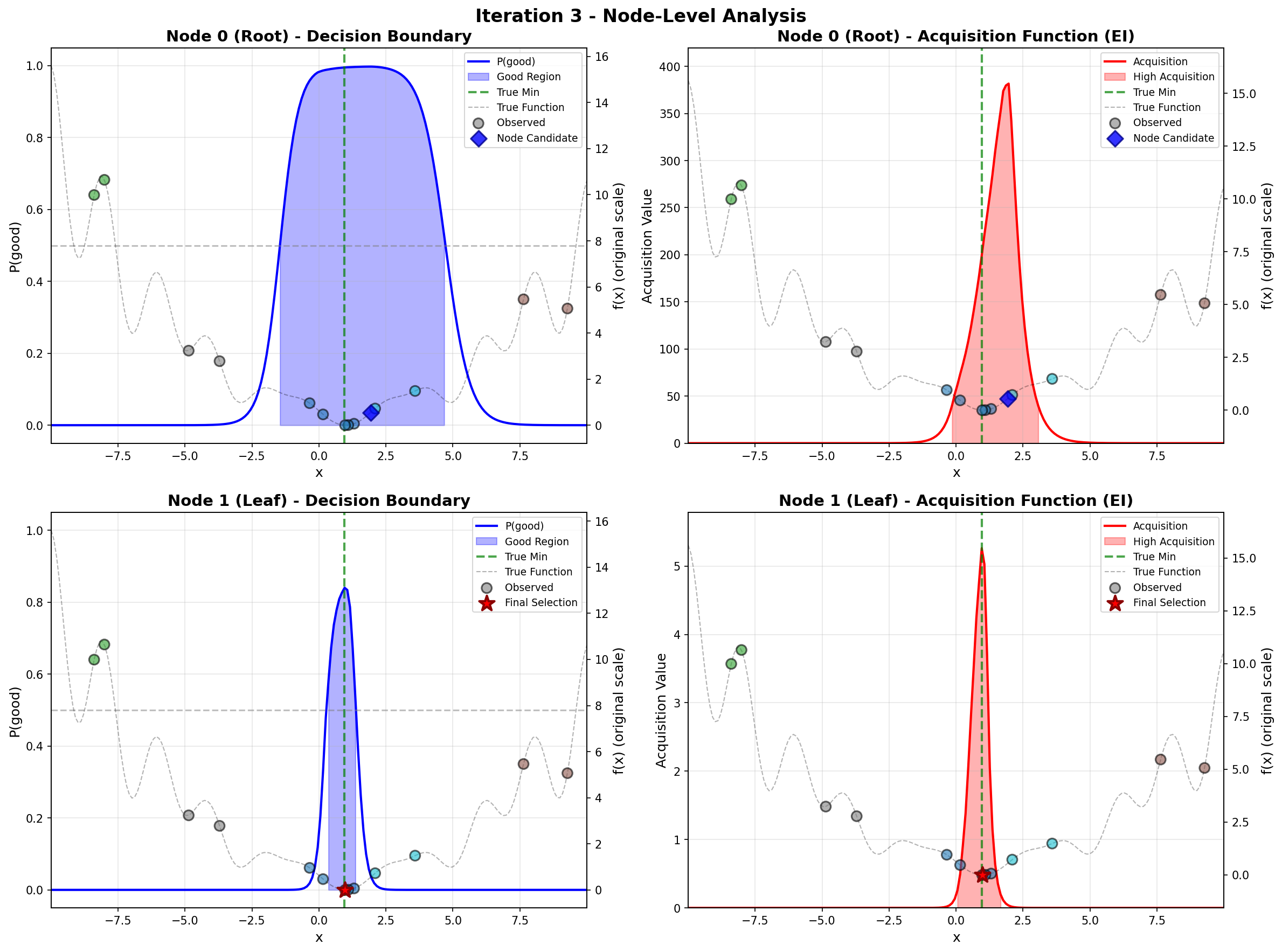}
        \caption{Refined AFs in LLMAT enable more efficient BO (Iteration 3).}
        \label{fig:LLMAT_refine_AFs_iteration3}
    \end{minipage}
\end{figure}
\begin{figure}[H]
    \centering
    \begin{minipage}[t]{0.49\linewidth}
        \centering
        \includegraphics[width=\linewidth]{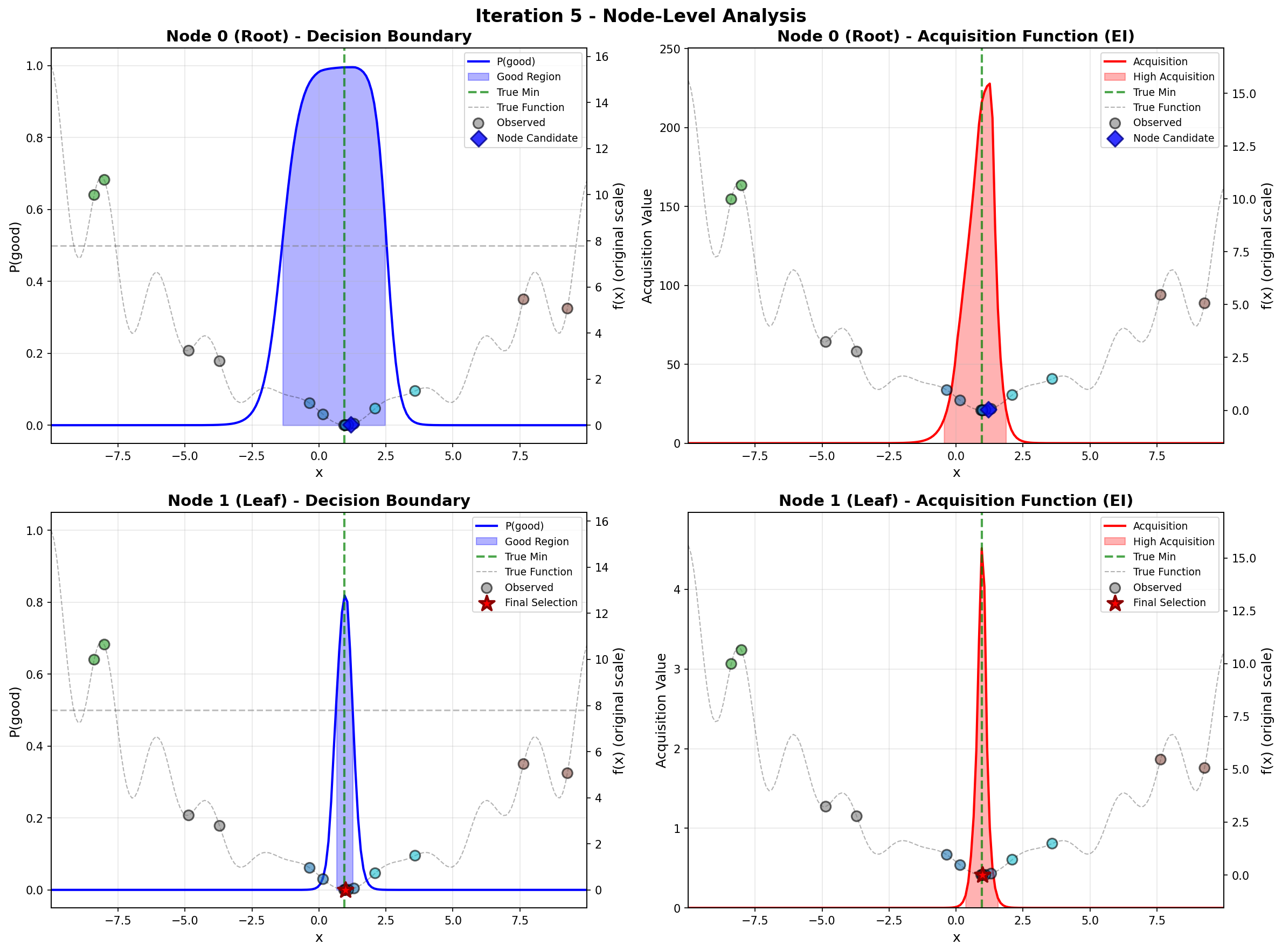}
        \caption{Refined AFs in LLMAT enable more efficient BO (Iteration 5).}
        \label{fig:LLMAT_refine_AFs_iteration5}
    \end{minipage}\hfill
    \begin{minipage}[t]{0.49\linewidth}
        \centering
        \includegraphics[width=\linewidth]{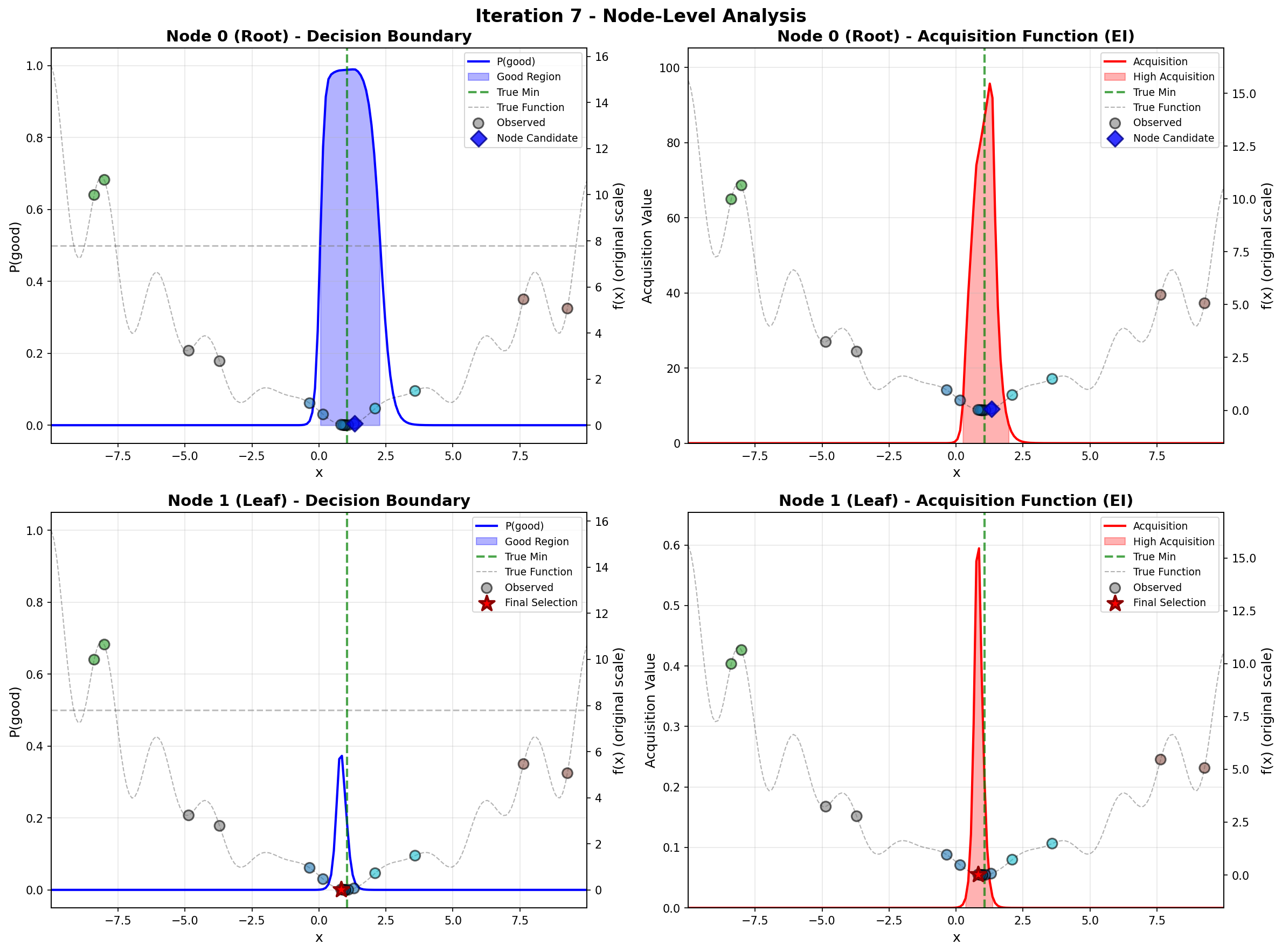}
        \caption{Refined AFs in LLMAT enable more efficient BO (Iteration 7).}
        \label{fig:LLMAT_refine_AFs_iteration7}
    \end{minipage}
\end{figure}
\subsubsection{Levy-10D Data}
\begin{figure}[H]
    \centering
    \includegraphics[width=0.6\linewidth]{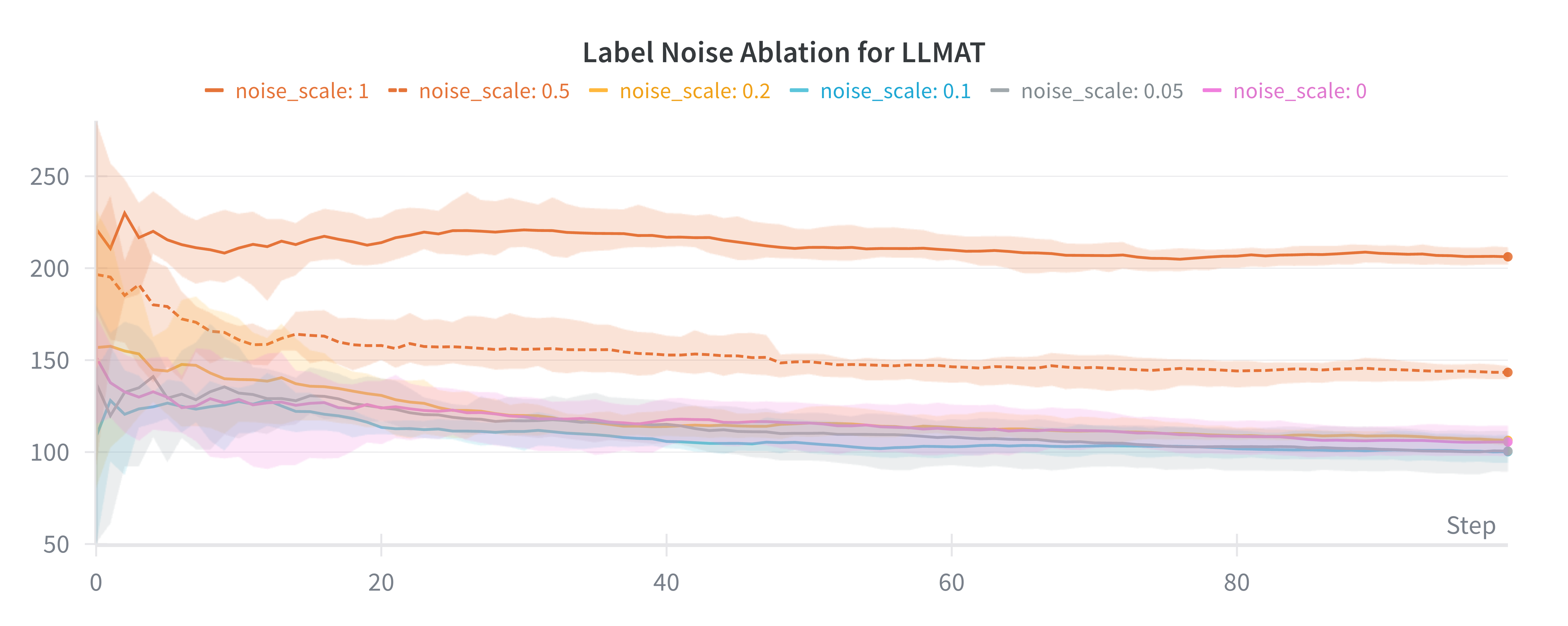}
    \caption{Illustration of label noise's effects on the average regret of LLMAT on the Levy-10D function.}
    \label{fig:Label Noise Ablation}
\end{figure}


\subsection{Ablation of \texorpdfstring{$\gamma$}{gamma}}
\begin{figure}[H]
    \centering
    \includegraphics[width=0.7\linewidth]{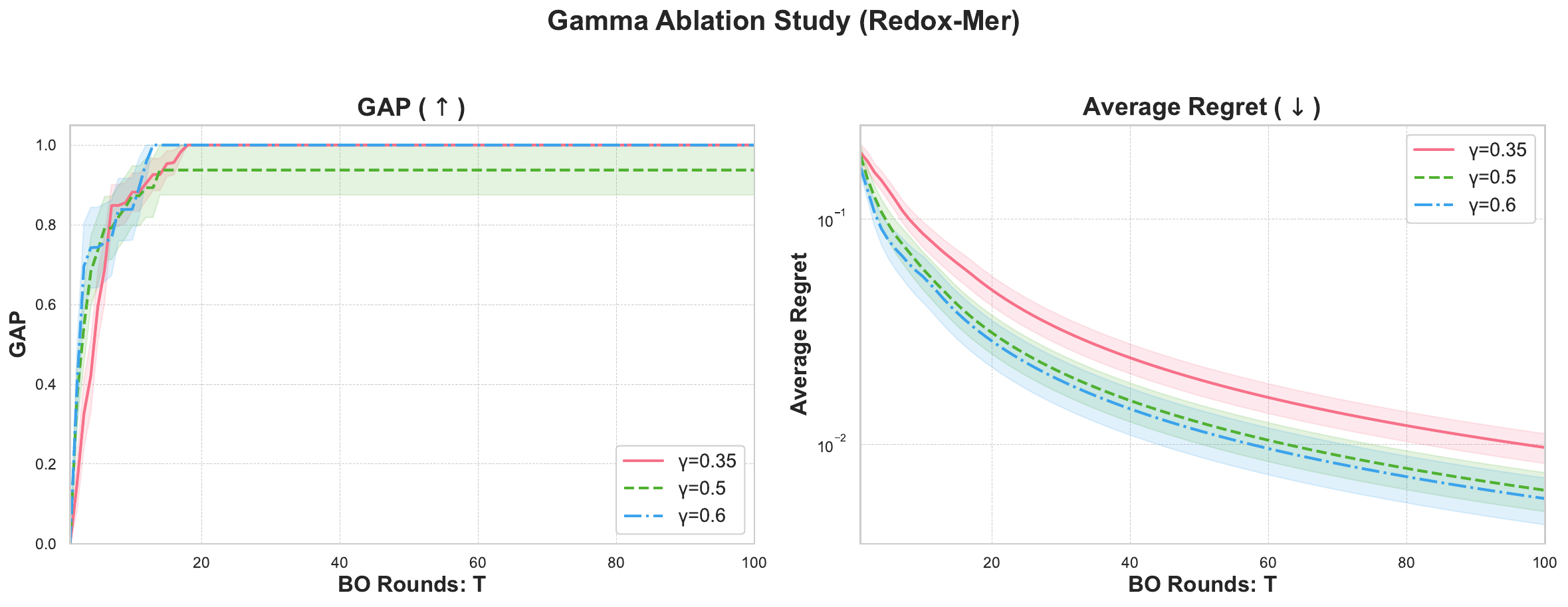}
    \caption{Illustration of LLMAT performance change w.r.t.\ different $\gamma$s.}
    \label{fig: Ablation of gamma.}
\end{figure}
\subsection{Additional results for \texorpdfstring{Llama-2-7b}{Llama-2-7b}}

\begin{figure}[H]
    \centering
    \includegraphics[width=0.7\linewidth]{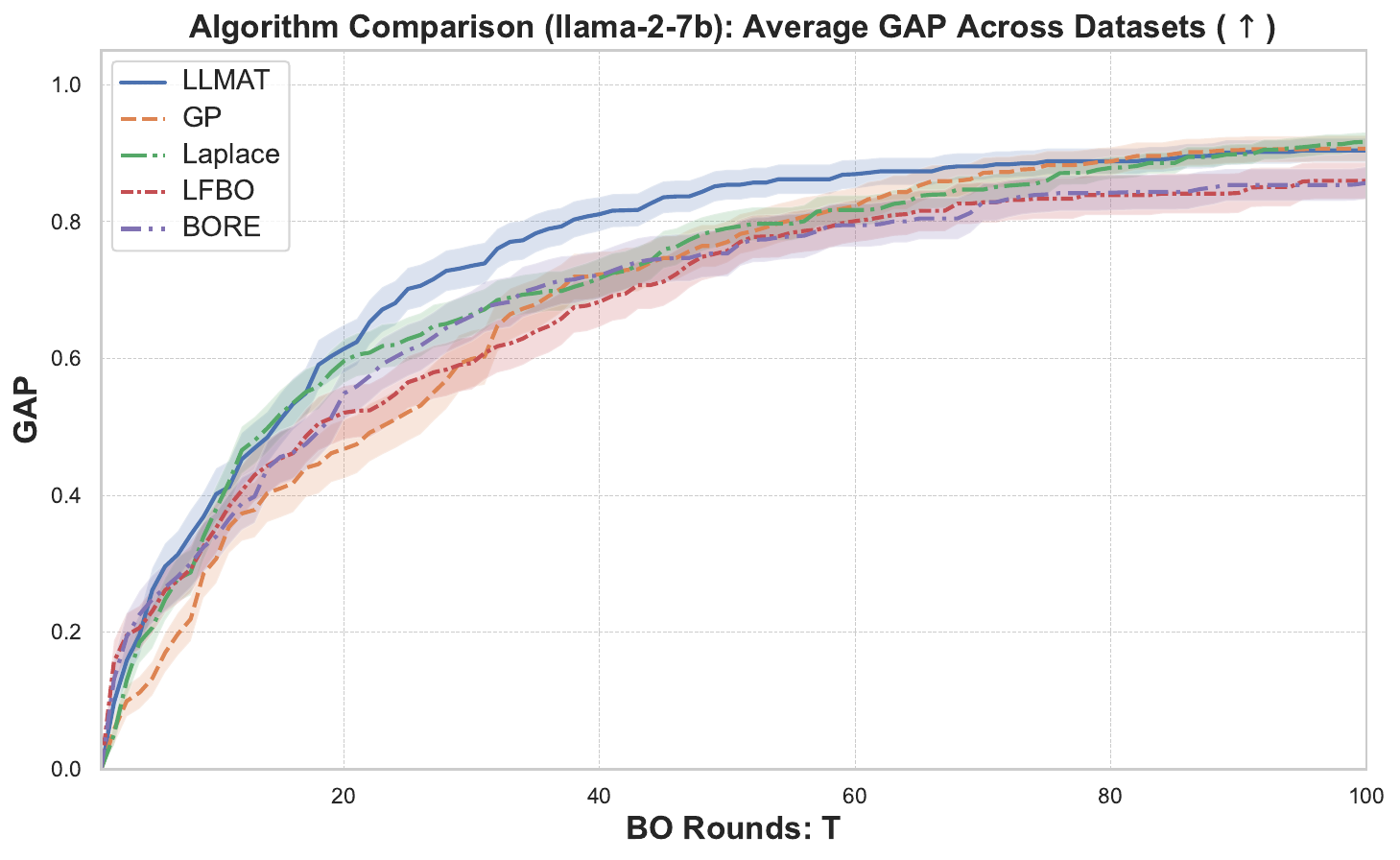}
    \caption{Performance comparison of various algorithms on Llama-2-7b across all datasets.}
    \label{fig: llma2.}
\end{figure}

\subsection{Multi-objective Optimization Extension}
\begin{figure}[H]
    \centering
    \includegraphics[width=0.45\linewidth]{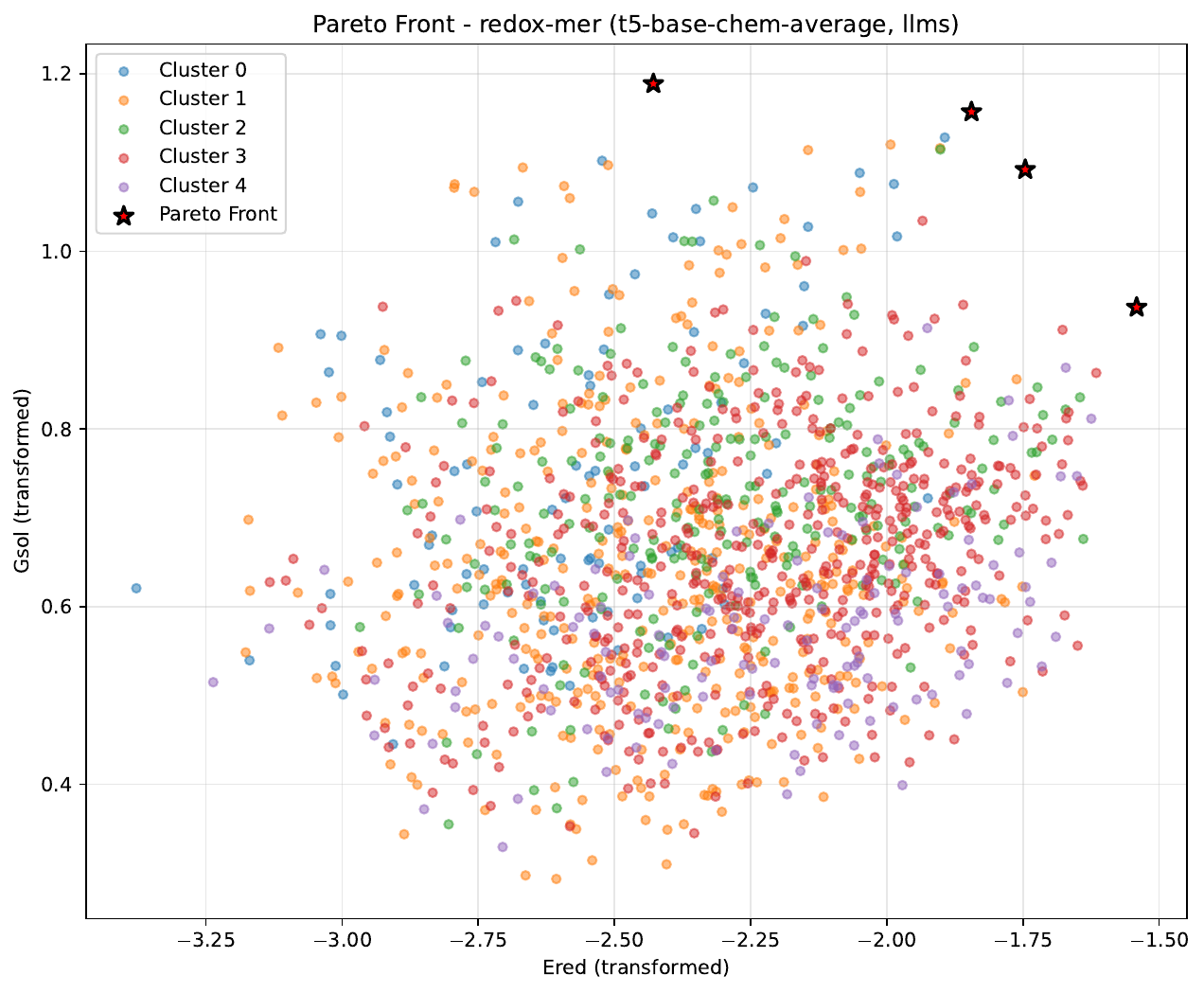}
        \includegraphics[width=0.45\linewidth]{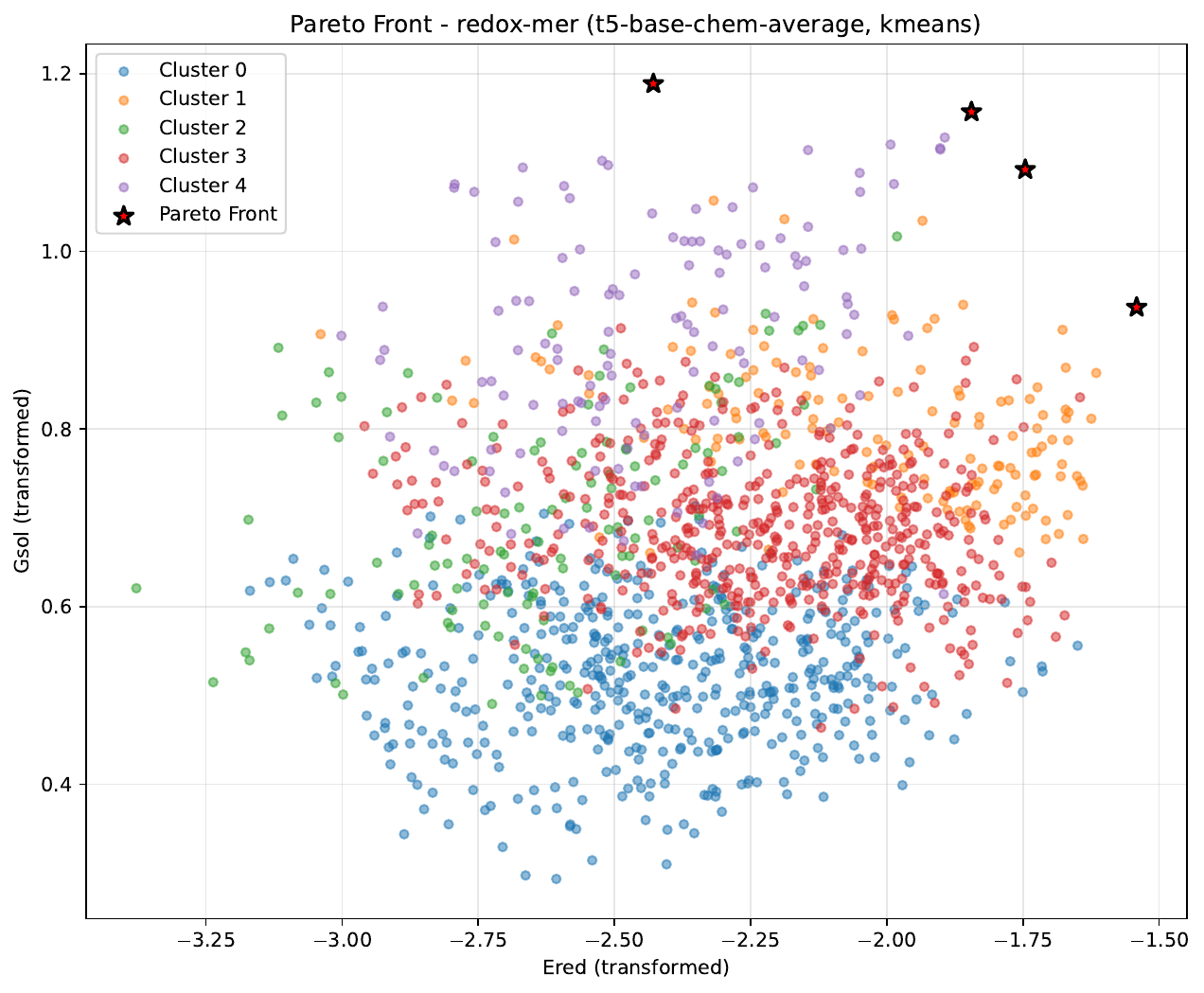}
    \caption{Pareto front with LLM (left) and K-means (right) clusters on Multi-Redox data.}
    \label{fig:Multi-Redox Pareto Front}
\end{figure}

\begin{figure}[H]
    \centering
    \includegraphics[width=0.45\linewidth]{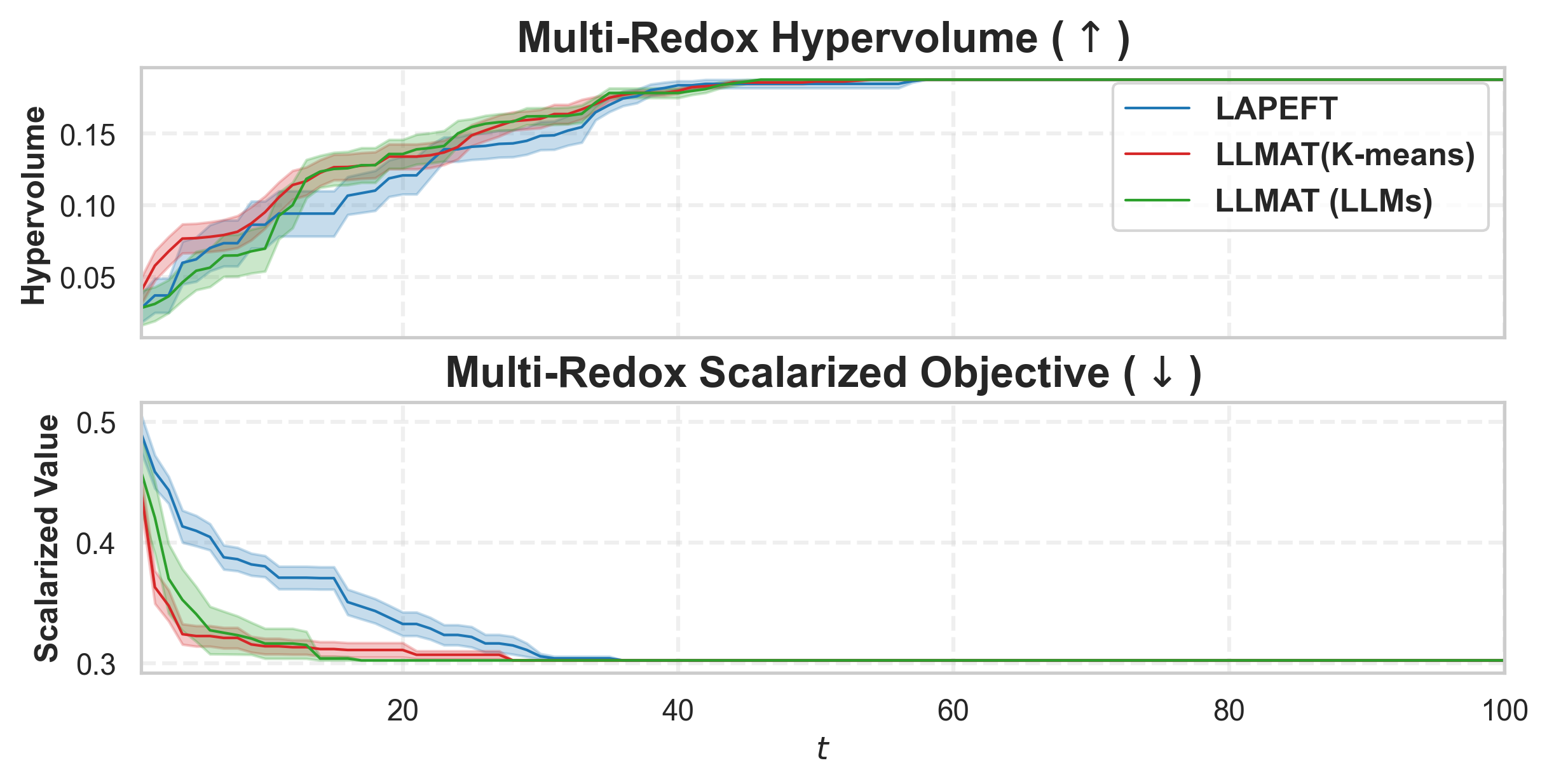}
    \includegraphics[width=0.45\linewidth]{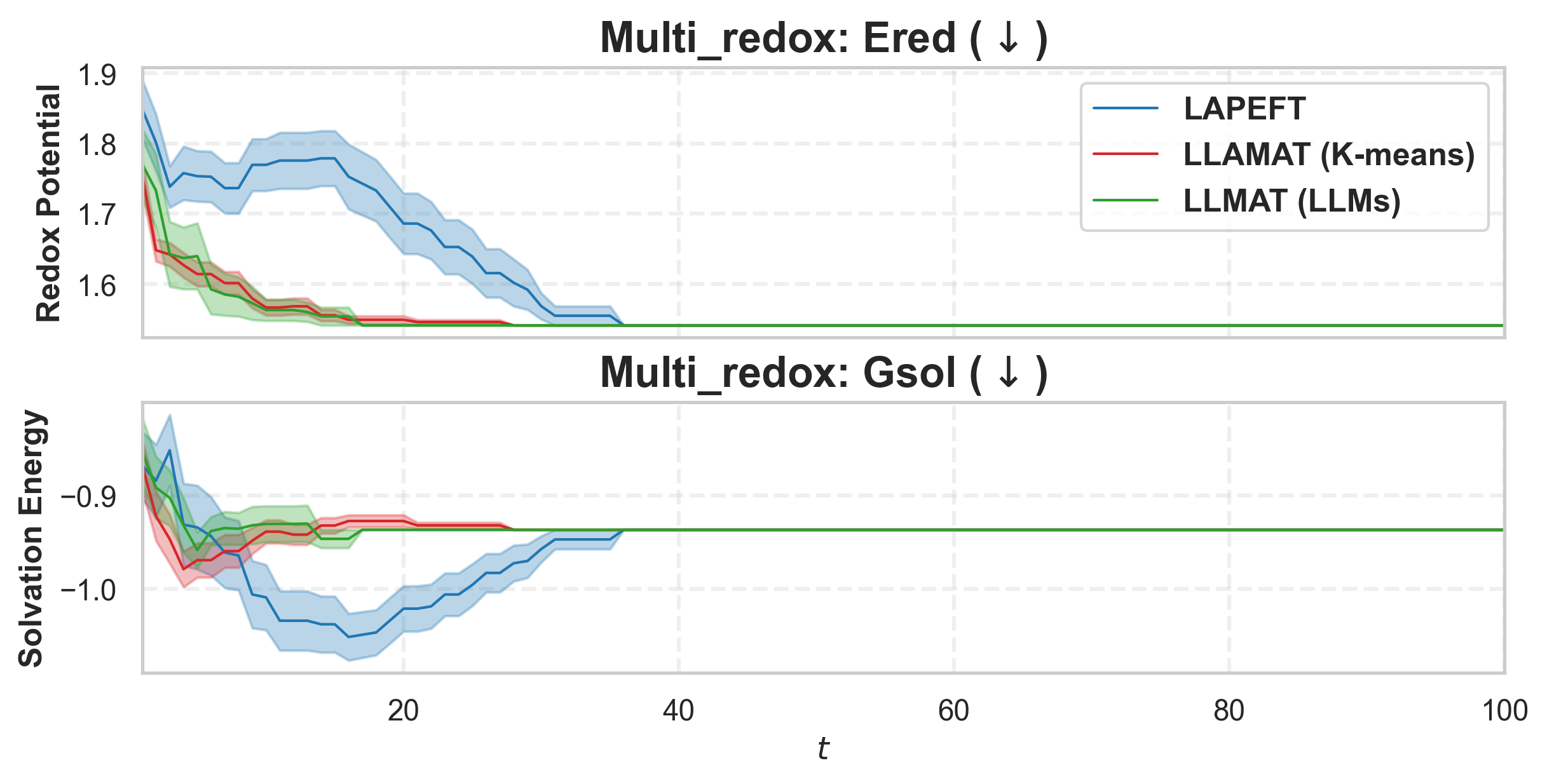}
\caption{Multi-objective optimization for LAPEFT vs. LLMAT on the Redox-mer dataset: \textbf{Upper \& Lower Left:} Hypervolume, scalarized objective. \textbf{Right:} Redox potential and solvation energy.}
    \label{fig:multi-objective extension}
\end{figure}

\begin{figure}[H]
    \centering
    \includegraphics[width=\linewidth]{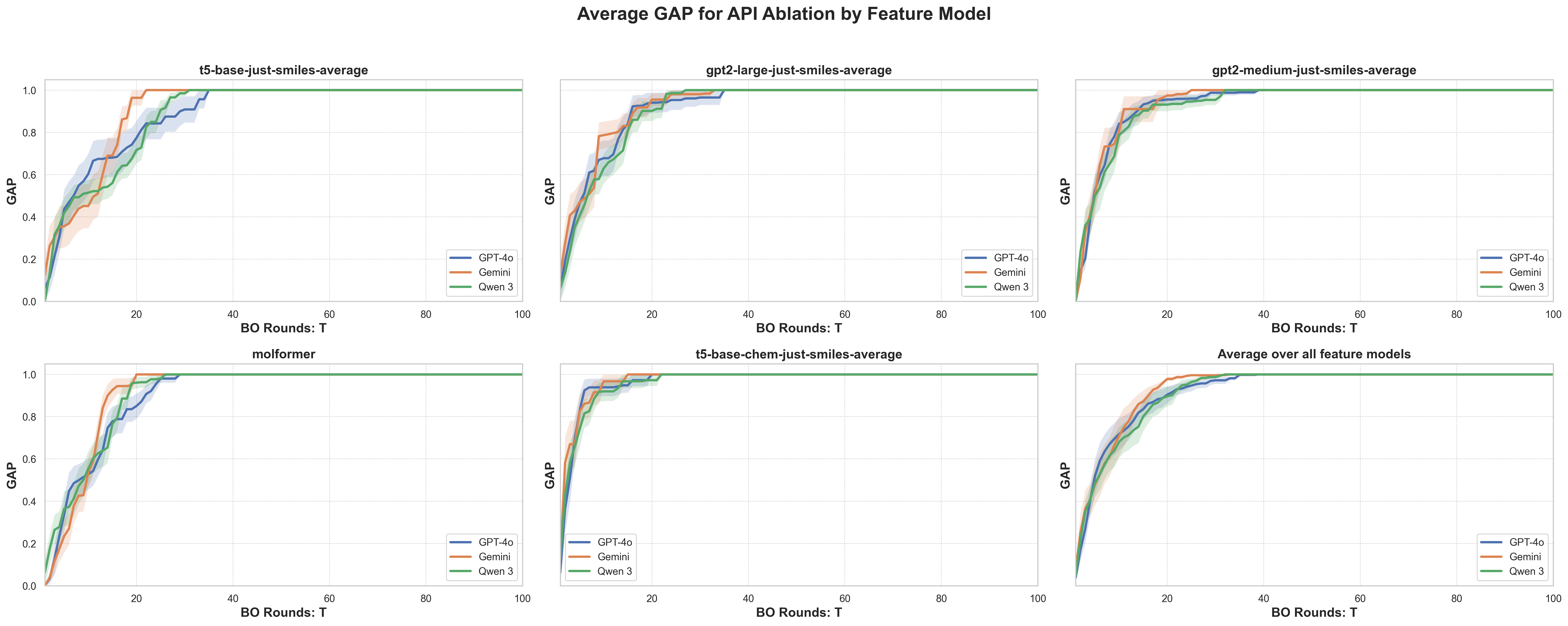}
        \includegraphics[width=\linewidth]{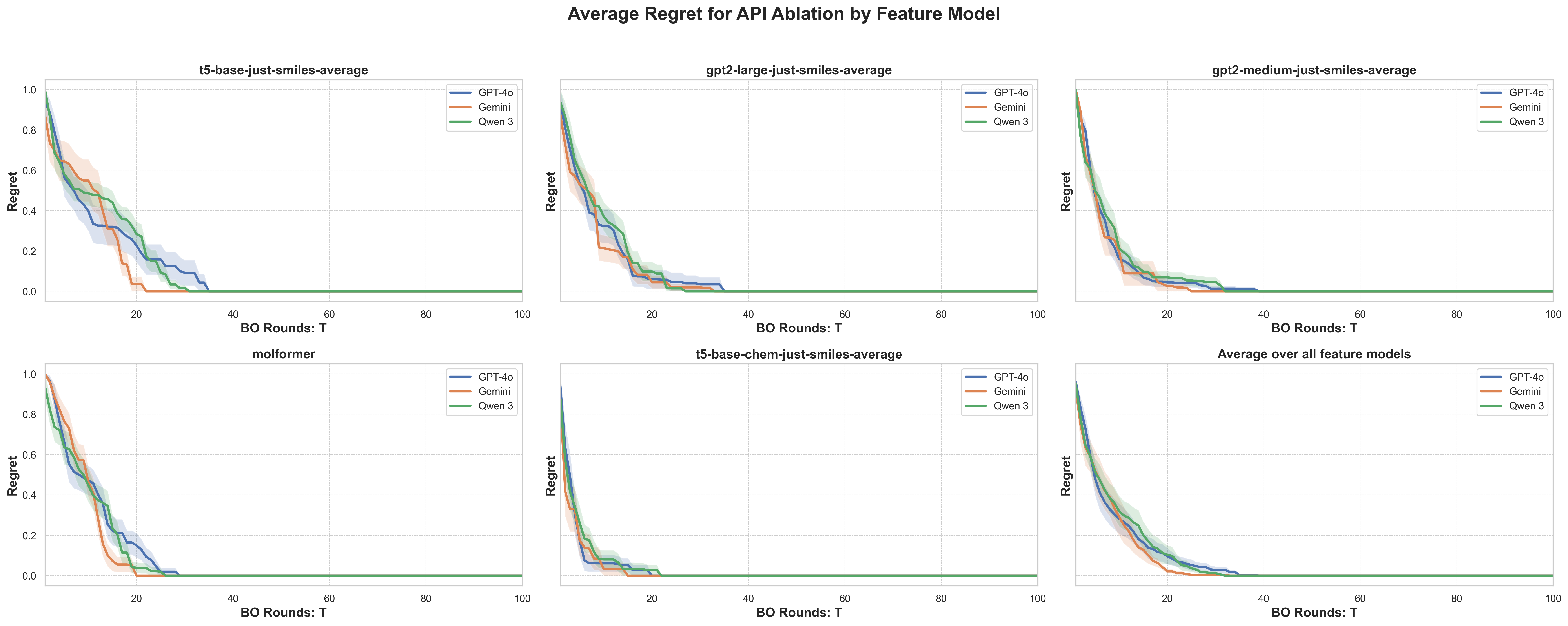}
    \caption{GAP (top) and average regret (bottom) of LLMAT on Redox-mer when LLM-based clustering uses different LLM APIs (GPT-4o, Gemini, Qwen3-Plus), across feature models.}
    \label{fig:api_ablation}
\end{figure}

We consider two objectives on the Redox-mer dataset, redox potential and solvation energy, and compare LLMAT with LAPEFT using the hypervolume and a scalarized objective (Fig.~\ref{fig:multi-objective extension}). Fig.~\ref{fig:Multi-Redox Pareto Front} shows the Pareto fronts obtained with LLM-based and K-means clusters.

\textbf{Setup.} Both objectives (redox potential and solvation energy) are minimized; we negate them so that the problem becomes maximization. LLMAT is extended by scalarization: at every round the objectives of each observation are combined with a linear scalarization with uniform weights $1/k$ ($k=2$), applied to the raw, unnormalized objective values, and this scalar value replaces $y$ in the original algorithm. The scalar values define the thresholds $\tau_k$ and binary labels of the node classifiers, so the tree construction, the local acquisition functions and the UCT-based selection are unchanged, and the candidate with the highest acquisition value is queried. This variant uses fixed features (T5-Chem), $10$ initial points, and $T=100$ rounds. After each query we update the Pareto front of all observations and report its hypervolume, as well as the best scalarized value on the current Pareto front. The hypervolume reference point is inferred with BoTorch's \texttt{infer\_reference\_point} from the Pareto front of the full candidate pool, and the maximal hypervolume is that of this ground-truth front.

\subsection{Ablation over different LLM APIs}

\subsection{Fixed GNN Embeddings (GIN)}\label{sec:gin}
To verify that LLMAT is not limited to SMILES/text representations, we evaluate it with fixed embeddings from a pretrained GIN \citep{xu_how_2019, hu_strategies_2020}, without finetuning during BO. We report the AUC of the normalized gap $\mathrm{Gap}_t=\frac{y_t^{\text{best}}-y_0}{y_{\max}-y_0}$ over the whole trajectory (Gap AUC, Sec.~\ref{sec:gap-auc}); higher is better. LLMAT achieves the best average Gap AUC ($79.11$) across the six datasets, with particularly strong gains on Redoxmer, Laser, and Photovoltaics, so its sample-efficiency advantage extends to pretrained graph-based molecular features.

\begin{table}[H]
\centering\small
\caption{Gap AUC ($\uparrow$) with fixed GIN embeddings (mean $\pm$ std). Best per row in bold.}
\label{tab:gin}
\begin{tabular}{lcccc}
\toprule
Dataset & LLMAT & GP & Laplace & Random \\
\midrule
Redoxmer & \textbf{92.3$\pm$0.9} & 85.4$\pm$2.5 & 75.2$\pm$3.1 & 30.9$\pm$6.4 \\
Solvation & 83.9$\pm$2.8 & \textbf{85.3$\pm$1.1} & 76.1$\pm$1.5 & 44.5$\pm$4.8 \\
Kinase & 77.1$\pm$4.7 & \textbf{78.8$\pm$1.8} & 67.9$\pm$4.4 & 47.4$\pm$6.6 \\
Laser & \textbf{74.3$\pm$3.4} & 47.8$\pm$4.8 & 63.9$\pm$3.8 & 30.2$\pm$5.6 \\
Photovoltaics & \textbf{68.6$\pm$6.4} & 55.0$\pm$6.6 & 59.3$\pm$6.3 & 41.1$\pm$6.3 \\
Photoswitches & 78.5$\pm$3.0 & \textbf{82.4$\pm$3.5} & 59.2$\pm$5.3 & 34.9$\pm$7.5 \\
\midrule
Average & \textbf{79.11} & 72.46 & 66.92 & 38.14 \\
\bottomrule
\end{tabular}
\end{table}

\subsection{Informativeness of Representations}\label{sec:informativeness}
A general-purpose LLM representation is not intrinsically a ``weak'' prior, so informativeness should be defined with respect to the downstream property. As a task-specific proxy, we fit the same linear probe on each representation and report cross-validated $R^2$ (Table~\ref{tab:linear-probe}). There is no universal ordering: foundation-model embeddings are generally more predictive on larger, more complex datasets such as Kinase, while Morgan fingerprints perform best on Photoswitches.

\begin{table}[H]
\centering\small
\caption{Linear-probe cross-validated $R^2$ ($\uparrow$) for different molecular representations. Best per row in bold.}
\label{tab:linear-probe}
\resizebox{\linewidth}{!}{%
\begin{tabular}{lrccccccc}
\toprule
Dataset & Size & T5-Chem & Fingerprints & MolFormer & T5-Base & GPT2-Med & GPT2-Large & LLaMA-2-7B \\
\midrule
Redoxmer & 1407 & \textbf{0.940} & 0.931 & 0.919 & 0.920 & 0.892 & 0.882 & 0.915 \\
Solvation & 1407 & \textbf{0.856} & 0.833 & 0.845 & 0.847 & 0.834 & 0.841 & 0.843 \\
Kinase & 10449 & \textbf{0.757} & 0.680 & 0.736 & 0.708 & 0.689 & 0.697 & 0.744 \\
Laser & 10000 & 0.574 & 0.598 & 0.548 & 0.514 & 0.566 & 0.570 & \textbf{0.629} \\
PCE & 10000 & \textbf{0.608} & 0.595 & 0.577 & 0.581 & 0.549 & 0.578 & 0.607 \\
Photoswitch & 392 & 0.441 & \textbf{0.516} & 0.352 & 0.320 & 0.303 & 0.261 & 0.335 \\
\bottomrule
\end{tabular}}
\end{table}

To assess sensitivity to the choice of representation, Table~\ref{tab:worst-case} compares each algorithm under its worst-performing representation. LLMAT attains the highest worst-case Gap AUC on five of six datasets and is comparable to Laplace on Laser ($56.78\pm4.96$ vs.\ $57.84\pm2.72$). Averaged over datasets, the worst-case Gap AUC is $72.28$ for LLMAT, versus $55.11$ for GP and $61.41$ for Laplace, so LLMAT is substantially more robust to the representation.

\begin{table}[H]
\centering\small
\caption{Worst-case Gap AUC ($\uparrow$) over representations; the representation attaining the worst value is in parentheses.}
\label{tab:worst-case}
\resizebox{\linewidth}{!}{%
\begin{tabular}{lccc}
\toprule
Dataset & GP worst & LLMAT worst & Laplace worst \\
\midrule
Redoxmer & 70.15$\pm$2.94 (GPT2-Med) & \textbf{82.09$\pm$2.29} (LLaMA-2-7B) & 72.86$\pm$2.76 (GPT2-Med) \\
Solvation & 56.85$\pm$3.37 (GPT2-Med) & \textbf{82.72$\pm$3.29} (GIN) & 66.36$\pm$1.38 (LLaMA-2-7B) \\
Kinase & 55.53$\pm$4.21 (Fingerprints) & \textbf{74.64$\pm$3.28} (MolFormer) & 50.83$\pm$4.07 (Fingerprints) \\
Laser & 42.74$\pm$3.10 (GPT2-Med) & 56.78$\pm$4.96 (Fingerprints) & \textbf{57.84$\pm$2.72} (Fingerprints) \\
Photovoltaics & 53.90$\pm$4.90 (GPT2-Med) & \textbf{69.44$\pm$3.36} (LLaMA-2-7B) & 57.47$\pm$3.06 (Fingerprints) \\
Photoswitches & 51.50$\pm$6.13 (GPT2-Med) & \textbf{68.02$\pm$6.22} (LLaMA-2-7B) & 63.12$\pm$4.10 (GIN) \\
\midrule
Average & 55.11 & \textbf{72.28} & 61.41 \\
\bottomrule
\end{tabular}}
\end{table}

\subsection{Morgan Fingerprints vs.\ Foundation-Model Embeddings}\label{sec:fp-vs-fm}
Table~\ref{tab:fp-vs-fm-qual} summarizes, by visual inspection of the BO trajectories, how foundation-model (LLM) embeddings compare with Morgan fingerprints (FP) for each algorithm. Foundation-model embeddings do not universally outperform Morgan fingerprints, but give clear gains on most datasets. Fingerprints are particularly strong on Photoswitches under LLMAT and GP, consistent with prior observations that standard molecular descriptors can be more than competitive with neural representations on this dataset \citep{griffiths_data-driven_2022}. The benefit of foundation-model representations is therefore task- and surrogate-dependent. One possible explanation for the differences across algorithms is that GP and LLMAT depend directly on the structure induced by the input representation (GP through kernel similarities, LLMAT through classifiers that partition the space into promising and non-promising regions), whereas Laplace applies a learned nonlinear transformation before uncertainty estimation and is thus less sensitive to the original geometry.

\begin{table}[H]
\centering\small
\caption{Qualitative comparison of foundation-model (LLM) embeddings and Morgan fingerprints (FP), from BO trajectories.}
\label{tab:fp-vs-fm-qual}
\resizebox{\linewidth}{!}{%
\begin{tabular}{llll}
\toprule
Dataset & LLMAT & GP & Laplace \\
\midrule
Redoxmer & LLM clearly better & LLM slightly better & Comparable \\
Solvation & LLM slightly better & LLM slightly better & LLM slightly better \\
Kinase & LLM better & LLM better & LLM better \\
Laser & LLM clearly better & LLM better & LLM clearly better \\
Photovoltaics & LLM better & LLM slightly better & LLM better \\
Photoswitches & FP clearly better & FP clearly better & LLM slightly better \\
\bottomrule
\end{tabular}}
\end{table}

To quantify these observations, we compute Gap AUC over all BO rounds (Table~\ref{tab:fp-vs-fm-quant}). With the best foundation-model embedding per dataset, LLMAT improves the mean Gap AUC over Morgan fingerprints by $9.7$ points and wins on $5/6$ datasets; averaging over all foundation-model representations, the improvement is $+2.6$ points, with the average representation beating fingerprints on $4/6$ datasets. LLMAT also obtains the highest Gap AUC among the three algorithms under all three representation summaries: with Morgan fingerprints it reaches $74.8$ (GP $72.0$, Laplace $65.3$); with the best foundation model per dataset $84.4$ (GP $78.7$, Laplace $81.2$); and averaged over all foundation models $77.4$ (GP $68.2$, Laplace $72.3$). The advantage therefore cannot be attributed solely to more expressive embeddings.

\begin{table}[H]
\centering\small
\caption{Gap AUC with Morgan fingerprints (FP) vs.\ foundation-model (FM) embeddings. ``Best FM'' selects the best FM per dataset; ``Average FM'' averages over all FMs. $\Delta$ is the difference to FP.}
\label{tab:fp-vs-fm-quant}
\resizebox{\linewidth}{!}{%
\begin{tabular}{lccccccc}
\toprule
Algorithm & Morgan FP & Best FM & $\Delta$ & Best FM beats FP & Average FM & $\Delta$ & Average FM beats FP \\
\midrule
LLMAT & 74.8 & 84.4 & +9.7 & 5/6 & 77.4 & +2.6 & 4/6 \\
GP & 72.0 & 78.7 & +6.7 & 5/6 & 68.2 & $-$3.8 & 1/6 \\
Laplace & 65.3 & 81.2 & +15.9 & 6/6 & 72.3 & +7.0 & 3/6 \\
\bottomrule
\end{tabular}}
\end{table}

\subsection{Ablation: Unified Classifier vs.\ Decoupled Partitioning and LFBO}\label{sec:decoupled}
To verify the benefit of coupling partitioning and acquisition in one classifier, we implement a \emph{decoupled} baseline under fixed features. At each node, K-means partitions the samples in the joint $(x,y)$ space and its labels train a routing classifier, while a separate LFBO classifier performs acquisition scoring; features, initialization, BO budget, and the LFBO objective are unchanged. Values in Table~\ref{tab:decoupled} are averaged over six fixed feature models per dataset. LLMAT wins on $26/36$ configurations, and among $11$ significant paired Wilcoxon comparisons ($p<0.05$), $10$ favor LLMAT. Coupling is thus not uniformly beneficial, but it improves the overall Gap AUC by $2.1$ points and is more parsimonious, reusing one classifier for both partitioning and acquisition.

\begin{table}[H]
\centering\small
\caption{Gap AUC ($\uparrow$) of the unified LLMAT classifier vs.\ decoupled partitioning + independent LFBO (mean $\pm$ std over runs and six fixed feature models). For this paired comparison, GAP is normalized with a baseline $y_0$ shared by both methods (the worst initial value over all runs of both methods) instead of each run's own $y_0$, which avoids penalizing runs whose first query already lands near the optimum; values are therefore larger than in tables that use per-run $y_0$.}
\label{tab:decoupled}
\begin{tabular}{lccc}
\toprule
Dataset & LLMAT & Decoupled & Gain \\
\midrule
Redoxmer & 94.6$\pm$2.4 & 89.6$\pm$4.4 & +5.0 \\
Solvation & 92.0$\pm$3.2 & 86.2$\pm$4.0 & +5.8 \\
Kinase & 88.7$\pm$3.7 & 88.2$\pm$3.9 & +0.4 \\
Laser & 82.5$\pm$4.8 & 83.5$\pm$2.9 & $-$1.0 \\
Photovoltaics & 88.8$\pm$3.3 & 88.2$\pm$4.5 & +0.6 \\
Photoswitches & 88.8$\pm$6.0 & 86.7$\pm$5.8 & +2.0 \\
\midrule
Overall & 89.2$\pm$5.3 & 87.1$\pm$4.5 & +2.1 \\
\bottomrule
\end{tabular}
\end{table}

\subsection{Ablation: UCT vs.\ Greedy Selection}\label{sec:greedy}
The ``w/o MCTS'' ablation (tree depth $0$) removes recursive partitioning, node-local acquisition functions, and UCT-based path selection at once. To isolate the effect of UCT, we add a variant that keeps the learned tree, recursive partitioning, and node-local acquisition functions unchanged, but replaces UCT with greedy selection. Table~\ref{tab:greedy} reports Gap AUC over 15 runs per dataset with MolFormer features. UCT gives a modest improvement over greedy selection. We attribute the small gap to the partitioning mechanism: node classifiers are trained on labels induced by the ranking of property values, so the learned partitions are already aligned with the optimization objective and greedy selection can exploit them effectively, leaving less room for the exploration bonus. UCT nevertheless remains beneficial when partitions are imperfect, by balancing exploration and exploitation.

\begin{table}[H]
\centering\small
\caption{Gap AUC ($\uparrow$) of LLMAT variants with MolFormer features (15 runs per dataset; mean $\pm$ s.e.m.\ over all $6\times15$ runs). Gap AUC integrates GAP over rounds $t=1,\dots,100$. ``w/o MCTS'' uses separate runs of the single-node (LFBO) variant, so it differs slightly from depth $0$ in Table~\ref{tab:depth}.}
\label{tab:greedy}
\begin{tabular}{lc}
\toprule
Config & Gap AUC \\
\midrule
LLMAT w/ LLM Clustering & \textbf{82.36$\pm$1.49} \\
LLMAT w/ K-means Clustering & 80.24$\pm$1.63 \\
LLMAT Base (w/o Clustering) & 80.53$\pm$1.58 \\
LLMAT w/ Greedy Selection (UCT$\to$Greedy) & 80.29$\pm$1.62 \\
LLMAT Base w/o Meta ($\eta=0$) & 79.75$\pm$1.76 \\
LLMAT Base w/o MCTS (tree depth $=0$) & 78.39$\pm$1.89 \\
\bottomrule
\end{tabular}
\end{table}

\subsection{Meta-learning Dynamics at Deeper Nodes}\label{sec:meta-dynamics}
We examine per-node training behavior with and without meta-learning (Table~\ref{tab:meta-dynamics}). Shallow nodes behave similarly, while meta-learning yields consistently lower loss over much of the training process at deeper, data-scarce nodes, with the clearest improvement at depth $2$. At depth $1$, where each node holds only about $11$ samples, meta-learning reduces the final loss from $0.4269$ to $0.4121$; at depth $2$, with only $7$--$8$ samples per node, the reduction rises to about $14\%$ ($0.3221\to0.2772$). At the root (depth $0$) we do not observe an advantage. This supports the intended role of meta-learning: transferring information across nodes is most useful when local adaptation is data-limited, rather than uniformly improving every node.


\begin{table}[H]
\centering\small
\caption{Per-node training statistics with and without meta-learning, by tree depth. $n_{\text{pos}}$/$n_{\text{neg}}$: average numbers of positive/negative labels per node.}
\label{tab:meta-dynamics}
\begin{tabular}{llcccc}
\toprule
Depth & Method & Avg.\ \# samples & Final loss & Avg.\ $n_{\text{pos}}$ & Avg.\ $n_{\text{neg}}$ \\
\midrule
0 & meta & 25.4 & 0.1871$\pm$0.0187 & 7.9 & 17.5 \\
0 & w/o meta & 23.2 & 0.1439$\pm$0.0155 & 8.5 & 14.7 \\
1 & meta & 11.5 & 0.4121$\pm$0.0267 & 3.4 & 8.1 \\
1 & w/o meta & 11.4 & 0.4269$\pm$0.0290 & 3.5 & 7.9 \\
2 & meta & 7.6 & 0.2772$\pm$0.0330 & 2.3 & 5.4 \\
2 & w/o meta & 7.4 & 0.3221$\pm$0.0461 & 2.2 & 5.2 \\
\bottomrule
\end{tabular}
\end{table}

\subsection{Tree Depth Ablation}\label{sec:depth-ablation}
With fixed MolFormer features, we vary the maximum depth $L$ from $0$ (no partitioning) to $4$; the number of nodes is $2^{L+1}-1$, ranging from $1$ to $31$ (Table~\ref{tab:depth}). Increasing depth does not cause severe degradation: moderate partitioning generally improves performance, and depth $3$ improves Gap AUC over depth $0$ from $78.12$ to $80.08$ and reduces Regret@100 from $7.29$ to $5.81$ (about $20\%$). Early on, depth $1$ achieves the best GAP@25, suggesting shallow partitions help when few observations are available. Going to depth $4$ gives no further gain and slightly degrades several metrics, consistent with the trade-off between localization granularity and per-node sample size; the degradation is mild rather than catastrophic. The main cost of depth is computational: runtime grows from $43.9$\,s (depth $0$) to $131.6$\,s (depth $3$) and $171.6$\,s (depth $4$). Differences among depths $1$--$4$ in Gap AUC ($78.00$--$80.08$) are small relative to the run-to-run variability ($\pm1.5$--$1.8$), so we do not claim a sharp optimum; a shallow tree ($L=1$) already attains $79.60$ at about half the runtime of depth $3$ ($69.4$\,s vs.\ $131.6$\,s).

\begin{table}[H]
\centering\small
\caption{Tree-depth ablation with MolFormer features (mean $\pm$ s.e.m.\ over all $6\times15$ runs). Best per column in bold. Here Gap AUC integrates GAP over rounds $t=1,\dots,99$, so it is about $1$ point lower than the $t=1,\dots,100$ integral used in Table~\ref{tab:greedy} for the same configuration; depth $0$ uses the depth-$0$ runs of the tree algorithm rather than the separate LFBO runs of Table~\ref{tab:greedy}.}
\label{tab:depth}
\resizebox{\linewidth}{!}{%
\begin{tabular}{cccccrr}
\toprule
Depth & Gap AUC ($\uparrow$) & GAP@25 & GAP@50 & GAP@100 & Regret@100 ($\downarrow$) & Time (s) \\
\midrule
0 & 78.12$\pm$1.77 & 0.7341$\pm$0.0334 & 0.8671$\pm$0.0256 & \textbf{0.9748$\pm$0.0055} & 7.2938$\pm$1.9920 & 43.9$\pm$4.2 \\
1 & 79.60$\pm$1.57 & \textbf{0.7813$\pm$0.0291} & 0.9103$\pm$0.0163 & 0.9554$\pm$0.0125 & 6.1819$\pm$1.9737 & 69.4$\pm$6.7 \\
2 & 78.00$\pm$1.73 & 0.7473$\pm$0.0334 & 0.8831$\pm$0.0234 & 0.9630$\pm$0.0098 & 6.9169$\pm$2.0112 & 105.7$\pm$10.1 \\
3 & \textbf{80.08$\pm$1.53} & 0.7652$\pm$0.0318 & 0.9056$\pm$0.0198 & 0.9725$\pm$0.0073 & \textbf{5.8139$\pm$1.7825} & 131.6$\pm$13.5 \\
4 & 79.91$\pm$1.67 & 0.7701$\pm$0.0314 & \textbf{0.9129$\pm$0.0183} & 0.9628$\pm$0.0107 & 6.2854$\pm$2.1980 & 171.6$\pm$18.5 \\
\bottomrule
\end{tabular}}
\end{table}

\subsection{Cost of LLM-based Clustering}\label{sec:llm-cost}
One complete clustering pass across the six datasets consumes about $616$K input and $102$K output tokens ($\approx$718K total), with a summed API latency of about $71$ minutes. At current API pricing, the total cost is about \$2.59 (GPT-4o), \$1.81 (Gemini-2.5-Pro), and \$0.47 (Qwen3-Plus). To judge when this overhead is worthwhile, we define the average GAP gain over all BO rounds, in percentage points (pp):
\[
\Delta\mathrm{AUC\text{-}GAP}=100\left[\frac{1}{T}\sum_{t=1}^{T}\mathrm{GAP}^{\text{LLM}}_t-\frac{1}{T}\sum_{t=1}^{T}\mathrm{GAP}^{\text{Kmeans}}_t\right],
\]
so $\Delta\mathrm{AUC\text{-}GAP}=2$ means LLMAT with LLM clustering ($p{=}0.01$) improves GAP by $2$\,pp on average across the $T=100$ rounds over LLMAT with K-means clustering ($p{=}0$, no pruning). The largest gains occur on Kinase ($+5.77$\,pp) and PCE ($+4.83$\,pp), while gains are small or negative on Redox-Mer, Solvation, and Photoswitch. The Kinase gain is the largest and not driven by outliers (median $+5.08$\,pp, positive in $9$ of $15$ seeds), whereas the PCE mean is influenced by a few seeds (median $+1.32$\,pp). With $15$ seeds these differences are not statistically significant ($p>0.3$, paired tests), so they should be read as trends. LLM clustering is thus most justified when it yields a substantial improvement in sample efficiency, particularly when oracle evaluations are expensive; for simpler problems where K-means performs similarly, the overhead may not be warranted.

\begin{table}[H]
\centering\small
\caption{Cost, latency, and benefit of one LLM-clustering pass. $N$: number of candidates. $\Delta$AUC-GAP compares LLMAT with LLM-based clustering ($p{=}0.01$) against LLMAT with K-means clustering ($p{=}0$, i.e., no statistical pruning), using MolFormer features, tree depth 3 in code ($L{=}1$), and 15 seeds; it therefore reflects both the clustering method and the pruning setting.}
\label{tab:llm-cost}
\begin{tabular}{lrrrrrr}
\toprule
Dataset & $N$ & GPT-4o & Gemini & Qwen3 & Latency & $\Delta$AUC-GAP (pp) \\
\midrule
Redox-Mer & 1,407 & \$0.103 & \$0.076 & \$0.018 & 3.3 min & $-$0.937 \\
Solvation & 1,407 & \$0.102 & \$0.075 & \$0.018 & 3.3 min & +0.089 \\
Kinase & 10,449 & \$0.647 & \$0.482 & \$0.111 & 22.1 min & +5.773 \\
Laser & 10,000 & \$1.042 & \$0.671 & \$0.203 & 20.8 min & +1.149 \\
PCE & 10,000 & \$0.670 & \$0.485 & \$0.118 & 20.8 min & +4.833 \\
Photoswitch & 392 & \$0.023 & \$0.018 & \$0.004 & 0.8 min & $-$0.101 \\
\midrule
Total & & \$2.59 & \$1.81 & \$0.47 & 71 min & \\
\bottomrule
\end{tabular}
\end{table}

\subsection{Scalability w.r.t.\ Candidate Pool Size}\label{sec:scalability}
LLMAT assumes a finite candidate pool and scores all unobserved candidates at the root node in every BO round. Table~\ref{tab:scalability} reports the average per-round runtime (seconds) across T5-Chem and MolFormer features for pool sizes from $392$ to about $10$K. Prediction cost scales roughly linearly with pool size, as expected from scoring each candidate once at the root: growing the pool from $1{,}407$ to $10{,}000$ ($7.1\times$) increases LLMAT prediction time from $0.097$\,s to $0.673$\,s ($6.9\times$) while training time rises only from $0.590$\,s to $0.925$\,s, so total per-round time grows from $0.773$\,s to $1.713$\,s ($2.2\times$). At the largest pool ($10{,}449$) LLMAT needs about $1.6$--$1.7$\,s per round and remains faster overall than GP and Laplace. Scoring a finite pool is shared by pool-based discrete BO methods, and enumerating arbitrarily large chemical spaces remains infeasible; this can be mitigated by pre-clustering or pre-screening to relevant regions, candidate subsampling, batched inference, hierarchical retrieval, and parallel batch BO.

\begin{table}[H]
\centering\small
\caption{Average per-round runtime (s) vs.\ candidate pool size (T5-Chem and MolFormer features).}
\label{tab:scalability}
\resizebox{\linewidth}{!}{%
\begin{tabular}{rccccccccc}
\toprule
Pool size & \multicolumn{3}{c}{LLMAT} & \multicolumn{3}{c}{GP} & \multicolumn{3}{c}{Laplace} \\
\cmidrule(lr){2-4}\cmidrule(lr){5-7}\cmidrule(lr){8-10}
 & Train & Predict & Total & Train & Predict & Total & Train & Predict & Total \\
\midrule
392 & 0.680 & 0.058 & 0.816 & 1.521 & 0.014 & 1.548 & 1.323 & 0.181 & 1.516 \\
1,407 & 0.590 & 0.097 & 0.773 & 1.424 & 0.035 & 1.485 & 1.542 & 0.597 & 2.167 \\
10,000 & 0.925 & 0.673 & 1.713 & 2.091 & 0.272 & 2.616 & 2.836 & 4.513 & 7.557 \\
10,449 & 0.828 & 0.701 & 1.636 & 2.350 & 0.318 & 2.944 & 3.733 & 6.764 & 10.801 \\
\bottomrule
\end{tabular}}
\end{table}

\section{Broader Impacts}\label{sec:impact}
\begin{itemize}
    \item \textbf{Potential Positive Impacts.} The proposed methods can accelerate molecular discovery by leveraging knowledge from both general LLMs and domain-specific foundation models. They can be combined with generative approaches to propose novel, unseen molecules for drug and material design, offering significant societal benefits: new drugs may save lives, and new materials could contribute to environmental protection.
    \item \textbf{Potential Negative Impacts.} However, the method also carries potential risks. For example, the same capabilities could be misused to create toxic or harmful compounds, and overreliance on AI predictions without rigorous experimental validation could lead to unintended consequences. Responsible use, careful oversight, and appropriate safety measures are therefore critical.
\end{itemize}

\section{Declaration of LLMs Usage}\label{sec:llm_use}
We use LLMs as grammar correction tools and for improving written text flow, combined with human proofreading. No LLM was used for the idea or primary text of the paper. However, as this paper investigates how LLMs can help inform acquisition functions for BO over molecular datasets, we incorporate LLMs as algorithmic modules, where we also prompt ChatGPT to generate proper LLM-clustering prompts. LLM-based coding assistants were also used to help write the plotting code for the figures; the authors reviewed the code and the resulting figures.

\section{Theoretical Results}\label{sec:theory}

A complete regret guarantee for LLMAT would require jointly analyzing adaptive tree construction, node selection, local sample allocation, and acquisition learning. We therefore do not claim a general convergence theorem for the full algorithm. Instead, we first give a density-ratio and divergence interpretation of the node-wise LFBO classifier and show why recursive local classifiers can outperform a global one at the population level. We then provide an approximation--estimation--routing analysis under the PI special case, followed by a cumulative-regret decomposition under the probabilistic least-squares setting of BORE++ \citep{oliveira2022batch}.

\subsection{Density-ratio and divergence interpretation of local classifiers}\label{sec:theory_divergence}

We consider the PI case, $u(y;\tau)=\oneb(y>\tau)$, where the connection is most transparent. For a tree node $\Omega_k$, let $\mathbf{D}_t^k := \mathbf{D}_t \cap \Omega_k = \{(x_i,y_i)\in\mathbf{D}_t \mid x_i\in\Omega_k\}$ denote the observations assigned to this node. Following Sec.~\ref{sec:BO}, define
\begin{equation*}
    g_k(x) = p(x \mid y>\tau_k, \mathbf{D}_t^k), \quad l_k(x) = p(x \mid y\leq\tau_k, \mathbf{D}_t^k), \quad \gamma_k = p(y>\tau_k \mid \mathbf{D}_t^k),
\end{equation*}
and let the marginal distribution of all samples in the node be $p_k(x) = p(x\mid\mathbf{D}_t^k) = \gamma_k g_k(x) + (1-\gamma_k) l_k(x)$. Since $\mathbb{E}_{p_k}[u(y;\tau_k)\log\pi_\theta(x)] = \gamma_k\mathbb{E}_{g_k}[\log\pi_\theta(x)]$, the LFBO loss \citep{song_general_2022} in this node becomes
\begin{equation}\label{eq:node_lfbo_pi}
    \Lcal^{\tau_k}_{\mathbf{D}_t^k}(\theta) = -\gamma_k\,\mathbb{E}_{g_k}\big[\log\pi_\theta(x)\big] - \mathbb{E}_{p_k}\big[\log(1-\pi_\theta(x))\big].
\end{equation}

\begin{proposition}[Density-ratio and divergence interpretation]\label{prop:divergence}
Let $w_k = \gamma_k/(1+\gamma_k)$ and $M_k = w_k g_k + (1-w_k)p_k$. Over all measurable classifiers, the minimizer of Eq.~\eqref{eq:node_lfbo_pi} is
\begin{equation}\label{eq:node_optimal_classifier}
    \pi_k^\star(x) = \frac{\gamma_k g_k(x)}{\gamma_k g_k(x) + p_k(x)}, \qquad \frac{\pi_k^\star(x)}{1-\pi_k^\star(x)} = \gamma_k\frac{g_k(x)}{p_k(x)} = p(y>\tau_k \mid x, \mathbf{D}_t^k).
\end{equation}
Moreover, the optimal loss satisfies
\begin{equation}\label{eq:node_js}
    -\Lcal^{\tau_k}_{\mathbf{D}_t^k}(\theta^\star) = c(\gamma_k) + (1+\gamma_k)\,\mathrm{JS}_{w_k}(g_k\,\|\,p_k),
\end{equation}
where $\mathrm{JS}_{w}(g\|p) := w\KL(g\|M) + (1-w)\KL(p\|M)$ is the $w$-weighted Jensen--Shannon divergence and $c(\gamma_k) = \gamma_k\log\gamma_k - (1+\gamma_k)\log(1+\gamma_k)$ depends only on $\gamma_k$.
\end{proposition}

\begin{proof}
Since $(1+\gamma_k)w_k = \gamma_k$ and $(1+\gamma_k)(1-w_k) = 1$, Eq.~\eqref{eq:node_lfbo_pi} can be rewritten as
\begin{equation*}
    \Lcal^{\tau_k}_{\mathbf{D}_t^k}(\theta) = -(1+\gamma_k)\Big[w_k\,\mathbb{E}_{g_k}\log\pi_\theta(x) + (1-w_k)\,\mathbb{E}_{p_k}\log(1-\pi_\theta(x))\Big].
\end{equation*}
Pointwise maximization of $w_k g_k(x)\log\pi + (1-w_k)p_k(x)\log(1-\pi)$ over $\pi\in(0,1)$ gives $\pi_k^\star = w_k g_k/M_k$, which equals Eq.~\eqref{eq:node_optimal_classifier}; its odds equal $\gamma_k g_k/p_k$, which is $p(y>\tau_k\mid x,\mathbf{D}_t^k)$ by Bayes' rule. Substituting $\pi_k^\star = w_k g_k/M_k$ and $1-\pi_k^\star = (1-w_k)p_k/M_k$ yields
\begin{equation*}
    -\frac{\Lcal^{\tau_k}_{\mathbf{D}_t^k}(\theta^\star)}{1+\gamma_k} = w_k\log w_k + (1-w_k)\log(1-w_k) + w_k\KL(g_k\|M_k) + (1-w_k)\KL(p_k\|M_k).
\end{equation*}
The last two terms are exactly $\mathrm{JS}_{w_k}(g_k\|p_k)$, and multiplying the first two terms by $1+\gamma_k$ gives $c(\gamma_k)$.
\end{proof}

Proposition~\ref{prop:divergence} shows that the node-wise classifier is not an arbitrary discriminator: at the population optimum, \textbf{its odds exactly recover the local PI acquisition function}. Equivalently, it measures how strongly promising samples are enriched relative to the distribution of all samples currently contained in the node. Recursive partitioning uses this density-ratio signal to progressively restrict the search to regions with a higher probability of improvement. In the divergence view, the LFBO classifier distinguishes the \textbf{promising-sample distribution} $p(x\mid y>\tau_k,\mathbf{D}_t^k)$ from the \textbf{distribution of all samples in the current node} $p(x\mid\mathbf{D}_t^k)$, and its optimal discriminative gain is their weighted Jensen--Shannon divergence. Unlike GANs, LLMAT does not adversarially minimize this divergence; instead, it exploits the corresponding density-ratio signal to identify promising regions and recursively localize acquisition estimation.

\subsection{Why recursive local classifiers can outperform a global model}\label{sec:theory_local_vs_global}

Fix a partition $\{\Omega_k\}_{k=1}^K$ of the candidate space and a target acquisition (i.e., fixed utility and thresholds). Let $P(\Omega_k) := p(x\in\Omega_k)$ denote the probability mass of region $\Omega_k$ under the input distribution, and define the population LFBO loss conditioned on that region as
\begin{equation*}
    \Lcal_{\Omega_k}(S) = -\mathbb{E}_{(x,y)\mid x\in\Omega_k}\Big[u(y;\tau)f'(S(x)) - f^*\big(f'(S(x))\big)\Big],
\end{equation*}
where $f$ is the convex generator of the variational $f$-divergence objective in LFBO \citep{song_general_2022} and $f^*$ is its convex conjugate.

\begin{proposition}[Localization never increases the best-in-class population loss]\label{prop:local_vs_global}
For any model class $\Fcal$,
\begin{equation}\label{eq:local_vs_global}
    \sum_{k=1}^K P(\Omega_k)\inf_{S_k\in\Fcal}\Lcal_{\Omega_k}(S_k) \;\leq\; \inf_{S\in\Fcal}\sum_{k=1}^K P(\Omega_k)\Lcal_{\Omega_k}(S).
\end{equation}
The inequality is strict if and only if no single $S\in\Fcal$ simultaneously attains (or approaches) $\inf_{S_k\in\Fcal}\Lcal_{\Omega_k}(S_k)$ for every $k$ with $P(\Omega_k)>0$.
\end{proposition}

\begin{proof}
The left-hand side is the value of the same objective when the single function $S$ is replaced by $K$ independently chosen functions $S_1,\dots,S_K\in\Fcal$. Choosing $S_k = S$ for all $k$ recovers any feasible point of the right-hand side, so the left-hand side is no larger. Equality holds exactly when the region-wise infima can be achieved by a common $S$.
\end{proof}

The left-hand side of Eq.~\eqref{eq:local_vs_global} is the population loss of a separate estimator in each region, whereas the right-hand side is that of a single global estimator. Since the Bayes-optimal loss is identical on both sides, Proposition~\ref{prop:local_vs_global} equivalently states that, for a fixed partition and target acquisition, localization cannot increase the best-in-class excess risk. Note that if $\Fcal$ contains all measurable functions, equality holds, since the LFBO optimum is attained pointwise. The benefit of localization therefore stems from the limited capacity of practical classifiers: when the acquisition landscape is heterogeneous or multimodal, a global model must represent all modes simultaneously, whereas each local model only needs to approximate the acquisition structure within a restricted region. In LLMAT, each node additionally uses its own threshold $\tau_k$, so the local target itself adapts to the region; the comparison above isolates the approximation effect for a fixed target.

This does not imply that deeper localization always improves finite-sample estimation, because smaller regions contain fewer observations and therefore incur higher estimation variance. LLMAT controls this trade-off through the maximal tree depth $L$ and the minimal leaf sample size $m$ (Algo.~\ref{alg:MCTS}), which is consistent with the tree-depth ablation in Fig.~\ref{fig:molformer_tree_depth}. Sec.~\ref{sec:theory_decomposition} makes this approximation--estimation trade-off explicit, together with the additional routing error introduced by partitioning.

\subsection{Approximation--estimation--routing decomposition}\label{sec:theory_decomposition}

Let $x^\star \in \argmax_{x\in\Omega} f(x)$ and $\Delta_f = \max_x f(x) - \min_x f(x) < \infty$. We make the following assumptions.

\begin{assumption}\label{asm:noise}
Observations satisfy $y = f(x) + \varepsilon$, where the noise CDF $F_\varepsilon$ is strictly increasing.
\end{assumption}

\begin{assumption}\label{asm:pi}
Each node $k$ uses a fixed threshold $\tau_k$, with population PI classifier $\pi_k^\star(x) = p(y > \tau_k \mid x) = 1 - F_\varepsilon(\tau_k - f(x))$. Hence $\pi_k^\star$ and $f$ induce the same ranking within $\Omega_k$. We assume the corresponding inverse link is $L_\varepsilon$-Lipschitz over the relevant probability range.
\end{assumption}

\begin{assumption}\label{asm:selection}
At round $t$, the tree selects $\Omega_{k_t}$ based on the BO history, and $x_{t+1} \in \argmax_{x\in\Omega_{k_t}} \hat{\pi}_{t,k_t}(x)$.
\end{assumption}

\begin{assumption}[Low-dimensional local structure]\label{asm:local_dim}
Each local problem has effective dimension $d_k$, and the adaptive samples assigned to node $k$ sufficiently cover its relevant feature directions. We use the schematic scaling $e_{t,k} = \widetilde{O}\big(\sqrt{d_k/n_{t,k}}\big)$, where $n_{t,k}$ is the number of samples available to node $k$. This is a structural assumption used to expose the approximation--estimation trade-off, not an i.i.d.\ generalization claim.
\end{assumption}

Define the correct-routing event $E_t = \{x^\star \in \Omega_{k_t}\}$, its failure probability $q_t = p(E_t^c)$, and decompose the local classifier error as
\begin{equation*}
    \big\|\hat{\pi}_{t,k} - \pi_k^\star\big\|_{\infty,\Omega_k} \leq a_k + e_{t,k},
\end{equation*}
where $a_k$ and $e_{t,k}$ are the approximation and estimation errors, respectively.

On $E_t$, since $x_{t+1}$ maximizes $\hat{\pi}_{t,k_t}$, we have $\pi_{k_t}^\star(x^\star) - \pi_{k_t}^\star(x_{t+1}) \leq 2(a_{k_t} + e_{t,k_t})$, and therefore $r_{t+1} \leq 2L_\varepsilon(a_{k_t} + e_{t,k_t})$. If routing fails, $r_{t+1} \leq \Delta_f$. Hence
\begin{equation}\label{eq:routing_regret}
    \mathbb{E}[r_{t+1}] \leq 2L_\varepsilon(1-q_t)\, \mathbb{E}\left[a_{k_t} + e_{t,k_t} \mid E_t\right] + q_t\Delta_f.
\end{equation}
For a single global classifier with approximation and estimation errors $a_0$ and $e_{t,0}$, $\mathbb{E}\big[r_{t+1}^{(0)}\big] \leq 2L_\varepsilon\, \mathbb{E}[a_0 + e_{t,0}]$. Therefore, localization has a tighter upper bound whenever
\begin{equation}\label{eq:localization_boundary}
    (1-q_t)\, \mathbb{E}\left[a_{k_t} + e_{t,k_t} \mid E_t\right] + \frac{\Delta_f}{2L_\varepsilon}\, q_t < \mathbb{E}[a_0 + e_{t,0}].
\end{equation}
This gives the basic boundary: localization can reduce approximation and statistical complexity, but it also fragments samples and introduces routing error.

Under Assumption~\ref{asm:local_dim}, and similarly writing $e_{t,0} = \widetilde{O}\big(\sqrt{d_0/n_t}\big)$ for the global classifier, the boundary becomes schematically
\begin{equation}\label{eq:dimension_boundary}
    (1-q_t)\, \mathbb{E}\left[a_{k_t} + \sqrt{\frac{d_{k_t}}{n_{t,k_t}}} \;\middle|\; E_t\right] + \frac{\Delta_f}{2L_\varepsilon}\, q_t < a_0 + \sqrt{\frac{d_0}{n_t}}.
\end{equation}
Ignoring approximation and routing errors, localization improves the estimation term when $d_k/n_{t,k} < d_0/n_t$. Letting $\rho_{t,k} = n_{t,k}/n_t$, this is equivalent to $d_k < \rho_{t,k}\, d_0$. Thus, the reduction in effective dimension must be large enough to compensate for sample fragmentation. Recursive localization is therefore most advantageous when the global problem is heterogeneous or piecewise low-dimensional, so that promising local regions satisfy $d_k \ll d_0$ while still receiving sufficient samples.

\subsection{Cumulative-regret decomposition}\label{sec:theory_cumulative}

Because BO observations are adaptively selected, Assumption~\ref{asm:local_dim} should not be interpreted as an ordinary i.i.d.\ generalization bound. For cumulative regret, we further specialize to the adaptive probabilistic least-squares (PLS)/RKHS setting analyzed in BORE++ \citep{oliveira2022batch}.

\begin{assumption}\label{asm:rkhs}
For each visited node $k$, $\pi_k^\star$ belongs to an RKHS $\mathcal{H}_k$ with bounded norm, and the local binary residuals form a conditionally sub-Gaussian martingale difference sequence. The local query and node-assignment decisions are predictable with respect to the BO filtration.
\end{assumption}

Applying the adaptive confidence result of \citet[Theorem~1]{oliveira2022batch} node-wise gives $\big|\hat{\pi}_{t,k}(x) - \pi_k^\star(x)\big| \leq \beta_{t,k}\sigma_{t,k}(x)$ with high probability. Following the BORE regret argument \citep[Theorem~2]{oliveira2022batch}, on correctly routed rounds, $r_{t+1} \leq L_\varepsilon \beta_{t,k_t}\big[\sigma_{t,k_t}(x^\star) + \sigma_{t,k_t}(x_{t+1})\big]$. Therefore,
\begin{equation}\label{eq:cumulative_regret}
    R_T \leq L_\varepsilon \sum_{t=0}^{T-1} \beta_{t,k_t}\big[\sigma_{t,k_t}(x^\star) + \sigma_{t,k_t}(x_{t+1})\big]\mathbf{1}_{E_t} + \Delta_f \sum_{t=0}^{T-1} \mathbf{1}_{E_t^c}.
\end{equation}
Thus, sublinear cumulative regret requires both the aggregate local uncertainty and the number of routing failures to be sublinear in $T$. Low-complexity local models can reduce the former when each promising region receives sufficient samples.

Importantly, LLMAT, like the original BORE algorithm, maximizes $\hat{\pi}$ rather than a classifier UCB. Hence the uncertainty term at $x^\star$ remains, exactly as in BORE \citep[Theorem~2]{oliveira2022batch}. BORE++ removes this term by maximizing a classifier UCB, yielding the stronger no-regret result in \citet[Theorem~3]{oliveira2022batch}.

We therefore present this as a \textbf{regret decomposition and boundary characterization}, rather than a general no-regret theorem. It identifies three conditions under which recursive localization can be beneficial: \textbf{lower local approximation error or effective dimension, sufficient local sampling, and sufficiently accurate routing}.


\end{document}